\documentclass[conference,letterpaper]{IEEEtran}

\usepackage[dvipsnames]{xcolor}
\usepackage{xspace}
\usepackage{graphicx}
\usepackage{amsmath}
\usepackage{amssymb}
\usepackage{booktabs}
\usepackage[numbers,sort&compress]{natbib}
\usepackage{etoolbox}
\usepackage[format=plain,labelformat=simple,labelsep=period,font=small,compatibility=false]{caption}
\usepackage[font=footnotesize,skip=3pt,subrefformat=parens]{subcaption}
\usepackage[hyphens]{url}
\usepackage[shortlabels,inline]{enumitem}
\setlist[itemize]{noitemsep,leftmargin=*,topsep=0em}
\setlist[enumerate]{noitemsep,leftmargin=*,topsep=0em}

\renewcommand{\dblfloatpagefraction}{0.99}

\makeatletter
\renewcommand{\thesection}{\arabic{section}}

\renewcommand{\section}{\@startsection{section}{1}{\z@}{-1.5ex plus -0.5ex minus -0.2ex}{0.8ex plus 0.2ex}{\normalfont\large\bfseries}}
\renewcommand{\subsection}{\@startsection{subsection}{2}{\z@}{-1.2ex plus -0.4ex minus -0.2ex}{0.6ex plus 0.2ex}{\normalfont\normalsize\bfseries}}
\renewcommand{\subsubsection}{\@startsection{subsubsection}{3}{\z@}{-1.0ex plus -0.3ex minus -0.2ex}{0.4ex plus 0.1ex}{\normalfont\normalsize\bfseries}}
\renewcommand{\paragraph}{\@startsection{paragraph}{4}{\z@}{1.0ex plus 0.3ex minus 0.2ex}{-1em}{\normalfont\normalsize\bfseries}}
\makeatother

\graphicspath{{figures/images/}}

\usepackage{algorithm}
\usepackage{algpseudocode}

\usepackage{amsthm}
\usepackage{thm-restate}   %
\usepackage{needspace}   %
\usepackage{stfloats}    %
\usepackage{tikz}        %
\usetikzlibrary{positioning,arrows.meta,shapes.multipart,calc}   %
\theoremstyle{plain}

\theoremstyle{definition}

\newtheorem{assumption}{Assumption}

\theoremstyle{plain}
\newtheorem{slemma}{Lemma}[section]
\newtheorem{sproposition}{Proposition}[section]
\newtheorem{scorollary}{Corollary}[section]
\theoremstyle{definition}

\newlist{propparts}{enumerate}{1}
\setlist[propparts]{label=(\alph*),ref=\theproposition(\alph*),leftmargin=*,nosep}
\newlist{thmparts}{enumerate}{1}
\setlist[thmparts]{label=(\alph*),ref=\thetheorem(\alph*),leftmargin=*,nosep}
\AtEndPreamble{%
  \crefname{assumption}{Assumption}{Assumptions}%
  \Crefname{assumption}{Assumption}{Assumptions}%
  \crefname{proppartsi}{Proposition}{Propositions}%
  \Crefname{proppartsi}{Proposition}{Propositions}%
  \crefname{thmpartsi}{Theorem}{Theorems}%
  \Crefname{thmpartsi}{Theorem}{Theorems}%
  \crefname{slemma}{Lemma}{Lemmas}%
  \Crefname{slemma}{Lemma}{Lemmas}%
  \crefname{sproposition}{Proposition}{Propositions}%
  \Crefname{sproposition}{Proposition}{Propositions}%
  \crefname{scorollary}{Corollary}{Corollaries}%
  \Crefname{scorollary}{Corollary}{Corollaries}%
  \crefname{sremark}{Remark}{Remarks}%
  \Crefname{sremark}{Remark}{Remarks}%
}

\newcommand{\runin}[1]{\textit{#1.}\ }
\newcommand{\intuition}{\par\noindent\runin{Intuition}}

\newcommand{\inSupp}{ in the supplementary material}

\definecolor{linkblue}{rgb}{0.21,0.49,0.74}
\usepackage[pagebackref,breaklinks,colorlinks,allcolors=linkblue]{hyperref}
\usepackage[capitalize]{cleveref}
\crefname{section}{Sec.}{Secs.}
\Crefname{section}{Section}{Sections}
\Crefname{table}{Table}{Tables}
\crefname{table}{Tab.}{Tabs.}

\title{Perception--Alignment Tradeoff in CFG Based Sampling}

\author{%
\IEEEauthorblockN{Gal Metzer}
\IEEEauthorblockA{Tel Aviv University}
\and
\IEEEauthorblockN{Adar Elad}
\IEEEauthorblockA{Tel Aviv University}
\and
\IEEEauthorblockN{Daniel Cohen-Or}
\IEEEauthorblockA{Tel Aviv University}
\and
\IEEEauthorblockN{Raja Giryes}
\IEEEauthorblockA{Tel Aviv University}
}

\begin{document}
\bstctlcite{IEEEexample:BSTcontrol}
\maketitle
\makeatletter\global\@topnum\z@\makeatother
\pagestyle{plain}

\begin{abstract}
Classifier-free guidance (CFG) is a standard and highly effective tool for conditioning diffusion models. Yet, its scale is largely chosen empirically: too little guidance weakens prompt adherence, while too much reduces diversity and introduces artifacts. Moderate guidance often improves image quality, but how much alignment does it gain, at what perceptual cost, and could other methods do better?
We address these questions through an \textbf{information-theoretic framework} and prove a sharp, \textbf{convex tradeoff}: greater condition alignment incurs an unavoidable, increasingly steep perceptual cost. This tradeoff is \textbf{intrinsic} to the model, computable directly from its conditional and unconditional denoisers without additional critics.
We extend this tradeoff to a third dimension, \textbf{diversity}, by measuring how much of the model's unconditional distribution remains covered under guidance. Experiments across diverse, widely adopted diffusion models show that standard CFG falls short of the optimal frontier and that alternative guidance methods can outperform it, turning guidance-scale selection from trial and error into a measurable tradeoff.
\end{abstract}
\section{Introduction}
\label{sec:intro}

\begin{figure}[t]
    \centering
    \includegraphics[width=\linewidth]{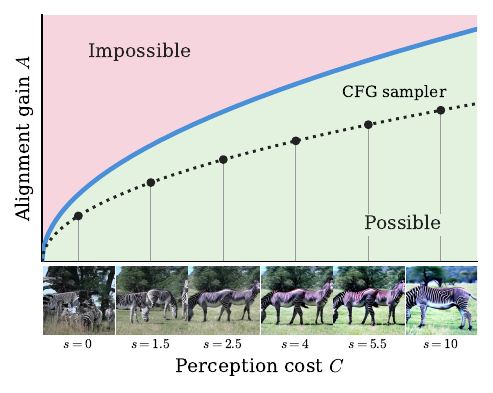}
    \vspace{-9mm}
    \caption{\textbf{No free lunch for guidance.}
    Stronger prompt alignment always costs naturalness, and every extra gain costs more than the last.
    Using only the two denoisers CFG already has, we trace the frontier between what guidance can and cannot achieve, and show that the CFG sampler falls short of it.}
    \label{fig:teaser}
\end{figure}

Diffusion models have demonstrated remarkable power in generating rich, realistic images from text. 
Yet their success leaves a central challenge: producing high-quality images that closely follow the requested prompt. 
These goals do not always improve together, and stronger prompt alignment can come at the expense of visual quality and diversity. 
Classifier-free guidance (CFG)~\cite{ho2022classifierfree} is a widely used means of strengthening alignment, controlled by a single guidance scale. 
Increasing this scale encourages closer adherence to the prompt, but can also narrow the range of generated images and introduce oversaturation or unnatural details. 

Despite the widespread use of CFG, the balance it governs remains incompletely understood: what determines the cost of stronger alignment, and how much of that cost is unavoidable? 
These questions make the choice of guidance scale more than an empirical tuning decision. 
They call for examining what the guidance changes in the generated distribution and the limits of the balance it achieves.

Beyond the fundamental questions raised by this tradeoff, understanding it also has a direct practical consequence on choosing and controlling the guidance scale. 
This seemingly simple parameter governs a complex balance, yet its value is typically chosen through trial and error, often using the same scale across conditions that may respond very differently to guidance. 
A clearer account of the tradeoff could therefore serve both purposes: reveal the underlying limits of stronger alignment and clarify the consequences of applying different guidance strengths to a given condition.

To study these questions, we utilize relationships between information and estimation, developing an information-theoretic account of the perception–alignment tradeoff in CFG.
Through \Cref{thm:renyi-immse}, we build on the connection
between denoiser disagreement and distributional divergence,
applying it to the conditional and unconditional denoisers
of CFG.
This allows us to quantify the distributional cost of stronger alignment, which we call the perception cost, and characterize the fundamental limits of this balance. 
The same framework gives the guidance scale a precise interpretation, clarifying how guidance strength shapes this tradeoff.

Our central contribution is formulating guidance as a perception--alignment tradeoff.
Inspired by the perception--distortion perspective~\cite{blau2018perception}, we measure alignment with the implicit classifier that CFG already uses, and cost as the divergence from the model's own conditional distribution.
Within this formulation, we show a sharp lower bound on the cost of a given alignment gain, and the distribution that attains it is the one whose score CFG follows.
Via the Gibbs variational bound~\cite{csiszar1975divergence}, we show the bound is convex, and identify the ideal tradeoff no practical sampler can surpass (\cref{fig:teaser}).
Building on these findings, we establish the optimal frontier of the tradeoff, compute it for practical models, and quantify
how far the CFG sampler falls short of this optimum.
In addition, our experiments provide empirical evidence for the behaviour predicted by the theory, revealing which conditions permit alignment gains at a lower distributional cost.
Together, the theoretical bound and its empirical support establish the perception--alignment tradeoff as a quantitative framework for understanding the consequences of guidance strength.

The same framework also answers a complementary question: how much of the broader image distribution does a condition (e.g. class or prompt) cover?
A condition may describe a broad range of possibilities or select a narrow portion of that range, and guidance narrows it further.
We develop a \emph{relative diversity index} that uses the information available in the denoisers to quantify this coverage, both for the condition itself and for what is lost as the guidance scale increases.
Diversity thus joins the same picture: what guidance gains in alignment, it pays for in both distributional cost and diversity, connecting the information-theoretic view of CFG to the practical interest in understanding diversity across classes and prompts.

\section{Related Work}
\label{sec:related}

\textbf{Guidance and its tradeoff.}
Classifier guidance adds the gradient of a classifier, trained on noisy images, to the score of a diffusion model~\cite{dhariwal2021diffusion}.
Classifier-free guidance (CFG) gets the same direction from a conditional and an unconditional denoiser~\cite{ho2022classifierfree}, and text-to-image models use it as standard~\cite{saharia2022photorealistic,rombach2022high}.
As with the GAN truncation trick, a higher guidance scale gives more quality and alignment but less diversity~\cite{brock2019large,ho2022classifierfree,karras2024guiding}, and very high scales cause oversaturation~\cite{saharia2022photorealistic,sadat2025eliminating}.
Some methods improve this balance: they anneal the condition~\cite{sadat2024cads}, limit guidance to middle noise levels~\cite{kynkaanniemi2024applying}, learn a guidance schedule~\cite{yehezkel2025navigating}, or swap the unconditional denoiser for a weaker model~\cite{karras2024guiding} or a better-trained one~\cite{phunyaphibarn2026unconditional}.
Other methods project the update~\cite{sadat2025eliminating}, keep it near the data manifold as inverse solvers do~\cite{chung2025cfgpp,chung2022manifold,chung2023posterior}, or iterate it toward consistent paths~\cite{wang2025golden}.
We do not propose a new sampler but give a limit for every output distribution of a fixed model, and we measure the position of CFG relative to this limit.

\textbf{CFG sampler.}
At each noise level, the CFG score equals the score of the conditional density reweighted to favour the condition, yet the CFG sampler does not produce this reweighted distribution~\cite{chidambaram2024guidance,bradley2025classifier}.
Instead, strong guidance pushes samples toward the edge of the class support~\cite{chidambaram2024guidance}, and CFG behaves like alternating denoising and sharpening steps~\cite{bradley2025classifier}.
Moufad et al.~\cite{moufad2025conditional} trace this gap to a missing R\'enyi divergence term and reduce it with repeated noising and denoising.
Feynman--Kac correctors reach the reweighted distribution with weighted, resampled particles~\cite{skreta2025feynman}.
Related reward-guided samplers use reward-aware particle
initialization~\cite{yoon2025psi} or steer the injected noise
rather than the reverse mean~\cite{hwang2026noisetilt}.
For Gaussian mixtures, guidance makes samples easier to classify but less diverse~\cite{wu2024theoretical}, although this sharpening vanishes in high dimension~\cite{pavasovic2025classifier}.
In concurrent work, Chen et al.~\cite{chen2026information} tune guidance schedules so that the sampler approaches a reweighted clean distribution, using a related balance of alignment and KL divergence as a fitting loss.
We instead use the fact that, at each noise level, the reweighted distribution gives the most alignment for any KL cost~\cite{csiszar1975divergence}, and show that the sampler's path stays on or below this limit.

\textbf{Information and estimation.}
The I-MMSE identity relates mutual information in a Gaussian channel to the minimum mean-square error~\cite{guo2005mutual}.
Its mismatched form relates relative entropy to the excess error of an estimator that assumes the wrong distribution~\cite{verdu2010mismatched}.
Both identities also hold pointwise, for a single noise path~\cite{venkat2012pointwise}.
Diffusion denoisers learn these estimators, and prior work uses them to compute likelihoods~\cite{kong2023information}, pointwise mutual information~\cite{kong2024interpretable,wang2025information}, perceptual distances~\cite{ohayon2026distance}, and zero-shot classifiers~\cite{li2023diffusion}.
MINDE estimates KL divergence and mutual information from score differences with the Girsanov theorem~\cite{franzese2024minde}, and the same theorem bounds diffusion likelihoods~\cite{song2021maximum}.
These works measure a fixed model; we use the same tools to measure what guidance changes.
The mismatched identity has a counterpart for R\'enyi divergences~\cite{vanerven2014renyi}, known for Langevin dynamics toward a fixed target~\cite{vempala2019rapid} and, closest to ours, for two densities under the same noise~\cite{shen2026renyi}, where it motivates a training objective.
Shen et al.~\cite{shen2026renyi} compare the data with a model and use the identity as a training objective. We apply it to the two denoisers of one guided model, where the order becomes the guidance scale, and use it to measure guidance rather than to train.

\textbf{Tradeoffs with a divergence cost.}
Blau and Michaeli~\cite{blau2018perception,blau2019rethinking} showed that image restoration and compression face a hard limit: making outputs more accurate eventually makes them look less natural.
We show that guidance faces a similar limit, where stronger alignment with the condition eventually costs naturalness.
In our case, the limit has a simple exact solution~\cite{csiszar1975divergence}, and CFG points directly at it, so the limit can be computed from the model itself.
Reward fine-tuning targets the same solution~\cite{korbak2022rl,domingoenrich2025adjoint}, but reaches it by retraining the model. Guidance aims at it during sampling, with no training.

\definecolor{igHead}{HTML}{353F47}\definecolor{igGold}{HTML}{B6883F}\definecolor{igOchre}{HTML}{A77020}\definecolor{igEq}{HTML}{E1E7EC}
\definecolor{igCream}{HTML}{F9EAC5}\definecolor{igOf}{HTML}{F1EDE9}\definecolor{igO}{HTML}{999999}\definecolor{igArr}{HTML}{363E46}
\colorlet{igG}{igHead}\colorlet{igGf}{igEq}\colorlet{igD}{igHead}\colorlet{igDf}{igEq}\colorlet{igF}{igHead}\colorlet{igFf}{igEq}
\begin{figure}[t]
    \centering
    \resizebox{\linewidth}{!}{%
    \begin{tikzpicture}[
      font=\scriptsize,
      card/.style n args={3}{rectangle split, rectangle split parts=2, rectangle split part fill={#1, #2}, draw=igOchre,
        rounded corners=2.5pt, line width=.8pt, align=center, inner xsep=3pt, inner ysep=4.5pt},   %
      ttl/.style={text=white, font=\scriptsize\bfseries, align=center},
      badge/.style={circle, fill=igGold, draw=white, line width=.9pt, text=white, font=\tiny\bfseries, inner sep=1pt,
        minimum size=3.6mm},
      ocard/.style={draw=igO, fill=igOf, rounded corners=2.5pt, line width=.6pt, align=left, inner sep=2.5pt},
      tag/.style={draw=igOchre, fill=igCream, rounded corners=2pt, font=\tiny\bfseries, inner sep=1.3pt, line width=.4pt, text=igHead},
      arr/.style={-{Stealth[length=3.5pt]}, line width=.7pt, igArr},
      ]
      \node[card={igG}{igGf}{igG}] (G) {\rule{0pt}{2.6mm}\textcolor{white}{\textbf{Denoiser}}\\\textcolor{white}{\textbf{disagreement}}%
        \nodepart{two}\rule[-1.9mm]{0pt}{5.6mm}\small$\mathbb E_{p_{\gamma,s}}\|\hat x_c-\hat x\|^2$};
      \node[card={igD}{igDf}{igD}, anchor=text split west] (D) at ($(G.text split east)+(11mm,0)$) {\rule{0pt}{2.6mm}\textcolor{white}{\textbf{R\'enyi}}\\\textcolor{white}{\textbf{divergence}}%
        \nodepart{two}\rule[-1.9mm]{0pt}{5.6mm}\small$D_{1+s}(p^c_\gamma\|p_\gamma)$};
      \node[card={igF}{igFf}{igF}, anchor=text split west] (F) at ($(D.text split east)+(11mm,0)$) {\rule{0pt}{2.6mm}\textcolor{white}{\textbf{Log-}}\\\textcolor{white}{\textbf{normalizer}}%
        \nodepart{two}\rule[-1.9mm]{0pt}{5.6mm}\small$F=\log Z_\gamma(s)$};
      \foreach \n/\c/\k in {G/igG/1, D/igD/2, F/igF/3} \node[badge=\c] at (\n.north) {\k};   %
      \draw[arr] (G.text split east) -- (D.text split west) node[tag=igG, midway, above=2.5pt] {Thm.~\ref{thm:renyi-immse}};
      \draw[arr] (D.text split east) -- (F.text split west) node[tag=igD, midway, above=2.5pt] {Prop.~\ref{prop:log-normalizer}};
      \node[ocard, anchor=west] (C) at ($(F.text split east)+(12mm,0)$) {\textbf{Cost (\cref{sec:tradeoff})}\\$C=s\partial_sF-F$};
      \node[ocard, above=1.5mm of C.north west, anchor=south west] (A) {\textbf{Alignment (\cref{sec:tradeoff})}\\$A=\partial_sF-\partial_sF|_{s=0}$};
      \node[ocard, below=1.5mm of C.south west, anchor=north west] (V) {\textbf{Coverage loss (\cref{sec:diversity})}\\$V=(1{+}s)\partial_sF-F$};
      \draw[arr] (F.text split east) to[out=0, in=180] (A.west);
      \draw[arr] (F.text split east) -- (C.west);
      \draw[arr] (F.text split east) to[out=0, in=180] (V.west);
    \end{tikzpicture}}
    \vspace{-5mm}
    \caption{\textbf{Theorem Flow.} The disagreement of the two denoisers is used to approximate the R\'enyi divergence (\cref{thm:renyi-immse}); the divergence is the log-normalizer of the tilt (\cref{prop:log-normalizer}); and the quantities of \cref{sec:tradeoff,sec:diversity} are $F$ and its derivatives in $s$.}
    \label{fig:identity-graph}
\end{figure}
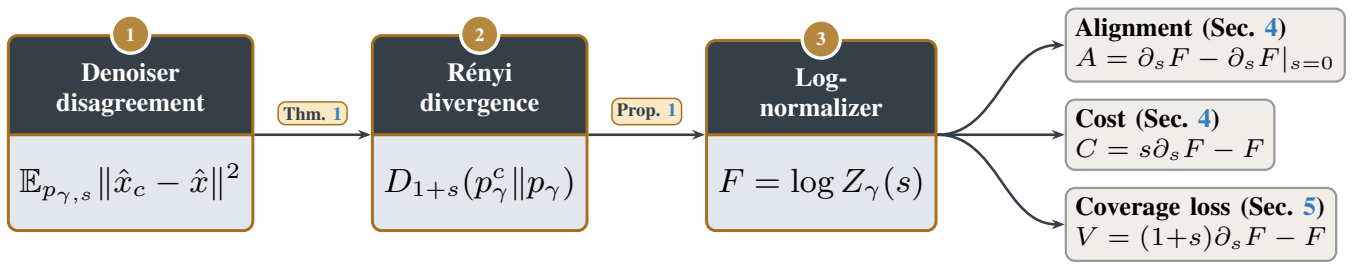

\section{An Information-Theoretic View of CFG}
\label{sec:cfg_info}
This section connects CFG to information theory.
First, we briefly introduce CFG.
Then, we present the main results that \Cref{sec:tradeoff,sec:diversity} build on; \cref{fig:identity-graph} summarizes how they relate. \Cref{thm:renyi-immse} and \Cref{prop:log-normalizer} are proved in full in \cref{sec:supp:theory} of the supplementary material.

It's worth noting that we use the word \emph{law} for the probability distribution of a random variable; for a law with a density, the two words have the same meaning.
\Cref{sec:tradeoff} compares the laws of full sampling trajectories, which do not have densities in the usual sense.

\noindent \textbf{Diffusion Models.}
A diffusion model is a generative procedure that iteratively cleans Gaussian noise to obtain the clean sample $X$,
\vspace{-2mm}
\begin{equation}
    Z_\sigma=X+\sigma\varepsilon,
    \qquad \varepsilon\sim\mathcal N(0,I),
    \label{eq:sigma-channel}
\end{equation}
To do so, it learns a \emph{denoiser} $\hat x(z,\sigma)=\mathbb E[X\mid Z_\sigma=z]$, or equivalently a noise predictor $\varepsilon_\theta(z,\sigma)=(z-\hat x)/\sigma$.
By Tweedie's formula~\cite{efron2011tweedie}, both give the score of the noisy density $p_\sigma$ of $Z_\sigma$: $\nabla_z\log p_\sigma(z)=(\hat x-z)/\sigma^2=-\varepsilon_\theta/\sigma$~\cite{song2021score}.
Sampling follows this score from large $\sigma$ down to $\sigma=0$.
Later, it will be useful to parameterize the noise by the signal-to-noise ratio (SNR) $\gamma=\sigma^{-2}$,

\vspace{-3mm}
\begin{equation}
    Y_\gamma=\gamma Z_\sigma=\gamma X+\sqrt\gamma\,\varepsilon,
    \label{eq:channel}
\end{equation}

which is a Gaussian channel: $\gamma=0$ is pure noise and $\gamma\to\infty$ is clean data.
We state the theory with $\gamma$, and we give the noise levels of the experiments as $\sigma$.

\noindent \textbf{Classifier-Free Guidance.}
A conditional model has two noisy densities at each $\gamma$ level, $p^c_\gamma$ for $X\sim p^c$ and $p_\gamma$ for $X\sim p$, and two denoisers, the \emph{conditional} $\hat x_c(y,\gamma)=\mathbb E[X\mid Y_\gamma=y,c]$ and the \emph{unconditional} $\hat x(y,\gamma)=\mathbb E[X\mid Y_\gamma=y]$ one; their difference $g=\hat x_c-\hat x$ is the \emph{CFG direction}.
CFG~\cite{ho2022classifierfree} combines the two noise predictions,
\begin{equation}
    \tilde\varepsilon_\theta
    =(1+s)\varepsilon_\theta(\cdot,c)
    -s\varepsilon_\theta(\cdot,\varnothing),
    \label{eq:cfg-noise}
\end{equation}
with guidance scale $s\ge0$ ($w=1+s$ in the common convention, so $s=0$ is conditional sampling), and the sampler uses this combination in place of the conditional prediction at every step.
Noise predictions are linear in scores, so the same combination gives the ideal CFG score:
\begin{align}
    v_s(y)
    &=(1+s)\nabla_y\log p^c_\gamma(y)
      -s\nabla_y\log p_\gamma(y)\notag\\
    &=\underbrace{\nabla_y\log p^c_\gamma(y)}_{\text{conditional score}}
      +s\underbrace{\nabla_y\log p(c\mid Y_\gamma=y)}_{\text{implicit classifier guidance}}
    \label{eq:cfg-score}
\end{align}

\noindent We call $p(c\mid Y_\gamma=y)$ the \emph{implicit classifier}.

\Cref{eq:cfg-score} also provides another realization: the \textit{tilted density} $p_{\gamma,s}$, whose score CFG follows:
\begin{equation}
    p_{\gamma,s}=\frac{p^c_\gamma e^{sT_\gamma}}{Z_\gamma(s)}
    \propto(p^c_\gamma)^{1+s}p_\gamma^{-s}.
    \label{eq:cfg-tilt}
\end{equation}

Here the \emph{alignment score} is the log-likelihood ratio $T_\gamma=\log(p^c_\gamma/p_\gamma)$, which, by Bayes' rule, equals $\log p(c\mid Y_\gamma=y)$ up to a constant.
The tilt is defined for $s>-1$ with a finite normalizer $Z_\gamma(s)=\mathbb E_{p^c_\gamma}e^{sT_\gamma}$.

\noindent \textbf{Assumptions.}
Our results assume that the two denoisers are the exact posterior means of the same two clean laws at all noise levels (\cref{asm:exact}), which trained models satisfy only approximately.
The remaining assumptions (\cref{asm:finite,asm:regular,asm:zero-snr,asm:girsanov}) are usual technical conditions.
All five are stated precisely in \cref{sec:supp:assumptions} of the supplementary material.

\noindent \textbf{Denoising Errors Measure Divergences.}
We were intrigued by the recent use of information-estimation identities with diffusion denoisers (IEM~\cite{ohayon2026distance}).
Its underlying identities are I-MMSE and its mismatched version~\cite{guo2005mutual, verdu2010mismatched, venkat2012pointwise}.
The mismatched version compares two laws through their denoisers, and CFG evaluates exactly such a pair.
For this pair we obtain not one divergence but a whole family, indexed by the guidance scale (\cref{thm:renyi-immse}).

\needspace{5\baselineskip}
\begin{restatable}[R\'enyi I-MMSE for CFG]{theorem}{thmRenyiIMMSE}
\label{thm:renyi-immse}
Measurable denoiser disagreement is a bridge for R\'enyi divergence.
Let \cref{asm:exact,asm:finite,asm:regular,asm:zero-snr} hold.
For each $s>-1$ for which the tilt exists,
\vspace{-2mm}
\begin{equation}
    \frac{d}{d\gamma}D_{1+s}(p^c_\gamma\|p_\gamma)
    =\frac{1+s}{2}\,\mathbb E_{p_{\gamma,s}}\|\hat x_c-\hat x\|^2,
    \label{eq:renyi-immse}
\end{equation}
where $D_{1+s}$ is the R\'enyi divergence of order $1+s$.
Taking the integral from zero SNR to a horizon $\Gamma$ gives
\vspace{-2mm}
\begin{equation}
    \boxed{
    D_{1+s}(p^c_\Gamma\|p_\Gamma)
    =\frac{1+s}{2}G(\Gamma,s),}
    \label{eq:renyi-integral}
\end{equation}
where $G(\Gamma,s)=\int_0^\Gamma\mathbb E_{p_{\gamma,s}}\|\hat x_c-\hat x\|^2\,d\gamma=\int_0^\Gamma\mathbb E_{p_{\gamma,s}}\|g\|^2\,d\gamma$ is the accumulated disagreement under the tilt.
\end{restatable}

\noindent Shen et al.~\cite{shen2026renyi} give this identity for two general densities, as a derivative in the noise variance (\cref{eq:supp:renyi-heat} in the supplementary material).
Our contribution is its reading for guidance: the order is the guidance scale, the average is under the CFG tilt, and the rate is the disagreement of the two denoisers that CFG already evaluates.
The reason this family matters for the applications becomes clear next.

\noindent \textbf{The Log-Normalizer.}
Due to the tilt \cref{eq:cfg-tilt} being an exponential family in $s$, its log-partition function $F(\gamma,s):=\log Z_\gamma(s)$ is the divergence that \cref{thm:renyi-immse} measures.

\begin{restatable}[The log-normalizer is a R\'enyi divergence]{proposition}{propLogNormalizer}
\label{prop:log-normalizer}
For each $s>-1$ with $Z_\gamma(s)<\infty$, and with \cref{eq:renyi-integral} for the second equality,
\vspace{-2mm}
\begin{equation}
    \boxed{
    F(\gamma,s)=sD_{1+s}(p^c_\gamma\|p_\gamma)=\frac{s(1+s)}2G(\gamma,s).}
    \label{eq:log-normalizer}
\end{equation}
\end{restatable}
\vspace{-0.5mm}
\runin{Why it matters}
As \Cref{fig:identity-graph} depicts, the alignment gain, perception cost and coverage loss of \cref{sec:tradeoff,sec:diversity} are each a combination of $F$ and its derivatives in $s$.
Thus, $D_{1+s}$ is the bridge that lets us evaluate $F$ through measurable disagreements of the two denoisers ($G$).
In practice, we use Langevin dynamics to sample the tilt at each $\gamma$, and sum the mean $\|\hat x_c-\hat x\|^2$ over $\gamma$ to get $G$.
We integrate up to a finite horizon $\Gamma$, where we believe the noisy laws already represent the clean laws that we wish to measure (\cref{alg:langevin,sec:supp:horizon} in the supplementary material).

\section{Perception--Alignment Tradeoff}
\label{sec:tradeoff}

Blau and Michaeli~\cite{blau2018perception} showed a perception--distortion tradeoff for image restoration, similar to rate--distortion theory~\cite{shannon1959coding,cover2006elements} for compression.
In their tradeoff, more accurate restoration pushes the output distribution away from natural images.
We show that guided sampling from a diffusion model has a tradeoff of the same type.
Stronger alignment with the condition always incurs a minimum cost in how natural the generated distribution stays.
The two denoisers of the model alone set the frontier of this tradeoff, so it bounds every way of sampling from the model.
The frontier is convex, and the guidance scale is the price along it (\cref{fig:teaser}).

\subsection{Theory}
\label{sec:tradeoff:theory}

We start with the perception--alignment definitions, identify the tilt as the ideal tradeoff distribution through the Gibbs variational bound, and state the shape of the ideal frontier.
We then move to path space to account for the sampler dynamics, which differ from the tilted distribution.
Finally, we prove the sampler's points lie on or below the ideal frontier.

\noindent \textbf{Alignment and cost.}
Fix a noise level $\gamma$ and a condition $c$.
For each candidate sampling law $q$ we ask two questions:
How much more do its observations agree with the condition?
How far did the law move from usual conditional sampling?

\begin{restatable}[Alignment gain, perception cost and frontier]{definition}{defGainCost}
\label{def:gain-cost}
For a law $q$ of observations at noise level $\gamma$, the \emph{alignment gain} and the \emph{perception cost} are
\begin{equation}
    \boxed{
    A[q]=\mathbb E_qT_\gamma-\mathbb E_{p^c_\gamma}T_\gamma,
    \qquad C[q]=\mathrm{KL}(q\|p^c_\gamma).}
    \label{eq:candidate-alignment-cost}
\end{equation}
The \emph{ideal frontier} is the largest gain for a cost budget $B$:
\begin{equation}
    A^\ast(B)=\sup_{q:\,C[q]\le B} A[q].
    \label{eq:alignment-budget}
\end{equation}
Square brackets mark a functional of a law; round brackets take a number, a scale $s$ or a budget $B$.
\end{restatable}

\noindent Up to a constant, $T_\gamma$ is the log-probability of the implicit classifier (\cref{eq:cfg-score}).
Thus, $A[q]$ is the increase in the mean log-probability that this classifier gives to $c$.
The KL in $C$ measures how far the generated image law departs from the data's conditional distribution.
Blau and Michaeli~\cite{blau2018perception} also measure perceptual quality as a divergence between similar distributions.

\runin{Relation to CFG}
The gain uses the implicit classifier whose gradient CFG adds (\cref{eq:cfg-score}), and each gain above usual conditional sampling has a price. With that said, we have yet to state the law that maximizes this tradeoff.
In the following paragraph, we identify the \textit{tilt} (\cref{eq:cfg-tilt}) as the maximizer.

\noindent\textbf{The ideal law.}
We write $A(s):=A[p_{\gamma,s}]$ and $C(s):=C[p_{\gamma,s}]$ for the gain and the cost of the tilt at scale $s$. \Cref{thm:ideal-tilt} shows that, at each cost, no law has a larger gain \textbf{than the tilt}, and that the tilt is the only law on the frontier at its cost.

\begin{restatable}[The CFG tilt is the ideal tradeoff]{theorem}{thmIdealTilt}
\label{thm:ideal-tilt}
Fix $\gamma$ and $s>0$, and let \cref{asm:finite} hold.
Let $q$ be a law with $C[q]<\infty$ and $\mathbb E_q|T_\gamma|<\infty$.
Then the Gibbs variational bound~\cite{csiszar1975divergence} holds:
\vspace{-2.5mm}
\begin{equation}
    \mathrm{KL}(q\|p^c_\gamma)
    \ge s\,\mathbb E_qT_\gamma-F(\gamma,s).
    \label{eq:gibbs-bound}
\end{equation}
Equality holds if and only if $q=p_{\gamma,s}$.
Thus, for the budget $B=C(s)$, the tilt is the only maximizer in \cref{eq:alignment-budget}, and $A^\ast(C(s))=A(s)$.
Since the tilt is an exponential family, $\partial_sF=\mathbb E_{p_{\gamma,s}}T_\gamma$, and the gain and the cost of the tilt are
\begin{align}
    A(s)&=\mathbb E_{p_{\gamma,s}}T_\gamma-\mathbb E_{p^c_\gamma}T_\gamma
        =\partial_sF(\gamma,s)-\partial_sF(\gamma,0),\notag\\
    C(s)&=\mathbb E_{p_{\gamma,s}}\big[sT_\gamma-F(\gamma,s)\big]
        =s\,\partial_sF(\gamma,s)-F(\gamma,s).
    \label{eq:alignment-cost}
\end{align}
\end{restatable}

\begin{figure}[t]
    \centering
    \includegraphics[width=\linewidth]{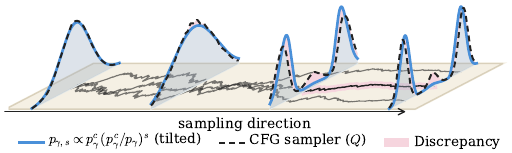}
    \vspace{-5mm}
    \caption{\textbf{Ideal tilts and sampler paths.} Illustrations over 1D GMM. $p^c_\gamma$, $p_\gamma$ and the tilt are known exactly at every noise level. Sampler trajectories (solid) and their marginals at single noise levels (dashed) versus the ideal tilts (blue), with their discrepancy shaded pink. The sampler keeps mass in a mode that the tilt removes (pink shading).}
    \label{fig:tilt-paths-concept}
\end{figure}

\runin{Relation to CFG}
The bound gives CFG a clear motivation: the law that best trades alignment for cost is the tilt \cref{eq:cfg-tilt}, exactly the law whose score CFG follows, and the guidance scale sets the budget.
Keep in mind that even though CFG follows the score of the tilt, it does not produce the law of the tilt at each noise level. This subtlety is depicted in \Cref{fig:tilt-paths-concept} and discussed later in this section.
In fact, no sampler, guidance rule or fine-tuned model can buy more gain for the same cost as the tilt. 

\noindent \textbf{The shape of the frontier.}
\Cref{thm:ideal-tilt} places the tilt on the frontier but says nothing about its shape.
Since the tilt is an exponential family in $s$, the shape follows from standard duality, and in guidance terms it reads as follows.

\Needspace{5\baselineskip}
\begin{restatable}[A convex frontier with the guidance scale as the price]{proposition}{propFrontierShape}
\label{prop:frontier-shape}
Let $\chi_\gamma(s)=\mathrm{Var}_{p_{\gamma,s}}(T_\gamma)=\partial_s^2F(\gamma,s)$.
Let \cref{asm:finite} hold on an interval of scales $s\ge0$ where $\chi_\gamma>0$.
Along the frontier $s\mapsto(A(s),C(s))$,
\vspace{-1.7mm}
\begin{equation}
    \boxed{
    \frac{dC}{dA}=s,
    \qquad 
    \frac{d^2C}{dA^2}=\frac1{\chi_\gamma(s)} > 0.}
    \label{eq:frontier-shape}
\end{equation}
\end{restatable}

\noindent The main point is that the frontier is convex: any gain beyond conditional sampling is paid for in distance from the conditional law, at an increasing rate.

This is why the guidance scale is a tradeoff and not a free improvement. A moderate $s$ can be a price worth paying, but convexity guarantees that the price is always there and that returns diminish as $s$ grows.

\noindent \textbf{The CFG sampler on path space.}
\Cref{thm:ideal-tilt} is about the exact tilt at one noise level.
Unfortunately, a practical CFG sampler carries its state across noise levels, while the tilt is defined fresh at each level, so the sampler does not visit it (\cref{fig:tilt-paths-concept}).
This requires redefining cost and alignment for \textit{path space}: the path law of practical samplers.
To describe the sampler exactly, we use the stochastic CFG dynamics, with the SNR as the time variable:
\begin{equation}
    dY_\gamma=\big(\hat x_c+\underbrace{s\,g}_{\text{CFG term}}\big)(Y_\gamma,\gamma)\,d\gamma+dB_\gamma,
    \quad Y_0=0.
    \label{eq:cfg-sde}
\end{equation}
The drift is the guided denoiser.
Let $Q_s$ be the law of the trajectories up to a horizon $\Gamma$, and $q_{\Gamma,s}$ the law of the end point $Y_\Gamma$, which is usually not the tilt $p_{\Gamma,s}$.

\begin{figure*}[t]
    \centering
    \includegraphics[width=\textwidth]{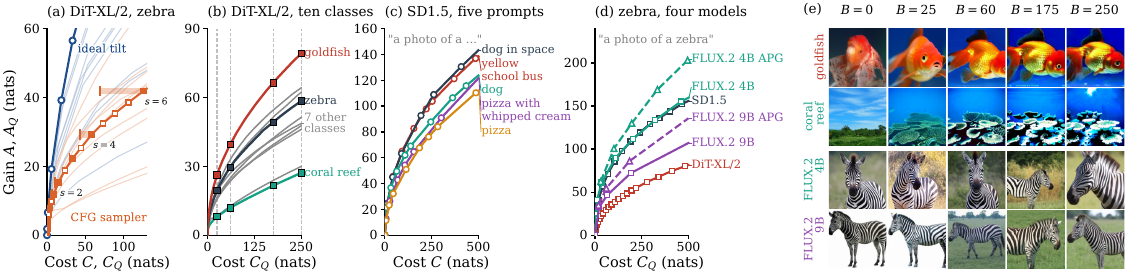}
    \vspace{-7mm}
    \caption{\textbf{Perception--alignment tradeoffs across conditions, models, and samplers.}
    (a)~Estimated ideal tilt (circles) and CFG path (squares) for DiT-XL/2 zebra; each bar is the range that holds the end-point cost of the sampler at that scale (\cref{eq:endpoint-cost-bound}). Faint curves: the same pair for the nine other classes of \cref{fig:supp-ac-dit} (supplementary).
    (b)~DiT-XL/2 CFG paths for ten ImageNet classes; dashed lines mark the shared cost budgets $B$ of (e).
    (c)~Ideal-tilt curves for five SD1.5 prompts ``a photo of a \ldots''.
    (d)~Zebra CFG paths across four models (``a photo of a zebra''; class label for DiT-XL/2), plus FLUX.2 4B and 9B with APG~\cite{sadat2025eliminating} (dashed triangles).
    (e)~Samples behind the curves. Top: DiT-XL/2 goldfish and coral reef at equal cost budget $B$, more classes in \cref{fig:supp-budget-images} (supplementary). Bottom: FLUX.2 Klein 4B and 9B samples for ``a photo of a zebra'' at the same guidance scale, from the CFG runs behind (d).}
    \label{fig:tradeoff-imagenet}
\end{figure*}

\begin{restatable}[Path-space gain and cost]{proposition}{propSampler}
\label{prop:sampler}
Let $\Lambda=\log(dQ_0/dQ_{-1})$ be the \emph{path alignment score}, the log-likelihood ratio between the conditional and unconditional samplers.
The \emph{path gain} and \emph{path cost} of the CFG sampler are a trajectory-based mirror of \cref{eq:candidate-alignment-cost}:
\begin{equation}
    A_Q(s)=\mathbb E_{Q_s}\Lambda-\mathbb E_{Q_0}\Lambda,
    \qquad
    C_Q(s)=\mathrm{KL}(Q_s\|Q_0).
    \label{eq:path-alignment-cost}
\end{equation}
Let $\hat x_c$ and $\hat x$ be two denoisers, exact or learned, and let \cref{asm:girsanov} hold.
Then, with $I_s=\int_0^\Gamma\mathbb E_{Q_s}\|g\|^2\,d\gamma$ the disagreement accumulated along the CFG trajectories,
\begin{equation}
    C_Q(s)=\frac{s^2}{2}I_s,\qquad
    A_Q(s)=\frac{1+2s}{2}I_s-\frac12I_0.
    \label{eq:sampler-point}
\end{equation}
\end{restatable}

\noindent The guided and conditional samplers differ only by the CFG term in the drift, so by Girsanov's theorem one number ($I_s$) gives both the gain and the cost.
In practice, CFG evaluates both denoisers ($\hat x_c, \hat x$) at each step, so calculating $I_s$ costs nothing beyond ordinary sampling.

\noindent \textbf{The sampler and the ideal frontier.}
The ideal frontier and the sampler are so far measured on different domains.
To place the sampler against the frontier, we must bring both to the same domain.

\begin{restatable}[The exact CFG sampler and its gap to the frontier]{theorem}{thmSamplerBelow}
\label{thm:sampler-below}
Let \cref{asm:exact,asm:finite,asm:girsanov} hold, and let $s>0$.
Let $s^\ast$ be the scale at which the ideal tilt reaches the same gain as the sampler, $A(s^\ast)=A_Q(s)$.
Then
\vspace{-1mm}
\begin{equation}
    \boxed{
    C_Q(s)=C(s^\ast)+\mathrm{KL}\big(q_{\Gamma,s}\big\|p_{\Gamma,s^\ast}\big)+S_s .}
    \label{eq:path-endpoint-bridge}
\end{equation}
The \emph{path slack} $S_s$ is, with $\kappa=\tfrac12s(1+s)$,
\begin{equation}
\begin{aligned}
    S_s&=\kappa\,\big(\tilde I_s-I_s\big)\ \ge0,\\
    \tilde I_s&=-\frac1\kappa\,\mathbb E_{Y_\Gamma\sim q_{\Gamma,s}}\log\mathbb E\big[e^{-\kappa\|g\|^2_{\tilde Y}}\,\big|\,Y_\Gamma\big],
\end{aligned}
    \label{eq:slack-formula}
\end{equation}
where $\|g\|^2_{\tilde Y}=\int_0^\Gamma\|g(\tilde Y_\gamma,\gamma)\|^2d\gamma$ is the disagreement along a Brownian bridge $\tilde Y$ from $0$ to the end point $Y_\Gamma$.
Hence the sampler pays at least the cost of the frontier:
\begin{equation}
    C(s^\ast)\ \le\ C[q_{\Gamma,s}]=C_Q(s)-S_s.
    \label{eq:endpoint-cost-bound}
\end{equation}
\end{restatable}

\noindent \Cref{eq:endpoint-cost-bound} places the cost of the generated images between the frontier and the $C_Q$ plotted in \Cref{fig:tradeoff-imagenet}.
The slack is exact, but its inner mean runs over all bridges to an end point.
With finitely many bridges the estimate is biased upward, so in practice we bound the slack from above, with the same two denoisers evaluated along paths re-noised from the sampler's end points.
We can thus check how much of the gap to the frontier it explains; in our experiments it is not all of it (\cref{sec:tradeoff:experiments}).
The supplementary material gives the proofs (\cref{sec:supp:tradeoff-theory}) and the estimators (\cref{sec:supp:tradeoff-estimators}).

\subsection{Experiments}
\label{sec:tradeoff:experiments}
We begin with ImageNet, as its simple nature serves as a controlled environment.
Although DiT-XL/2~\cite{peebles2023scalable} is no longer a state-of-the-art generator, this setting lets us examine the tradeoff before introducing the complexity of text conditioning.
We then turn to Stable Diffusion 1.5~\cite{rombach2022high} as an archetype of text-conditioned diffusion, and finally to the modern FLUX.2 family~\cite{bfl2025flux2} and Adaptive Projected Guidance (APG)~\cite{sadat2025eliminating} to examine how the picture changes with the model and sampling method.
The supplementary material holds the experimental procedures and the estimator limitations (\cref{sec:supp:details}), additional results on SD3.5~\cite{esser2024scaling} (\cref{sec:supp:tradeoff}), and the figures with prefix ``S'' (S1, S2, \dots{}) cited below.
In particular, all sampler points are measured on trajectories of the stochastic sampler of \cref{eq:cfg-sde} (\cref{sec:supp:details-paths} of the supplementary material).
Deterministic samplers are still bounded by \cref{thm:ideal-tilt} at their end point, but the path estimator of \cref{prop:sampler} does not apply to them.

\noindent\textbf{The sampler gap in practice.}
\Cref{fig:tradeoff-imagenet}(a) shows the gap between the CFG sampler and the ideal tilt for the zebra class of DiT-XL/2.
Each bar is the range that \cref{thm:sampler-below} leaves for the cost of the generated images once the path slack is accounted for, and at every scale the whole bar is to the right of the ideal curve: the sampler falls short of the frontier, and slack does not explain the gap.
The faint curves repeat the comparison for nine more ImageNet classes, and the supplementary material gives the same analysis with slack for more classes and prompts (\cref{fig:supp-ac-dit}).

\noindent \textbf{Condition difficulty.}
\Cref{fig:tradeoff-imagenet}(b) compares sampler curves between different ImageNet classes.
The ``flatness'' of the curve indicates how difficult it is to strengthen a condition: fast-rising curves (e.g. goldfish) reflect high alignment at low cost; slow-rising curves (e.g. coral reef) reflect difficulty in the model's ability to increase alignment no matter how much budget is invested.
This phenomenon is depicted visually in \cref{fig:tradeoff-imagenet}(e) (top).
The goldfish gets its features at low $B$, while the coral reef has to reach high distortion levels (cost) to increase alignment by growing more corals. More visual examples are depicted in~\cref{fig:supp-budget-images} of the supplementary material.
The SD1.5 frontiers in \cref{fig:tradeoff-imagenet}(c) extend this comparison to text prompts. A different phenomenon is observed: the unusual prompts (e.g. ``a dog in space'') are easier to strengthen compared to their more usual version (e.g. ``a dog''). A possible explanation could be that a less usual scene offers more room for improvement than a familiar one.
This behaviour seems to be consistent within models but not across different ones, as shown in~\cref{fig:supp-zebra-chain} of the supplementary material.

\noindent \textbf{Model and sampler comparison.}
\Cref{fig:tradeoff-imagenet}(d) compares the zebra curves across models and samplers.
APG lifts both FLUX.2 curves above ordinary CFG, as expected from a better sampler, while remaining close to ordinary CFG on the simpler models (\cref{fig:supp-ac-dit} of the supplementary material).

One surprising outcome is that the FLUX.2 4B curve lies above the 9B curve.
\Cref{fig:tradeoff-imagenet}(e) (bottom) gives a fuller picture: the 4B images show more realistic zebras in natural surroundings, resembling the photographs we typically associate with the prompt, compared to 9B images that place zebras against a plain white background, making them look like isolated cutouts rather than photographs of zebras in nature.
We hypothesize that this visual advantage is what the smaller model's higher alignment curve reflects.

\section{A Diversity Index for CFG}
\label{sec:diversity}
The perception--alignment frontier measures what guidance gains and what it costs, relative to usual conditional sampling.
A second question remains: how much of the unconditional distribution does the guided law still cover?
The answer has two parts: the coverage that a condition has before guidance, and the coverage that guidance removes.
We show that the same log-normalizer gives both, and that the coverage removed by guidance is exactly the alignment gain plus the perception cost.

\subsection{Theory}
\label{sec:diversity:theory}

\begin{restatable}[Coverage index]{definition}{defCoverage}
\label{def:coverage}
The \emph{coverage loss} of the tilt is its KL divergence from the unconditional law.
The \emph{coverage index} $N(s)$ is the exponential of $-V(s)$:
\begin{equation}
    \boxed{
    V(s)=\mathrm{KL}(p_{\gamma,s}\|p_\gamma),
    \qquad N(s)=e^{-V(s)}.}
    \label{eq:guided-coverage}
\end{equation}
\end{restatable}

\intuition
The KL divergence is a cross-entropy minus an entropy:
\begin{equation}
    V(s)=\underbrace{-H(p_{\gamma,s})}_{\text{narrowing}}
        +\underbrace{\mathrm{CE}(p_{\gamma,s}\|p_\gamma)}_{\text{moving}}.
    \label{eq:coverage-entropy}
\end{equation}
Entropy $H$ alone counts any spread, including outcomes that the unconditional law rarely produces.
The cross-entropy term corrects this, so $V$ measures only the variety that stays natural under $p_\gamma$.
In particular, a law that keeps a subset of $p_\gamma$ with its relative probabilities has a constant density ratio to it, so $N$ is the probability of that subset.

\Cref{eq:coverage-entropy} holds at every scale, and we read it in two parts.
The \emph{baseline} $V(0)=\mathrm{KL}(p^c_\gamma\|p_\gamma)$ is the coverage that the condition itself gives up, before any guidance.
The rest, $V(s)-V(0)$, is the coverage that guidance removes on top of it.

\noindent\textbf{Baseline and loss due to guidance.}
With exact denoisers (\cref{asm:exact,asm:finite,asm:regular,asm:zero-snr}), the mismatched I-MMSE~\cite{verdu2010mismatched} gives the baseline at a horizon $\gamma=\Gamma$ as:
\vspace{-2mm}

\begin{equation}
    \boxed{
    V(0)=\frac12\int_0^\Gamma
        \mathbb E_{p^c_\gamma}\|\hat x_c-\hat x\|^2\,d\gamma,}
    \label{eq:coverage-immse}
\end{equation}
and the coverage lost to guidance, $\Delta V$, as:
\vspace{-2mm}
\begin{equation}
    \boxed{\Delta V(s):=V(s)-V(0)=A(s)+C(s).}
    \label{eq:coverage-alignment-cost}
\end{equation}
For the CFG sampler, with $V_Q(s)=\mathrm{KL}(Q_s\|Q_{-1})$ its path divergence from the unguided sampler, the same identity is developed in a similar way (\cref{asm:girsanov}):
\begin{equation}
    \Delta V_Q(s):=V_Q(s)-V_Q(0)=A_Q(s)+C_Q(s).
    \label{eq:sampler-coverage}
\end{equation}

\noindent\textbf{Two complementary readings.}
The baseline compares conditions: a broad condition covers much of the unconditional law and has a small $V(0)$, and a specific one has a large $V(0)$.
In contrast, $\Delta V$ compares scales for one condition, and it is the gain plus the cost of the frontier.
Together, $N(s)=N(0)\,e^{-\Delta V(s)}$: the coverage that the condition starts with, times the fraction that guidance keeps (with $\Delta V_Q$ for the sampler).
Two conditions with the same gain and cost can thus end with different coverage.
The derivations are in \cref{sec:supp:diversity-theory} of the supplementary material.

\subsection{Experiments}
\label{sec:diversity:experiments}

We test the two parts of the index: the baseline $V(0)$ and the loss due to guidance $\Delta V$.
We begin with ImageNet, where both agree with established diversity measures.
We then put more emphasis on state-of-the-art text-conditioned models in common settings, e.g.\ guidance scale range and prompt complexity, showing that our measure is not only interesting in theory, but also applies to real settings that practitioners encounter.
On FLUX.2, $\Delta V_Q$ follows the visible loss of sample diversity as the scale grows.
On Stable Diffusion 1.5, the baseline grows as prompts get more specific, and it tracks the share of real images that match each prompt.
Complete experimental procedures, estimator settings and further experiments, including SD3.5, are in the supplementary material (\cref{sec:supp:diversity,sec:supp:details}), where we also show a quantitative comparison to existing external references~\cite{friedman2023vendi}, grounding our measure in well-established metrics.

\noindent\textbf{ImageNet.}
Across 100 ImageNet classes, the index correlates with the Vendi score~\cite{friedman2023vendi}: Spearman $0.60$--$0.63$ for the baseline $V(0)$, and for the loss $\Delta V(s)$, $0.75$--$0.79$ for the ideal tilt and $0.46$--$0.53$ for the sampler.
Classes with greater coverage are more diverse, and classes that lose more coverage also lose more diversity.
Complete quantitative and qualitative results are in \cref{sec:supp:diversity-imagenet} of the supplementary material.

\begin{figure*}[t]
    \centering
    \includegraphics[width=\linewidth]{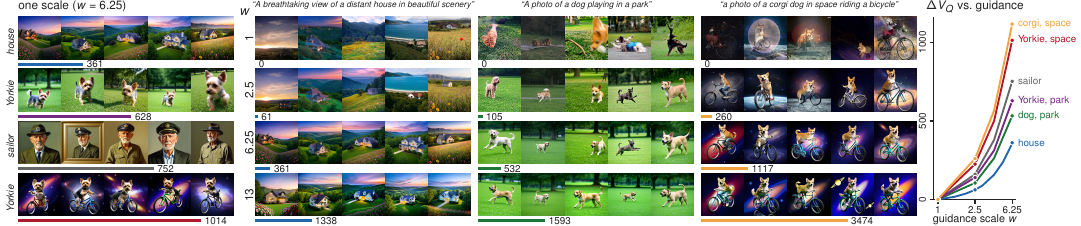}\
    \vspace{-6.5mm}
    \caption{\textbf{Guidance effect on sample diversity in FLUX.2.}
    Samples of FLUX.2 Klein-base 4B with the CFG sampler, at the usual guidance scale $w=1+s$.
    The bars and curves show the coverage the sampler loses relative to unguided conditional sampling, $\Delta V_Q$ (\cref{eq:sampler-coverage}).
    \emph{Left:} four prompts at $w=6.25$, sorted by $\Delta V_Q$.
    \emph{Middle:} three prompts at four scales; within a prompt, each column follows the same sampling path in every row, and the bars share one scale.
    \emph{Right:} $\Delta V_Q$ against $w$ for all six prompts; dots mark the middle panel's scales.
    Complete experimental settings are in \cref{sec:supp:details} of the supplementary material.}
    \label{fig:diversity-flux}
\end{figure*}

\noindent\textbf{Guidance effect on diversity.}
Practitioners raise $s$ for a given prompt in search of better condition adherence. 
Our theory shows that this comes with direct loss of diversity.
\Cref{fig:diversity-flux} shows this effect on FLUX.2 4B, with the usual guidance scale $w=1+s$.
Without extrapolated guidance ($w=1$), the samples of each prompt vary widely in layout, light, pose and style.
As $w$ grows, they become alike, but not at the same speed.
The houses keep different buildings and framing up to $w=13$.
By $w=6.25$, the corgis in space share one pose on a bicycle in front of a galaxy, and only the bicycle and the light still change.

In addition, in the left part of \Cref{fig:diversity-flux} we see a comparison of different conditions at the same guidance scale ($w=6.25$, a recommended scale for this model).
The relative value of $\Delta V_Q$ orders the generated images by their apparent diversity, as expected from a diversity measure, supporting it as an intrinsic and theoretically grounded value.

\begin{figure}[t]
    \centering
    \includegraphics[width=\linewidth]{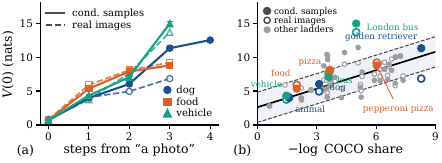}
    \vspace{-7mm}
    \caption{\textbf{Coverage along prompt ladders and COCO image shares (SD1.5).}
    (a) $V(0)$ for conditional samples (solid, standard-error bars) and matched real images (dashed).
    (b) $V(0)$ versus negative log matching-image share: conditional samples (filled) and real images (hollow).
    The ladders of (a) are in colour and the prompts of ten more ladders in grey.
    The line is a robust (RANSAC) fit to the conditional samples, and the band is its inlier threshold.}
    \label{fig:diversity-prompts}
\end{figure}

\noindent\textbf{Prompt specificity: a condition diversity baseline.}
As $V(0)$ measures the condition's coverage baseline, we expect to see an increase in $V(0)$ as prompts get more specific.
For example, moving from ``an animal'' to ``a dog'' and then to ``a golden retriever'' restricts the images that can satisfy the prompt, thus reducing the baseline diversity.
In this experiment, we inspect prompt ladders that progressively add detail to a description, testing whether coverage captures gradual restrictions on content.

In \cref{fig:diversity-prompts}(a), each added detail increases baseline coverage loss, both on conditional samples and on matching real photographs.
Conditional samples are generated by SD1.5 and real photographs are queried results from COCO~\cite{lin2014microsoft} that textually match the prompt.
For more experiments and grounding to external measures see \cref{sec:supp:diversity-chains} in the supplementary material.
This experiment supports our theoretical predictions on a crafted scenario where prompts get semantically more specific, suggesting a way to compare baseline diversity for given conditions.

\noindent\textbf{Grounding $V(0)$ in real image shares.}
To check that $V(0)$ is correct, we ground it in a distribution of images that we understand.
We assume a universal distribution of natural images, in which each prompt keeps a share that shrinks as the prompt gets more specific, e.g.\ from ``an animal'' to ``a dog'' to ``a golden retriever''.
The model is trained on a large and diverse image set, so its unconditional law stands for this distribution; a prompt that selects its matching images keeps a fraction $N(0)$ of this distribution, so we compare $V(0)=-\log N(0)$ with the negative logarithm of its share.
COCO is also a set of everyday photographs, which we treat as a sample of the same distribution, with no bias assumed towards our prompts.
Its annotations give the share of images matching each prompt by counting, with no generated images and no image embedder.

\Cref{fig:diversity-prompts}(b) compares the two: a quantity computed from two denoisers on noisy latents, with no access to COCO, tracks the share that each prompt keeps along the subset-selection diagonal.
COCO and the training set are different samples of the universal distribution, so the two differ by a common offset, and the comparison is of the relative loss between prompts.
This grounds coverage as relative diversity in a counted fraction of real images, not a feature-space proxy.
The supplementary material derives the share (\cref{lem:supp:subset}) and gives the COCO comparison in detail (\cref{sec:supp:diversity-coco}).

\section{Discussion and Conclusion}
\label{sec:conclusion}

We have developed a formal framework for understanding the balance between prompt alignment and the distribution of images produced by diffusion models. 
Grounded in information theory, our formulation shows that stronger alignment comes at an unavoidable distributional cost, with a computable frontier that we prove the CFG sampler cannot surpass. %
This gives a precise account of the intuitive and familiar observation: pushing generation toward closer adherence to a prompt changes the quality and range of the images it produces. 
Moreover, the conditional and unconditional CFG denoisers alone provide this information, making our results an intrinsic property of the model.

The practical value of this framework is to make the consequences of guidance strength more interpretable. 
Increasing guidance always comes at both a distributional and diversity cost, and improving the sampling algorithms has a proved computable ceiling.
With that said, each condition has different cost responses that often align with its semantic meaning.
Our framework allows these responses to be measured and compared, giving the guidance scale meaning beyond a parameter chosen through trial and error. 
In addition, the relative diversity index completes the picture, measuring what a condition selects from the model's image distribution, as well as its decrease by the guidance scale.
Together, these tools help clarify what stronger guidance gains, and more importantly what it costs. 
Our theory characterizes the tradeoff rather than removing it, providing a basis for further study of guidance and its practical use.

\section*{Acknowledgments}
This work was supported by the Center for AI and Data Science at Tel Aviv University (TAD), and in part by the Israel Science Foundation (grants No.~3441/21 and 1473/24) and by Len Blavatnik and the Blavatnik Family Foundation.

{
    \small
    \bibliographystyle{IEEEtranN}
    \bibliography{main}
}

\renewcommand{\inSupp}{}
\clearpage
\twocolumn[\centering\Large\textbf{Supplementary Material}\vspace{1.5em}]

\setcounter{section}{0}
\renewcommand{\thesection}{\Alph{section}}
\setcounter{figure}{0}
\counterwithin{equation}{section}
\renewcommand{\thefigure}{S\arabic{figure}}
\renewcommand{\theassumption}{S\arabic{assumption}}
\renewcommand{\thealgorithm}{S\arabic{algorithm}}

\section*{Index Summary}
\newcommand{\supptocline}[4]{\par\noindent\hspace*{#1}\hyperref[#3]{\makebox[#2][l]{\ref*{#3}}#4}\,\dotfill\,\pageref*{#3}\par}
\newcommand{\supptocsec}[3]{\par\addvspace{4pt}\noindent\hyperref[#1]{\textbf{\makebox[1.4em][l]{\ref*{#1}}#2}}\,\dotfill\,\textbf{\pageref*{#1}}\par
    {\leftskip=1.4em\noindent#3\par}\vspace{1pt}}
\newcommand{\supptocdesc}[2]{{\leftskip=#1\footnotesize\noindent#2\par}}
\newcommand{\supptocsub}[3]{\supptocline{1.4em}{2.3em}{#1}{#2}\supptocdesc{1.4em}{#3}}
\newcommand{\supptocsubsub}[3]{\supptocline{2.6em}{2.9em}{#1}{#2}\supptocdesc{2.6em}{#3}}
\begin{small}
\supptocsec{sec:supp:experiments}{Additional Experiments}{Experiments beyond the main text, with images next to every quantitative comparison, and outside checks of alignment (\cref{fig:supp-classifier}) and cost (\cref{fig:supp-cost-kid}).}
\supptocsub{sec:supp:tradeoff}{Perception--Alignment Tradeoff}{The frontier gap across more conditions, models and guidance methods, and how the best sampler changes with the condition.}
\supptocsubsub{sec:supp:tradeoff-more}{More Conditions, Gaps and Samplers}{Frontiers and sampler curves for ten ImageNet classes and for text prompts, other guidance rules, and images at equal cost.}
\supptocsubsub{sec:supp:slack-bound}{How Much of the Gap Is Path Slack}{The slack bound for ten classes and for two FLUX.2 prompts.}
\supptocsubsub{sec:supp:tradeoff-flatness}{Prompt Specificity and the CFG Tradeoff}{Sampler curves along a chain of nested prompts, on three models.}
\supptocsubsub{sec:supp:tradeoff-classification}{The Alignment Score as a Classifier}{Outside check: the score classifies held-out real images by label.}
\supptocsubsub{sec:supp:tradeoff-cost-kid}{The Cost Seen by an Outside Network}{Outside check: within each condition, the cost and the DINOv2 KID to the unguided samples order the guidance scales alike.}
\supptocsub{sec:supp:diversity}{Diversity Index}{How coverage relates to visible sample variety across conditions and guidance scales.}
\supptocsubsub{sec:supp:diversity-models}{The Baseline Is Shared Across Models}{Two models with different architectures give nearly the same baseline for 100 classes.}
\supptocsubsub{sec:supp:diversity-chains}{More Prompt Ladders}{Thirteen ladders of nested prompts, compared with the Vendi score.}
\supptocsubsub{sec:supp:diversity-coco}{COCO Shares in Detail}{The retained share against the share of matching COCO images.}
\supptocsubsub{sec:supp:diversity-imagenet}{ImageNet: Coverage and Vendi Diversity}{The index against Vendi across 100 classes and along the guidance scale.}
\supptocsubsub{sec:supp:diversity-visual}{What Diversity Loss Looks Like}{Samples for FLUX.2 and SD3.5 prompts, ordered by the index.}
\supptocsec{sec:supp:horizon}{The Noise Horizon}{Why every quantity is reported up to a finite noise horizon, $\sigma\approx3$, rather than the clean limit: the choice follows the information-estimation metric~\cite{ohayon2026distance}, which at a comparable horizon agrees with human judgments on three benchmarks. We give the horizon of each model in standard deviations of its latents and measure how our conclusions depend on it (\cref{fig:supp:horizon-sweep}).}
\supptocsec{sec:supp:theory}{Derivations}{Each main-text result restated and proved, in three parts that follow Sections 3 to 5; each proof is one derivation, and known theorems are stated and cited.}
\supptocsub{sec:supp:common}{Preliminaries}{Setting, CFG, assumptions and the known I-MMSE identities; proofs of the R\'enyi I-MMSE theorem and the log-normalizer identity; the two estimators (\cref{alg:langevin,alg:sampler}) and the theory behind them.}
\supptocsub{sec:supp:tradeoff-theory}{Perception--Alignment Tradeoff}{Why the tilt is the ideal law, the shape of the frontier, the sampler on path space, and the split of its gap to the frontier into a shortfall of the end-point law and a path slack, with a bound on the slack.}
\supptocsub{sec:supp:diversity-theory}{A Diversity Index for CFG}{The meaning of the coverage index, its baseline, and the proof that the coverage lost to guidance is alignment gain plus perception cost, for the tilt and for the sampler.}
\supptocsec{sec:supp:toy}{Numerical Check on Gaussian Mixtures}{Gaussian mixtures, where everything has a closed form, as a ground truth for the identities and the estimators.}
\supptocsub{sec:supp:toy-closed}{Closed Form}{The identities checked as equalities of numbers by quadrature.}
\supptocsub{sec:supp:toy-exact}{Exact Denoisers}{The paper's estimators, run with exact denoisers, recover the closed form.}
\supptocsub{sec:supp:toy-trained}{Trained Denoisers}{The same pipeline on learned denoisers, separating numerical error from head error.}
\supptocsec{sec:supp:details}{Experimental Details}{The complete settings of all experiments, organized by setting, so each setting is stated once.}
\supptocsub{sec:supp:details-models}{Models and Latent Spaces}{The five models, their latent spaces and noise grids.}
\supptocsub{sec:supp:details-conditions}{Conditions and Image Pools}{The ImageNet classes, the text prompts and the pools of real images.}
\supptocsub{sec:supp:details-horizon}{Noise Grids and Horizon}{The grid of each model and the noise level at which quantities are reported.}
\supptocsub{sec:supp:details-paths}{CFG Trajectory Runs}{Trajectories and guidance scales of the sampler runs, the slack bound and the outside distance.}
\supptocsub{sec:supp:details-guidance}{Other Guidance Rules}{APG, CFG++ and Annealing Guidance in our sampler.}
\supptocsub{sec:supp:details-langevin}{Ideal-Frontier Runs}{The Langevin chains and the reconstruction of the frontier.}
\supptocsub{sec:supp:details-pools}{Estimates on Real Images}{The estimates on fixed pools of images and the reference classifiers.}
\supptocsub{sec:supp:details-vendi}{Diversity Measures and Statistics}{The Vendi score, the settings of the ImageNet study and the fit of the Vendi gauge.}
\supptocsub{sec:supp:details-images}{Displayed Images}{How the shown images are chosen.}
\supptocsub{sec:supp:details-accuracy}{Numerical Accuracy and Limitations}{Step-size and mixing checks.}
\supptocsub{sec:supp:details-compute}{Compute}{Network evaluations and GPU-hours of the core computations.}

\addvspace{6pt}
\noindent\begin{minipage}{\columnwidth}\small
\textbf{Conventions.}
Figures, assumptions and algorithms of this supplement are numbered S1, S2, \dots, tables are numbered from 1, and equations carry the letter of their section, as in (C.1).
Results restated from the main text keep their main-text numbers.
Throughout, $s=0$ is the conditional baseline, so the usual CFG weight is $1+s$.
Gain, cost and coverage are in nats, are computed in the latent space of each model, and are reported up to the noise horizon of \cref{sec:supp:horizon}.
\end{minipage}
\end{small}
\section{Additional Experiments}
\label{sec:supp:experiments}

\subsection{Perception--Alignment Tradeoff}
\label{sec:supp:tradeoff}

This section extends the tradeoff experiments of \cref{sec:tradeoff:experiments}.
\Cref{sec:supp:tradeoff-more} compares samplers with the estimated ideal frontier for more conditions, models and guidance rules, \cref{sec:supp:slack-bound} splits the distance to the frontier into a shortfall of the end-point law and a path slack, and \cref{sec:supp:tradeoff-flatness} follows the tradeoff as a prompt becomes more specific.

The last two experiments check the two quantities against outside references.
The alignment score $T_\gamma$ classifies held-out real images by their true labels, with accuracy up to $0.91$ (\cref{sec:supp:tradeoff-classification}).
Within each condition, the path cost $C_Q$ orders the guidance scales as the DINOv2 KID between guided and unguided samples does (\cref{sec:supp:tradeoff-cost-kid}).

\subsubsection{More Conditions, Gaps and Samplers}
\label{sec:supp:tradeoff-more}

We extend the tradeoff comparison across conditions, models, and guidance methods, examining the gap to the estimated ideal frontier and how the highest-gain sampler varies with the condition.
Paired images illustrate how these measured differences appear in the generated content.

\begin{figure}[!h]
    \centering
    {\footnotesize \textbf{DiT-XL/2}: ImageNet class \textit{clog}}\\[1pt]
    \includegraphics[width=\linewidth]{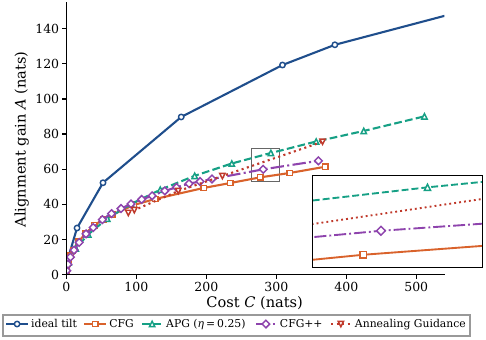}
    \caption{\textbf{DiT-XL/2: guidance methods on one class.}
    CFG, APG~\cite{sadat2025eliminating}, CFG++~\cite{chung2025cfgpp} and Annealing Guidance~\cite{yehezkel2025navigating}, all conditioned on the ImageNet class \textit{clog}.
    Inset: zoom on $C=265$--$305$.}
    \label{fig:supp-ac-methods}
\end{figure}

In \cref{fig:supp-ac-methods}, the four guidance methods achieve different alignment gains at matched cost for the ImageNet clog class using DiT-XL/2, while all remain below the estimated ideal frontier.
In the highlighted cost range, APG achieves the largest gain, followed by Annealing Guidance, CFG++, and CFG.
The separation shows that changing the guidance method can improve alignment at the same cost, although the estimated ideal frontier still offers substantially greater gains.

\begin{figure*}[!t]
    \centering
    {\footnotesize \textbf{DiT-XL/2}: ImageNet class labels}\\[1pt]
    \includegraphics[width=\linewidth]{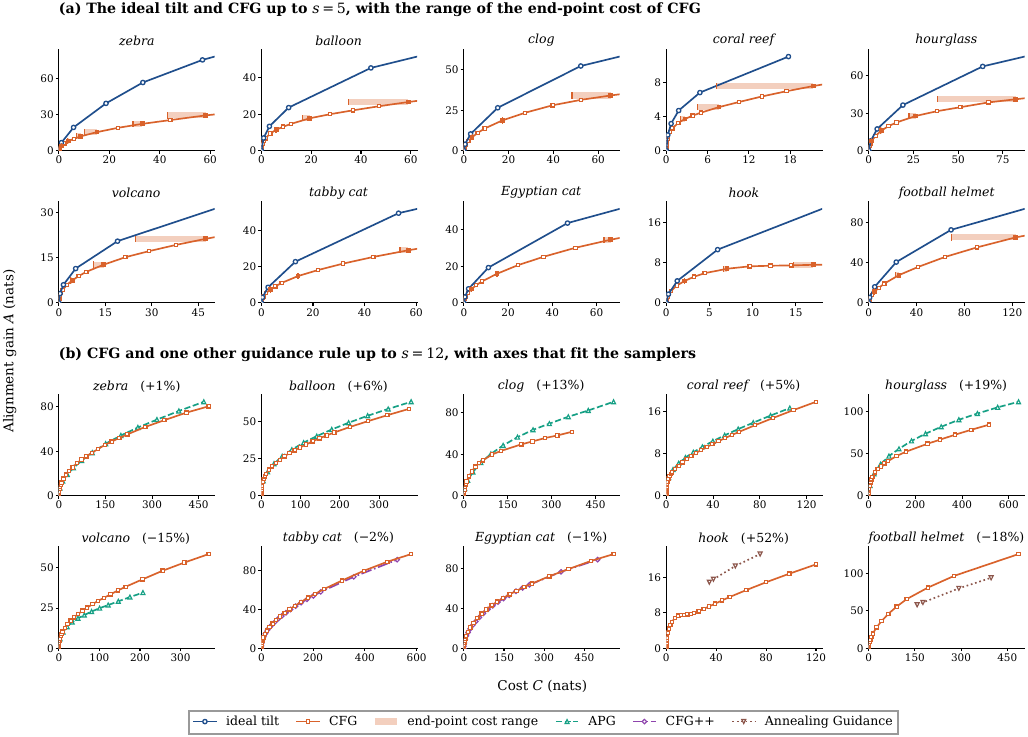}
    \caption{\textbf{DiT-XL/2: class-dependent sampler gains.}
    (a)~CFG and the estimated ideal frontier for ten classes, up to $s=5$.
    At each filled square, the bar is the range that holds the end-point cost $C[q_{\Gamma,s}]$ of the sampler (\cref{eq:endpoint-cost-bound}): from the larger of the frontier cost at the same gain and $C_Q$ minus the slack bound of \cref{eq:supp:slack-bound}, up to the path cost $C_Q$.
    Bars at $s\in\{1,2,4\}$; zebra also at $s\in\{0.25,0.5,1.5,3\}$, hourglass also at $s=0.5$.
    (b)~The same CFG curves up to $s=12$, next to one other guidance rule and without the ideal frontier, so that the axes fit the samplers: APG~\cite{sadat2025eliminating} (first six panels), CFG++~\cite{chung2025cfgpp} (cats), or Annealing Guidance~\cite{yehezkel2025navigating} (hook, helmet).
    Percentages: mean relative gain over CFG at matched cost, at the costs that CFG reaches for $s=6$ to $7$ ($6$ to $8$ for the football helmet, and over the whole Annealing Guidance curve for the hook); negative values favor CFG.}
    \label{fig:supp-ac-dit}
\end{figure*}

Across the ten classes in \cref{fig:supp-ac-dit}, the estimated ideal curve lies above the sampled curves over their common cost range, extending the main paper's comparison to additional conditions.
The flattening of the ideal gain as cost grows illustrates the increasing price of alignment in \cref{prop:frontier-shape}.
Panel~(b) also shows why sampler improvements should be measured per condition: APG gives larger gains on clog and hourglass than on balloon and coral reef, whereas volcano favors CFG.
Each gain compares APG and CFG on the same noise.
Five of the six differ from zero by more than five standard errors, so the differences between classes are not noise; only the zebra gain is too small to tell.
Over the six classes the gain is between $-15\%$ and $+19\%$, about $5\%$ on average, and it changes sign; on the three FLUX.2 conditions of \cref{fig:supp-flux-apg-images} it is not negative at any scale, and it is about $20\%$ to $30\%$ in the same range of scales.
In this sense APG stays close to CFG on DiT-XL/2.
The cat panels compare CFG++ with CFG: their curves remain close, with CFG++ slightly below CFG at the costs where the percentage is taken.

\Cref{fig:supp-ac-dit-images} shows this class dependence in the samples themselves: the sampler with the highest alignment gain in \cref{fig:supp-ac-dit} changes with the class, and the measured gain has a visible counterpart in the generated content.
Among the displayed settings, the red frames identify CFG for football helmet, CFG++ for coral reef, Annealing Guidance for hook, and APG for zebra.
CFG brings the helmet into close-up while the other methods show players on a field.
CFG++ produces a more colorful reef with prominent purple coral.
Annealing Guidance produces a recognizable metal hook while CFG returns a blank image, and APG depicts the zebra with less oversaturated surroundings than CFG and CFG++.
Each column uses the same noise across methods.
These selected examples connect the measured gains to visible class content and appearance, while the curves in \cref{fig:supp-ac-dit} provide the comparisons at matched cost.

\Cref{fig:supp-budget-images} extends the equal-budget samples of \cref{fig:tradeoff-imagenet}(e) to the zebra and four more classes of \cref{fig:supp-ac-dit}, chosen for curves far apart and ordered by the area under the CFG curve.
For the easiest classes the content changes early: by $B=25$ a player in a helmet and a clean, centered hourglass replace the cluttered scenes at $B=0$, and by $B=175$ the helmet fills the frame.
For the harder classes, balloon and volcano, the same budgets change the content more slowly: a single balloon emerges against a sunset, and the snowy mountain starts to erupt only at $B=110$.
At the largest budgets several rows show oversaturated colors, such as the glowing hourglass, the purple tint of the zebras, and the lava.

\newcommand{\ditpick}[4]{%
  {\setlength{\fboxsep}{0pt}\setlength{\fboxrule}{1pt}\fcolorbox{#4}{white}{\includegraphics[width=.20\linewidth]{dit_picks/#1}}}}
\newcommand{\ditrow}[1]{\raisebox{.085\linewidth}{\rotatebox{90}{\makebox[0pt][c]{\scriptsize #1}}}}
\begin{figure}[!t]
    \centering
    \setlength{\tabcolsep}{1pt}%
    \renewcommand{\arraystretch}{0}%
    \begin{tabular}{@{}c@{\hspace{2pt}}cccc@{}}
      & {\scriptsize\itshape football helmet}\rule[-2pt]{0pt}{9pt} & {\scriptsize\itshape coral reef} & {\scriptsize\itshape hook} & {\scriptsize\itshape zebra} \\
      \ditrow{CFG} &
        \ditpick{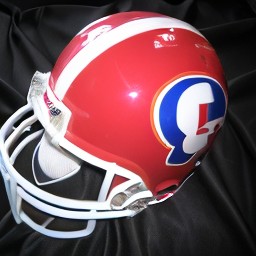}{$s{=}4$}{$C{=}123$, $A{=}65$}{red} &
        \ditpick{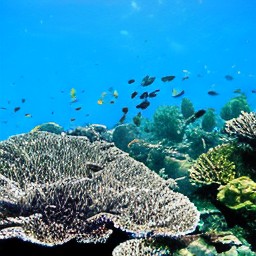}{$s{=}4$}{$C{=}21$, $A{=}7.5$}{white} &
        \ditpick{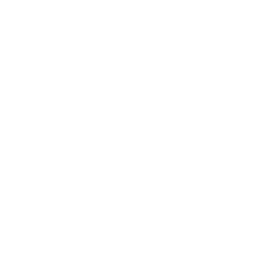}{$s{=}6$}{$C{=}30$, $A{=}8.8$}{white} &
        \ditpick{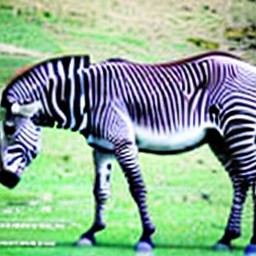}{$s{=}12$}{$C{=}482$, $A{=}80$}{white} \\[1pt]
      \ditrow{CFG++} &
        \ditpick{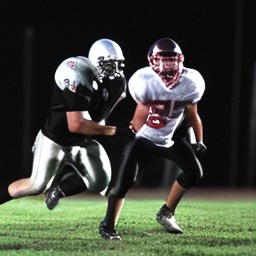}{$\lambda{=}0.5$}{$C{=}83$, $A{=}50$}{white} &
        \ditpick{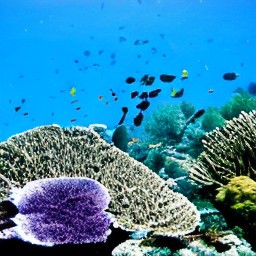}{$\lambda{=}0.7$}{$C{=}26$, $A{=}8.6$}{red} &
        \ditpick{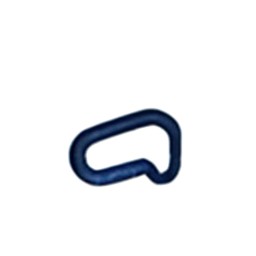}{$\lambda{=}0.85$}{$C{=}30$, $A{=}9.4$}{white} &
        \ditpick{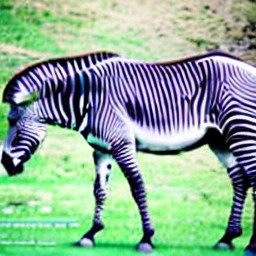}{$\lambda{=}1.4$}{$C{=}469$, $A{=}80$}{white} \\[1pt]
      \ditrow{AG} &
        \ditpick{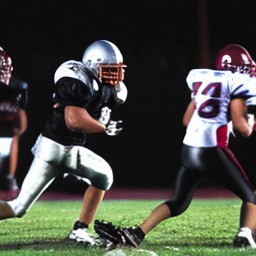}{$\lambda{=}0.05$}{$C{=}157$, $A{=}58$}{white} &
        \ditpick{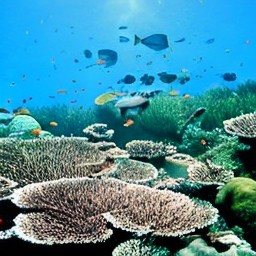}{$\lambda{=}0.7$}{$C{=}23$, $A{=}8.3$}{white} &
        \ditpick{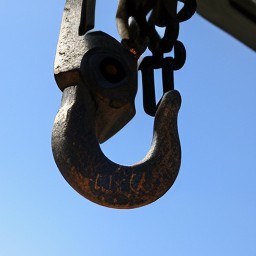}{$\lambda{=}0.05$}{$C{=}35$, $A{=}15$}{red} &
        \ditpick{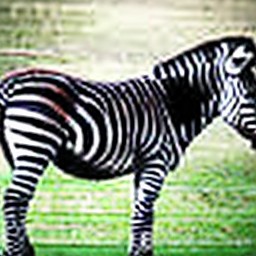}{$\lambda{=}1$}{$C{=}356$, $A{=}71$}{white} \\[1pt]
      \ditrow{APG} &
        \ditpick{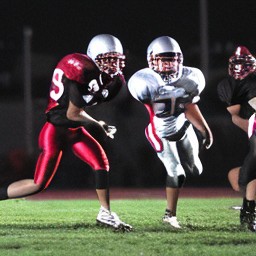}{$s{=}4$}{$C{=}85$, $A{=}48$}{white} &
        \ditpick{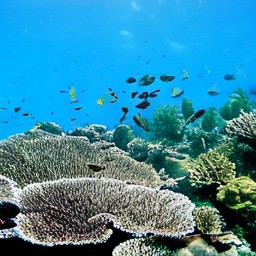}{$s{=}4$}{$C{=}16$, $A{=}7.2$}{white} &
        \ditpick{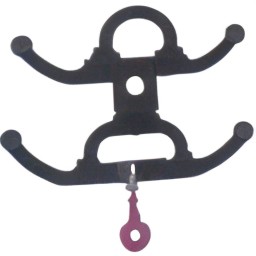}{$s{=}5$}{$C{=}30$, $A{=}13$}{white} &
        \ditpick{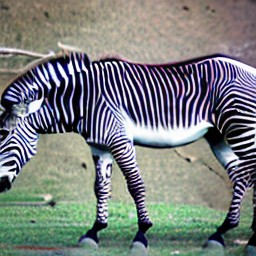}{$s{=}12$}{$C{=}467$, $A{=}84$}{red} \\
    \end{tabular}
    \caption{\textbf{The highest-gain sampler varies by class, with visible improvements.}
    Red frames mark the highest $A$ at the selected settings; each column uses the same noise.
    The blank CFG \textit{hook} is not a failed run: CFG saturates this class towards white ($98\%$ of the pixels at $s=6$).}
    \label{fig:supp-ac-dit-images}
\end{figure}

\begin{figure}[!b]
    \centering
    \includegraphics[width=\columnwidth]{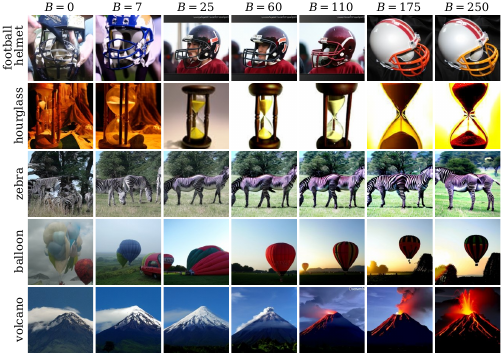}
    \caption{\textbf{DiT-XL/2 samples at equal cost budget for more classes.}
    As in \cref{fig:tradeoff-imagenet}(e) (top): each row is one class at the guidance scales where it reaches the budget $B$, with the same noise along the row.
    Rows are ordered by the area under the CFG curve in \cref{fig:supp-ac-dit} up to $B=250$, from the easiest class to strengthen (top) to the hardest.}
    \label{fig:supp-budget-images}
\end{figure}

\begin{figure*}[!t]
    \centering
    \includegraphics[width=\textwidth]{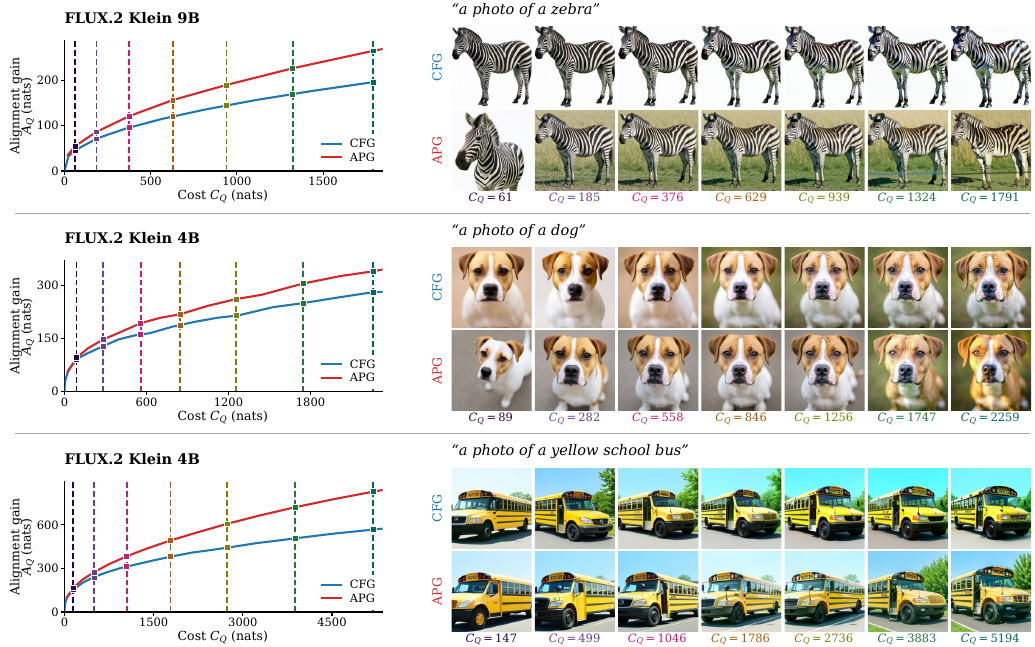}
    \caption{\textbf{FLUX.2 Klein: CFG and APG~\cite{sadat2025eliminating} at matched cost $C_Q$}.
    Top to bottom: zebra (9B), dog (4B), and yellow school bus (4B).
    Left: alignment gain $A_Q$ versus cost $C_Q$; colored vertical lines mark the seven pictured costs.
    Right: image grids with CFG above APG, one matched-cost pair per column and increasing cost from left to right.
    Each case uses one seed across all images.}
    \label{fig:supp-flux-apg-images}
\end{figure*}

Since APG generally achieves the highest gain at matched cost in the highlighted range, we next examine its tradeoff across conditions and models.
The left-hand plots in \cref{fig:supp-flux-apg-images} show that APG achieves greater measured alignment than CFG at matched cost for all three displayed FLUX conditions.
The corresponding image grids on the right pair CFG above APG at seven increasing costs, marked by the vertical lines in each plot.
Across these grids, CFG produces increasingly saturated colors at higher costs, while APG generally reduces this oversaturation at matched costs.
The difference is particularly visible in the school-bus images in the bottom case, linking APG's higher alignment curve to a visible improvement in the generated images.
The zebra images in the top case also show a change in scene composition: APG places the animal in a grassy photographic setting, while CFG retains the isolated animal against a white background.
This is consistent with the main paper's interpretation that alignment concerns the full scene associated with a prompt, including its setting.
The paired examples illustrate how changing the sampler can improve the use of a cost budget.

\begin{figure*}[t]
    \centering
    \includegraphics[width=\textwidth]{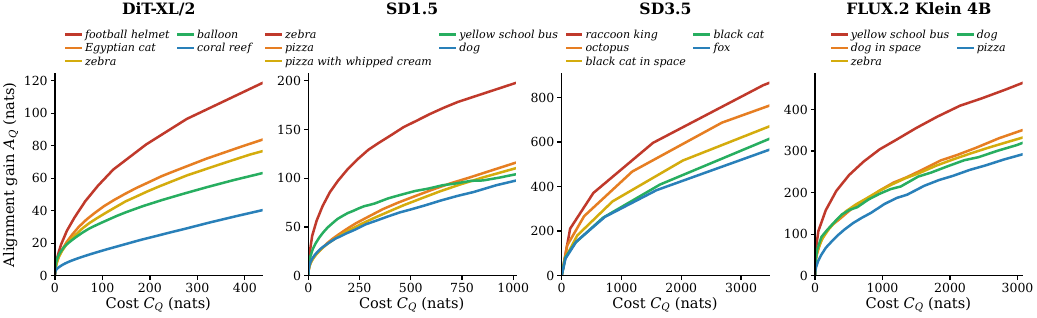}
    \caption{\textbf{CFG paths across conditions, per model}.
    Five conditions per model spanning the spread of the measured ones (30 ImageNet classes for DiT-XL/2, 10 prompts for SD1.5, all 5 for SD3.5 and FLUX.2 Klein 4B).}
    \label{fig:supp-conditions}
\end{figure*}

\Cref{fig:supp-conditions} shows a substantial spread of CFG operating curves within each model.
For DiT-XL (ImageNet), football helmet gains alignment much faster with cost than coral reef.
For SD1.5, yellow school bus gains more alignment than dog at shared costs, and for SD3.5 the composed scene of a raccoon king gains more than fox.
For FLUX.2 4B, yellow school bus has the largest gain over the displayed range.
These comparisons extend the main paper's argument that the response to guidance depends on the condition.
Here, an easier condition means one for which guidance buys more gain relative to its own conditional baseline.
These are sampler curves, so their ordering reflects both the condition and the sampler's ability to exploit it, and each model is evaluated in its own latent space.

\begin{figure*}[b]
    \centering
    {\footnotesize \textbf{SD1.5}: \textit{``a photo of a \dots''}}\\[1pt]
    \includegraphics[width=\linewidth]{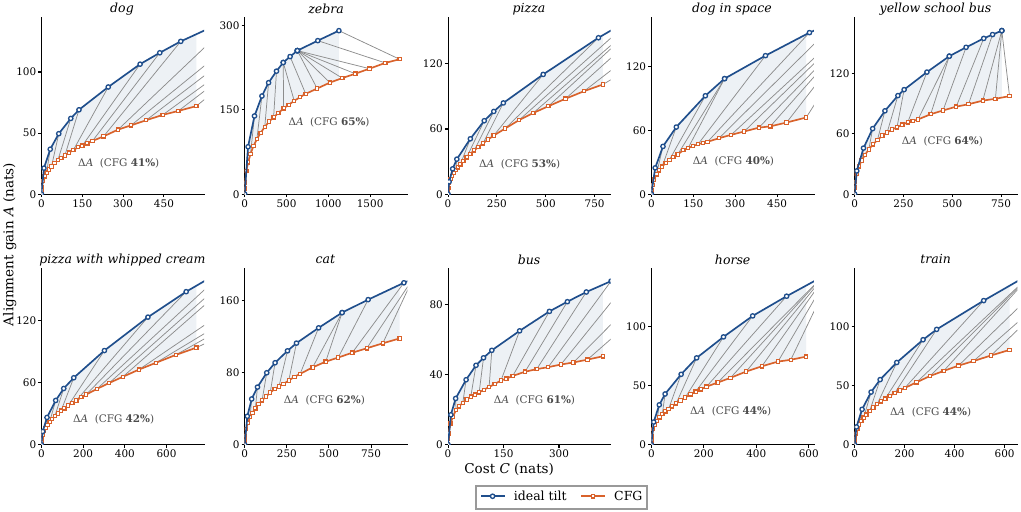}
    \caption{\textbf{SD1.5: ideal tilt and CFG.}
    Shading: the gain gap at matched cost. Gray lines join CFG and ideal points with the same $s$. Labels $\Delta A$: gain comparison at the same scale $s=6$, with bold percentages giving CFG's gain as a share of the ideal gain, $100 A_{\mathrm{CFG}}(6)/A_{\mathrm{ideal}}(6)$.}
    \label{fig:supp-ac-sd15}
\end{figure*}

For all ten SD1.5 prompts in \cref{fig:supp-ac-sd15}, CFG remains below the estimated ideal curve at matched cost.
The cat, bus, horse, and train panels extend this pattern to further everyday subjects, alongside the unusual combinations of dog in space and pizza with whipped cream.
The same ordering appears for two FLUX.2 Klein 4B prompts in \cref{fig:supp-flux-slack}, one of them compositional, suggesting that the distinction is relevant beyond simple class labels and familiar scenes.
The curves connect the two main observations of the paper: stronger alignment becomes increasingly expensive, and the sampler need not achieve the best available gain at that expense.

\subsubsection{How Much of the Gap Is Path Slack}
\label{sec:supp:slack-bound}

The sampler points are plotted at the path cost $C_Q$, while the ideal curve is the frontier of the end-point cost.
\Cref{thm:sampler-below} splits the horizontal distance between the two into a shortfall of the end-point law and a path slack, and \cref{eq:supp:slack-bound} bounds the slack from above with $n$ bridges per end point.
\Cref{fig:supp-ac-dit}(a) reports this bound for ten DiT-XL/2 classes, as bars like those of \cref{fig:tradeoff-imagenet}(a), and \cref{fig:supp-flux-slack} reports it for two text prompts on FLUX.2 Klein 4B; the settings are in \cref{sec:supp:details-paths}.
At each scale, the end-point cost $C[q_{\Gamma,s}]$ of the sampler lies between $C_Q$ minus the bound of \cref{eq:supp:slack-bound} and $C_Q$, and it cannot be below the frontier cost at the same gain.
Each bar spans this range.

Up to $s=2$ the lower bound is above the frontier cost in every class, by at least $2.7$ standard errors, so the end-point law of the sampler is itself off the frontier.
For zebra at $s=2$, for example, its cost is at least $10.1$ nats, where the tilt with the same gain costs $3.9$.
By \cref{eq:path-endpoint-bridge} the difference is a lower bound on the divergence of the end-point law from that tilt.
At these scales the shortfall is at least about $40\%$ of the plotted distance between the sampler and the frontier in every class.
For the two cats and the clog, the lower bound reaches $C_Q$, so the whole distance is shortfall and there is no measurable slack.

At $s=4$ the split depends on the class.
For the two cats and the hook, the shortfall is still more than $80\%$ of the distance, and for zebra and the clog it is about $60\%$ to $70\%$ (for zebra, at least $43$ nats where the tilt costs $11$).
For the balloon it is about half.
For the hourglass, the volcano, the coral reef and the football helmet, the shortfall that the bound shows is at most a quarter of the distance, and for the last three the lower bound is within two standard errors of the frontier, so there the bound cannot tell how much of the distance is slack.
The bars of \cref{fig:supp-ac-dit}(a) stop at $s=4$, because at $s=8$ the bound is informative only for the two cats and the hook.
In the other classes it is then noisy or below the frontier: the weights $e^{-\kappa\tilde W_r}$ are dominated by a few bridges, and more bridges would be needed.
Where the lower bound is above $C_Q$, which happens at $s\le2$ by up to about two standard errors, the bar shrinks to the square: the bound is taken on $64$ of the $256$ end points, while $C_Q$ is over all of them.
The two zebra samplers whose slack is zero by construction give a check of the bridges: the mean of $\tilde W_r$ minus $I_s$ is $-0.20\pm0.12$ at $s=0$ and $0.01\pm0.15$ at $s=-1$.
On FLUX.2 Klein 4B (\cref{fig:supp-flux-slack}), the lower bound is above the frontier cost at every scale for both prompts.
For the zebra prompt, the shortfall is $60\%$ to $80\%$ of the distance at every scale, and from $s=2$ on the lower bound is above the frontier cost by at least $4.7$ standard errors.
For the corgi prompt, the shortfall is $40\%$ to $65\%$ of the distance up to $s=7$, with at least $3.2$ standard errors from $s=2$ to $7$, and $16\%$ at $s=8$.
From $s=6$ on, the effective sample size of the resampled bridges is $2\%$ to $5\%$ of $n$, so there the slack bound is likely too large and the shortfall is understated.
The bound assumes exact denoisers (\cref{asm:exact}).

\needspace{6\baselineskip}   %
\subsubsection{Prompt Specificity and the CFG Tradeoff}
\label{sec:supp:tradeoff-flatness}

We next examine how adding details to a prompt changes the alignment gain available at a given cost.
Comparing nested descriptions in SD1.5, SD3.5 and FLUX.2 lets us assess whether greater prompt specificity produces a consistent ordering of the measured CFG tradeoffs.

\begin{figure*}[!t]
    \centering
    \includegraphics[width=\textwidth]{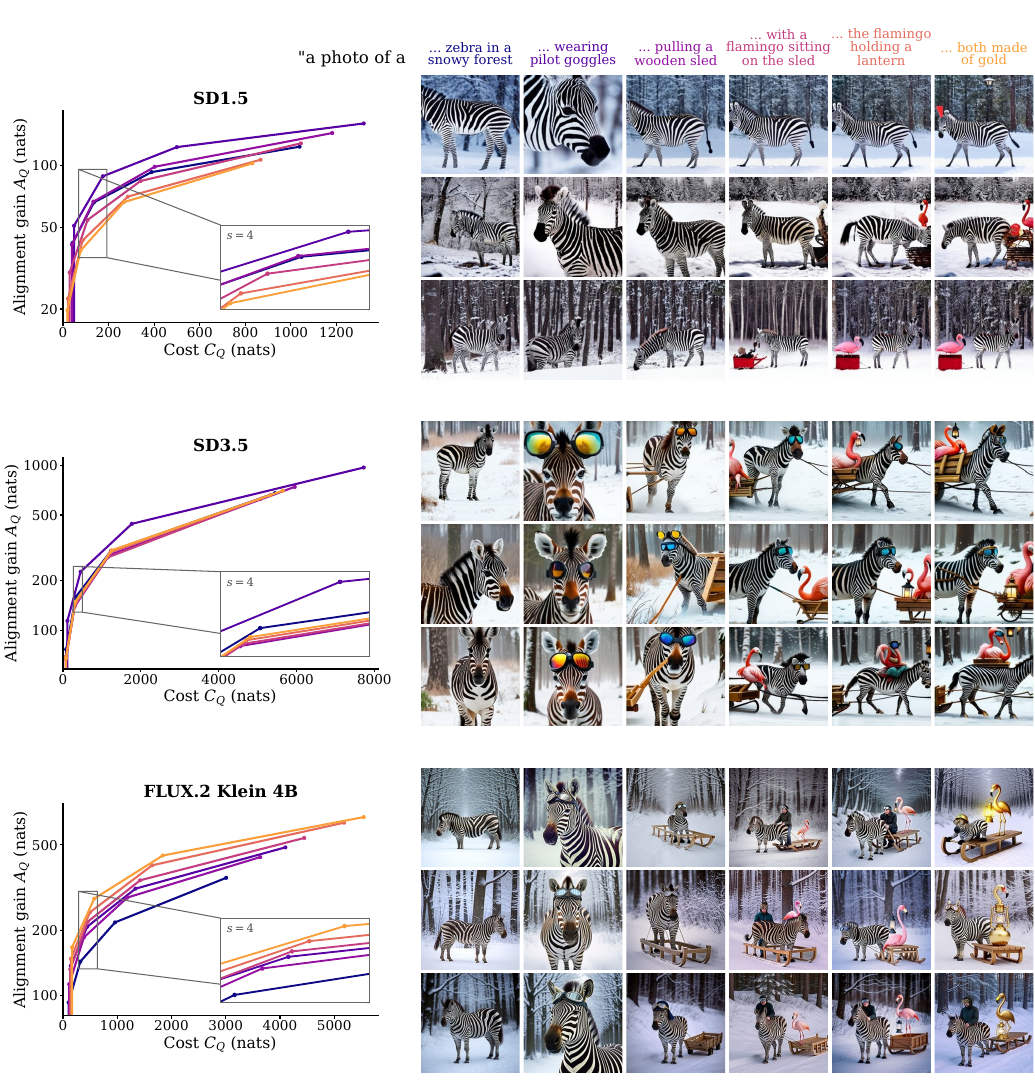}
    \caption{\textbf{Zebra prompt chain.} Left: measured CFG gain--cost curves for six nested prompts in SD1.5, SD3.5 and FLUX.2 Klein 4B, with a logarithmic $A_Q$ axis and dots at $s=2,4,8,16$. Insets magnify the region around the $s=4$ points. Right: each column adds the clause shown above it, with header color matching the corresponding curve. Within each model, each row uses the same noise across prompts. Images are shown at $s=4$ for SD1.5 and SD3.5, and $s=8$ for FLUX.2 Klein 4B.}
    \label{fig:supp-zebra-chain}
\end{figure*}

\begin{figure}[t]
    \centering
    {\footnotesize \textbf{FLUX.2 Klein 4B}}\\[1pt]
    \includegraphics[width=\linewidth]{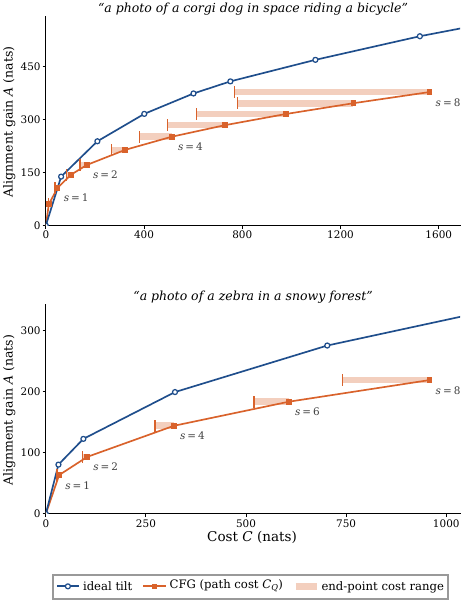}
    \caption{\textbf{FLUX.2 Klein 4B: frontier, CFG and the slack bound.}
    Estimated ideal frontier (circles) and CFG at the path cost $C_Q$ (squares) up to $s=8$, for two prompts.
    At each square, the bar is the range that holds the end-point cost $C[q_{\Gamma,s}]$ of the sampler, as in \cref{fig:supp-ac-dit}(a): from the larger of the frontier cost at the same gain and $C_Q$ minus the slack bound of \cref{eq:supp:slack-bound}, up to $C_Q$.
    The bound uses $n=64$ resampled bridges on $8$ end points (\cref{sec:supp:details-paths}).}
    \label{fig:supp-flux-slack}
\end{figure}

\Cref{fig:supp-zebra-chain} shows different responses to added prompt detail across models.
At matched cost, more elaborate descriptions generally lie higher in FLUX.2 Klein 4B, whereas SD1.5 shows a mostly downward ordering after the goggles clause.
SD3.5 also places the goggles prompt highest, but the curves for its longer descriptions lie close together, as the inset makes clear.
The insets magnify the region around the $s=4$ points; comparisons at matched cost follow a vertical slice through the curves, since equal guidance scales need not incur equal costs.
Thus, adding detail does not produce a universal ordering of the CFG tradeoffs.

Downward shifts can also accompany difficulty rendering an added relation.
In FLUX, adding ``pulling a wooden sled'' lowers the curve relative to the goggles prompt, and the displayed samples often place the zebra on the sled rather than pulling it.
In SD1.5, the lower curves for the later descriptions accompany samples that omit several requested details, including the lantern and gold appearance.
These examples suggest a connection between the response to guidance and compositional difficulty, but do not establish a general test of prompt satisfaction.
Here, a downward shift compares different prompts at matched cost: each individual curve still rises over the measured guidance scales.
A higher curve means more alignment gain relative to that prompt's own conditional baseline; it does not guarantee that every clause is satisfied.

\subsubsection{The Alignment Score as a Classifier}
\label{sec:supp:tradeoff-classification}

Throughout the paper we read $T_\gamma$ as an alignment score: how strongly the network judges that a noisy input satisfies the condition.
The gain $A$ is the mean change of this score under guidance, so its meaning as alignment rests on this reading.
This experiment checks the reading against a reference outside the model, the true labels of held-out real images, and asks two questions.
First, does $T_\gamma$ behave as a measure of alignment should?
Writing the condition explicitly, $T_\gamma(y;c)$ is, up to a constant, the log-probability $\log p(c\mid Y_\gamma=y)$ of the implicit classifier, so it should be highest for the condition an image actually satisfies.
We test this by classification against the true labels of held-out images, following the diffusion classifier of Li et al.~\cite{li2023diffusion}, who showed that the conditional denoisers of a diffusion model classify real images well: we noise an image to level $\gamma$, score it under each of $K$ classes, and predict the class with the largest score.
Classification accuracy is a proxy for alignment and not a direct test of it, but a score that separates the classes this way is one that responds to the content the condition asks for.
Second, is the result a property of a particular network, or of the data?
With exact denoisers, the answer is fixed by the data distribution alone, so comparing different networks tells us how close each comes to it.

Two facts make the test meaningful.
The unconditional law cancels when classes are compared at the same input,
\begin{align}
    T_\gamma(y;c)-T_\gamma(y;c')
    &=\log\frac{p^c_\gamma(y)}{p^{c'}_\gamma(y)}\notag\\
    &=\log\frac{p(c\mid Y_\gamma=y)}{p(c'\mid Y_\gamma=y)},
    \label{eq:supp:cls-bayes}
\end{align}
where the second equality holds under a uniform prior over the $K$ classes.
Hence $\arg\max_cT_\gamma(y;c)$ is the Bayes classifier of the class from the noisy observation $Y_\gamma=y$; it uses only the two denoisers, with no external classifier.
The comparison does not assume that the unconditional law is the mixture of the class laws: it cancels in \cref{eq:supp:cls-bayes}, and only the conditional laws enter.
With exact denoisers its accuracy is the Bayes accuracy of the Gaussian channel,
\begin{equation}
    \mathrm{acc}^\star(\gamma)
    =\mathbb E\Big[\max_c\,p(c\mid Y_\gamma)\Big],
    \label{eq:supp:cls-acc}
\end{equation}
where $Y_\gamma=\gamma X+W_\gamma$ and $X$ is an image of a uniformly drawn class.
\Cref{eq:supp:cls-acc} involves only the class-conditional data laws and the noise level: two exact models trained on the same data have the same accuracy curve.
It is also nondecreasing in $\gamma$, since a lower SNR is a further-noised version of a higher one.
A learned model departs from $\mathrm{acc}^\star$ only through the error of its heads.
We compute $T_\gamma$ along the noise path of each image with the pointwise identity \cref{eq:supp:pointwise}, for ten ImageNet classes and two models with different architectures and latent spaces, DiT-XL/2~\cite{peebles2023scalable} and EDM2-XS~\cite{karras2024analyzing} (details in \cref{sec:supp:details-pools}).

\begin{figure}[t]
    \centering
    \includegraphics[width=\columnwidth]{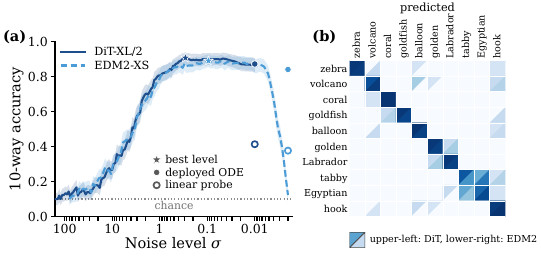}
    \caption{\textbf{The alignment score as a classifier.}
    (a) Ten-way accuracy of $\arg\max_cT_\gamma(y;c)$ on 300 held-out ImageNet images as a function of the noise level, for DiT-XL/2 and EDM2-XS (bands: $\pm2$ binomial standard errors; stars: best level).
    At the clean end, filled dots give the deployed-ODE classifier (on 320 images) and hollow dots a linear probe on clean latents.
    (b) Confusions of the deployed-ODE classifier, row-normalized (rows: true class); each cell is split between the two models.}
    \label{fig:supp-classifier}
\end{figure}

\Cref{fig:supp-classifier}(a) shows the accuracy as a function of the noise level.
At high noise it is at chance; it rises steeply between $\sigma\approx10$ and $\sigma\approx1$, where the class becomes visible, and peaks at $0.91$ for DiT ($\sigma\approx0.3$) and $0.89$ for EDM2 ($\sigma\approx0.1$), against $0.41$ and $0.38$ for a linear probe on the clean latents.
The two curves nearly coincide: down to $\sigma=0.01$ they differ by at most $0.06$, within two standard errors of their difference at every level, although the models share neither architecture nor latent space.
This answers both questions: $T_\gamma$ ranks the true class first for most images, and the agreement between the models is what \cref{eq:supp:cls-acc} predicts when both approach the same data-determined curve.
Since $\mathrm{acc}^\star$ cannot decrease as the noise falls, the declines in the measured curves point to model error rather than to the data:
the mild decline after the peak, to $0.87$ and $0.86$ at $\sigma=0.01$, and EDM2's last levels, $\sigma\le0.004$, where its path estimate falls to chance; the deployed classifier below shows that its class information is still present there.
Half the accumulated squared disagreement, $\frac12\int_0^\Gamma\|g\|^2d\gamma$, does not classify on its own ($0.05$ and $0.07$): the class evidence comes from how the image itself matches each class's predictions along the path.

The same comparison can be made with the densities that the models' conditional probability-flow ODEs assign to the clean image, $\arg\max_c\log q^{\mathrm{ODE}}(x\mid c)$.
This deployed classifier reaches $0.87$ on DiT and $0.84$ on EDM2 (filled dots in \cref{fig:supp-classifier}(a)), including EDM2's clean end where the path estimate has collapsed; a larger EDM2 conditional head gives $0.85$.
Its confusions in \cref{fig:supp-classifier}(b) are shared by the two models: most errors fall in the same few blocks, led by the two cat breeds (17 of DiT's 41 errors and 22 of EDM2's 51) and followed by the two retrievers.
These pairs also look alike, consistent with the score erring where the classes themselves overlap.
The classifier reading extends to text conditions: on SD1.5~\cite{rombach2022high}, the same deployed classifier sorts real COCO images among four captions with accuracy $0.89$, with most errors on dog images.

\subsubsection{The Cost Seen by an Outside Network}
\label{sec:supp:tradeoff-cost-kid}

The previous experiment checks the alignment score against the true labels.
Here we check the cost against a network that has no access to the denoisers.
The path cost $C_Q$ of \cref{eq:path-alignment-cost} is computed by the model itself, from the disagreement of its two denoisers along the trajectories, and measures how far guidance moves the sampler from its own conditional law.
A network that has no access to the denoisers gives a second view of the same move.
We embed the decoded CFG samples at scale $s$ and the model's unguided samples of the same condition with DINOv2~\cite{oquab2024dinov}, and measure the distance between the two sets with the kernel inception distance (KID)~\cite{bińkowski2018demystifying}.
Like $C_Q$, the KID is zero when the two sets come from the same law.
If $C_Q$ measures the departure from the conditional law, the outside distance should grow with it.

\begin{figure}[t]
    \centering
    \includegraphics[width=\columnwidth]{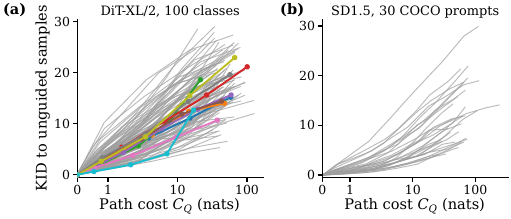}
    \caption{\textbf{The cost seen by an outside network.}
    For each condition, the KID between DINOv2 features of CFG samples at scale $s$ and of the model's unguided samples of the same condition, against the path cost $C_Q(s)$, for $s$ from $0$ to $4$; one curve per condition.
    (a) DiT-XL/2, 100 ImageNet classes (described in \cref{sec:supp:details-paths}); the ten classes of \cref{fig:tradeoff-imagenet}(b) in color.
    (b) SD1.5, 30 COCO prompts.}
    \label{fig:supp-cost-kid}
\end{figure}

\Cref{fig:supp-cost-kid} follows both quantities along the guidance scale, one curve per condition.
On DiT-XL/2, in all 100 ImageNet classes, the KID grows at every increase of the scale, so it orders the scales of each class exactly as $C_Q$ does.
On SD1.5, the same holds for 28 of the 30 COCO prompts; in the other two, one pair of neighboring scales is swapped (rank correlation $0.98$).
The zero point is shared as well: two independent sets of unguided samples of the same class are at KID $\approx0$ (median $-0.01$), against $2$ to $19$ for the guided samples already at $s=1$.
The curves leave zero smoothly, so even small costs appear as a measurable change in the samples.
How much distance one nat of cost buys depends on the condition, since a feature distance depends on how each condition is spread in feature space.
The cost is computed by the model in its own latent space, from its own denoisers; the distance is measured by a network with a different architecture, training and feature space.
Their agreement within each condition is a cross-validation of the cost as a measure of departure from the conditional law, in the same spirit as the agreement of the two models in \cref{sec:supp:tradeoff-classification}.
Settings are in \cref{sec:supp:details-paths}.

\subsection{Diversity Index}
\label{sec:supp:diversity}

Several experiments in this section compare coverage with the Vendi score~\cite{friedman2023vendi}, an established measure of sample diversity.
Vendi is computed from the pairwise similarities of a set of images; we use DINOv2 features.
We use it as an external sanity check: we neither aim to reproduce Vendi's values nor optimize agreement with it.
Our index has an intrinsic meaning in terms of the model's conditional and unconditional laws, and it is estimated from the denoisers without an external embedder.

\subsubsection{The Baseline Is Shared Across Models}
\label{sec:supp:diversity-models}

The baseline $V(0)=\mathrm{KL}(p^c_\gamma\|p_\gamma)$ is defined by the data laws, but we measure it with learned denoisers.
This experiment asks whether the measured value is a property of the condition or of the network that computes it.
If the baseline describes how much of the data a condition keeps, it should not depend on which model we ask.

\begin{figure}[t]
    \centering
    \includegraphics[width=\columnwidth]{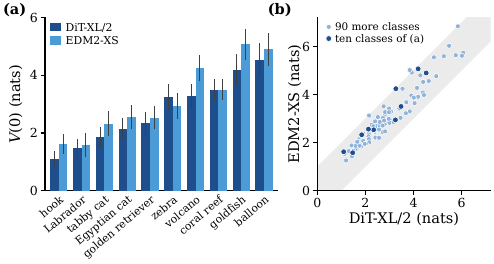}
    \caption{\textbf{The baseline is shared across models.}
    (a) Baseline $V(0)$ of ten ImageNet classes for DiT-XL/2 and EDM2-XS (bars: $\pm2$ standard errors over 256 class images), sorted by the DiT value.
    (b) DiT against EDM2 for 100 classes, with the ten classes of (a) in dark blue; the gray band marks differences within one nat.}
    \label{fig:supp-baseline-models}
\end{figure}

With exact denoisers, the two heads are the posterior means of the data, $\hat x_c=\mathbb E[X\mid Y_\gamma,c]$ and $\hat x=\mathbb E[X\mid Y_\gamma]$, so \cref{eq:coverage-immse} reads
\begin{equation}
    V(0)
    =\frac12\int_0^\Gamma
    \mathbb E_{p^c_\gamma}\Big\|\mathbb E[X\mid Y_\gamma,c]-\mathbb E[X\mid Y_\gamma]\Big\|^2\,d\gamma,
    \label{eq:supp:baseline-data}
\end{equation}
which involves only the class-conditional data laws and the noise.
Two exact models trained on the same data therefore give the same baseline for each class, and a gap between them is model error.
The two models we compare add noise in different latent spaces ($32{\times}32{\times}4$ for DiT-XL/2~\cite{peebles2023scalable} and $64{\times}64{\times}4$ for EDM2-XS~\cite{karras2024analyzing}), so equal values are expected rather than guaranteed by \cref{eq:supp:baseline-data}.
We estimate $V(0)$ for the ten classes of \cref{sec:supp:tradeoff-classification} with the pointwise identity \cref{eq:supp:pointwise}, along noise paths of the same ImageNet images for both models (details in \cref{sec:supp:details-pools}).

\begin{figure}[t]
    \centering
    \includegraphics[width=\linewidth]{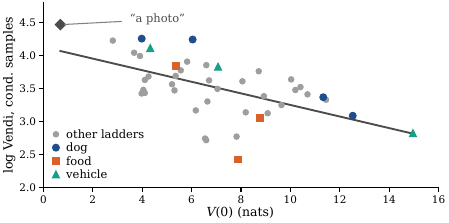}
    \caption{\textbf{Coverage and Vendi along prompt ladders (SD1.5).}
    $V(0)$ versus conditional-sample log Vendi for the three ladders of \cref{fig:diversity-prompts}(a) (color) and the prompts of ten more ladders (gray), with a least-squares line through all prompts.}
    \label{fig:supp-diversity-prompts-vendi}
\end{figure}

\Cref{fig:supp-baseline-models}(a) shows the baselines, $1.1$--$4.5$ nats for DiT and $1.6$--$5.1$ for EDM2.
The two models give nearly the same value per class: they differ by at most $0.98$ nats (volcano), by less than $0.5$ for seven of the ten classes, and order the classes alike (Spearman $0.95$).
Broad classes such as hook, the two retrievers and the two cats have the smallest baselines, and balloon and goldfish the largest, in both models.
\Cref{fig:supp-baseline-models}(b) plots the two models against each other: all ten classes lie within one nat of equality.
Two networks with different architectures and latent spaces thus recover nearly the same baseline, as \cref{eq:supp:baseline-data} predicts when both approach the data: the baseline is a property of the condition, not of the model.

\begin{table*}[t]
    \centering
    \small
    \setlength{\tabcolsep}{5pt}
    \begin{tabular}{llrccc}
        \toprule
        Prompt (``a photo of \dots'') & COCO rule & Matches & $-\log$ share & $V(0)$, cond.\ samples & $V(0)$, COCO images \\
        \midrule
        ``a photo'' (root) & all images & 118\,287 & $0$ & $0.7$ & $0.8$ \\
        \midrule
        an animal & supercategory \emph{animal} & 23\,989 & $1.6$ & $4.0$ & $4.1$ \\
        a dog & caption: dog & 5\,240 & $3.1$ & $6.0$ & $5.0$ \\
        a golden retriever & caption: golden retriever & 30 & $8.3$ & $11.3$ & $6.8$ \\
        a golden retriever puppy on grass & caption: golden retriever puppy & 0 & -- & $12.5$ & -- \\
        \midrule
        food & supercategory \emph{food} & 16\,255 & $2.0$ & $5.4$ & $5.9$ \\
        a pizza & caption: pizza & 3\,067 & $3.7$ & $7.9$ & $8.1$ \\
        a pepperoni pizza on a wooden table & caption: pepperoni & 285 & $6.0$ & $8.8$ & $9.2$ \\
        \midrule
        a vehicle & supercategory \emph{vehicle} & 27\,358 & $1.5$ & $4.3$ & $3.7$ \\
        a bus & caption: bus & 3\,230 & $3.6$ & $7.1$ & $7.6$ \\
        a red double-decker bus in London & caption: double decker & 804 & $5.0$ & $15.0$ & $13.7$ \\
        \bottomrule
    \end{tabular}
    \caption{\textbf{COCO shares of the three main prompt ladders (SD1.5).}
    Matches are COCO train2017 images (of 118\,287) selected by the rule; a caption rule keeps an image when any of its five captions contains the words, with an optional plural.
    $V(0)$ is in nats, estimated on the model's conditional samples and on the matched COCO images.
    The root and the puppy prompt (no matching caption) have no share below one and are left out of \cref{fig:diversity-prompts}(b).}
    \label{tab:supp-coco-prompts}
\end{table*}

The agreement is not limited to these ten classes.
Over 100 classes, these ten, 40 more chosen to span narrow and broad classes, and 50 drawn at random (\cref{fig:supp-baseline-models}(b)), the two models differ by less than one nat for 99 classes, with a median difference of $0.26$ nats, and order the classes alike (Spearman $0.95$, with a $95\%$ bootstrap interval of $[0.92,0.97]$, and $0.91$, $[0.82,0.96]$, on the 50 random classes alone; \cref{sec:supp:details-vendi}).

\subsubsection{More Prompt Ladders}
\label{sec:supp:diversity-chains}

A prompt that adds a detail narrows the set of images that satisfy it, so its baseline coverage loss $V(0)$ should grow, and its samples should become less varied.
\Cref{fig:diversity-prompts}(a) shows the first part on three prompt ladders.
This experiment asks two further questions.
First, does the trend hold beyond three ladders that we chose ourselves?
Second, does $V(0)$ agree with an external measure of variety, the Vendi score of the samples?
Vendi serves only as a sanity check here: $V(0)$ is computed from the two denoisers, with no image embedder.

The three main ladders start at ``a photo'' and add one detail at a time: an animal, a dog, a golden retriever, a golden retriever puppy on grass; food, a pizza, a pepperoni pizza on a wooden table; a vehicle, a bus, a red double-decker bus in London.
To avoid choosing ladders that suit the measure, we added ten more ladders that an LLM suggested as plausible, natural sequences of the same shape: a broad category, an object, an attribute, then a context.
For example, furniture, a couch, a leather couch in a living room; a person, a woman, a woman holding an umbrella, in the rain; a vehicle, a train, a steam train, with smoke.
Six of these ladders open new categories (furniture, electronics, a kitchen appliance, a person, and two sports), and four branch from the animal, food and vehicle prompts of the main ladders to a cat, a cake, an airplane and a train.
All prompts are listed in \cref{sec:supp:details-conditions}.

\Cref{fig:supp-diversity-prompts-vendi} plots $V(0)$ against the Vendi score of each prompt's conditional samples.
Prompts with greater coverage loss produce less varied samples, and the points of all thirteen ladders fall around a single downward line.
Along a ladder, Vendi falls at most added details.
The prompts of the ten ladders scatter more around the line than those of the main ladders: prompts with similar $V(0)$ can differ in visible variety.
Coverage therefore orders diversity most reliably within a ladder, and as a broad trend across unrelated prompts.
Under guidance, coverage keeps the prompt ordering and continues to rank diversity.

\subsubsection{COCO Shares in Detail}
\label{sec:supp:diversity-coco}

This subsection details the comparison of \cref{fig:diversity-prompts}(b), where $V(0)$ of each prompt is set against the negative log share of the COCO images that match it (\cref{eq:supp:subset}).

\textbf{Matching prompts to annotations.}
Shares are counted on the 118\,287 images of COCO train2017~\cite{lin2014microsoft}, from its annotations alone.
A prompt is matched by an instance rule, which keeps the images that contain an object of a given COCO supercategory or category, by a caption rule, which keeps the images for which any of the five captions contains given words, or by both.
On the three main ladders (\cref{tab:supp-coco-prompts}), the broad prompts use the supercategory (animal, food, vehicle) and the others a caption rule on their key words; the London-bus prompt, for example, keeps the images whose captions mention a double decker.
On the ten ladders, a category prompt uses the COCO category (couch, laptop, cat, \dots), and each more specific prompt adds caption words to the rule of the prompt before it: ``a leather couch in a living room'' keeps the couch images whose captions mention leather and a living room.
These rules nest, so the share can only shrink along a ladder, and each prompt of these ten ladders keeps at least 64 images.
On the main ladders, the golden-retriever prompt keeps only 30 (\cref{tab:supp-coco-prompts}).

\textbf{Counting shares.}
The share of a prompt is its number of matching images divided by 118\,287.
``A photo'' matches every image, so its share is one, and the puppy prompt matches none, since no caption mentions a golden retriever puppy; both are left out, which leaves nine prompts from the main ladders and 32 from the ten ladders.
$V(0)$ is estimated on two pools per prompt, the model's conditional samples and the COCO images matched by the same rule (\cref{sec:supp:details-pools}).

\textbf{Result.}
Prompts that keep a smaller share of COCO have a larger baseline coverage loss.
A robust line through the conditional samples has slope $0.91$, close to one: halving the share adds about $\log2$ nats to $V(0)$, as subset selection predicts, so differences in $V(0)$ between prompts measure differences in the share they keep.
The line is shifted up by a common offset of about $2.6$ nats.
A shift is expected: the shares are counted in COCO, while $V(0)$ is measured against the unconditional law of the model, which was trained on a broader collection of images.
Even ``a photo'', which keeps all of COCO, measures $0.7$ nats.
The agreement is strongest within a ladder, where each added detail shrinks the share and raises $V(0)$ at almost every step; across ladders, prompts with similar shares can differ by several nats.
The matched COCO images follow the same trend as the conditional samples, and slightly more closely, since they carry the variety of the matching images rather than the model's rendering of the prompt.
As in any empirical comparison, not every prompt falls on the line, but such prompts are a minority: 31 of the 41 lie within the band, so the offset is shared rather than prompt specific.

\subsubsection{ImageNet: Coverage and Vendi Diversity}
\label{sec:supp:diversity-imagenet}

\begin{figure*}[t]
    \centering
    \includegraphics[width=\textwidth]{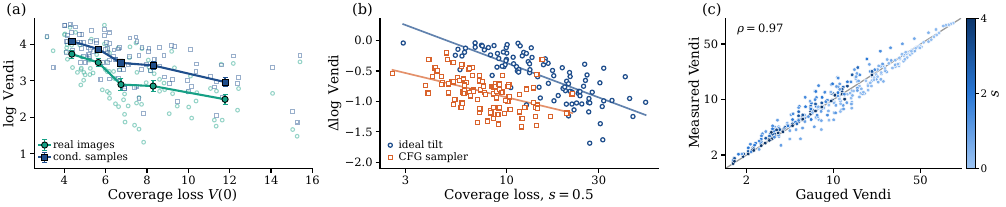}
    \caption{\textbf{ImageNet coverage and Vendi diversity.}
    (a) Baseline $V(0)$ versus log Vendi for real images and conditional samples. Faint markers: the 100 classes; filled markers: means over quintiles of $V(0)$, with standard errors.
    (b) Coverage lost from $s=0$ to $s=0.5$, $A+C$ for the ideal tilt and $A_Q+C_Q$ for the CFG sampler, versus the change in log Vendi, with least-squares lines.
    (c) Measured Vendi of CFG samples against the prediction of \cref{eq:vendi-gauge} from $A_Q+C_Q$, for 30 classes and $0<s\le4$.
    Coverage is in nats.}
    \label{fig:diversity-imagenet}
\end{figure*}

To check whether coverage captures visible variety, we compare it with the Vendi score on ImageNet.
For the class-conditioned DiT model, we evaluate it separately on equally sized pools of real training images and ordinary conditional samples within each class.
These pools test whether coverage reflects variety in both the data and the model's outputs.
The quantitative comparison is in \cref{tab:diversity-checks} and \cref{fig:diversity-imagenet}, and samples that show the loss of diversity under guidance, for SD3.5 and FLUX.2, follow in \cref{sec:supp:diversity-visual}.

\paragraph{Across classes.}
The top of \cref{tab:diversity-checks} and \cref{fig:diversity-imagenet}(a) rank the 100 classes by their baseline coverage.
Classes with greater coverage have more diverse real images and more diverse conditional samples, so the two denoisers recover the ordering of class variety before any guidance.
The middle of \cref{tab:diversity-checks} and \cref{fig:diversity-imagenet}(b) rank the same classes by the coverage they lose from $s=0$ to $s=0.5$, for the ideal tilt and for the CFG sampler.
Classes that lose more coverage also lose more diversity according to Vendi, strongly for the tilt and moderately for the sampler, whose index is a path-space quantity.
The same holds at $s=1$, with slightly lower correlations for both.
All coverage terms in these rows are read at the horizon $\sigma=2$; the reason for this horizon is given in \cref{sec:supp:horizon}.

\begin{table}[t]
    \centering
    \small
    \setlength{\tabcolsep}{3pt}
    \begin{tabular}{lclc}
        \toprule
        Diversity of & $s$ & Predictor & Spearman [95\% CI] \\
        \midrule
        \multicolumn{4}{l}{\emph{Baseline: class diversity before guidance, 100 classes}} \\
        Real images & $0$ & $-V(0)$ & $0.60$ [$0.45,0.72$] \\
        Cond.\ samples & $0$ & $-V(0)$ & $0.63$ [$0.49,0.74$] \\
        \midrule
        \multicolumn{4}{l}{\emph{Change: diversity lost from $0$ to $s$, 100 classes}} \\
        Ideal tilt & $0.5$ & $-(A+C)$ & $0.79$ [$0.71,0.85$] \\
                   & $1$   & $-(A+C)$ & $0.75$ [$0.64,0.83$] \\
        CFG sampler & $0.5$ & $-(A_Q+C_Q)$ & $0.53$ [$0.37,0.66$] \\
                    & $1$   & $-(A_Q+C_Q)$ & $0.46$ [$0.27,0.62$] \\
        \midrule
        \multicolumn{4}{l}{\emph{Along guidance: 30 classes}} \\
        Per class (median) & $0$--$4$ & $-(A_Q+C_Q)$ & $1.00$ [$1.00,1.00$] \\
        All (class, $s$) & $0$--$4$ & Gauge~\eqref{eq:vendi-gauge} & $0.97$ [$0.95,0.98$] \\
        \bottomrule
    \end{tabular}
    \caption{\textbf{Coverage as a diversity measure.}
    Spearman rank correlation between coverage predictors and DINOv2 Vendi of ImageNet classes, with 95\% bootstrap intervals over classes.
    Top: diversity before guidance. Middle: the loss of diversity from $s=0$ to the listed scale; coverage in both is read at the horizon $\sigma=2$.
    Bottom: along the guidance scale.
    The predictors carry a minus sign because a larger coverage loss means fewer distinct outcomes, so a positive correlation means coverage orders the classes as Vendi does.
    The gauge of \cref{eq:vendi-gauge} predicts Vendi's shape across classes and scales.}
    \label{tab:diversity-checks}
\end{table}

\paragraph{Within a class, along the guidance scale.}
Practitioners raise $s$ for a given class and see its samples become alike.
We therefore follow 30 classes from $s=0$ to $s=4$, and compare the coverage loss of the CFG sampler with the Vendi of its samples at every scale.
Within each class, $-(A_Q+C_Q)$ orders the scales as Vendi does in 28 of the 30 classes (bottom of \cref{tab:diversity-checks}).
Vendi's values, however, are not on our scale: $e^{-(A_Q+C_Q)}$ is the fraction of baseline coverage that guidance keeps (\cref{eq:guided-coverage}), whereas Vendi counts images its embedding tells apart, and it levels off at a class-dependent floor $\mathrm{Vendi}_\infty$ of variety, such as pose and background, that guidance does not remove.
Comparing values therefore requires a gauge.
The subset-selection reading of $N(s)$ suggests that the removable variety shrinks with the coverage kept, so we model the count above the floor as
\begin{equation}
    \mathrm{Vendi}(s)
        =\mathrm{Vendi}_\infty
        +\bigl[\mathrm{Vendi}(0)-\mathrm{Vendi}_\infty\bigr]\,
        e^{-\nu\,[A_Q(s)+C_Q(s)]},
    \label{eq:vendi-gauge}
\end{equation}
with one rate $\nu\approx0.11$ for all classes, the share of coverage nats that the embedding resolves.
This gauged Vendi, with $\nu$ fitted on held-out classes, matches measured Vendi across all 320 (class, scale) cells ($\rho=0.97$, \cref{fig:diversity-imagenet}(c); fitting details in \cref{sec:supp:details-vendi}).

\begin{table*}[t]
    \centering
    \footnotesize
    \setlength{\tabcolsep}{4pt}
    \newcommand*{\rr}{\raggedright\hspace{0pt}}%
    \begin{tabular}{@{}c p{.17\textwidth} p{.25\textwidth} p{.15\textwidth} p{.32\textwidth}@{}}
        \toprule
        Set & \rr Scenery (``A breathtaking view of a distant \dots'') & \rr Plain photo (``A photo of \dots'') & \rr Painting (``An oil painting of \dots'') & \rr Composite scene (``a photo of \dots'') \tabularnewline
        \midrule
        1 & \rr castle on a misty hill & \rr a \{tabby / black\} cat napping on a windowsill & \rr a lighthouse keeper & \rr a \{tabby / black\} cat astronaut floating inside a space station \tabularnewline
        2 & \rr sailboat on a calm sea & \rr a \{sea turtle / jellyfish\} swimming over a coral reef & \rr a pirate captain & \rr \{an octopus / a seahorse\} playing the piano in an underwater concert hall \tabularnewline
        3 & \rr city skyline at dusk & \rr a \{red vintage car / Vespa scooter\} parked on a cobblestone street & \rr a street musician & \rr a hot air balloon shaped like \{a teapot / an elephant\} floating over Paris \tabularnewline
        4 & \rr waterfall in a rainforest & \rr \{a fox / an owl\} sitting in the snow & \rr a medieval knight & \rr a \{raccoon / hedgehog\} wearing a crown on a throne made of pizza \tabularnewline
        5 & \rr windmill in a tulip field & \rr \{a bowl of ramen / a stack of pancakes\} on a wooden table & \rr an old baker & \rr \{an astronaut / a panda\} riding a horse on the moon \tabularnewline
        \bottomrule
    \end{tabular}
    \caption{\textbf{Prompt sets of the SD3.5 and FLUX.2 diversity figures.}
    Each set follows the pattern of the prompts of \cref{fig:diversity-flux}: a scenery view, a plain photo, an oil painting and a composite scene; the photo and the composite scene each come in two versions that differ only in the subject (braces).
    \Cref{fig:supp-diversity-sets-sd35,fig:supp-diversity-sets-flux} show sets 1--4 with the scenery view and the first subject of each pair; \cref{fig:supp-diversity-one-scale} shows all sets with the scenery view, the painting and the second subject of each pair.}
    \label{tab:supp:diversity-prompt-sets}
\end{table*}

\paragraph{Motivating the Vendi gauge.}\label{sec:supp:vendi-gauge}
\Cref{eq:vendi-gauge} is a fitted model, not a derived identity.
By \cref{eq:sampler-coverage}, the fraction of baseline coverage that the sampler keeps at scale $s$ is $N_Q(s)/N_Q(0)=e^{-[A_Q(s)+C_Q(s)]}$, with $N_Q=e^{-V_Q}$ (\cref{eq:supp:VQ}).
By analogy with subset selection (\cref{lem:supp:subset}), we read this as the share of the baseline outcomes still produced.
The analogy is not exact: under \cref{asm:girsanov} the sampler laws are mutually absolutely continuous, so none is a restriction of another to a subset.
Vendi counts images that a feature embedding tells apart.
It does not fall to one, because pose, viewpoint and background vary within a class even when guidance has fixed the content; this is the class-dependent floor $\mathrm{Vendi}_\infty$.
The embedding also resolves only part of the removed variety, so the count above the floor falls as a power $\nu\in(0,1]$ of the retained fraction, which gives the exponent $-\nu\,[A_Q(s)+C_Q(s)]$ of \cref{eq:vendi-gauge}.
The floor is fitted per class and $\nu$ is shared (\cref{sec:supp:details-vendi}).
The rate $\nu$ depends on the embedding and is not a property of the model.
The gauge describes only the range of scales where Vendi decreases with $s$.

\subsubsection{What Diversity Loss Looks Like}
\label{sec:supp:diversity-visual}

\renewcommand{\dblfloatpagefraction}{0.8}

\begin{figure*}[p]
    \centering
    \includegraphics[width=\textwidth]{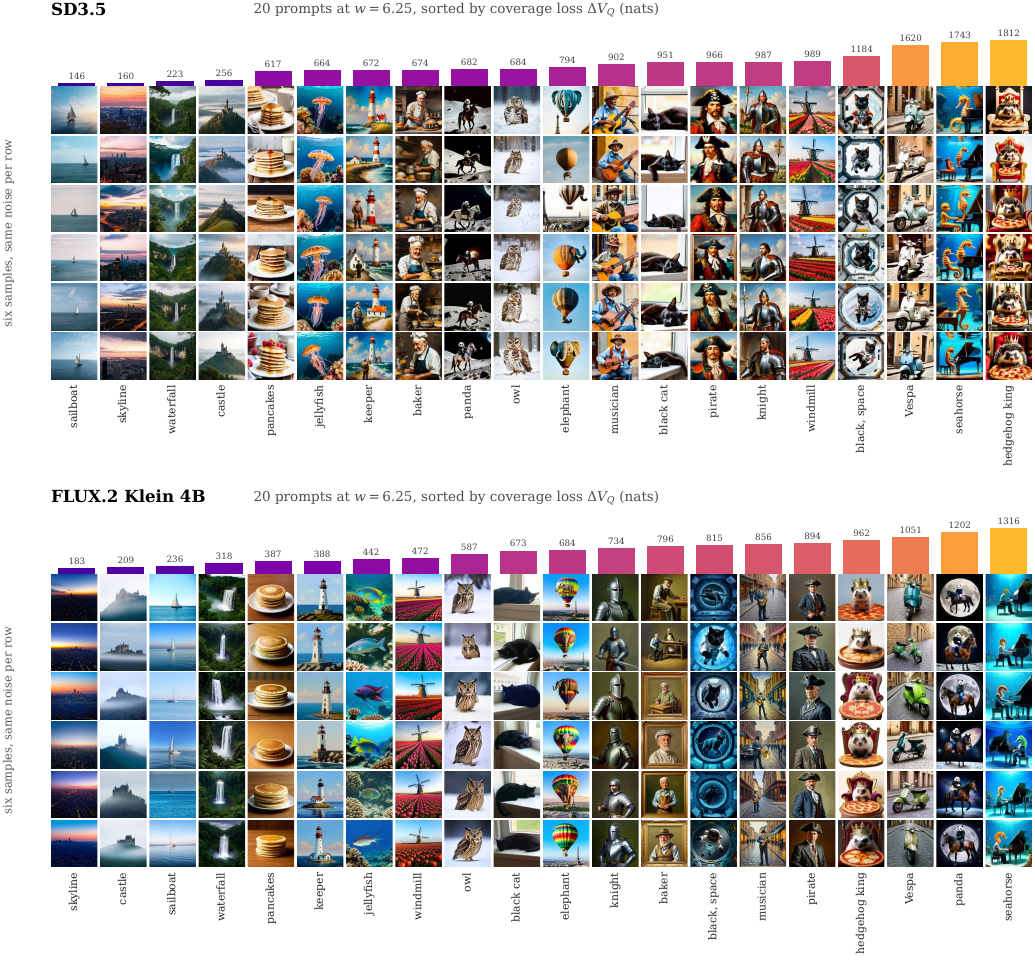}
    \caption{\textbf{Coverage loss ranks visible variety across twenty prompts.}
    SD3.5 (top) and FLUX.2 Klein 4B (bottom) with the CFG sampler at $w=6.25$.
    Each column shows six samples of one prompt, with the same noise in every row; columns are sorted by the coverage loss $\Delta V_Q=A_Q+C_Q$ at that scale (\cref{eq:sampler-coverage}), given in nats by the bar above each column.
    The twenty prompts are scenery views, plain photos, oil paintings and composite scenes, listed in \cref{tab:supp:diversity-prompt-sets}; the bar color follows $\Delta V_Q$.}
    \label{fig:supp-diversity-one-scale}
\end{figure*}

\Cref{fig:supp-diversity-one-scale} sorts twenty prompts by their coverage loss at one guidance scale for SD3.5 and FLUX.2: in both models the scenery prompts with the smallest loss keep varied framing, light and layout across the six samples, while the prompts with the largest loss, mostly composite scenes, repeat one composition.

\begin{figure*}[p]
    \centering
    \includegraphics[width=\textwidth]{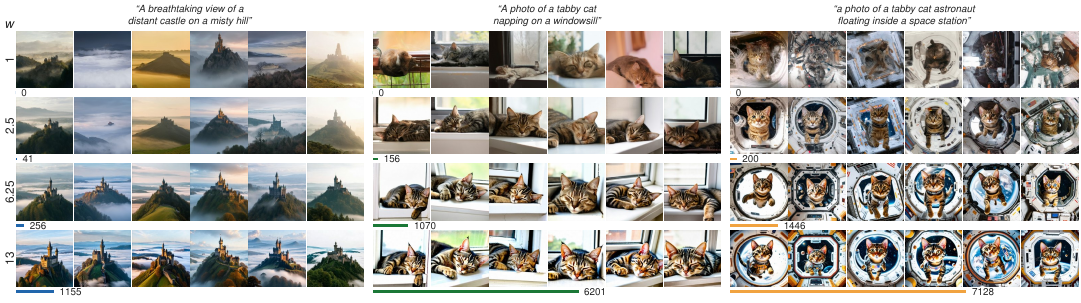}\\[0pt]
    \includegraphics[width=\textwidth]{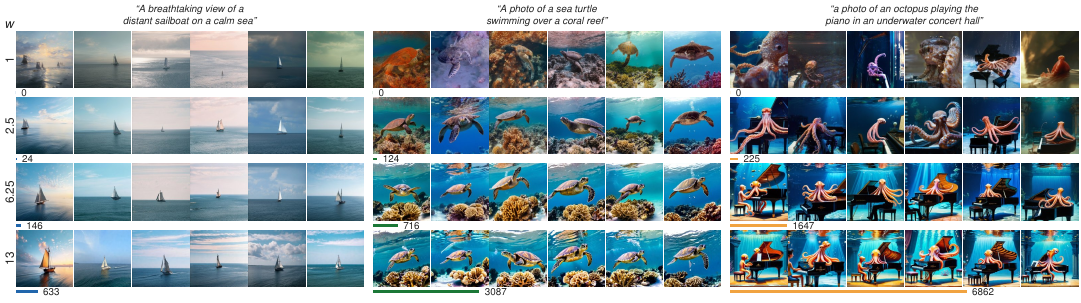}\\[0pt]
    \includegraphics[width=\textwidth]{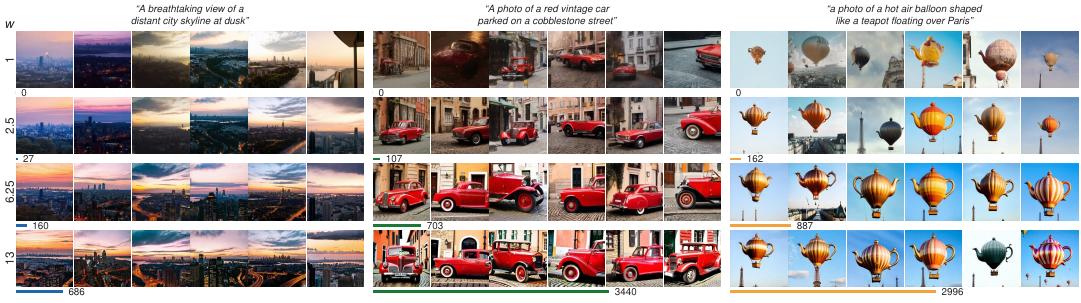}\\[0pt]
    \includegraphics[width=\textwidth]{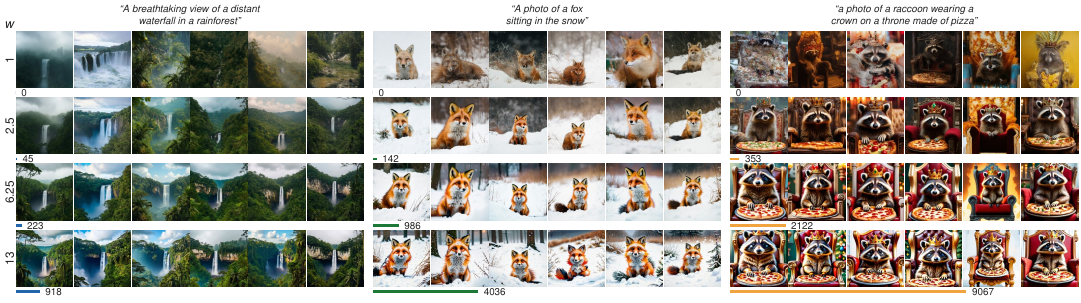}
    \caption{\textbf{Diversity falling with guidance, SD3.5.}
    Sets 1--4 of \cref{tab:supp:diversity-prompt-sets}, each with its scenery view, plain photo and composite scene, at $w=1, 2.5, 6.25, 13$; within a prompt, each column uses the same noise in every row, and the bars give the coverage loss $\Delta V_Q=A_Q+C_Q$ in nats (\cref{eq:sampler-coverage}), on one shared scale per set.}
    \label{fig:supp-diversity-sets-sd35}
\end{figure*}

\begin{figure*}[p]
    \centering
    \includegraphics[width=\textwidth]{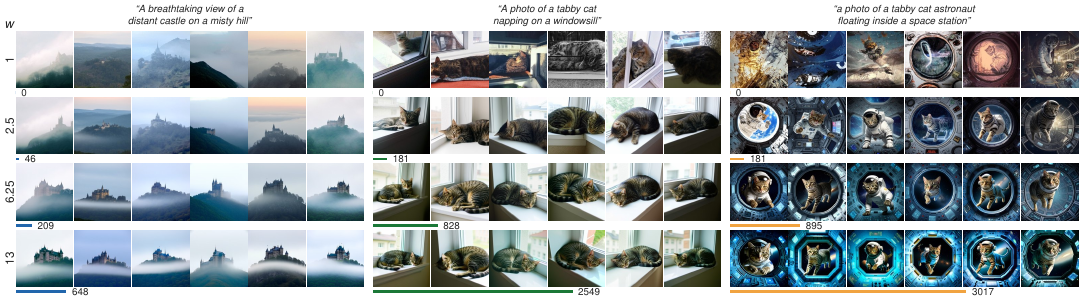}\\[0pt]
    \includegraphics[width=\textwidth]{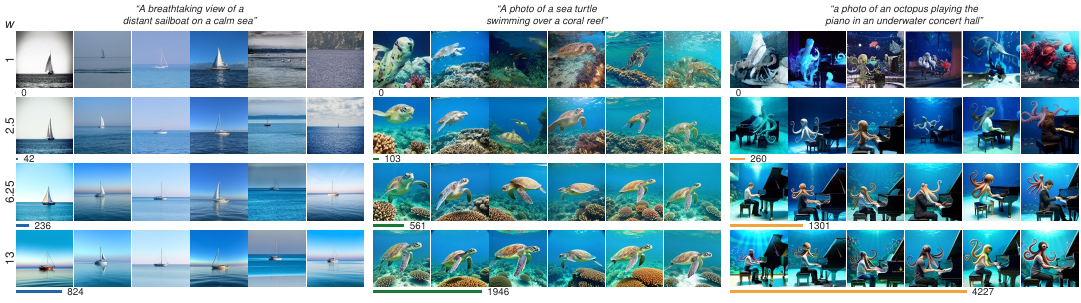}\\[0pt]
    \includegraphics[width=\textwidth]{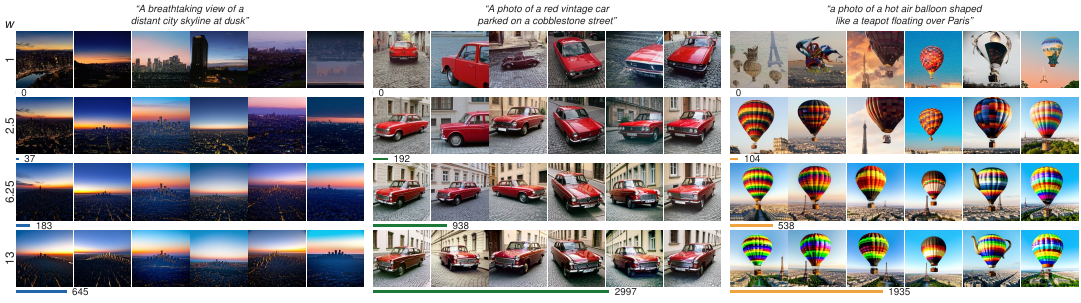}\\[0pt]
    \includegraphics[width=\textwidth]{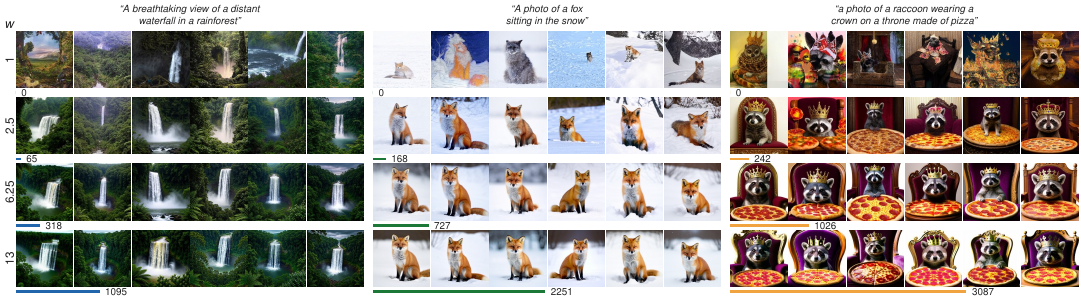}
    \caption{\textbf{Diversity falling with guidance, FLUX.2 Klein 4B.}
    Sets 1--4 of \cref{tab:supp:diversity-prompt-sets}, each with its scenery view, plain photo and composite scene, at $w=1, 2.5, 6.25, 13$; within a prompt, each column uses the same noise in every row, and the bars give the coverage loss $\Delta V_Q=A_Q+C_Q$ in nats (\cref{eq:sampler-coverage}), on one shared scale per set.}
    \label{fig:supp-diversity-sets-flux}
\end{figure*}

\Cref{fig:supp-diversity-sets-sd35,fig:supp-diversity-sets-flux} follow four prompt sets (\cref{tab:supp:diversity-prompt-sets}) across guidance scales in the layout of \cref{fig:diversity-flux}: as $w$ rises, the samples of every prompt converge toward one composition, and the prompt whose coverage loss grows fastest converges first.
\clearpage
\section{The Noise Horizon}
\label{sec:supp:horizon}

\begin{figure*}[t]
    \centering
    \includegraphics[width=\linewidth]{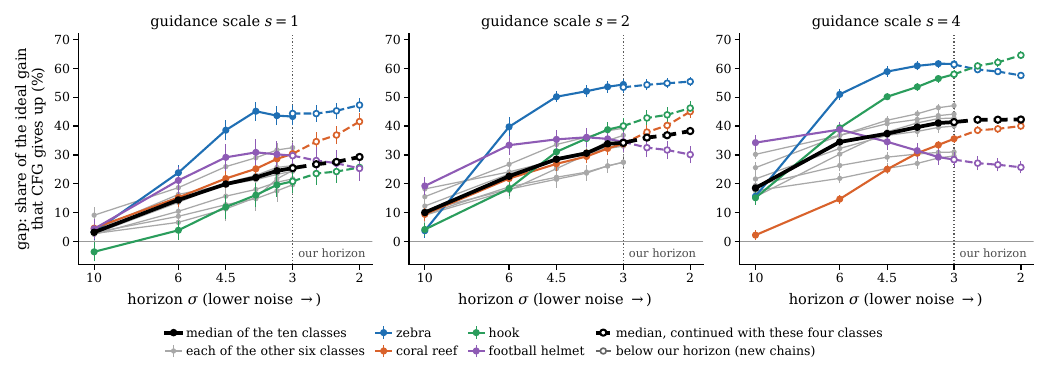}
    \caption{\textbf{Sensitivity of the gap between CFG and the frontier to the noise horizon.}
    Each point repeats the full analysis of \cref{fig:supp-ac-dit} with the integrals stopped at the horizon on the horizontal axis; the noise decreases to the right, and the dotted line marks our horizon, $\sigma=3$.
    The vertical axis is the gap: the share of the ideal gain that CFG gives up at equal path cost.
    Each panel is one guidance scale.
    Each line is one of the ten DiT-XL/2 classes (the bars are one standard error), and the solid black line is their median.
    Four classes, in color, are also followed below our horizon, to $\sigma=2$, with a frontier estimated from additional Langevin chains (dashed lines, open markers; \cref{sec:supp:details-horizon}).
    The dashed black line continues the median there: it starts from the median of the ten classes at $\sigma=3$ and adds the median change of these four classes since $\sigma=3$.
    The gap is not specific to our horizon: it is smallest at $\sigma=10$, grows as lower noise levels are included, and is larger at a larger guidance scale at every horizon.
    Below $\sigma=3$ the gap remains in all four classes and they keep their order; its size still moves, up for coral reef and hook and down for football helmet, by more than it does for zebra.}
    \label{fig:supp:horizon-sweep}
\end{figure*}

\paragraph{Choice of the horizon.}
Every quantity in the paper is defined at a finite SNR $\Gamma$, which we call the \emph{horizon}.
The divergence of \cref{eq:renyi-integral}, the gain and the cost of \cref{sec:tradeoff}, and the coverage of \cref{sec:diversity} all compare noisy laws at level $\Gamma$; the clean laws are reached only in the limit $\Gamma\to\infty$.
The horizon is therefore where we choose to meet the end point of the sampler.
We stop at $\sigma\approx3$, that is $\Gamma\approx1/9$, and not at the lowest noise level of the model.
We choose this horizon so that our quantities compare the laws at a scale that matters for perception, and we base the choice on previous work.
Ohayon et al.~\cite{ohayon2026distance} define the information-estimation metric (IEM), which integrates a squared difference of denoiser outputs over the SNR, as our quantities do.
They stop the integral at a noise level four times the standard deviation of their images, although their denoiser is trained at noise levels about two thousand times lower~\cite[App.~E.2]{ohayon2026distance}.
At this horizon the metric has ``a strikingly high Spearman's rank correlation coefficient'' with the mean opinion scores of human observers on three standard image-quality benchmarks, TID2013~\cite{ponomarenko2015image}, LIVE~\cite{sheikh2005live} and CSIQ~\cite{larson2010most}~\cite[Sec.~3.2]{ohayon2026distance}.
This is our reason for the place of the cut: a quantity of the same form, read at this horizon, correlates highly with how humans perceive the clean images.
We use one horizon for all models.
In units of the standard deviation of the latents (about $0.8$, $0.9$ and $1.0$ for DiT-XL/2, SD1.5 and FLUX.2), it is $3.6$, $3.3$ and $2.9$, close to the four of the IEM.
We also measure how our main conclusions depend on the horizon (\cref{fig:supp:horizon-sweep}).

\begin{figure*}[t]
    \centering
    \includegraphics[width=\linewidth]{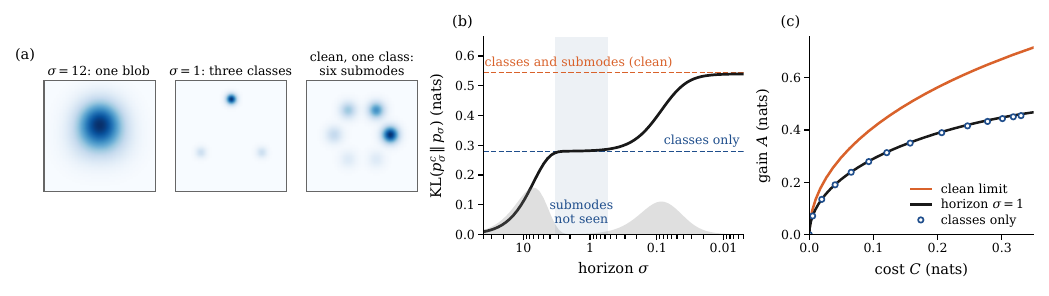}
    \caption{\textbf{The horizon as a regularizer, on a Gaussian mixture with two scales}: three classes far apart, each made of six narrow submodes (the parameters are in \cref{sec:supp:details-horizon}).
    The condition prefers one class and, inside every class, some of the submodes.
    (a) The conditional law at three noise levels. At high noise it is a single blob; at an intermediate horizon the three classes are apart; the submodes of a class appear only near the clean limit (the last image is a close-up of one class).
    (b) The KL divergence between the conditional and the unconditional laws against the horizon. It rises in two steps: first to the value of the classes alone, where it stays over the shaded range, and only at much lower noise to its clean value, which also counts the submodes. The gray area is the KL gained per unit of log-SNR, on its own scale: it peaks once where the classes separate and once where the submodes separate.
    (c) At a horizon inside the shaded range, the frontier coincides with the frontier of the classes alone (markers). At the clean limit it is a different curve.}
    \label{fig:supp:toy-horizon}
\end{figure*}

\paragraph{Finite horizon and clean limit.}
A finite horizon does more than cut the integral: it sets the resolution at which the two laws are compared.
Adding noise convolves both laws with the same Gaussian, which smooths them.
Near the clean limit, a divergence between the laws reacts to every difference between images, down to single pixels, much as a pixel-wise $L_2$ distance does.
At very high noise the two laws are the same Gaussian, and nothing is seen.
A useful measure lies between the two, and the horizon sets where: raising the SNR adds finer and finer differences between the laws, and these need not be differences that a viewer sees.
Ohayon et al.\ read the horizon of the IEM in the same way: it controls ``the finest resolution at which the metric is adapted to the density''~\cite[Sec.~4]{ohayon2026distance}, and in a mixture the modes merge at low SNR and separate only as the SNR grows~\cite[Fig.~1]{ohayon2026distance}.
In this sense the horizon acts as a regularizer that sets the scale of the attributes being compared; the example at the end of this section shows this effect (\cref{fig:supp:toy-horizon}).

Ideally, the horizon is high enough that what remains to the clean limit is negligible.
By data processing, the divergence between the conditional and the unconditional laws at the horizon is a lower bound on the clean one, and it does not decrease as $\Gamma$ grows (\cref{sec:supp:info-measures}).
For the gain, the frontier and the gap between a sampler and the frontier, we measure the dependence on the horizon (\cref{fig:supp:horizon-sweep}).

\paragraph{Estimates at low noise.}
With exact denoisers, the horizon could be moved to the clean limit at no cost.
With learned denoisers it cannot.
Once the class is decided, the disagreement should fade, yet per unit of log-SNR the measured disagreement stays large down to the lowest noise levels, and the integrals keep growing.
The difference of the two denoisers enters there with a large SNR weight, so a small error in it is amplified and cannot be told from a real difference in fine detail.

\paragraph{What the horizon covers.}
A useful horizon must include the noise levels at which the condition shapes the sample, and two measurements locate them.
The first is the alignment score used as a classifier (\cref{sec:supp:tradeoff-classification}): its accuracy is at chance at high noise, rises steeply between $\sigma\approx10$ and $\sigma\approx1$, that is for $\gamma$ between $0.01$ and $1$, and then saturates (\cref{fig:supp-classifier}).
The second is the disagreement $g$ of the two denoisers, which drives every quantity in the paper.
Per unit of log-SNR, the mean disagreement $\gamma\,\mathbb E_{p^c_\gamma}\|g\|^2$ grows by more than an order of magnitude over the same band.
Both suggest that the attributes specific to the condition are decided in this band, while lower noise levels mainly add finer detail.
At $\sigma=3$ the classifier is already far above chance (accuracy $0.58$ against $0.1$ for the ten classes of DiT-XL/2), and the levels below refine the class and add detail.

One exception to the common horizon is the ImageNet diversity study of \cref{sec:supp:diversity-imagenet}, which reads coverage at $\sigma=2$.
That study compares coverage with the variety of images within a class, which is decided at lower noise than the identity of the class.
At $\sigma=2$ the baseline is still a property of the class: two different models order the classes alike there, whereas near the clean end they do not.

\paragraph{Sensitivity of the gap to the horizon.}
Does the main finding of \cref{sec:tradeoff}, the gap between CFG and the frontier, depend on where we stop?
\Cref{fig:supp:horizon-sweep} answers this by repeating the whole analysis of the ten DiT-XL/2 classes of \cref{fig:supp-ac-dit} with the integrals stopped earlier, at horizons from $\sigma=10$ down to our $\sigma=3$.
At each horizon we recompute the frontier, the gain and the cost, and read the gap: the share of the ideal gain that CFG gives up at equal cost.
Moving to the right in a panel therefore means including lower noise levels, and the dotted line is the horizon used in the paper.

The figure shows three things.
First, the gap is not a product of our horizon.
It is smallest at $\sigma=10$, where the class is hardly decided yet and there is little gain to lose; it is clearly open at $\sigma=6$, and it grows as lower noise levels are included.
Second, the growth slows towards our horizon: most of the gap of the median class (black) is already there at $\sigma=4.5$, and the last steps to $\sigma=3$ add a few points.
Third, the dependence on the guidance scale is the same at every horizon: a stronger guidance gives up a larger share.
The classes (thin lines) differ in the size of their gap, and most of them follow this pattern.

The frontier below $\sigma=3$ needs new Langevin chains, which we ran for four classes: zebra, coral reef, hook and football helmet (in color, open markers).
Down to $\sigma=2$ the gap remains in all four, between $25\%$ and $65\%$, and the four classes keep the order they have at our horizon.
Its size still moves: by at most four points for zebra, up by as much as eleven points for coral reef, up for hook and down for football helmet.
Across the horizons we tested, the size of the gap changes and the conclusion does not.

\paragraph{An example with two scales.}
\Cref{fig:supp:toy-horizon} shows the regularizing effect of the horizon on an example in which every quantity can be computed down to the clean limit.
It is a Gaussian mixture with two scales: three classes that lie far apart, which stand for the meaning of the condition, each made of six narrow submodes, which stand for fine detail.
The classes are about a hundred times farther apart than the submodes.
The condition prefers one class, and inside every class it also prefers some of the submodes.
As the horizon is lowered, the KL divergence between the conditional and the unconditional laws rises in two steps: it first reaches the value that the classes alone would give and stays there over a wide range of horizons, and only at much lower noise do the submodes add their part.
At a horizon inside this range, guidance sees the classes only: the frontier is the one that the classes alone would give.
At the clean limit the frontier is a different curve, because there the submodes shape it as well.

The theory does not depend on the horizon: our results hold at every SNR $\Gamma$, so they apply at the horizon as they would at the clean limit.
The frontier itself is a statement at one horizon: the law that attains it at $\Gamma$ is not, in general, the law that attains it at another level, because tilting and smoothing do not commute (\cref{sec:supp:cfg-identities}).
The horizon applies only to the reported quantities: the displayed images are decoded from trajectories that run to the final noise level of the model (\cref{sec:supp:details-horizon}).
The truncation is a compromise that learned denoisers force on us in practice, not an assumption of the analysis.
The choice of the horizon remains an open problem, as Ohayon et al.\ also note for the IEM~\cite[Sec.~4]{ohayon2026distance}, and estimating these quantities reliably down to the clean limit is an interesting direction for future work.
\clearpage
\providecommand{\replayed}[1]{{\renewcommand{\vspace}[1]{}#1}}
\section{Derivations}
\label{sec:supp:theory}

This section proves the results of \cref{sec:cfg_info,sec:tradeoff,sec:diversity}, in three subsections that follow these three sections.
Each numbered result of the main text is restated with its number and then proved in one derivation; known theorems are stated and cited.
The few auxiliary results that do not appear in the main text are numbered with the letter of this section.
The proofs use \cref{asm:exact,asm:finite,asm:regular,asm:zero-snr,asm:girsanov}, which \cref{sec:supp:assumptions} states precisely.
Two distinctions hold throughout: the ideal tilt $p_{\gamma,s}$ is a density at one noise level, while the CFG sampler $Q_s$ is a law on trajectories; and each identity is for the exact posterior means of two fixed clean laws, which a learned denoiser only approximates.
\Cref{tab:supp:proof-map} lists where each result is proved, the known results its proof builds on, and where it is checked numerically on Gaussian mixtures (\cref{sec:supp:toy}).

\begin{table*}[t]
    \centering
    \footnotesize
    \setlength{\tabcolsep}{5pt}
    \begin{tabular}{@{}lllll@{}}
        \toprule
        Result & Content for guidance & Proof & Builds on & Numerical check \\
        \midrule
        \Cref{thm:renyi-immse} & R\'enyi I-MMSE at order $1+s$ & \cref{sec:supp:proof-renyi-immse} & identity of~\cite{shen2026renyi}, via \cref{eq:supp:heat,eq:snr-scores} & \cref{tab:supp:toy-closed}(a), \cref{fig:supp:toy-closed}(c) \\
        \Cref{prop:log-normalizer} & $F=sD_{1+s}=\tfrac{s(1+s)}2G$ & \cref{sec:supp:proof-log-normalizer} & definition of $D_{1+s}$, \cref{thm:renyi-immse} & \cref{tab:supp:toy-exact}(c), \cref{tab:supp:toy-trained} \\
        \Cref{thm:ideal-tilt} & The CFG tilt attains the gain--cost frontier & \cref{sec:supp:ideal-law} & Gibbs bound~\cite{csiszar1975divergence}, \cref{lem:supp:moments} & \cref{tab:supp:toy-closed}(b) \\
        \Cref{prop:frontier-shape} & Guidance scale as slope, curvature & \cref{sec:supp:frontier-shape} & exponential-family duality (\cref{lem:supp:moments}) & \cref{tab:supp:toy-closed}(b), \cref{fig:supp:toy-closed}(a) \\
        \Cref{prop:sampler} & Sampler gain and cost & \cref{sec:supp:pathspace} & Girsanov (\cref{lem:supp:girsanov}) & \cref{sec:supp:toy-exact} \\
        \Cref{thm:sampler-below} & Sampler gap: frontier, shortfall and slack & \cref{sec:supp:proof-sampler-below,sec:supp:slack-formula} & \cref{lem:supp:endpoint}, \cref{eq:supp:gibbs-identity}, KL chain rule & \cref{tab:supp:toy-exact}(b), \cref{fig:supp:toy-bridge} \\
        \Cref{eq:coverage-immse} & Baseline from the two denoisers & \cref{sec:supp:coverage-baseline} & mismatched I-MMSE \cref{eq:kl-immse} & \cref{tab:supp:toy-trained}, \cref{fig:supp:toy-trained-profile} \\
        \Cref{eq:coverage-alignment-cost} & Coverage loss is gain plus cost & \cref{sec:supp:coverage-frontier} & \cref{lem:supp:moments}, \cref{eq:alignment-cost} & \cref{fig:supp:toy-closed}(b) \\
        \Cref{eq:sampler-coverage} & Coverage of the sampler & \cref{sec:supp:coverage-sampler} & \cref{eq:supp:path-kl}, \cref{prop:sampler} & \cref{fig:supp:toy-levels} \\
        \bottomrule
    \end{tabular}
    \caption{\textbf{Where each result of the main text is proved,} the known results its proof builds on, and the numerical check of \cref{sec:supp:toy} that tests it.
    The first four results read known identities for guidance; the sampler results (\cref{prop:sampler,thm:sampler-below}) and the coverage results are specific to this paper.}
    \label{tab:supp:proof-map}
\end{table*}

\subsection{Preliminaries}
\label{sec:supp:common}

This subsection follows \cref{sec:cfg_info}.
It fixes the setting, recalls classifier-free guidance and the known identities between information and denoising errors, and states the assumptions.
It then proves \cref{thm:renyi-immse,prop:log-normalizer} and gives the estimators that the experiments use.

\subsubsection{Setting and Notation}
\label{sec:supp:setting}

\paragraph{Clean laws, channel and denoisers.}
Fix a condition $c$.
The clean laws $p^c$ and $p$ are densities on $\mathbb R^d$ with finite second moments, observed through the channel of \cref{eq:channel}, $Y_\gamma=\gamma X+\sqrt\gamma\,\varepsilon$.
Let $\varphi_\gamma$ be the density of $\mathcal N(0,\gamma I)$.
The observation densities are
\begin{align}
    p^c_\gamma(y)&=\int p^c(x)\,\varphi_\gamma(y-\gamma x)\,dx,\notag\\
    p_\gamma(y)&=\int p(x)\,\varphi_\gamma(y-\gamma x)\,dx,
    \label{eq:supp:observation-density}
\end{align}
and the posterior density of $X$ given $Y_\gamma=y$ is
\begin{equation}
    p(x\mid y)=\frac{p(x)\,\varphi_\gamma(y-\gamma x)}{p_\gamma(y)},
    \label{eq:supp:posterior}
\end{equation}
with $p^c(x\mid y)$ the same expression with $p^c$ and $p^c_\gamma$ in place of $p$ and $p_\gamma$.
The two denoisers are the posterior means,
\begin{align}
    \hat x_c(y,\gamma)&=\mathbb E[X\mid Y_\gamma=y,c]=\int x\,p^c(x\mid y)\,dx,\notag\\
    \hat x(y,\gamma)&=\mathbb E[X\mid Y_\gamma=y]=\int x\,p(x\mid y)\,dx,
    \label{eq:supp:denoisers}
\end{align}
and their difference $g=\hat x_c-\hat x$ is the CFG direction.
The alignment score is the log-likelihood ratio $T_\gamma=\log(p^c_\gamma/p_\gamma)$.
For a guidance scale $s\ge-1$ at which the normalizer is finite, the tilt and its normalizer are those of \cref{eq:cfg-tilt},
\begin{equation}
    p_{\gamma,s}=\frac{p^c_\gamma\,e^{sT_\gamma}}{Z_\gamma(s)},
    \qquad
    Z_\gamma(s)=\mathbb E_{p^c_\gamma}e^{sT_\gamma},
    \label{eq:supp:tilt}
\end{equation}
and $F(\gamma,s)=\log Z_\gamma(s)$ is the log-normalizer.
At $s=-1$ the normalizer is $Z_\gamma(-1)=\int p_\gamma\,dy=1$ and the tilt is $p_\gamma$.

\paragraph{Sigma and gamma.}
Diffusion models are written and trained with the noise level $\sigma$, as in \cref{eq:sigma-channel}.
Our identities are written with the SNR $\gamma$, as in \cref{eq:channel}.
The two describe the same observation, up to a known rescaling.
We use $\gamma$ only for convenience: the information identities are shorter in it, because the SNR is the variable in which information accumulates.
The dictionary below translates each object once, so that an equation in $\gamma$ can be evaluated with a model given in $\sigma$.

The two observations are related by
\begin{align}
    z&=\frac{y}{\gamma},
    \qquad
    \sigma=\gamma^{-1/2},\notag\\
    Z_\sigma&=\frac{Y_\gamma}{\gamma}=X+\sigma\varepsilon .
    \label{eq:supp:coordinates}
\end{align}
Let $q^c_\sigma$ and $q_\sigma$ be the densities of $Z_\sigma$ under $p^c$ and $p$.
The map $y=\gamma z$ is linear with Jacobian $\gamma^d$, so
\begin{equation}
    q^c_\sigma(z)=\gamma^d\,p^c_\gamma(\gamma z),
    \qquad
    q_\sigma(z)=\gamma^d\,p_\gamma(\gamma z).
    \label{eq:supp:jacobian}
\end{equation}
The Jacobians cancel in a ratio, so the alignment score is the same function in both coordinates, $\log(q^c_\sigma/q_\sigma)(z)=T_\gamma(\gamma z)$.
The denoisers estimate the same $X$ from the same observation, so
\begin{align}
    D_c(z,\sigma)&=\hat x_c(\gamma z,\gamma),\notag\\
    D_u(z,\sigma)&=\hat x(\gamma z,\gamma),\notag\\
    (D_c-D_u)(z,\sigma)&=g(\gamma z,\gamma).
    \label{eq:supp:denoiser-dictionary}
\end{align}

\emph{Divergences and the normalizer.}
Substituting $y=\gamma z$ and distributing the Jacobian $\gamma^d=(\gamma^d)^\alpha(\gamma^d)^{1-\alpha}$ over the two factors,
\begin{align}
    &\int(p^c_\gamma)^\alpha p_\gamma^{1-\alpha}\,dy\notag\\
    &\quad=\int p^c_\gamma(\gamma z)^\alpha\,p_\gamma(\gamma z)^{1-\alpha}\,\gamma^d\,dz\notag\\
    &\quad=\int\big(\gamma^dp^c_\gamma(\gamma z)\big)^\alpha\big(\gamma^dp_\gamma(\gamma z)\big)^{1-\alpha}dz\notag\\
    &\quad=\int(q^c_\sigma)^\alpha q_\sigma^{1-\alpha}\,dz .
    \label{eq:supp:substitution}
\end{align}
Taking $\frac1{\alpha-1}\log$ of both sides gives the invariance of the R\'enyi divergence (\cref{eq:renyi-definition}), and $\alpha=1+s$ gives the normalizer, since $p^c_\gamma e^{sT_\gamma}=(p^c_\gamma)^{1+s}p_\gamma^{-s}$:
\begin{align}
    D_\alpha(p^c_\gamma\|p_\gamma)&=D_\alpha(q^c_\sigma\|q_\sigma),\notag\\
    Z_\gamma(s)&=\int(q^c_\sigma)^{1+s}q_\sigma^{-s}\,dz .
    \label{eq:supp:invariance}
\end{align}
The same substitution gives $\mathrm{KL}(p^c_\gamma\|p_\gamma)=\mathrm{KL}(q^c_\sigma\|q_\sigma)$.
The tilt in the coordinate $z$ is therefore
\begin{equation}
    q_{\sigma,s}(z)=\gamma^d\,p_{\gamma,s}(\gamma z)
    =\frac{q^c_\sigma(z)^{1+s}\,q_\sigma(z)^{-s}}{Z_\gamma(s)},
    \label{eq:supp:sigma-tilt}
\end{equation}
with the same normalizer, and an average under the tilt is the same number in both coordinates:
\begin{equation}
    \mathbb E_{p_{\gamma,s}}\|g\|^2=\mathbb E_{q_{\sigma,s}}\|D_c-D_u\|^2 .
    \label{eq:supp:average-dictionary}
\end{equation}

\emph{Scores.}
A score in $z$ is $\gamma=\sigma^{-2}$ times the score in $y$.
The scores are derived in $\gamma$ form in \cref{sec:supp:cfg-identities}, and their $\sigma$ forms are given there, in \cref{eq:sigma-scores}.

\emph{Integrals over the noise level.}
Since $\gamma=\sigma^{-2}$, we have $d\gamma=-2\sigma^{-3}\,d\sigma$, and zero SNR is $\sigma=\infty$.
An integral over the SNR up to a horizon $\Gamma$ is therefore
\begin{equation}
    \int_0^\Gamma f(\gamma)\,d\gamma
    =\int_{\Gamma^{-1/2}}^{\infty}f\big(\sigma^{-2}\big)\,\frac{2}{\sigma^{3}}\,d\sigma .
    \label{eq:supp:dgamma}
\end{equation}

\emph{From noise prediction to denoisers.}
A model need not output a denoiser.
Let a model observe $U_t=a_tX+b_t\varepsilon$, for any schedule with $a_t,b_t>0$, and predict the noise by $\varepsilon_\theta$.
Dividing by $a_t$ gives $U_t/a_t=X+(b_t/a_t)\,\varepsilon$, which is the observation $z$ of \cref{eq:supp:coordinates} at $\sigma=b_t/a_t$.
Solving $U_t=a_tX+b_t\varepsilon$ for $X$ with the predicted noise gives the two heads
\begin{align}
    D_c&=\frac{U_t-b_t\,\varepsilon_\theta(U_t,c)}{a_t},\notag\\
    D_u&=\frac{U_t-b_t\,\varepsilon_\theta(U_t,\varnothing)}{a_t}.
    \label{eq:supp:heads}
\end{align}
The same conversion applied to the CFG combination $\tilde\varepsilon_\theta$ of \cref{eq:cfg-noise} gives the CFG denoiser.
The two weights sum to one, $(1+s)-s=1$, so the term $U_t$ can be split between the two heads:
\begin{align}
    D_s
    &=\frac{U_t-b_t\big[(1+s)\varepsilon_\theta(U_t,c)-s\,\varepsilon_\theta(U_t,\varnothing)\big]}{a_t}\notag\\
    &=(1+s)\,\frac{U_t-b_t\varepsilon_\theta(U_t,c)}{a_t}
      -s\,\frac{U_t-b_t\varepsilon_\theta(U_t,\varnothing)}{a_t}\notag\\
    &=(1+s)D_c-sD_u\notag\\
    &=D_c+s\,(D_c-D_u).
    \label{eq:cfg-denoiser}
\end{align}
This step is algebra and holds for any two heads, exact or learned.
A flow model with velocity target $\varepsilon-X$ on $z_t=(1-t)X+t\varepsilon$ converts in the same way, with $\sigma=t/(1-t)$, observation $z_t/(1-t)$ and $D=z_t-t\,v_\theta(z_t,t)$.
In $\gamma$ form, \cref{eq:cfg-denoiser} is $\hat x_s=\hat x_c+s\,g$.

From here on, everything is written in $\gamma$; \cref{eq:supp:coordinates,eq:supp:invariance,eq:sigma-scores,eq:supp:dgamma} translate any line back to $\sigma$.

\subsubsection{Classifier-Free Guidance}
\label{sec:supp:cfg-identities}

\paragraph{Classifier guidance.}
Guidance began with an explicit classifier.
Dhariwal and Nichol~\cite{dhariwal2021diffusion} train a classifier $p_\phi(c\mid Y_\gamma=y)$ on noised samples and add its gradient, scaled by a weight $w$, to the score of the diffusion model:
\begin{equation}
    \nabla_y\log p_\gamma(y)+w\,\nabla_y\log p_\phi(c\mid Y_\gamma=y).
    \label{eq:supp:classifier-guidance}
\end{equation}
This is the score of the density proportional to $p_\gamma(y)\,p_\phi(c\mid Y_\gamma=y)^{w}$.
If the classifier is exact, Bayes' rule gives, as a function of $y$,
\begin{equation}
    p(c\mid Y_\gamma=y)\propto\frac{p^c_\gamma(y)}{p_\gamma(y)},
    \label{eq:supp:bayes}
\end{equation}
so $w=1$ gives the conditional density $p^c_\gamma$, and $w>1$ gives more weight to the points that the classifier assigns to $c$.

\paragraph{Classifier-free guidance.}
Ho and Salimans~\cite{ho2022classifierfree} remove the classifier.
The proportionality constant in \cref{eq:supp:bayes} does not depend on $y$, so it drops out of the gradient of $\log p(c\mid Y_\gamma=y)$:
\begin{align}
    \nabla_y\log p(c\mid Y_\gamma=y)
    &=\nabla_y\log p^c_\gamma(y)-\nabla_y\log p_\gamma(y)\notag\\
    &=\nabla_yT_\gamma(y).
    \label{eq:supp:implicit-classifier}
\end{align}
The classifier is therefore implicit in the two scores of a conditional model, and no separate network is needed.
In a trained model the two scores come from two heads, and nothing forces the unconditional law $p$ to be the mixture $\sum_{c}\pi_c\,p^{c}$ of the conditional laws.
The implicit classifier is then defined by the ratio $p^c_\gamma/p_\gamma$ itself: a log-likelihood ratio of the model's two laws, which is a normalized class probability only for a Bayes-consistent pair.
Substituting \cref{eq:supp:implicit-classifier} into \cref{eq:supp:classifier-guidance} and collecting the two scores,
\begin{align}
    &\nabla_y\log p_\gamma+w\,\big(\nabla_y\log p^c_\gamma-\nabla_y\log p_\gamma\big)\notag\\
    &\quad=w\,\nabla_y\log p^c_\gamma-(w-1)\,\nabla_y\log p_\gamma .
    \label{eq:supp:cg-to-cfg}
\end{align}
With $s=w-1$ this is the CFG score $v_s$ of \cref{eq:cfg-score}: the conditional score plus $s$ times the gradient of the implicit classifier.
Ho and Salimans write the rule for noise predictions, with weights $(1+s)$ and $-s$ on the conditional and unconditional predictions; this is \cref{eq:cfg-noise}, the convention of the main text.
Thus, $s=0$ is plain conditional sampling and $s=-1$ is unconditional sampling, while a weight $w$ on the classifier gradient added to the unconditional score, as in \cref{eq:supp:classifier-guidance}, corresponds to $s=w-1$.
Noise predictions are linear in scores, and the two weights sum to one, so \cref{eq:cfg-noise,eq:cfg-score,eq:cfg-denoiser} are the same rule in three parametrizations.

\paragraph{How the sampling step changes.}
A sampler uses the model at each noise level only through a score.
We compute the three scores from the denoisers, under \cref{asm:exact}.
The Gaussian kernel satisfies $\nabla_y\varphi_\gamma(y-\gamma x)=\big(x-\frac y\gamma\big)\varphi_\gamma(y-\gamma x)$.
Differentiating \cref{eq:supp:observation-density} under the integral and recognizing the posterior \cref{eq:supp:posterior},
\begin{align}
    \nabla_y\log p_\gamma(y)
    &=\frac{1}{p_\gamma(y)}\int p(x)\,\nabla_y\varphi_\gamma(y-\gamma x)\,dx\notag\\
    &=\int\frac{p(x)\,\varphi_\gamma(y-\gamma x)}{p_\gamma(y)}\Big(x-\frac y\gamma\Big)dx\notag\\
    &=\int p(x\mid y)\,x\,dx-\frac y\gamma\notag\\
    &=\hat x(y,\gamma)-\frac y\gamma .
    \label{eq:supp:tweedie}
\end{align}
This is Tweedie's formula~\cite{efron2011tweedie} in $\gamma$ form.
The same computation with $p^c$ gives the conditional score, and the term $y/\gamma$ cancels in the difference:
\begin{align}
    \nabla_y\log p^c_\gamma&=\hat x_c-\frac y\gamma,
    \qquad
    \nabla_y\log p_\gamma=\hat x-\frac y\gamma,\notag\\
    \nabla_yT_\gamma&=\nabla_y\log p^c_\gamma-\nabla_y\log p_\gamma=g .
    \label{eq:snr-scores}
\end{align}
The CFG direction is thus the gradient of the alignment score.
Now take the logarithm of the tilt \cref{eq:supp:tilt}, $\log p_{\gamma,s}=\log p^c_\gamma+sT_\gamma-F(\gamma,s)$.
The log-normalizer $F(\gamma,s)$ does not depend on $y$, so its gradient is zero and drops out:
\begin{align}
    \nabla_y\log p_{\gamma,s}
    &=\nabla_y\log p^c_\gamma+s\,\nabla_yT_\gamma-\nabla_yF(\gamma,s)\notag\\
    &=\nabla_y\log p^c_\gamma+s\,\nabla_yT_\gamma\notag\\
    &=\hat x_c+s\,g-\frac y\gamma .
    \label{eq:supp:tilt-score}
\end{align}
The second row is the CFG score \cref{eq:cfg-score}, so at each noise level the score that CFG follows is the score of the tilt.
Comparing the last row with the conditional score in \cref{eq:snr-scores}, the guided step is the conditional step with the denoiser $\hat x_c$ replaced by $\hat x_s=\hat x_c+s\,g$, and nothing else changes.
In particular, the sampler never needs the normalizer $Z_\gamma(s)$.
In SNR time this is the dynamics \cref{eq:cfg-sde}, whose drift is $\hat x_c+s\,g$ and which reduces to the conditional sampler at $s=0$.

\par\noindent\emph{In sigma coordinates.}
By \cref{eq:supp:jacobian} and the chain rule, a score in $z$ is $\gamma=\sigma^{-2}$ times the score in $y$:
\begin{equation}
    \nabla_z\log q_\sigma(z)=\gamma\,\nabla_y\log p_\gamma(y)\big|_{y=\gamma z},
    \label{eq:supp:score-chain}
\end{equation}
and the same holds for $q^c_\sigma$ and for the tilt $q_{\sigma,s}$ of \cref{eq:supp:sigma-tilt}.
Multiplying \cref{eq:snr-scores,eq:supp:tilt-score} by $\gamma$, setting $y=\gamma z$, and using \cref{eq:supp:denoiser-dictionary},
\begin{align}
    \nabla_z\log q^c_\sigma&=\frac{D_c-z}{\sigma^2},\notag\\
    \nabla_z\log\frac{q^c_\sigma}{q_\sigma}&=\frac{D_c-D_u}{\sigma^2},\notag\\
    \nabla_z\log q_{\sigma,s}&=\frac{D_s-z}{\sigma^2},
    \label{eq:sigma-scores}
\end{align}
where $D_s=D_c+s\,(D_c-D_u)$ is the CFG denoiser of \cref{eq:cfg-denoiser}.
The first row is Tweedie's formula as the main text states it~\cite{efron2011tweedie}.

\paragraph{The combined tilt.}
Since $p^c_\gamma e^{sT_\gamma}=(p^c_\gamma)^{1+s}p_\gamma^{-s}$, the tilt is the normalized geometric mixture of the two observation densities,
\begin{align}
    p_{\gamma,s}&=\frac{(p^c_\gamma)^{1+s}\,p_\gamma^{-s}}{Z_\gamma(s)},\notag\\
    p_{\gamma,0}&=p^c_\gamma,
    \qquad
    p_{\gamma,-1}=p_\gamma .
    \label{eq:supp:geometric}
\end{align}
For $s>0$ it moves past the conditional density, away from the unconditional one.
The tilt is defined separately at each noise level.
Tilting and Gaussian smoothing do not commute in general, so the family $\{p_{\gamma,s}\}_\gamma$ need not be the noised versions of any one clean law.
A sampler that follows the score \cref{eq:supp:tilt-score} across levels therefore need not have $p_{\gamma,s}$ as its marginal at level $\gamma$.
We keep the two apart: the tilt $p_{\gamma,s}$ is a density at one noise level, and the CFG sampler defines a law $Q_s$ on trajectories, which \cref{sec:supp:pathspace} treats.

\subsubsection{Information Measures}
\label{sec:supp:info-measures}

For densities $q$ and $p$ with the required integrals finite, the entropy, the cross-entropy and the KL divergence are
\begin{align}
    H(q)&=-\mathbb E_q\log q,
    \qquad
    \mathrm{CE}(q\|p)=-\mathbb E_q\log p,\notag\\
    \mathrm{KL}(q\|p)&=\mathbb E_q\log\frac qp=\mathrm{CE}(q\|p)-H(q).
    \label{eq:supp:info-notation}
\end{align}
The entropy measures the spread of $q$, and the KL divergence is the extra cost of describing samples of $q$ with $p$.
All logarithms are natural.
For two densities $P$ and $Q$ and an order $\alpha>0$ with $\alpha\ne1$, the R\'enyi divergence~\cite{vanerven2014renyi} is
\begin{equation}
    D_\alpha(P\|Q)
    =\frac{1}{\alpha-1}\log\int P(x)^\alpha Q(x)^{1-\alpha}\,dx,
    \label{eq:renyi-definition}
\end{equation}
and $D_1=\mathrm{KL}$.
The KL divergence is the limit of $D_\alpha$ as $\alpha\uparrow1$, always, and as $\alpha\downarrow1$ when $D_\beta(P\|Q)<\infty$ for some $\beta>1$.
The same definitions apply to laws without a Lebesgue density, such as the laws on trajectories of \cref{sec:supp:pathspace}.
Let $P$ and $Q$ be two laws with densities $p$ and $q$ with respect to a common dominating measure $\mu$.
Then
\begin{align}
    D_\alpha(P\|Q)
    &=\frac{1}{\alpha-1}\log\int p^\alpha q^{1-\alpha}\,d\mu,\notag\\
    \mathrm{KL}(P\|Q)
    &=\mathbb E_P\log\frac{dP}{dQ}
    \quad\text{if }P\ll Q,
    \label{eq:supp:renyi-general}
\end{align}
and $\mathrm{KL}(P\|Q)=+\infty$ if $P$ is not absolutely continuous with respect to $Q$.
The value does not depend on the choice of $\mu$, and \cref{eq:renyi-definition} is the case where $\mu$ is the Lebesgue measure.
The order sets how strongly the divergence weights the regions where the ratio $P/Q$ is large.
In this paper the order is $\alpha=1+s$, so each guidance scale $s>-1$ has its own divergence.

We use three properties, all from~\cite{vanerven2014renyi}; they hold in the generality of \cref{eq:supp:renyi-general}.
The divergence is nonnegative, $D_\alpha(P\|Q)\ge0$, with equality only when $P=Q$.
It is nondecreasing in the order $\alpha$.
It satisfies data processing: if the same Markov kernel $K$ is applied to both laws, then $D_\alpha(KP\|KQ)\le D_\alpha(P\|Q)$.
The map from a trajectory to its end point is such a kernel, so a divergence between two laws on trajectories is at least the divergence between their end-point laws; \cref{sec:supp:proof-sampler-below} uses this consequence.

\subsubsection{From Information--Estimation Identities to CFG}
\label{sec:supp:assumptions}

The information-theory literature gives bridges between quantities of information and denoising errors in a Gaussian channel.
On one side is a mutual information or a divergence; on the other is a squared error of a denoiser.
The second side is what a diffusion model computes.
CFG evaluates a specific pair of denoisers at every step, the conditional $\hat x_c$ and the unconditional $\hat x$, and each is the wrong denoiser for the law of the other.
The rest of this subsection develops these bridges for this pair.
\Cref{sec:supp:info-background} states the known identities.
\Cref{sec:supp:proof-renyi-immse,sec:supp:proof-log-normalizer} then extend them from the conditional law to the whole tilt family, which gives the log-normalizer $F$ from the disagreement $\|g\|^2$ of the two denoisers.
The two applications are built on $F$ and its derivatives in $s$: the perception--alignment tradeoff in \cref{sec:supp:tradeoff-theory} and the diversity index in \cref{sec:supp:diversity-theory}.

\paragraph{Assumptions.}
The main text summarizes the assumptions informally; the precise statements follow.
\begin{assumption}[Exact denoisers]
\label{asm:exact}
The two denoisers are the exact posterior means \cref{eq:supp:denoisers} of the same two clean laws, $p^c$ and $p$, at all noise levels.
\end{assumption}
\begin{assumption}[Finite quantities]
\label{asm:finite}
All normalizers, moments and divergences in a result are finite, at each $(\gamma,s)$ in the result, with $s\ge-1$ and $Z_\gamma(s)<\infty$.
For a derivative in $s$ at $s_0$, $\mathbb E_{p^c_\gamma}e^{sT_\gamma}<\infty$ for all $s$ in an open interval around $s_0$, which permits differentiation under the integral; a finite moment at one end point is not sufficient.
\end{assumption}
\begin{assumption}[Regularity]
\label{asm:regular}
In the heat-flow computation in the proof of \cref{thm:renyi-immse} and in the direct computation of the KL case that follows that proof (\cref{sec:supp:proof-renyi-immse}), we can differentiate under the integral sign, and the boundary terms of the integration by parts are zero.
The resulting derivative of the divergence is continuous in $\gamma$ on $(0,\Gamma]$, so that it can be integrated.
\end{assumption}
\begin{assumption}[No divergence at zero SNR]
\label{asm:zero-snr}
For every integral that starts at $\gamma=0$, $D_{1+s}(p^c_\gamma\|p_\gamma)\to0$ as $\gamma\downarrow0$.
\end{assumption}
\begin{assumption}[Girsanov conditions]
\label{asm:girsanov}
For each scale $s\ge-1$ in a result (the scale $s=-1$ is the unconditional sampler, with drift $\hat x$), \cref{eq:cfg-sde} has a unique law, and Novikov's condition holds for the drift differences on $[0,\Gamma]$ under the reference law of each comparison, so that Girsanov's theorem (\cref{lem:supp:girsanov}) applies.
The drift differences are also square integrable: $\mathbb E\int_0^\Gamma\|g\|^2\,d\gamma<\infty$ under each sampler law in a result.
\end{assumption}

\Cref{asm:zero-snr} is an assumption only for orders above one.
For the KL divergence it follows from the finite second moments.
Let $(X,X')$ be any coupling of $p^c$ and $p$.
Given $(X,X')$, the two observations are $\mathcal N(\gamma X,\gamma I)$ and $\mathcal N(\gamma X',\gamma I)$, whose KL divergence is $\|\gamma X-\gamma X'\|^2/(2\gamma)$, and the KL divergence is jointly convex, so
\begin{equation}
    \mathrm{KL}(p^c_\gamma\|p_\gamma)
    \le\frac\gamma2\,\mathbb E\|X-X'\|^2
    \xrightarrow[\gamma\downarrow0]{}0 .
    \label{eq:supp:zero-snr-kl}
\end{equation}
The R\'enyi divergence is nondecreasing in its order, so the same limit holds for every order $1+s\le1$.
For $s>0$ the limit is assumed.
Statements about the clean laws use the opposite limit, $\gamma\to\infty$.
This limit is taken in the coordinate $z$ of \cref{eq:supp:coordinates}, where $q^c_\sigma\to p^c$ and $q_\sigma\to p$ weakly as $\sigma\to0$.
The divergence is the same in both coordinates, $D_\alpha(p^c_\gamma\|p_\gamma)=D_\alpha(q^c_\sigma\|q_\sigma)$ (\cref{eq:supp:invariance}).
The observation $Z_\sigma$ is $X$ passed through the same Gaussian kernel under both laws, so data processing gives $D_\alpha(q^c_\sigma\|q_\sigma)\le D_\alpha(p^c\|p)$ for every $\sigma$.
The R\'enyi divergence is lower semicontinuous in the pair of laws under weak convergence~\cite{vanerven2014renyi}, so $\liminf_{\sigma\to0}D_\alpha(q^c_\sigma\|q_\sigma)\ge D_\alpha(p^c\|p)$.
Hence $D_\alpha(p^c_\gamma\|p_\gamma)\to D_\alpha(p^c\|p)$ as $\gamma\to\infty$, for every order.
The limit can be infinite; a result that needs a finite clean divergence assumes it through \cref{asm:finite}.

The KL case $s=0$ is derived directly where possible and otherwise as the limit $s\to0$.
With learned denoisers, \cref{asm:exact} becomes the assumption that both heads are posterior means of the same two laws at all levels.
It cannot be checked directly, and it is why the paper keeps measured path quantities and fixed-level estimates apart.

\subsubsection{I-MMSE Identities}
\label{sec:supp:info-background}

Three known identities concern the channel of \cref{eq:channel}.
They say the same thing with increasing generality: a quantity of information is an accumulated denoising error.
We state each one, say what it means, and say where it is used.
The mismatched and pointwise identities were stated for scalar inputs; we use them in vector form, as Ohayon et al.~\cite{ohayon2026distance} do.

\paragraph{I-MMSE.}
Guo, Shamai and Verd\'u~\cite{guo2005mutual} showed that the mutual information between the input and the output of the channel grows with the SNR at a rate equal to half the smallest error with which the input can be estimated from the output.
For a law with posterior mean $\hat x$,
\begin{equation}
    \frac{d}{d\gamma}I(X;Y_\gamma)=\frac12\,\mathbb E\|X-\hat x(Y_\gamma,\gamma)\|^2 .
    \label{eq:supp:immse}
\end{equation}
The right side is the minimum mean-square error (MMSE), the error of the posterior mean, and the posterior mean is what a diffusion model learns as its denoiser.
Thus, the left side is a quantity of information, and the right side is a quantity that a trained denoiser measures.
We do not use \cref{eq:supp:immse} directly.
It concerns one law and its own denoiser, and it is the template for the identities below, which have the same form for a pair of laws and their two denoisers.

\paragraph{Mismatched I-MMSE.}
Verd\'u~\cite{verdu2010mismatched} asked what is lost when the data follow a law $P$ but the denoiser is the posterior mean of another law $Q$.
Let $P_\gamma$ and $Q_\gamma$ be the two observation densities and $m_P$ and $m_Q$ the two posterior-mean denoisers, evaluated at $(Y_\gamma,\gamma)$.
The loss is a divergence:
\begin{align}
    &\frac{d}{d\gamma}\mathrm{KL}(P_\gamma\|Q_\gamma)\notag\\
    &\quad=\frac12\Big[\mathbb E_P\|X-m_Q\|^2-\mathbb E_P\|X-m_P\|^2\Big]\notag\\
    &\quad=\frac12\,\mathbb E_{P_\gamma}\|m_P-m_Q\|^2 .
    \label{eq:mismatched-immse}
\end{align}
The first row of the right side is the excess error of the wrong denoiser over the right one, with $X\sim P$.
The second row holds because the error $X-m_P$ of the posterior mean is orthogonal, under $P$, to every function of $Y_\gamma$, and $m_P-m_Q$ is such a function.
The two CFG denoisers are a mismatched pair.
With $P=p^c$, $Q=p$, $m_P=\hat x_c$, $m_Q=\hat x$ and $m_P-m_Q=g$,
\begin{equation}
    \frac{d}{d\gamma}\mathrm{KL}(p^c_\gamma\|p_\gamma)
    =\frac12\,\mathbb E_{p^c_\gamma}\|g\|^2 .
    \label{eq:kl-immse}
\end{equation}
In expectation under $p^c_\gamma$, $\|g\|^2$ is therefore the excess error of the unconditional denoiser, and half of this expectation, accumulated over the SNR, is the KL divergence between the two observation laws.
\Cref{eq:kl-immse} is used twice.
It is the order-one case, $s=0$, of \cref{thm:renyi-immse}.
And integrated from zero SNR to the horizon $\Gamma$ it gives the coverage baseline \cref{eq:coverage-immse} (\cref{sec:supp:coverage-baseline}).

\paragraph{Pointwise form.}
Both identities are averages of statements that hold for single noise paths.
Couple all noise levels by one Brownian motion, $Y_\gamma=\gamma X+W_\gamma$, with $X\sim P$.
Venkat and Weissman~\cite{venkat2012pointwise} showed that the log-likelihood ratio at the horizon differs from half the excess squared error accumulated along the path by an It\^o integral against the noise:
\begin{align}
    &\log\frac{P_\Gamma(Y_\Gamma)}{Q_\Gamma(Y_\Gamma)}
    =\int_0^\Gamma(m_P-m_Q)\cdot dW_\gamma\notag\\
    &\quad+\frac12\int_0^\Gamma\big[\|X-m_Q\|^2-\|X-m_P\|^2\big]\,d\gamma .
    \label{eq:supp:pointwise}
\end{align}
The It\^o integral has mean zero when $\mathbb E\int_0^\Gamma\|m_P-m_Q\|^2\,d\gamma<\infty$, and then the mean of \cref{eq:supp:pointwise} is the integral of \cref{eq:mismatched-immse}.
For the CFG pair this expectation is $2\,\mathrm{KL}(p^c_\Gamma\|p_\Gamma)$, by \cref{eq:kl-immse} integrated from zero SNR, and it is finite by \cref{eq:supp:zero-snr-kl}, which bounds the KL divergence at every $\gamma$.
The pointwise form is what makes the identities usable along trajectories.
For the CFG pair its left side is the alignment score $T_\Gamma(Y_\Gamma)$, and \cref{eq:supp:T-path} in \cref{sec:supp:proof-sampler-below} is this identity written along the path.

\paragraph{The information-estimation metric.}
\label{sec:supp:iem}
Ohayon et al.~\cite{ohayon2026distance} use these identities with trained diffusion denoisers.
From the denoiser of one law they build a distance between two clean points, which are noised with the same Brownian path.
Their two-law form is the mismatched identity \cref{eq:mismatched-immse} integrated over all SNRs, $\mathrm{KL}(p_1\|p_2)=\frac12\int_0^\infty\mathbb E_{p_{1,\gamma}}\|m_1-m_2\|^2\,d\gamma$, so two denoisers give a divergence between two laws with no density evaluated.

\paragraph{What CFG needs beyond these identities.}
The known identities stop at the conditional law: \cref{eq:kl-immse} averages $\|g\|^2$ under $p^c_\gamma$.
Guidance needs two more averages of the same disagreement.
Averaging $\|g\|^2$ under the tilt $p_{\gamma,s}$ gives the R\'enyi divergence of order $1+s$ (\cref{thm:renyi-immse}); the denoisers are still those of the two original laws, not of the tilt.
Averaging it along CFG trajectories gives the path quantity $I_s$ of \cref{eq:sampler-point}, which needs no exactness assumption (\cref{prop:sampler}).
\Cref{tab:supp:identities} places the identities side by side.

\begin{table}[t]
    \centering
    \footnotesize
    \setlength{\tabcolsep}{2pt}
    \begin{tabular}{@{}llll@{}}
        \toprule
        Identity & Information & Squared error & Averaged under \\
        \midrule
        I-MMSE~\cite{guo2005mutual} & $I(X;Y_\gamma)$ & $\|X-\hat x\|^2$ & data and noise \\
        Mismatched~\cite{verdu2010mismatched} & $\mathrm{KL}(P_\gamma\|Q_\gamma)$ & $\|m_P-m_Q\|^2$ & $P_\gamma$ \\
        IEM~\cite{ohayon2026distance} & $\mathrm{KL}(p_1\|p_2)$ & $\|m_1-m_2\|^2$ & noised $p_1$ \\
        R\'enyi I-MMSE & $D_{1+s}(p^c_\gamma\|p_\gamma)$ & $\|\hat x_c-\hat x\|^2$ & tilt $p_{\gamma,s}$ \\
        \bottomrule
    \end{tabular}
    \caption{\textbf{Information as accumulated denoising error.}
    In the I-MMSE, mismatched and R\'enyi rows, the rate in the SNR of the quantity in the second column is a multiple of the squared error in the third, averaged under the law in the fourth.
    The IEM row is the mismatched identity integrated over all SNRs: the quantity itself, not its rate, is half the accumulated error.
    The last row is \cref{thm:renyi-immse}: the two denoisers are those of the mismatched pair, and only the law of the average changes.}
    \label{tab:supp:identities}
\end{table}

\subsubsection{The R\'enyi I-MMSE Theorem}
\label{sec:supp:proof-renyi-immse}

\replayed{\thmRenyiIMMSE*}

The proof first sets up four things: the integral that defines the divergence, the reason to differentiate it in the noise level, the heat equation, and the scores.
The computation then takes one chain of equalities.

\begin{proof}[Proof of \cref{thm:renyi-immse}]
Fix a scale $s>-1$ with $s\neq0$, and write $\alpha=1+s$ for the order, so that $\alpha>0$ and $\alpha\neq1$.
The case $s=0$ is the known mismatched I-MMSE and is treated at the end of the proof.

\paragraph{Step 1: the objects $h_t$ and $H(t)$.}
We work in the coordinate $z$ of \cref{eq:supp:coordinates} (\cref{sec:supp:setting}), with the noise variance as the clock:
\begin{equation}
    z=\frac y\gamma=X+\sqrt t\,\varepsilon,
    \qquad
    t=\sigma^2=\frac1\gamma .
    \label{eq:supp:renyi-clock}
\end{equation}
Let $P_t$ be the density of $z$ when $X\sim p^c$, and $Q_t$ the density of $z$ when $X\sim p$.
These are the two observation densities of \cref{eq:supp:jacobian} in the clock $t=\sigma^2$,
\begin{equation}
    P_t=q^c_\sigma,
    \qquad
    Q_t=q_\sigma,
    \label{eq:supp:renyi-PQ}
\end{equation}
so that $P_t(z)=\gamma^d\,p^c_\gamma(\gamma z)$ and $Q_t(z)=\gamma^d\,p_\gamma(\gamma z)$.
Here $Q_t$ is a density at one noise level, not the law $Q_s$ of the sampler's trajectories, which this proof does not use.
Both are smooth and positive for $t>0$, so their logarithms and gradients are defined.
The R\'enyi divergence of order $\alpha$ (\cref{eq:renyi-definition}) is built from one integral.
We name its integrand and its value:
\begin{equation}
    h_t=P_t^{\alpha}\,Q_t^{1-\alpha},
    \qquad
    H(t)=\int h_t(z)\,dz .
    \label{eq:supp:renyi-hH}
\end{equation}
The function $h_t$ is the unnormalized geometric mean of the two densities, and $H(t)$ is its mass.
With them,
\begin{equation}
    D_\alpha(P_t\|Q_t)=\frac{1}{\alpha-1}\,\log H(t).
    \label{eq:supp:renyi-logH}
\end{equation}
The only requirement is $0<H(t)<\infty$.
For $\alpha<1$ this always holds, since H\"older's inequality gives $H(t)\le1$.
For $\alpha>1$ it is part of \cref{asm:finite}; it is the condition that the tilt exists.
Then $h_t/H(t)$ is a probability density, the normalized geometric mean.
Step 5 shows that it is the tilt $p_{\gamma,s}$, written in the coordinate $z$.

\paragraph{Step 2: why we differentiate in $t$.}
The left side of \cref{eq:renyi-immse} is the rate at which the divergence changes with the noise level.
By \cref{eq:supp:invariance} the divergence does not depend on the coordinate,
\begin{equation}
    D_\alpha(p^c_\gamma\|p_\gamma)=D_\alpha(P_t\|Q_t),
    \qquad t=\frac1\gamma,
    \label{eq:supp:renyi-same}
\end{equation}
so we can compute the rate in the clock $t$ and convert it to $\gamma$ with the chain rule at the end.
By \cref{eq:supp:renyi-logH}, the rate in $t$ is
\begin{equation}
    \frac{d}{dt}D_\alpha(P_t\|Q_t)=\frac{H'(t)}{(\alpha-1)\,H(t)} .
    \label{eq:supp:renyi-rate}
\end{equation}
The whole proof is therefore the computation of $H'(t)$.

\paragraph{Step 3: the heat equation.}
We use $t$ and not $\gamma$ because in this clock the two densities obey a simple evolution law.
In \cref{eq:supp:renyi-clock}, $z$ is the clean sample plus independent Gaussian noise of variance $t$.
The density of a sum of independent variables is the convolution of their densities, so
\begin{equation}
    P_t(z)=\int p^c(x)\,\varphi_t(z-x)\,dx,
    \label{eq:supp:renyi-conv}
\end{equation}
where $\varphi_t(u)=(2\pi t)^{-d/2}e^{-\|u\|^2/(2t)}$ is the density of $\mathcal N(0,tI)$, and $Q_t$ is the same with $p$ in place of $p^c$.
A direct differentiation of the Gaussian kernel gives
\[
    \partial_t\varphi_t
    =\Big(\frac{\|u\|^2}{2t^2}-\frac{d}{2t}\Big)\varphi_t
    =\frac12\,\Delta\varphi_t ,
\]
where $\Delta$ is the Laplacian in $u$.
In \cref{eq:supp:renyi-conv} only the kernel depends on $t$ and $z$.
Differentiating under the integral (\cref{asm:regular}) passes the equation of the kernel to both densities:
\begin{equation}
    \partial_tP_t=\tfrac12\Delta P_t,
    \qquad
    \partial_tQ_t=\tfrac12\Delta Q_t .
    \label{eq:supp:heat}
\end{equation}
This is the heat equation: adding Gaussian noise diffuses a density as heat diffuses in space.
It helps because it replaces a derivative in the noise level by a Laplacian in $z$, and integration by parts turns a Laplacian into squared gradients.
The squared gradient that remains at the end is the squared CFG direction.

\paragraph{Step 4: the scores.}
The gradients that appear are the scores of the two densities and of their geometric mean.
All gradients are in $z$:
\begin{gather}
    \mathbf s_P=\nabla\log P_t,
    \qquad
    \mathbf s_Q=\nabla\log Q_t,\notag\\
    \mathbf s_h=\nabla\log h_t=\alpha\,\mathbf s_P+(1-\alpha)\,\mathbf s_Q .
    \label{eq:supp:renyi-scores}
\end{gather}
They are bold so that they are not confused with the guidance scale $s$.
The last equality holds because $\log h_t=\alpha\log P_t+(1-\alpha)\log Q_t$.
Two facts about the scores are used.
First, the Laplacian of a density is written through its score.
Since $\nabla P_t=P_t\,\mathbf s_P$, the product rule gives
\begin{align}
    \Delta P_t
    &=\nabla\!\cdot\big(P_t\,\mathbf s_P\big)\notag\\
    &=\nabla P_t\cdot\mathbf s_P+P_t\,\nabla\!\cdot\mathbf s_P\notag\\
    &=P_t\big(\nabla\!\cdot\mathbf s_P+\|\mathbf s_P\|^2\big),
    \label{eq:supp:renyi-laplacian}
\end{align}
and likewise $\Delta Q_t=Q_t\big(\nabla\!\cdot\mathbf s_Q+\|\mathbf s_Q\|^2\big)$.
Second, the three squared norms combine into one squared difference:
\begin{align}
    &\alpha\|\mathbf s_P\|^2+(1-\alpha)\|\mathbf s_Q\|^2-\|\mathbf s_h\|^2\notag\\
    &\quad=\alpha(1-\alpha)\big(\|\mathbf s_P\|^2+\|\mathbf s_Q\|^2
      -2\,\mathbf s_P\!\cdot\mathbf s_Q\big)\notag\\
    &\quad=\alpha(1-\alpha)\,\|\mathbf s_P-\mathbf s_Q\|^2 .
    \label{eq:supp:renyi-square}
\end{align}
The first equality expands $\|\mathbf s_h\|^2=\alpha^2\|\mathbf s_P\|^2+(1-\alpha)^2\|\mathbf s_Q\|^2+2\alpha(1-\alpha)\,\mathbf s_P\!\cdot\mathbf s_Q$ and uses $\alpha-\alpha^2=(1-\alpha)-(1-\alpha)^2=\alpha(1-\alpha)$.

\paragraph{Step 5: the computation.}
\runin{(a) The derivative of $H$}
We differentiate $H(t)=\int h_t\,dz$ and simplify, one operation per row:
\begin{align}
    H'(t)
    &=\int\partial_th_t\,dz\notag\\
    &=\int h_t\Big[\alpha\,\frac{\partial_tP_t}{P_t}
      +(1-\alpha)\,\frac{\partial_tQ_t}{Q_t}\Big]dz\notag\\
    &=\frac12\int h_t\Big[\alpha\,\frac{\Delta P_t}{P_t}
      +(1-\alpha)\,\frac{\Delta Q_t}{Q_t}\Big]dz\notag\\
    &=\frac12\int h_t\big[\nabla\!\cdot\mathbf s_h
      +\alpha\|\mathbf s_P\|^2+(1-\alpha)\|\mathbf s_Q\|^2\big]dz\notag\\
    &=\frac12\int h_t\big[\alpha\|\mathbf s_P\|^2
      +(1-\alpha)\|\mathbf s_Q\|^2-\|\mathbf s_h\|^2\big]dz\notag\\
    &=\frac{\alpha(1-\alpha)}2\int h_t\,\|\mathbf s_P-\mathbf s_Q\|^2\,dz .
    \label{eq:supp:renyi-Hprime}
\end{align}
The rows use, in order, the following rules.
Row 1 differentiates under the integral (\cref{asm:regular}).
Row 2 is $\partial_th_t=h_t\,\partial_t\log h_t$ with $\log h_t=\alpha\log P_t+(1-\alpha)\log Q_t$.
Row 3 is the heat equation \cref{eq:supp:heat}.
Row 4 inserts \cref{eq:supp:renyi-laplacian} for both densities and collects the two divergences, $\alpha\,\nabla\!\cdot\mathbf s_P+(1-\alpha)\,\nabla\!\cdot\mathbf s_Q=\nabla\!\cdot\mathbf s_h$.
Row 5 integrates the divergence term by parts.
The boundary term at infinity is zero (\cref{asm:regular}), and $\nabla h_t=h_t\,\mathbf s_h$, so the divergence term becomes a squared norm:
\begin{align}
    \int h_t\,\nabla\!\cdot\mathbf s_h\,dz
    &=-\int\nabla h_t\cdot\mathbf s_h\,dz\notag\\
    &=-\int h_t\,\|\mathbf s_h\|^2\,dz .
    \label{eq:supp:renyi-parts}
\end{align}
Row 6 is the identity \cref{eq:supp:renyi-square}.

\runin{(b) The rate in the clock $t$}
We insert \cref{eq:supp:renyi-Hprime} into \cref{eq:supp:renyi-rate}.
The factor $(1-\alpha)/(\alpha-1)=-1$ fixes the sign for orders below and above one, and $h_t/H(t)$ is the normalized geometric mean of Step 1:
\begin{align}
    &\frac{d}{dt}D_\alpha(P_t\|Q_t)\notag\\
    &\quad=\frac{\alpha(1-\alpha)}{2(\alpha-1)}
      \int\frac{h_t}{H(t)}\,\|\mathbf s_P-\mathbf s_Q\|^2\,dz\notag\\
    &\quad=-\frac\alpha2\,\mathbb E_{h_t/H(t)}\|\mathbf s_P-\mathbf s_Q\|^2 .
    \label{eq:supp:renyi-heat}
\end{align}
This identity holds for any two densities smoothed by the same Gaussian noise; it is Theorem~3.1 of Shen et al.~\cite{shen2026renyi}, and we prove it here in our notation to keep the supplement self-contained.
It says that noise removes R\'enyi divergence at a rate given by the squared score difference, averaged under the normalized geometric mean.
Nothing so far used what the two clean laws are.

\runin{(c) Inserting the CFG laws}
We now set $\alpha=1+s$ and identify the two ingredients of \cref{eq:supp:renyi-heat} with the dictionary of \cref{sec:supp:setting}.
The first is the law of the average.
By \cref{eq:supp:renyi-PQ}, $h_t=(q^c_\sigma)^{1+s}q_\sigma^{-s}$, and its mass is $H(t)=Z_\gamma(s)$ by \cref{eq:supp:invariance}.
Hence $h_t/H(t)=q_{\sigma,s}$, the tilt in the coordinate $z$ (\cref{eq:supp:sigma-tilt}): the normalized geometric mean is the tilt.
The second ingredient is the score difference.
By the chain rule \cref{eq:supp:score-chain} and $\nabla_yT_\gamma=g$ (\cref{eq:snr-scores}, which uses \cref{asm:exact}), it is $\mathbf s_P-\mathbf s_Q=(D_c-D_u)/\sigma^2=\gamma\,(D_c-D_u)$, the second row of \cref{eq:sigma-scores}.
The average in \cref{eq:supp:renyi-heat} is therefore an average under the tilt, and \cref{eq:supp:average-dictionary} moves it to the coordinate $y$:
\begin{align}
    &\mathbb E_{h_t/H(t)}\|\mathbf s_P-\mathbf s_Q\|^2\notag\\
    &\quad=\mathbb E_{q_{\sigma,s}}\big\|\gamma\,(D_c-D_u)\big\|^2\notag\\
    &\quad=\gamma^2\,\mathbb E_{q_{\sigma,s}}\|D_c-D_u\|^2\notag\\
    &\quad=\gamma^2\,\mathbb E_{p_{\gamma,s}}\|g\|^2 .
    \label{eq:supp:renyi-average}
\end{align}

\runin{(d) Back to the SNR}
We combine \cref{eq:supp:renyi-same}, the chain rule with $dt/d\gamma=-\gamma^{-2}$, \cref{eq:supp:renyi-heat} and \cref{eq:supp:renyi-average}:
\begin{align*}
    &\frac{d}{d\gamma}D_{1+s}(p^c_\gamma\|p_\gamma)\\
    &\quad=\frac{dt}{d\gamma}\,\frac{d}{dt}D_{1+s}(P_t\|Q_t)\\
    &\quad=\big(-\gamma^{-2}\big)\Big(-\frac{1+s}2\Big)\,
      \gamma^2\,\mathbb E_{p_{\gamma,s}}\|g\|^2\\
    &\quad=\frac{1+s}2\,\mathbb E_{p_{\gamma,s}}\|g\|^2 .
\end{align*}
The two minus signs cancel, and the factor $\gamma^2$ of the score difference cancels the factor $\gamma^{-2}$ of the change of clock.
With $g=\hat x_c-\hat x$ this is \cref{eq:renyi-immse}.
For the integral form, integrate \cref{eq:renyi-immse} over $[\gamma_0,\Gamma]$ with $0<\gamma_0<\Gamma$:
\begin{align}
    D_{1+s}(p^c_\Gamma\|p_\Gamma)
    &=D_{1+s}(p^c_{\gamma_0}\|p_{\gamma_0})\notag\\
    &\quad+\frac{1+s}2\int_{\gamma_0}^\Gamma
      \mathbb E_{p_{\gamma,s}}\|g\|^2\,d\gamma .
    \label{eq:supp:finite-interval}
\end{align}
This step needs the divergence to be finite on the whole interval and its derivative to be integrable there.
For $\gamma'<\gamma$, the observation $Y_{\gamma'}/\gamma'$ is $Y_\gamma/\gamma$ plus independent Gaussian noise, under both laws, so $Y_{\gamma'}$ is a degraded version of $Y_\gamma$.
By data processing (\cref{sec:supp:info-measures}), $D_{1+s}(p^c_{\gamma'}\|p_{\gamma'})\le D_{1+s}(p^c_\gamma\|p_\gamma)$, so a divergence that is finite at $\Gamma$ is finite on $(0,\Gamma]$, and the tilt exists at every level of the integral.
The derivative \cref{eq:renyi-immse} is continuous on $(0,\Gamma]$ by \cref{asm:regular}, so the fundamental theorem of calculus gives \cref{eq:supp:finite-interval}.
Now let $\gamma_0\downarrow0$.
The first term on the right goes to zero by \cref{asm:zero-snr}.
The integrand is nonnegative, so by monotone convergence the integral increases to $G(\Gamma,s)$, which gives \cref{eq:renyi-integral}.

\paragraph{The case $s=0$.}
At $s=0$ the order is one, $D_1=\mathrm{KL}$, the tilt is $p^c_\gamma$ itself, and \cref{eq:renyi-immse} reads $\frac{d}{d\gamma}\mathrm{KL}(p^c_\gamma\|p_\gamma)=\frac12\mathbb E_{p^c_\gamma}\|g\|^2$.
This is the mismatched I-MMSE identity \cref{eq:mismatched-immse} for the pair $P=p^c$, $Q=p$, that is, \cref{eq:kl-immse}.
Its integral over $[\gamma_0,\Gamma]$ and the limit $\gamma_0\downarrow0$ are as above.
Here the limit at zero SNR needs no assumption, since \cref{eq:supp:zero-snr-kl} gives it from the finite second moments.
\end{proof}

\par\noindent\runin{Relation to the mismatched I-MMSE}
The mismatched identity \cref{eq:mismatched-immse} is formally the limit $\alpha\to1$ ($s\to0$) of \cref{eq:supp:renyi-heat}: the geometric mean $h_t/H(t)$ becomes $P_t$ itself, the factor $\alpha/2$ becomes $1/2$, and the average is under the true law.
The same three tools give it directly, and \cref{asm:regular} covers this computation as well.
With $\int\partial_tP_t\,dz=0$ in the first row, the heat equation \cref{eq:supp:heat} in the second, and integration by parts with $\nabla P_t=P_t\,\mathbf s_P$ and $\nabla(P_t/Q_t)=(P_t/Q_t)(\mathbf s_P-\mathbf s_Q)$ in the third (the two integrals give its two lines),
\begin{align*}
    &\frac{d}{dt}\mathrm{KL}(P_t\|Q_t)\\
    &\quad=\int\partial_tP_t\,\log\frac{P_t}{Q_t}\,dz
      -\int\frac{P_t}{Q_t}\,\partial_tQ_t\,dz\\
    &\quad=\frac12\int\Delta P_t\,\log\frac{P_t}{Q_t}\,dz
      -\frac12\int\frac{P_t}{Q_t}\,\Delta Q_t\,dz\\
    &\quad=-\frac12\int P_t\,\mathbf s_P\!\cdot(\mathbf s_P-\mathbf s_Q)\,dz\\
    &\qquad+\frac12\int P_t\,\mathbf s_Q\!\cdot(\mathbf s_P-\mathbf s_Q)\,dz\\
    &\quad=-\frac12\,\mathbb E_{P_t}\|\mathbf s_P-\mathbf s_Q\|^2 ,
\end{align*}
and parts (c) and (d) of Step 5 with $s=0$ convert this to \cref{eq:kl-immse}.
Thus \cref{thm:renyi-immse} keeps the two denoisers of the mismatched identity and changes only the law of the average, from $p^c_\gamma$ to the tilt $p_{\gamma,s}$.

\par\noindent\runin{Finite noise range}
A model covers a finite range of noise levels, so its denoisers give only the increment \cref{eq:supp:finite-interval} between the smallest SNR $\gamma_0$ of the model and $\Gamma$.
The boundary term $D_{1+s}(p^c_{\gamma_0}\|p_{\gamma_0})$ must be supplied separately; \cref{sec:supp:tradeoff-estimators} describes how.

\par\noindent\runin{Numerical check}
Block (a) of \cref{tab:supp:toy-closed} compares the two sides of the heat-flow identity \cref{eq:supp:renyi-heat} by quadrature, for orders below and above one.
Panel (c) of \cref{fig:supp:toy-closed} checks the integral form \cref{eq:renyi-integral} on a two-dimensional law.

\subsubsection{The Log-Normalizer}
\label{sec:supp:proof-log-normalizer}

\replayed{\propLogNormalizer*}

\begin{proof}[Proof of \cref{prop:log-normalizer}]
The tilt factor is a power of the likelihood ratio, $e^{sT_\gamma}=(p^c_\gamma/p_\gamma)^s$.
Inserting it into the normalizer of \cref{eq:cfg-tilt} and comparing with \cref{eq:renyi-definition} at order $1+s$,
\begin{align}
    Z_\gamma(s)
    &=\int p^c_\gamma\,e^{sT_\gamma}\,dy\notag\\
    &=\int p^c_\gamma\Big(\frac{p^c_\gamma}{p_\gamma}\Big)^s dy\notag\\
    &=\int(p^c_\gamma)^{1+s}\,p_\gamma^{-s}\,dy\notag\\
    &=e^{sD_{1+s}(p^c_\gamma\|p_\gamma)} .
    \label{eq:supp:Z-renyi}
\end{align}
The last row is the definition \cref{eq:renyi-definition}, which is stated for $s\neq0$.
Taking the logarithm gives the first equality of \cref{eq:log-normalizer}, $F(\gamma,s)=\log Z_\gamma(s)=sD_{1+s}(p^c_\gamma\|p_\gamma)$.
At $s=0$ the first row gives $Z_\gamma(0)=1$, so $F(\gamma,0)=0=0\cdot\mathrm{KL}(p^c_\gamma\|p_\gamma)$, where the KL divergence is finite by \cref{eq:supp:zero-snr-kl}.
The second equality holds under the assumptions of \cref{thm:renyi-immse}.
By data processing, $Z_{\gamma'}(s)<\infty$ for every $\gamma'\le\gamma$ (\cref{sec:supp:proof-renyi-immse}), so the tilt exists at every level below $\gamma$.
Replace the divergence by its integral form \cref{eq:renyi-integral} at the horizon $\gamma$:
\begin{align*}
    F(\gamma,s)
    &=s\,D_{1+s}(p^c_\gamma\|p_\gamma)\\
    &=s\cdot\frac{1+s}2\int_0^{\gamma}\mathbb E_{p_{\gamma',s}}\|g\|^2\,d\gamma'\\
    &=\frac{s(1+s)}2\,G(\gamma,s).
    \qedhere
\end{align*}
\end{proof}

The applications use $F$ through its derivatives in $s$, so we record them once here; \cref{sec:supp:ideal-law,sec:supp:frontier-shape,sec:supp:coverage-definition} cite the next lemma.

\begin{slemma}[Moments of the alignment score]
\label{lem:supp:moments}
Let \cref{asm:finite} hold.
Then $F(\gamma,\cdot)$ is infinitely differentiable on the interior of the set $\{s:Z_\gamma(s)<\infty\}$, and there
\begin{align}
    \partial_sF(\gamma,s)&=\mathbb E_{p_{\gamma,s}}T_\gamma,\notag\\
    \partial_s^2F(\gamma,s)&=\mathrm{Var}_{p_{\gamma,s}}(T_\gamma)\ge0,
    \label{eq:normalizer-derivatives}
\end{align}
and
\begin{equation}
    \partial_s^3F(\gamma,s)
    =\mathbb E_{p_{\gamma,s}}\big(T_\gamma-\mathbb E_{p_{\gamma,s}}T_\gamma\big)^3 .
    \label{eq:supp:third-derivative}
\end{equation}
Thus $F(\gamma,\cdot)$ is convex, and $\partial_s^2F(\gamma,\cdot)$ is locally Lipschitz on this interior.
Moreover $F(\gamma,0)=0$.
If $Z_\gamma(s_1)<\infty$ for some $s_1>0$, which is part of \cref{asm:finite}, then $s=0$ is an interior point and $\partial_sF(\gamma,0)=\mathrm{KL}(p^c_\gamma\|p_\gamma)$.
\end{slemma}
\begin{proof}
Let $s$ be an interior point and choose $\delta>0$ with $Z_\gamma(s\pm\delta)<\infty$.
For every $k\ge0$ there is a constant $C_k$ with $|T|^k\le C_k\,(e^{-\delta T}+e^{\delta T})$, so
\begin{equation}
    |T_\gamma|^k\,e^{sT_\gamma}
    \le C_k\big(e^{(s-\delta)T_\gamma}+e^{(s+\delta)T_\gamma}\big),
    \label{eq:supp:domination}
\end{equation}
and the right side is integrable under $p^c_\gamma$.
The same bound, with a smaller $\delta$, dominates the integrand in a neighborhood of $s$, so we can differentiate $Z_\gamma(s)=\int p^c_\gamma\,e^{sT_\gamma}\,dy$ under the integral at every order.
Each derivative brings down one factor $T_\gamma$, and dividing by $Z_\gamma(s)$ turns the integral into an average under the tilt \cref{eq:cfg-tilt}:
\begin{align*}
    \partial_sF
    &=\frac{\partial_sZ_\gamma(s)}{Z_\gamma(s)}\\
    &=\int\frac{p^c_\gamma\,e^{sT_\gamma}}{Z_\gamma(s)}\,T_\gamma\,dy
    =\mathbb E_{p_{\gamma,s}}T_\gamma,\\
    \partial_s^2F
    &=\frac{\partial_s^2Z_\gamma(s)}{Z_\gamma(s)}
    -\Big(\frac{\partial_sZ_\gamma(s)}{Z_\gamma(s)}\Big)^2\\
    &=\mathbb E_{p_{\gamma,s}}T_\gamma^2-\big(\mathbb E_{p_{\gamma,s}}T_\gamma\big)^2\\
    &=\mathrm{Var}_{p_{\gamma,s}}(T_\gamma)\ge0 .
\end{align*}
One more derivative of $\log Z_\gamma(s)$ gives, with $m_k=\mathbb E_{p_{\gamma,s}}T_\gamma^k=\partial_s^kZ_\gamma(s)/Z_\gamma(s)$,
\begin{align*}
    \partial_s^3F
    &=m_3-3\,m_2m_1+2\,m_1^3\\
    &=\mathbb E_{p_{\gamma,s}}(T_\gamma-m_1)^3 .
\end{align*}
It is finite and continuous in $s$, so $\partial_s^2F$ is locally Lipschitz.
At $s=0$ the tilt is $p^c_\gamma$, so $Z_\gamma(0)=\int p^c_\gamma\,dy=1$ and $F(\gamma,0)=0$.
Since $Z_\gamma(s)\le1$ for $-1\le s\le0$ by H\"older's inequality, $s=0$ is an interior point as soon as $Z_\gamma(s_1)<\infty$ for some $s_1>0$, and then $\partial_sF(\gamma,0)=\mathbb E_{p^c_\gamma}T_\gamma=\mathrm{KL}(p^c_\gamma\|p_\gamma)$.
\end{proof}

\par\noindent\runin{Numerical check}
Block (b) of \cref{tab:supp:toy-closed} computes the cost of the tilt from $s^2\partial_sD_{1+s}$, which uses the first equality of \cref{eq:log-normalizer} and \cref{lem:supp:moments}, and from the densities directly.
Block (c) of \cref{tab:supp:toy-exact} estimates $sD_{1+s}$ from paths, and \cref{tab:supp:toy-trained} repeats the estimate with trained denoisers.

\subsubsection{Estimators}
\label{sec:supp:tradeoff-estimators}

The experiments use two estimators: \cref{alg:langevin} for the ideal tilt and \cref{alg:sampler} for the CFG sampler.
Both are stated in SNR coordinates.
This part gives the theory behind each one, following the lines of the algorithm: what it targets, why it is correct, and what it assumes.
It proves nothing new, and the hyperparameters are in \cref{sec:supp:details}.

\begin{algorithm}[t]
\caption{Ideal frontier by fixed-level Langevin sampling}
\label{alg:langevin}
\begin{algorithmic}[1]
\Require positive SNR ladder to $\Gamma$, scale grid including $0$, step sizes $h_\gamma$, burn-in and recording lengths
\For{each scale $s$ and level $\gamma$}
    \State Initialize particles $y$ with data noised to level $\gamma$
    \For{each Langevin step}
        \State Evaluate $\hat x_c,\hat x$ at $(y,\gamma)$; $g\gets\hat x_c-\hat x$
        \State $v\gets\hat x_c+s\,g-y/\gamma$
        \State $y\gets y+h_\gamma v+\sqrt{2h_\gamma}\,\xi$, $\xi\sim\mathcal N(0,I)$
        \State After burn-in, record $\|g\|^2$
    \EndFor
    \State $H_{\gamma,s}\gets$ mean over particles and recorded steps
\EndFor
\State Integrate $H_{\gamma,s}$ over $\gamma$, including the initial interval, to estimate $G(\Gamma,s)$
\State $F(s)\gets s(1+s)G(\Gamma,s)/2$
\State Estimate $\partial_sF$ across the scale grid
\Ensure $(A(s),C(s))$ from \cref{eq:alignment-cost}
\end{algorithmic}
\end{algorithm}

\begin{algorithm}[t]
\caption{CFG trajectory gain, cost and slack bound}
\label{alg:sampler}
\begin{algorithmic}[1]
\Require SNR grid $0=\gamma_0<\cdots<\gamma_K=\Gamma$, scale grid including $0$, $M$ paths, initial drift limit; each $\xi$ is a new draw from $\mathcal N(0,I)$
\For{each scale $s$ and path $j$}
    \State $Y\gets0$; $I^{(j)}\gets0$
    \For{$k=0,\ldots,K-1$}
        \State $\Delta\gets\gamma_{k+1}-\gamma_k$
        \State $g\gets\hat x_c-\hat x$ at $(Y,\gamma_k)$; initial drift limit at $k=0$
        \State $I^{(j)}\gets I^{(j)}+\|g\|^2\Delta$
        \State $Y\gets Y+(\hat x_c+s\,g)\Delta+\sqrt\Delta\,\xi$
    \EndFor
    \For{$r=1,\ldots,n$} \Comment{optional, for the slack bound}
        \State $\tilde Y\gets Y$; $\tilde W^{(j,r)}\gets0$
        \For{$k=K-1,\ldots,0$}
            \State $\tilde Y\gets\frac{\gamma_k}{\gamma_{k+1}}\,\tilde Y+\sqrt{\gamma_k(1-\gamma_k/\gamma_{k+1})}\,\xi$
            \State $g\gets\hat x_c-\hat x$ at $(\tilde Y,\gamma_k)$, as above
            \State $\tilde W^{(j,r)}\gets\tilde W^{(j,r)}+\|g\|^2(\gamma_{k+1}-\gamma_k)$
        \EndFor
    \EndFor
    \State $L^{(j)}\gets\log\frac1n\sum_{r}e^{-\kappa\tilde W^{(j,r)}}$, $\kappa=\tfrac12s(1+s)$
\EndFor
\State $I_s\gets M^{-1}\sum_j I^{(j)}$ for each scale
\Ensure $(A_Q(s),C_Q(s))$ from \cref{eq:sampler-point}; with exact denoisers, the slack bound $-\kappa I_s-M^{-1}\sum_jL^{(j)}$ of \cref{eq:supp:slack-bound}
\end{algorithmic}
\end{algorithm}

\paragraph{\Cref{alg:langevin}: fixed-level Langevin sampling (the ideal tilt).}
\par\noindent\runin{Target}
The algorithm estimates the accumulated disagreement under the tilt, which gives the divergence by \cref{eq:renyi-integral}:
\begin{align}
    D_{1+s}(p^c_\Gamma\|p_\Gamma)
    &=\frac{1+s}2\,G(\Gamma,s),\notag\\
    G(\Gamma,s)
    &=\int_0^\Gamma\mathbb E_{p_{\gamma,s}}\|g\|^2\,d\gamma .
    \label{eq:supp:D-from-G}
\end{align}
The average at each level is under the tilt $p_{\gamma,s}$ of that level.
A CFG run does not supply these averages: its law at level $\gamma$ is the marginal of $Q_s$, which is in general not the tilt (the two coincide at $s=0$ and $s=-1$).
The outer loop therefore samples each pair of a scale and a level on its own.

\par\noindent\runin{Why Langevin dynamics}
Let $\pi$ be a positive smooth density on $\mathbb R^d$.
The Langevin diffusion of $\pi$, in a fictitious time $r$, is
\begin{equation}
    dz_r=\nabla\log\pi(z_r)\,dr+\sqrt2\,dB_r .
    \label{eq:supp:langevin-diffusion}
\end{equation}
Its density $\rho_r$ obeys the Fokker--Planck equation
\begin{align}
    \partial_r\rho_r
    &=-\nabla\!\cdot\big(\rho_r\,\nabla\log\pi\big)+\Delta\rho_r\notag\\
    &=\nabla\!\cdot\big(\nabla\rho_r-\rho_r\,\nabla\log\pi\big).
    \label{eq:supp:fokker-planck}
\end{align}
At $\rho_r=\pi$ the bracket is $\nabla\pi-\pi\,\nabla\log\pi=0$, so $\pi$ is an invariant density of \cref{eq:supp:langevin-diffusion}.
The diffusion uses $\pi$ only through its score.
This is what makes it fit the tilt: the score of the tilt is known from the two denoisers, $\nabla_y\log p_{\gamma,s}=\hat x_c+s\,g-y/\gamma$ (\cref{eq:supp:tilt-score}), while its normalizer $Z_\gamma(s)$ is not known and is not needed.

\par\noindent\runin{The update}
The two inner lines of \cref{alg:langevin} compute this score, $v=\hat x_c+s\,g-y/\gamma$, and take one Euler step of size $h_\gamma$ of \cref{eq:supp:langevin-diffusion} with $\pi=p_{\gamma,s}$.
Models are given in the coordinate $z=y/\gamma$, where the score of the tilt is $\nabla_z\log q_{\sigma,s}=(D_s-z)/\sigma^2=\gamma\,(\hat x_c+s\,g-z)$ by \cref{eq:sigma-scores}, with the denoisers evaluated at $(\gamma z,\gamma)$.
One Euler step of size $h$ of \cref{eq:supp:langevin-diffusion} with $\pi=q_{\sigma,s}$ is
\begin{equation}
    z\leftarrow z+h\,\gamma\,(\hat x_c+s\,g-z)+\sqrt{2h}\,\xi,
    \qquad\xi\sim\mathcal N(0,I).
    \label{eq:supp:langevin-step}
\end{equation}
The two forms are the same step: multiplying \cref{eq:supp:langevin-step} by $\gamma$ gives $y\leftarrow y+h_\gamma(\hat x_c+s\,g-y/\gamma)+\sqrt{2h_\gamma}\,\xi$ with $h_\gamma=\gamma^2h$, the update of the algorithm.

\par\noindent\runin{Initialization, burn-in and the average}
The particles start from data of the condition, noised to the level $\gamma$.
These are samples of $p^c_\gamma$, the tilt at $s=0$.
At $s=0$ the chains thus start from their target, and for $s\neq0$ the burn-in steps move them from $p^c_\gamma$ toward $p_{\gamma,s}$.
After burn-in, the mean $H_{\gamma,s}$ of $\|g\|^2$ over particles and recorded steps estimates $\mathbb E_{p_{\gamma,s}}\|g\|^2$.

\par\noindent\runin{What is guaranteed}
The claim is limited.
With exact denoisers (\cref{asm:exact}), the update is the Euler step of the diffusion $dz=\nabla_z\log q_{\sigma,s}\,dr+\sqrt2\,dB_r$, and this diffusion has the tilt $q_{\sigma,s}$ as an invariant density.
Two errors of the estimate are not controlled: the bias of the discretization at step $h$, and the incomplete mixing of a finite run.
Bounds on the first are known under conditions that we do not assume, a log-Sobolev inequality and a bounded Hessian of $\log q_{\sigma,s}$~\cite{vempala2019rapid}, and a smaller step does not reduce the second.
We therefore treat the convergence of the chains as an assumption of the estimator, and check it against quadrature in \cref{sec:supp:toy-exact}.
With learned denoisers the drift need not be the gradient of a log-density, and the chains target the stationary law of their own drift (\cref{sec:supp:toy-trained}).

\par\noindent\runin{Why not importance weights}
The alternative is to keep conditional samples and reweight them by $w=e^{sT_\gamma}$.
Since $\mathbb E_{p^c_\gamma}w^k=Z_\gamma(ks)=e^{F(\gamma,ks)}$, the relative variance of these weights, when $Z_\gamma(2s)$ is finite, is
\begin{equation}
    \frac{\mathrm{Var}_{p^c_\gamma}(w)}{(\mathbb E_{p^c_\gamma}w)^2}
    =\frac{Z_\gamma(2s)}{Z_\gamma(s)^2}-1
    =e^{F(\gamma,2s)-2F(\gamma,s)}-1 ;
    \label{eq:supp:is-variance}
\end{equation}
otherwise the variance is infinite.
The exponent is nonnegative by the convexity of $F$ with $F(\gamma,0)=0$ (\cref{lem:supp:moments}), and it grows quickly with $s$.
The Langevin averages need no weights.

\par\noindent\runin{The integral over noise levels}
The next line of the algorithm integrates $H_{\gamma,s}$ over the levels.
Models and noise grids are given in $\sigma$, so the integral is taken in $\sigma$.
Write $\sigma_*=\Gamma^{-1/2}$ for the horizon.
The divergence is the same in both coordinates (\cref{eq:supp:invariance}), the level average is the same in both coordinates (\cref{eq:supp:average-dictionary}), and \cref{eq:supp:dgamma} changes the variable of the integral, with the factor $\frac{1+s}2\cdot2=1+s$:
\begin{align}
    D_{1+s}(q^c_{\sigma_*}\|q_{\sigma_*})
    &=\frac{1+s}2\int_0^\Gamma\mathbb E_{p_{\gamma,s}}\|g\|^2\,d\gamma\notag\\
    &=(1+s)\int_{\sigma_*}^\infty
    \frac{\mathbb E_{q_{\sigma,s}}\|D_c-D_u\|^2}{\sigma^3}\,d\sigma .
    \label{eq:sigma-renyi-integral}
\end{align}
The level averages are computed on a ladder of noise levels, interpolated between the levels, and integrated by quadrature.
The quadrature error of a finite ladder is a further numerical error that the theory does not bound.
The part of the integral above the largest noise of the model is the initial interval of the algorithm, treated at the end of this part.

\par\noindent\runin{From $G$ to $F$}
No further sampling is needed.
By \cref{eq:log-normalizer}, the log-normalizer at the horizon is
\begin{equation}
    F(\Gamma,s)=s\,D_{1+s}(p^c_\Gamma\|p_\Gamma)=\frac{s(1+s)}2\,G(\Gamma,s),
    \label{eq:supp:F-from-G}
\end{equation}
evaluated at every scale $s$ of the grid on which the chains were run.

\par\noindent\runin{Derivatives of $F$ in $s$}
The last computation of the algorithm is the derivative of $F$ across the scale grid.
Differentiating \cref{eq:supp:F-from-G} by the product rule, with $\partial_s\big[s(1+s)\big]=1+2s$,
\begin{equation}
    \partial_sF(\Gamma,s)
    =\frac{1+2s}2\,G(\Gamma,s)+\frac{s(1+s)}2\,\partial_sG(\Gamma,s).
    \label{eq:supp:dsF-from-G}
\end{equation}
At $s=0$ the second term vanishes, so
\begin{equation}
    \partial_sF(\Gamma,0)=\frac12\,G(\Gamma,0),
    \label{eq:supp:dsF-zero}
\end{equation}
which needs no derivative in $s$; by \cref{lem:supp:moments} it is $\mathrm{KL}(p^c_\Gamma\|p_\Gamma)$.
For $s>0$ the derivative across scales is a finite difference on the scale grid: we take the slopes of the estimated $F$ between neighboring scales, which estimate the right-hand side of \cref{eq:supp:dsF-from-G} at the midpoints, and anchor them at \cref{eq:supp:dsF-zero}.
The second derivative $\partial_s^2F$ is the finite difference of these slopes.
Because $F$ is convex in $s$ (\cref{lem:supp:moments}), the slopes must be nondecreasing.
The reconstruction imposes this, by a monotone (isotonic) fit of the slopes, and takes $F$ as the integral of the fitted slopes, so that $F$ and $\partial_sF$ come from one convex function.
The shape of an estimated curve is therefore not independent evidence of convexity.
The output line of the algorithm turns $F$ and $\partial_sF$ into the gain and the cost of the tilt by \cref{eq:alignment-cost}, which \cref{sec:supp:ideal-law} derives; \cref{sec:supp:coverage-frontier} does the same for coverage.

\paragraph{\Cref{alg:sampler}: CFG trajectories (the sampler).}
\par\noindent\runin{Target}
The algorithm estimates the disagreement accumulated along the trajectories of the CFG sampler,
\begin{equation}
    I_s=\int_0^\Gamma\mathbb E_{Q_s}\|g\|^2\,d\gamma,
    \label{eq:supp:Is-target}
\end{equation}
where $Q_s$ is the law of the trajectories of the CFG dynamics \cref{eq:cfg-sde}.
The average is under the law that the sampler produces, not under the tilt, so here no separate chain is needed: the algorithm runs the sampler itself.

\par\noindent\runin{The update}
The inner loop simulates \cref{eq:cfg-sde} by the Euler--Maruyama method in SNR time.
On a grid $0=\gamma_0<\dots<\gamma_K=\Gamma$ with steps $\Delta_k=\gamma_{k+1}-\gamma_k$, starting from $Y=0$, one step is
\begin{equation}
    Y\leftarrow Y+(\hat x_c+s\,g)\,\Delta_k+\sqrt{\Delta_k}\,\xi,
    \qquad\xi\sim\mathcal N(0,I),
    \label{eq:supp:em-step}
\end{equation}
with the two denoisers evaluated at $(Y,\gamma_k)$.
The drift is the guided denoiser $\hat x_s=\hat x_c+s\,g$ of \cref{eq:cfg-denoiser}.
By the dictionary of \cref{sec:supp:setting}, dividing \cref{eq:supp:em-step} by $\gamma_{k+1}$ gives the same step for $z=Y/\gamma$ in $\sigma$ time, $z\leftarrow(1-\lambda)\,z+\lambda\,D_s+\sigma_{k+1}\sqrt\lambda\,\xi$ with $\lambda=1-\sigma_{k+1}^2/\sigma_k^2$, evaluated at $(z,\sigma_k)$.
This is the usual stochastic (ancestral) sampling step~\cite{ho2020denoising}: a draw of $Z_{\sigma_{k+1}}$ from its Gaussian posterior given $Z_{\sigma_k}=z$ and a clean sample equal to the guided denoiser $D_s$.
The law $Q_s$ is the law of the trajectories of the continuous dynamics, which the simulated trajectories approximate.

\par\noindent\runin{The estimator}
Before each step the algorithm adds $\|g\|^2\Delta_k$ to a running sum, so each trajectory carries a left-point sum of the integrand of \cref{eq:supp:Is-target}.
The estimate is the mean over $M$ independent trajectories,
\begin{equation}
\begin{aligned}
    \hat I_s&=\frac1M\sum_{j=1}^M\sum_{k}\big\|g^{(j)}_k\big\|^2\,\Delta_k,\\
    \Delta_k&=\gamma_{k+1}-\gamma_k,
\end{aligned}
\label{eq:supp:Is-hat}
\end{equation}
where $g^{(j)}_k$ is the denoiser difference at step $k$ of trajectory $j$.
This is a plain Monte-Carlo average.
It is unbiased for the discretized integral, that is, for the mean of the left-point sum under the law of the simulated trajectories.
It uses no importance weights and no density.
It costs nothing beyond sampling, since a guided run evaluates both heads at every step anyway.

\par\noindent\runin{The slack bound}
The optional loop of the algorithm estimates the slack of \cref{eq:slack-formula} from above, by the right side of \cref{eq:supp:slack-bound} (derived in \cref{sec:supp:slack-formula}),
\begin{equation}
    -\kappa\,I_s-\mathbb E\log\Big(\frac1n\sum_{r=1}^{n}e^{-\kappa\tilde W_r}\Big),
    \qquad\kappa=\tfrac12s(1+s),
    \label{eq:supp:slack-target}
\end{equation}
where $\tilde W_r$ is the disagreement $\|g\|^2_{\tilde Y^{(r)}}$ accumulated along the $r$-th of $n$ independent Brownian bridges $\tilde Y^{(1)},\dots,\tilde Y^{(n)}$ from $0$ to the end point $Y_\Gamma$ of one CFG trajectory, computed on the level grid.
No sampler is run for it.
A bridge is drawn backwards from the end point, level by level: given its state at $\gamma_{k+1}$, its state at $\gamma_k$ is Gaussian with mean $\frac{\gamma_k}{\gamma_{k+1}}\tilde Y_{\gamma_{k+1}}$ and covariance $\gamma_k(1-\gamma_k/\gamma_{k+1})I$.
In the noisy-image variable $z=Y/\gamma$ this is plain re-noising, one level at a time, from the horizon back to the largest noise level:
\begin{equation}
    \tilde z_{\sigma_k}=\tilde z_{\sigma_{k+1}}+\sqrt{\sigma_k^2-\sigma_{k+1}^2}\;\xi,
    \qquad\tilde z_{\sigma_K}=z_{\sigma_\Gamma},
    \label{eq:supp:renoise}
\end{equation}
with fresh noise $\xi\sim\mathcal N(0,I)$ at each step.
Each bridge is one such path, and the bound needs the sum $\tilde W_r$ along the whole path, so the noise of one bridge is carried across its levels and is not redrawn independently per level.
The sum $\tilde W_r$ is the left-point sum of \cref{eq:supp:Is-hat}, on the same grid and with the same weights $\Delta_k$ as $\hat I_s$.
For each end point the algorithm takes the logarithm of the mean of $e^{-\kappa\tilde W_r}$ over its $n$ bridges, and then the mean over the end points.
This is a plain Monte-Carlo average of independent terms, and it is unbiased for \cref{eq:supp:slack-target} at the given $n$; no correction for the number of bridges is needed, because the bound holds for every $n$.
The loop costs $n$ more passes of the two denoisers over the levels per end point, the same as $n$ more CFG runs.
Unlike $\hat I_s$, the bound needs exact denoisers, because it rests on the Brownian bridge of the conditional sampler (\cref{lem:supp:endpoint}).
Two samplers give a check: at $s=0$ and at $s=-1$ the factor $\kappa$ is zero, the bound is zero for every $n$, and the mean of $\tilde W_r$ must equal $I_s$, since the bridges of these two samplers are Brownian.

\par\noindent\runin{Why one number is enough}
All CFG samplers have the same noise, and their drifts differ by a multiple of the same field: $(\hat x_c+s\,g)-(\hat x_c+s'g)=(s-s')\,g$.
By Girsanov's theorem (\cref{lem:supp:girsanov}), the KL divergence between the path laws of two such diffusions is half the integrated squared difference of their drifts, averaged under the first law.
Every path divergence between two CFG samplers is therefore a multiple of $I_s$, and the same theorem gives the mean of the path alignment score $\Lambda$ from $I_s$ and $I_0$.
\Cref{sec:supp:pathspace} proves both statements, which are \cref{eq:sampler-point}, and the output line of the algorithm applies \cref{eq:sampler-point} to $\hat I_s$ and $\hat I_0$.
The gain needs $I_0$, so the $s=0$ run uses the same noise $\xi$ as the guided runs, which reduces the variance of the difference.

\par\noindent\runin{What is guaranteed}
Girsanov's theorem uses only the two drifts, not what $\hat x_c$ and $\hat x$ are.
The path identities \cref{eq:sampler-point} therefore need only \cref{asm:girsanov}, and not \cref{asm:exact}: with learned denoisers, $\hat I_s$ still estimates the gain and the cost of the sampler that is actually run.
This is the contrast with \cref{alg:langevin}, whose reading as a R\'enyi divergence needs exact denoisers and converged chains.
One error is not controlled.
The Euler--Maruyama step has a time-discretization error, in the law of the trajectories and in the left-point sum, and the theory, which is stated for the continuous dynamics, does not bound it.

\paragraph{The start at finite noise (both algorithms).}
Both targets are integrals from $\gamma=0$, while a trained model is available only for $\sigma\le\sigma_{\max}$, that is for $\gamma\ge\gamma_0=\sigma_{\max}^{-2}$.
As $\gamma\to0$ the observation carries no information, and the exact denoisers tend to the two prior means,
\begin{equation}
    \mu_c=\mathbb E[X\mid c],
    \qquad
    \mu=\mathbb E[X] .
    \label{eq:supp:prior-means}
\end{equation}
In practice $\mu_c$ and $\mu$ are the mean predictions of the two heads at $\sigma_{\max}$.
On $[0,\gamma_0]$ we extend both denoisers by these constants, $\hat x_c\equiv\mu_c$ and $\hat x\equiv\mu$, so that $g\equiv\mu_c-\mu$ there.
A constant integrand has the same average under every law, so the extension contributes
\begin{equation}
    \int_0^{\gamma_0}\|\mu_c-\mu\|^2\,d\gamma=\gamma_0\,\|\mu_c-\mu\|^2
    \label{eq:supp:prior-extension}
\end{equation}
to $G(\Gamma,s)$ for every $s$, and likewise to the path integral $I_s$ of the sampler.
In \cref{alg:langevin} this is the initial interval of the integration line.
In \cref{alg:sampler} it is the first step: the grid of the algorithm starts at zero, its first positive level is the smallest SNR of the model, and the initial drift limit is the pair $(\mu_c,\mu)$.
For the tilt, $\frac{1+s}2$ times \cref{eq:supp:prior-extension} stands in for the boundary term $D_{1+s}(p^c_{\gamma_0}\|p_{\gamma_0})$ of \cref{eq:supp:finite-interval}.
It is exact only if the denoiser difference is constant below $\gamma_0$; at the largest noise both observation laws are close to Gaussian, which makes a constant extension reasonable.

\par\noindent\runin{Numerical check}
Block (a) of \cref{tab:supp:toy-exact} compares the Langevin averages of \cref{eq:supp:langevin-step} under the tilt with quadrature, at two levels and two scales.
\Cref{sec:supp:toy-exact} also compares path quantities accumulated along simulated trajectories with their closed forms.

\subsection{Perception--Alignment Tradeoff}
\label{sec:supp:tradeoff-theory}

This subsection follows \cref{sec:tradeoff:theory}, in the same order: what is traded, which law wins the trade, the shape of the frontier, what the sampler does, and how far the sampler is from the frontier, which it cannot beat.

\subsubsection{Overview}
\label{sec:supp:tradeoff-definitions}

Blau and Michaeli~\cite{blau2018perception} showed that image restoration has a perception--distortion tradeoff: a more accurate restoration pushes the law of the outputs away from the law of natural images.
Guidance has a tradeoff of the same type.
More alignment with the condition $c$ moves the sampling law away from the conditional law, and the two denoisers alone set the frontier of this trade.
To state it, we fix a noise level $\gamma$ and a condition $c$, and ask two questions about a candidate sampling law $q$: how much more its observations agree with the condition, and how far it moved from usual conditional sampling.

\replayed{\defGainCost*}

\paragraph{Why the gain is a sensible measure of alignment.}
By Bayes' rule, the probability that a classifier gives to $c$ at the observation $y$ is proportional to the density ratio, $p(c\mid Y_\gamma=y)\propto p^c_\gamma(y)/p_\gamma(y)$.
The alignment score $T_\gamma(y)$ is therefore the log-probability $\log p(c\mid Y_\gamma=y)$ of this classifier, up to a constant that does not depend on $y$.
The classifier is the one implied by the two denoisers, and its gradient is the direction that CFG adds (\cref{eq:cfg-score}).
The constant cancels in the difference of the two means in \cref{eq:candidate-alignment-cost}:
\begin{equation}
    A[q]=\mathbb E_q\log p(c\mid Y_\gamma)
    -\mathbb E_{p^c_\gamma}\log p(c\mid Y_\gamma).
    \label{eq:supp:gain-bayes}
\end{equation}
Thus, the gain is the increase of the mean log-probability that the implicit classifier gives to $c$, relative to usual conditional sampling.
It needs no external classifier.
This reading does not require the model's two laws to be Bayes-consistent, $p=\sum_c\pi_c\,p^c$, and we do not assume it.
In general $T_\gamma$ is the log-likelihood ratio of the two laws, and the classifier it defines does not need to be normalized over the conditions.

\paragraph{Why the cost is a sensible measure of perception.}
The cost $C[q]=\mathrm{KL}(q\|p^c_\gamma)$ is a divergence from the conditional law, which is the law of the data for the condition $c$ at this noise level.
As in Blau and Michaeli~\cite{blau2018perception}, it is a property of the distribution $q$ and not of single samples.
A KL divergence is zero only for equal laws, so $C[q]=0$ only at $q=p^c_\gamma$, where $A[q]=0$ as well.
Zero cost therefore forces zero gain, and every gain has a price.
The rest of this subsection finds the smallest price (\cref{thm:ideal-tilt}) and how it grows (\cref{prop:frontier-shape}).

\subsubsection{The Ideal Tilt}
\label{sec:supp:ideal-law}

\replayed{\thmIdealTilt*}

\begin{proof}[Proof of \cref{thm:ideal-tilt}]
\emph{The identity behind the bound.}
Taking the logarithm of the definition of the tilt, \cref{eq:cfg-tilt}, with $F=\log Z_\gamma$,
\begin{equation}
    \log p_{\gamma,s}=\log p^c_\gamma+s\,T_\gamma-F(\gamma,s).
    \label{eq:supp:log-tilt}
\end{equation}
Let $q$ be a law with $C[q]<\infty$ and $\mathbb E_q|T_\gamma|<\infty$.
We insert \cref{eq:supp:log-tilt} into the KL divergence from $q$ to the tilt, and split the mean into its three terms:
\begin{align}
    &\mathrm{KL}(q\|p_{\gamma,s})
    =\mathbb E_q\big[\log q-\log p_{\gamma,s}\big]\notag\\
    &\quad=\mathbb E_q\big[\log q-\log p^c_\gamma-s\,T_\gamma+F(\gamma,s)\big]\notag\\
    &\quad=\mathrm{KL}(q\|p^c_\gamma)-s\,\mathbb E_qT_\gamma+F(\gamma,s).
    \label{eq:supp:gibbs-identity}
\end{align}
The two conditions on $q$ make each term on the last row finite.

\emph{The Gibbs variational bound.}
A KL divergence is nonnegative, and it is zero only for equal laws.
Applied to the left side of \cref{eq:supp:gibbs-identity}, this gives \cref{eq:gibbs-bound}, with equality if and only if $q=p_{\gamma,s}$.
Equivalently,
\begin{equation}
    F(\gamma,s)=\max_q\Big\{s\,\mathbb E_qT_\gamma-\mathrm{KL}(q\|p^c_\gamma)\Big\},
    \label{eq:supp:gibbs-formula}
\end{equation}
and the tilt is the only maximizer.
This is the Gibbs variational principle, also called the Donsker--Varadhan formula: the log-moment generating function $F(\gamma,s)=\log\mathbb E_{p^c_\gamma}e^{sT_\gamma}$ is the convex dual of the KL divergence from $p^c_\gamma$~\cite{csiszar1975divergence}, \cite[Sec.~4.9]{boucheron2013concentration}.
We use it as a known result.

\emph{The tilt is the best law at its budget.}
The tilt itself satisfies the two conditions on $q$, because under \cref{asm:finite} the score $T_\gamma$ has finite moments under $p_{\gamma,s}$.
At $q=p_{\gamma,s}$ the left side of \cref{eq:supp:gibbs-identity} is zero and the first term on its right is $C(s)$, so
\begin{equation}
    C(s)=s\,\mathbb E_{p_{\gamma,s}}T_\gamma-F(\gamma,s).
    \label{eq:supp:gibbs-equality}
\end{equation}
We subtract \cref{eq:supp:gibbs-equality} from \cref{eq:gibbs-bound}, so that $F$ cancels.
Then we add and subtract $\mathbb E_{p^c_\gamma}T_\gamma$ inside the bracket to obtain the two gains of \cref{eq:candidate-alignment-cost}:
\begin{align}
    C[q]-C(s)
    &\ge s\big(\mathbb E_qT_\gamma-\mathbb E_{p_{\gamma,s}}T_\gamma\big)\notag\\
    &=s\big(A[q]-A(s)\big).
    \label{eq:supp:budget-step}
\end{align}
Now let $q$ respect the budget $B=C(s)$, that is, $C[q]\le C(s)$, and let $\mathbb E_q|T_\gamma|<\infty$.
Then
\begin{equation}
    0\ \ge\ C[q]-C(s)\ \ge\ s\big(A[q]-A(s)\big),
    \label{eq:supp:budget-chain}
\end{equation}
and $s>0$ gives $A[q]\le A(s)$.
If $A[q]=A(s)$, both inequalities in \cref{eq:supp:budget-chain} are equalities; the second one is the equality case of \cref{eq:gibbs-bound}, which holds only for $q=p_{\gamma,s}$.

The supremum in \cref{eq:alignment-budget} runs over every law with $C[q]\le B$, also those with $\mathbb E_q|T_\gamma|=\infty$.
These laws do not change the result.
Write $T_\gamma^+=\max(T_\gamma,0)$.
The Gibbs bound~\cite{csiszar1975divergence} for the nonnegative function $s\,T_\gamma^+$, together with $e^{sT_\gamma^+}\le1+e^{sT_\gamma}$, gives
\begin{align}
    s\,\mathbb E_qT_\gamma^+
    &\le C[q]+\log\mathbb E_{p^c_\gamma}e^{sT_\gamma^+}\notag\\
    &\le C[q]+\log\big(1+Z_\gamma(s)\big)<\infty .
    \label{eq:supp:positive-part}
\end{align}
The positive part of $T_\gamma$ is thus integrable under every law of finite cost, and $\mathbb E_qT_\gamma$ is well defined in $[-\infty,\infty)$.
A law with $\mathbb E_qT_\gamma=-\infty$ has $A[q]=-\infty$, so it affects neither the supremum nor the uniqueness of the maximizer.
Hence
\begin{equation}
    A^\ast\big(C(s)\big)=\sup_{q:\,C[q]\le C(s)}A[q]=A(s),
    \label{eq:supp:frontier-attained}
\end{equation}
and the tilt is the only law that attains the supremum.

\emph{Gain and cost through $F$.}
It remains to show \cref{eq:alignment-cost}.
We start from the definitions in \cref{eq:candidate-alignment-cost} at $q=p_{\gamma,s}$.
By \cref{lem:supp:moments}, the mean score under the tilt is the first derivative of the log-normalizer, $\partial_sF(\gamma,s)=\mathbb E_{p_{\gamma,s}}T_\gamma$ (\cref{eq:normalizer-derivatives}).
At $s=0$ the tilt is the conditional law, $p_{\gamma,0}=p^c_\gamma$, so the same identity gives $\partial_sF(\gamma,0)=\mathbb E_{p^c_\gamma}T_\gamma$.
The lemma applies at $s=0$ because $0$ is an interior point of the set where $F(\gamma,\cdot)$ is finite: $Z_\gamma(s')\le1$ for $s'\in(-1,0)$ by H\"older's inequality, and $Z_\gamma$ is finite at the scale $s>0$ of the theorem, hence on $[0,s]$ by the convexity of $F(\gamma,\cdot)$.
Therefore
\begin{align}
    A(s)
    &=A[p_{\gamma,s}]\notag\\
    &=\mathbb E_{p_{\gamma,s}}T_\gamma-\mathbb E_{p^c_\gamma}T_\gamma\notag\\
    &=\partial_sF(\gamma,s)-\partial_sF(\gamma,0).
    \label{eq:supp:A-from-F}
\end{align}
The cost was computed in \cref{eq:supp:gibbs-equality}.
By \cref{eq:supp:log-tilt} it is the mean of $\log(p_{\gamma,s}/p^c_\gamma)=s\,T_\gamma-F(\gamma,s)$ under the tilt, and the mean of $T_\gamma$ is again $\partial_sF$:
\begin{align}
    C(s)
    &=\mathbb E_{p_{\gamma,s}}\big[s\,T_\gamma-F(\gamma,s)\big]\notag\\
    &=s\,\mathbb E_{p_{\gamma,s}}T_\gamma-F(\gamma,s)\notag\\
    &=s\,\partial_sF(\gamma,s)-F(\gamma,s).
    \label{eq:supp:C-from-F}
\end{align}
\Cref{eq:supp:A-from-F,eq:supp:C-from-F} are the two rows of \cref{eq:alignment-cost}.
\end{proof}

\paragraph{Reading the bound.}
Dividing \cref{eq:supp:gibbs-formula} by $s$ shows that the tilt maximizes $\mathbb E_qT_\gamma-\frac1s\mathrm{KL}(q\|p^c_\gamma)$, the mean alignment score minus a penalty for the distance from the conditional law.
This is the objective of KL-regularized fine-tuning~\cite{korbak2022rl}, with the implicit classifier as the reward and $1/s$ as the weight of the penalty.
The theorem places the tilt on the frontier at each budget $C(s)$ that a tilt attains; it does not say that a tilt with a finite scale exists for every budget $B$.

\paragraph{Gain and cost from the accumulated disagreement.}
\Cref{eq:alignment-cost} writes the gain and the cost through $F$, and \cref{eq:log-normalizer} writes $F$ at the horizon $\Gamma$ through the accumulated disagreement, $F(\Gamma,s)=\frac{s(1+s)}2G(\Gamma,s)$.
Together they give the gain and the cost in closed form through $G$.
We write $G=G(\Gamma,s)$ and $\partial_sG=\partial_sG(\Gamma,s)$.
The derivative of $F$ in $s$ is \cref{eq:supp:dsF-from-G},
\[
    \partial_sF(\Gamma,s)=\frac{1+2s}2\,G+\frac{s(1+s)}2\,\partial_sG,
\]
and at $s=0$ it is $\partial_sF(\Gamma,0)=\frac12G(\Gamma,0)$.
For the gain, we insert both values into the first row of \cref{eq:alignment-cost}:
\begin{align}
    A(s)
    &=\partial_sF(\Gamma,s)-\partial_sF(\Gamma,0)\notag\\
    &=\frac{1+2s}2\,G+\frac{s(1+s)}2\,\partial_sG
      -\frac12\,G(\Gamma,0).
    \label{eq:supp:A-from-G}
\end{align}
For the cost, we insert \cref{eq:supp:dsF-from-G,eq:log-normalizer} into the second row of \cref{eq:alignment-cost}, and collect the two terms with $G$, using $s(1+2s)-s(1+s)=s^2$:
\begin{align}
    C(s)
    &=s\,\partial_sF(\Gamma,s)-F(\Gamma,s)\notag\\
    &=\frac{s(1+2s)}2\,G+\frac{s^2(1+s)}2\,\partial_sG\notag\\
    &\quad-\frac{s(1+s)}2\,G\notag\\
    &=\frac{s^2}2\,G+\frac{s^2(1+s)}2\,\partial_sG.
    \label{eq:supp:C-from-G}
\end{align}
In summary, at the horizon $\Gamma$,
\begin{align}
    A(s)&=\frac{1+2s}2\,G(\Gamma,s)-\frac12\,G(\Gamma,0)\notag\\
        &\quad+\frac{s(1+s)}2\,\partial_sG(\Gamma,s),\notag\\
    C(s)&=\frac{s^2}2\,G(\Gamma,s)
         +\frac{s^2(1+s)}2\,\partial_sG(\Gamma,s).
    \label{eq:supp:AC-from-G}
\end{align}
The estimated ideal frontier does not evaluate $\partial_sG$ (\cref{alg:langevin}).
It forms $F(\Gamma,s)=\frac{s(1+s)}2G(\Gamma,s)$ on the scale grid, and takes $\partial_sF$ from finite-difference slopes of $F$ across the grid, with a monotone fit anchored at $\partial_sF(\Gamma,0)=\frac12G(\Gamma,0)$ (\cref{sec:supp:tradeoff-estimators}).
Gain and cost are then computed from \cref{eq:alignment-cost} with these estimates of $F$ and $\partial_sF$; \cref{eq:supp:AC-from-G} is the equivalent closed form in terms of $G$.

\par\noindent\runin{Numerical check}
Block (b) of \cref{tab:supp:toy-closed} computes the cost of the tilt by three routes and its gain by two: from the definitions in \cref{eq:candidate-alignment-cost}, and from the derivatives of $F$ in \cref{eq:alignment-cost}.

\subsubsection{The Shape of the Frontier}
\label{sec:supp:frontier-shape}

\replayed{\propFrontierShape*}

The gain and the cost of the tilt are built from $F$ and $\partial_sF$ by \cref{eq:alignment-cost}, so their derivatives in $s$ are multiples of the curvature $\chi_\gamma(s)=\partial_s^2F(\gamma,s)=\mathrm{Var}_{p_{\gamma,s}}(T_\gamma)$ of \cref{eq:normalizer-derivatives}.
In the cost, the product rule produces $\partial_sF$ twice with opposite signs, and the two terms cancel.

\begin{proof}[Proof of \cref{prop:frontier-shape}]
Write $\chi(s)=\chi_\gamma(s)$ and let primes denote derivatives in $s$ at the fixed level $\gamma$.
Under \cref{asm:finite}, $F(\gamma,\cdot)$ is twice differentiable on the interval (\cref{lem:supp:moments}).
Differentiating the two lines of \cref{eq:alignment-cost}, the second by the product rule,
\begin{align}
    A'(s)&=\partial_s^2F(\gamma,s)=\chi(s),\notag\\
    C'(s)&=\partial_sF(\gamma,s)+s\,\partial_s^2F(\gamma,s)-\partial_sF(\gamma,s)\notag\\
         &=s\,\chi(s).
    \label{eq:supp:frontier-derivatives}
\end{align}
Since $A'=\chi>0$ on the interval, $s\mapsto A(s)$ is strictly increasing and can be inverted, with $ds/dA=1/\chi(s)$.
The chain rule then gives the slope of the cost as a function of the gain,
\begin{equation}
    \frac{dC}{dA}=\frac{C'(s)}{A'(s)}=\frac{s\,\chi(s)}{\chi(s)}=s,
    \label{eq:supp:frontier-slope}
\end{equation}
and differentiating the slope once more in $A$,
\begin{equation}
    \frac{d^2C}{dA^2}=\frac{d}{dA}\,s=\frac{ds}{dA}=\frac1{A'(s)}=\frac1{\chi(s)}>0 .
    \label{eq:supp:frontier-curvature}
\end{equation}
This is \cref{eq:frontier-shape}.
The second derivative is positive, so $C$ is a strictly convex function of $A$.
For $s>0$ the cost is also strictly increasing in $s$, and the inverse slope $dA/dC=1/s$ decreases along the frontier, so $A$ is a concave function of $C$.
\end{proof}

\begin{scorollary}[The small-$s$ law]
\label{cor:supp:small-s}
Under the assumptions of \cref{prop:frontier-shape}, with $\chi_\gamma(0)>0$, as $s\downarrow0$,
\begin{equation}
\begin{aligned}
    C&=\frac{A^2}{2\chi_\gamma(0)}+O(A^3),\\
    A&=\sqrt{2\chi_\gamma(0)\,C}\,\big(1+O(s)\big).
\end{aligned}
\label{eq:supp:small-s}
\end{equation}
\end{scorollary}
\begin{proof}
By \cref{eq:alignment-cost}, $A(0)=0$ and $C(0)=-F(\gamma,0)=0$, because $Z_\gamma(0)=1$.
By \cref{lem:supp:moments}, $F(\gamma,\cdot)$ is infinitely differentiable on the interior of $\{s:Z_\gamma(s)<\infty\}$.
Under the assumptions of the corollary this interior contains $0$, so $\chi=\partial_s^2F$ is differentiable at $0$, and $\chi(u)=\chi(0)+O(u)$.
Integrating \cref{eq:supp:frontier-derivatives} from $0$ with this expansion,
\begin{align*}
    A(s)&=\int_0^s\chi(u)\,du=\chi(0)\,s+O(s^2),\\
    C(s)&=\int_0^su\,\chi(u)\,du=\frac{\chi(0)}2\,s^2+O(s^3).
\end{align*}
The first line gives $s=A/\chi(0)+O(A^2)$.
Substituting it into the second line gives the first part of \cref{eq:supp:small-s}.
For the second part, $2\chi(0)\,C=\chi(0)^2s^2\,(1+O(s))$, whose square root for $s\ge0$ is $\chi(0)\,s\,(1+O(s))=A\,(1+O(s))$.
\end{proof}
\Cref{eq:supp:small-s} is the second-order expansion of the frontier at the origin, where the slope is $0$ and the curvature is $1/\chi_\gamma(0)$.
The first nats of gain are the cheapest: the marginal price is $dC/dA=s\approx A/\chi_\gamma(0)$, so a larger $\chi_\gamma(0)=\mathrm{Var}_{p^c_\gamma}(T_\gamma)$ makes them cheaper.
Equivalently, $C(s)=F(\gamma,0)-F(\gamma,s)-(0-s)\,\partial_sF(\gamma,s)$ is the Bregman divergence of the convex function $F(\gamma,\cdot)$ from the scale $s$ to the scale $0$.

\par\noindent\runin{Numerical check}
The last column of block (b) of \cref{tab:supp:toy-closed} gives the slope $dC/dA$ by finite differences, which equals $s$ at each point.
Panel (a) of \cref{fig:supp:toy-closed} draws the tangents on a two-dimensional law, and panel (b) shows \cref{cor:supp:small-s}.
For an equal-covariance Gaussian pair the frontier has a closed form, the parabola of \cref{prop:supp:gaussian} in \cref{sec:supp:toy}; the shift pair of \cref{tab:supp:toy-laws} is an instance, and \cref{sec:supp:toy-exact} checks its parabolic frontier on simulated trajectories.

\subsubsection{Path Space}
\label{sec:supp:pathspace}

\par\noindent\runin{Why path space}
\Cref{thm:ideal-tilt} is about the exact tilt $p_{\gamma,s}$ at one noise level.
A sampler is a different object.
It carries one state across levels, so its law at level $\gamma$ is the marginal of its own trajectory.
Following the tilted score \cref{eq:supp:tilt-score} at every level would produce the tilts $p_{\gamma,s}$ only if they were the noising trajectory of one clean law.
In general they are not, because tilting and smoothing do not commute (\cref{sec:supp:cfg-identities}).
The end-point law $q_{\Gamma,s}$ of a CFG sampler therefore differs from $p_{\Gamma,s}$ in general.
The gain and cost of the sampler must be measured on the law that it does produce, which is the law $Q_s$ of its trajectories.

\par\noindent\runin{Definitions}
The main text defines the gain and cost of the sampler on path space and states their values.
\replayed{\propSampler*}

\par\noindent\runin{The dynamics}
The law $Q_s$ is the law on $[0,\Gamma]$ of the solution of \cref{eq:cfg-sde}, which we write with its drift $b_s$:
\begin{equation}
\begin{aligned}
    dY_\gamma&=b_s(Y_\gamma,\gamma)\,d\gamma+dB_\gamma,
    \qquad Y_0=0,\\
    b_s&=\hat x_c+s\,g .
\end{aligned}
\label{eq:supp:path-sde}
\end{equation}
Two scales give the two samplers that define $\Lambda$.
The scale $s=0$ gives the conditional sampler $Q_0$, with drift $b_0=\hat x_c$.
The scale $s=-1$ gives the unconditional sampler $Q_{-1}$, with drift $b_{-1}=\hat x_c-g=\hat x$.
All these diffusions start at $Y_0=0$ and have the same unit noise; they differ only in the drift, and every drift difference is a multiple of the same field $g$:
\begin{equation}
\begin{aligned}
    b_s-b_{s'}&=(s-s')\,g,\\
    b_s-b_0&=s\,g,\\
    b_0-b_{-1}&=g .
\end{aligned}
\label{eq:supp:drift-diff}
\end{equation}
In the $\sigma$ coordinates of \cref{sec:supp:setting}, on any interval of positive SNR, \cref{eq:supp:path-sde} is the usual reverse-time stochastic sampler driven by the guided score.

\par\noindent\runin{Girsanov's theorem}
Two diffusions with the same start and the same noise have a likelihood ratio on path space that depends only on the difference of their drifts.
This is Girsanov's theorem, which we state in the form used below and do not prove.
\begin{slemma}[Girsanov~{\cite[Ch.~8]{oksendal2003stochastic}}]
\label{lem:supp:girsanov}
For $i=1,2$, let $Q^{(i)}$ be the law on $[0,\Gamma]$ of
\[
    dY_\gamma=b_i(Y_\gamma,\gamma)\,d\gamma+dB_\gamma,
    \qquad Y_0=0,
\]
and let \cref{asm:girsanov} hold for this pair: each equation has a unique law, and Novikov's condition holds under the reference law $Q^{(2)}$,
\[
    \mathbb E_{Q^{(2)}}\exp\Big(\frac12\int_0^\Gamma\|b_1-b_2\|^2\,d\gamma\Big)<\infty .
\]
Let $B^{(1)}_\gamma=Y_\gamma-\int_0^\gamma b_1\,du$, which is a Brownian motion under $Q^{(1)}$.
Then $Q^{(1)}$ and $Q^{(2)}$ are mutually absolutely continuous, and $Q^{(1)}$-almost surely
\begin{align}
    \log\frac{dQ^{(1)}}{dQ^{(2)}}
    &=\int_0^\Gamma(b_1-b_2)\cdot dB^{(1)}_\gamma\notag\\
    &\quad+\frac12\int_0^\Gamma\|b_1-b_2\|^2\,d\gamma .
    \label{eq:supp:girsanov}
\end{align}
If also $\mathbb E_{Q^{(1)}}\int_0^\Gamma\|b_1-b_2\|^2\,d\gamma<\infty$, then
\begin{equation}
    \mathrm{KL}\big(Q^{(1)}\big\|Q^{(2)}\big)
    =\frac12\int_0^\Gamma\mathbb E_{Q^{(1)}}\|b_1-b_2\|^2\,d\gamma .
    \label{eq:supp:girsanov-kl}
\end{equation}
\end{slemma}
\noindent The textbook statement is written with the Brownian motion of the reference law, $B^{(2)}_\gamma=Y_\gamma-\int_0^\gamma b_2\,du$.
Under Novikov's condition, the density
\[
    \exp\Big(\int_0^\Gamma(b_1-b_2)\cdot dB^{(2)}_\gamma
    -\frac12\int_0^\Gamma\|b_1-b_2\|^2\,d\gamma\Big)
\]
with respect to $Q^{(2)}$ defines a law equivalent to $Q^{(2)}$, under which $Y$ solves the equation with drift $b_1$.
The uniqueness in law of \cref{asm:girsanov} identifies this law with $Q^{(1)}$.
Substituting $dB^{(2)}_\gamma=dB^{(1)}_\gamma+(b_1-b_2)\,d\gamma$ in the exponent gives \cref{eq:supp:girsanov}.
The KL formula needs in addition only the square integrability of the drift difference under $Q^{(1)}$, which is the second part of \cref{asm:girsanov}: the It\^o integral in \cref{eq:supp:girsanov} then has mean zero, and taking $\mathbb E_{Q^{(1)}}$ gives \cref{eq:supp:girsanov-kl}.

The lemma uses only the two drifts, not what $\hat x_c$ and $\hat x$ are.
The path identities below therefore hold for learned denoisers under \cref{asm:girsanov}.

\begin{proof}[Proof of \cref{prop:sampler}]
\emph{Cost.}
Apply \cref{eq:supp:girsanov-kl} to $Q^{(1)}=Q_s$ and $Q^{(2)}=Q_{s'}$.
By \cref{eq:supp:drift-diff} the drift difference is $(s-s')\,g$, and the constant $(s-s')^2$ leaves the integral:
\begin{align}
    \mathrm{KL}(Q_s\|Q_{s'})
    &=\frac12\int_0^\Gamma\mathbb E_{Q_s}\big\|(s-s')\,g\big\|^2\,d\gamma\notag\\
    &=\frac{(s-s')^2}2\int_0^\Gamma\mathbb E_{Q_s}\|g\|^2\,d\gamma\notag\\
    &=\frac{(s-s')^2}2\,I_s .
    \label{eq:supp:path-kl}
\end{align}
The last row is the definition of $I_s$.
The path cost is the case $s'=0$,
\begin{equation}
    C_Q(s)=\mathrm{KL}(Q_s\|Q_0)=\frac{s^2}2\,I_s .
    \label{eq:supp:path-cost}
\end{equation}

\emph{Gain.}
The path alignment score is \cref{eq:supp:girsanov} with $Q^{(1)}=Q_0$ and $Q^{(2)}=Q_{-1}$, whose drift difference is $g$ by \cref{eq:supp:drift-diff}.
The Brownian motion of the first law is the innovation of the conditional sampler, $dB^{(0)}_\gamma=dY_\gamma-\hat x_c\,d\gamma$, so
\begin{equation}
    \Lambda
    =\int_0^\Gamma g\cdot dB^{(0)}_\gamma+\frac12\int_0^\Gamma\|g\|^2\,d\gamma .
    \label{eq:supp:Lambda}
\end{equation}
This writes $\Lambda$ as a function of the trajectory $Y$.
The laws $Q_s$ and $Q_0$ have a likelihood ratio by \cref{lem:supp:girsanov}, so the same function is defined on the trajectories of $Q_s$, and we can average it there.
Under $Q_s$ the trajectory solves \cref{eq:supp:path-sde}, $dY_\gamma=b_s\,d\gamma+dB^{(s)}_\gamma$ with $B^{(s)}$ a Brownian motion, so the innovation of the conditional sampler is no longer pure noise:
\begin{align}
    dB^{(0)}_\gamma
    &=dY_\gamma-\hat x_c\,d\gamma\notag\\
    &=(b_s-\hat x_c)\,d\gamma+dB^{(s)}_\gamma\notag\\
    &=s\,g\,d\gamma+dB^{(s)}_\gamma .
    \label{eq:supp:innovation}
\end{align}
Substitute \cref{eq:supp:innovation} into \cref{eq:supp:Lambda} and collect the two integrals of $\|g\|^2$:
\begin{align}
    \Lambda
    &=\int_0^\Gamma g\cdot\big(s\,g\,d\gamma+dB^{(s)}_\gamma\big)\notag\\
    &\quad+\frac12\int_0^\Gamma\|g\|^2\,d\gamma\notag\\
    &=\int_0^\Gamma g\cdot dB^{(s)}_\gamma
    +s\int_0^\Gamma\|g\|^2\,d\gamma\notag\\
    &\quad+\frac12\int_0^\Gamma\|g\|^2\,d\gamma\notag\\
    &=\int_0^\Gamma g\cdot dB^{(s)}_\gamma
    +\frac{1+2s}2\int_0^\Gamma\|g\|^2\,d\gamma .
    \label{eq:supp:Lambda-under-s}
\end{align}
The coefficient $\frac{1+2s}2$ is the sum of $s$, from the CFG term inside the innovation, and $\frac12$, which was already in \cref{eq:supp:Lambda}.
Now take the mean under $Q_s$.
The It\^o integral has mean zero, because $B^{(s)}$ is a Brownian motion under $Q_s$ and $g$ is square integrable (\cref{asm:girsanov}), and the remaining integral is $I_s$:
\begin{align}
    \mathbb E_{Q_s}\Lambda
    &=\frac{1+2s}2\int_0^\Gamma\mathbb E_{Q_s}\|g\|^2\,d\gamma\notag\\
    &=\frac{1+2s}2\,I_s .
    \label{eq:supp:E-Lambda}
\end{align}
At $s=0$ this is $\mathbb E_{Q_0}\Lambda=\frac12I_0$.
Subtracting it gives the path gain,
\begin{align}
    A_Q(s)
    &=\mathbb E_{Q_s}\Lambda-\mathbb E_{Q_0}\Lambda\notag\\
    &=\frac{1+2s}2\,I_s-\frac12\,I_0 .
    \label{eq:supp:path-gain}
\end{align}
\Cref{eq:supp:path-cost,eq:supp:path-gain} are \cref{eq:sampler-point}.
\end{proof}

\par\noindent\runin{Computing $I_s$}
Both the gain and the cost of the sampler are functions of the single number $I_s$.
It is estimated by \cref{eq:supp:Is-hat} (\cref{alg:sampler}; theory in \cref{sec:supp:tradeoff-estimators}), from the record of $\|g\|^2$ along the run, and $A_Q$, $C_Q$ follow from \cref{eq:sampler-point}.

\par\noindent\runin{Numerical check}
\Cref{sec:supp:toy-exact} compares path quantities accumulated along simulated trajectories with their closed forms.
The comparison includes a pair with denoisers that are not gradients, for which \cref{eq:supp:path-kl} still holds.

\subsubsection{The Sampler and Its Gap to the Frontier}
\label{sec:supp:proof-sampler-below}

\replayed{\thmSamplerBelow*}

\noindent The proof rests on one idea: a trajectory is where it ends plus the route it took there.
The alignment score reads only where a trajectory ends, while the path cost also charges for the route.
The identity \cref{eq:path-endpoint-bridge} follows from this in three steps.
First, with exact denoisers the conditional and the unconditional sampler take the same routes to a given end point, so the path alignment score is the alignment score of the end point, and the path gain is the gain of the end-point law (\cref{lem:supp:endpoint}).
This is the only place where \cref{asm:exact} enters.
Second, the path cost is the cost of the end-point law plus the cost of the routes, which is the slack (\cref{eq:supp:cost-ineq}).
Third, the cost of the end-point law is the cost of the frontier at the same gain plus a shortfall (\cref{eq:supp:cost-split}).
The second and third steps add up to \cref{eq:path-endpoint-bridge}.
The statement defines the slack by a formula, $S_s=\kappa(\tilde I_s-I_s)$.
The proof defines it by what it is, the cost of the routes (\cref{eq:supp:slack}), and \cref{sec:supp:slack-formula} then derives the formula and the estimate of $\tilde I_s$.

\par\noindent\runin{End points and routes}
Let $Y_\gamma=\gamma X+W_\gamma$, $0\le\gamma\le\Gamma$, be a data path: $X$ is drawn from a clean law, and $W$ is a Brownian motion independent of $X$.
Every such path splits into a straight line to its end point and a remainder:
\begin{equation}
\begin{aligned}
    Y_\gamma&=\frac\gamma\Gamma\,Y_\Gamma+\beta_\gamma,\\
    \beta_\gamma&=Y_\gamma-\frac\gamma\Gamma\,Y_\Gamma
    =W_\gamma-\frac\gamma\Gamma\,W_\Gamma .
\end{aligned}
    \label{eq:supp:bridge-split}
\end{equation}
The remainder $\beta$ starts at $0$ and returns to $0$ at $\Gamma$.
It is a Brownian bridge, and we write
\[
    \mathsf{Br}_\Gamma=\text{the law of }\beta=\big(W_\gamma-\tfrac\gamma\Gamma\,W_\Gamma\big)_{0\le\gamma\le\Gamma}.
\]
The proof uses two properties of the bridge.
First, the bridge is pure noise: $\beta$ is built from $W$ alone, so its law $\mathsf{Br}_\Gamma$ is the same for every law of $X$.
Second, the bridge is independent of the end point.
Indeed, $\beta$ is uncorrelated with $W_\Gamma$, since $\mathrm{Cov}(\beta_\gamma,W_\Gamma)=\gamma-\frac\gamma\Gamma\,\Gamma=0$, and uncorrelated jointly Gaussian variables are independent.
As $X$ is independent of $W$, the bridge is independent of the pair $(X,W_\Gamma)$, and hence of $X$ and of the end point $Y_\Gamma=\Gamma X+W_\Gamma$.

A data path is thus made of two independent parts, an end point and a bridge, and the law of $X$ enters only the first.
The end point has the density $p^c_\Gamma$ if $X\sim p^c$ and the density $p_\Gamma$ if $X\sim p$ (\cref{eq:supp:observation-density}).
The bridge has the law $\mathsf{Br}_\Gamma$ in both cases:
\begin{equation}
\begin{aligned}
    X\sim p^c:&\quad Y_\Gamma\sim p^c_\Gamma,\quad\beta\sim\mathsf{Br}_\Gamma,\quad\text{independent},\\
    X\sim p:&\quad Y_\Gamma\sim p_\Gamma,\quad\beta\sim\mathsf{Br}_\Gamma,\quad\text{independent}.
\end{aligned}
    \label{eq:supp:bridge-product}
\end{equation}
The two rows differ only in the law of the end point.
The two data paths therefore differ in where they end, and not in how they get there: given the end point, the bridge that is added to the line has the same law in both.

We fix three words for the rest of the proof.
The \emph{bridge} of a path is its remainder $\beta$, which goes from $0$ to $0$.
A \emph{route} to an end point $y$ is the line to $y$ plus a bridge, $\gamma\mapsto\frac\gamma\Gamma\,y+\beta_\gamma$.
The \emph{reference routes} to $y$ are the routes with $\beta\sim\mathsf{Br}_\Gamma$; such a route is the Brownian bridge $\tilde Y$ from $0$ to the end point in the statement of \cref{thm:sampler-below}.

\paragraph{Step 1: the alignment score reads only the end point.}
The first step is a lemma, because later sections use it as well.

\begin{slemma}[The alignment score reads only the end point]
\label{lem:supp:endpoint}
Let \cref{asm:exact,asm:girsanov} hold, and let $W$ be a Brownian motion independent of $X$.
Then $Q_0$ is the law of the data path $Y_\gamma=\gamma X+W_\gamma$, $0\le\gamma\le\Gamma$, with $X\sim p^c$, and $Q_{-1}$ is the law of the same path with $X\sim p$.
Recall the two alignment scores: the path score $\Lambda=\log(dQ_0/dQ_{-1})$ of \cref{prop:sampler}, which is a function of a whole trajectory, and the score $T_\Gamma=\log(p^c_\Gamma/p_\Gamma)$ of \cref{sec:supp:setting}, which is a function of a point at the level $\Gamma$.
The first is the second, evaluated at the end point of the trajectory:
\begin{equation}
    \boxed{\Lambda=T_\Gamma(Y_\Gamma)}
    \qquad\text{almost surely.}
    \label{eq:supp:endpoint}
\end{equation}
Consequently, for every scale $s$, the path gain is the gain of the end-point law,
\begin{equation}
    \boxed{A_Q(s)=A[q_{\Gamma,s}] .}
    \label{eq:supp:gain-equal}
\end{equation}
\end{slemma}
\begin{proof}
The proof has three parts.
The first is technical: it shows that the two samplers produce data paths.
The second is the idea: given the end point, the two samplers take the same routes, so the routes cancel in the likelihood ratio.
The third reads the gain from the second.

\emph{Data paths solve the samplers' equations.}
Let $X\sim p^c$.
It is a classical fact of filtering that a data path solves the equation of the sampler with the exact denoiser~\cite[Prop.~1.1]{montanari2023sampling}: there is a Brownian motion $B^{(0)}$ with
\begin{equation}
    dY_\gamma=\hat x_c(Y_\gamma,\gamma)\,d\gamma+dB^{(0)}_\gamma,
    \qquad Y_0=0 .
    \label{eq:supp:data-innovation}
\end{equation}
The drift is the posterior mean of $X$ given the current point $Y_\gamma$, which is the exact denoiser of \cref{asm:exact}.
The fact needs only $\mathbb E\|X\|^2<\infty$ (\cref{sec:supp:setting}); it is the innovations theorem~\cite[Lemma 2.2]{fujisaki1972stochastic}, see also~\cite{liptser2001statistics}, applied to the observation $Y$ of the signal $X$.
\Cref{eq:supp:data-innovation} is \cref{eq:cfg-sde} at $s=0$, so by the uniqueness in \cref{asm:girsanov} the law of the data path is $Q_0$.
With $X\sim p$ the same fact gives the drift $\hat x=\hat x_c-g$, which is \cref{eq:cfg-sde} at $s=-1$, so the law is $Q_{-1}$.

\emph{The routes cancel.}
A path is its end point and its bridge, $Y\leftrightarrow(Y_\Gamma,\beta)$ (\cref{eq:supp:bridge-split}).
By the first part and \cref{eq:supp:bridge-product}, both path laws factor into an end-point density and the same bridge law:
\[
\begin{aligned}
    dQ_0(Y)&=p^c_\Gamma(Y_\Gamma)\,dY_\Gamma\;d\mathsf{Br}_\Gamma(\beta),\\
    dQ_{-1}(Y)&=p_\Gamma(Y_\Gamma)\,dY_\Gamma\;d\mathsf{Br}_\Gamma(\beta).
\end{aligned}
\]
In the ratio the routes cancel:
\begin{align}
    \frac{dQ_0}{dQ_{-1}}(Y)
    &=\frac{p^c_\Gamma(Y_\Gamma)\;d\mathsf{Br}_\Gamma(\beta)}
           {p_\Gamma(Y_\Gamma)\;d\mathsf{Br}_\Gamma(\beta)}\notag\\
    &=\frac{p^c_\Gamma(Y_\Gamma)}{p_\Gamma(Y_\Gamma)}
    =e^{T_\Gamma(Y_\Gamma)},
    \label{eq:supp:endpoint-ratio}
\end{align}
where $p_\Gamma>0$ everywhere.
Taking the logarithm gives \cref{eq:supp:endpoint}.
In words: the two samplers disagree about which end points are likely, but they take the same routes to a given end point.
A comparison of their paths therefore needs only the end point.

\emph{The gain.}
\Cref{eq:supp:endpoint} also holds almost surely under $Q_s$, because $Q_s$ and $Q_0$ are mutually absolutely continuous (\cref{lem:supp:girsanov}).
The path gain of \cref{eq:path-alignment-cost} is the difference of the means of $\Lambda$ under $Q_s$ and under $Q_0$.
A mean of a function of the end point is a mean over the law of the end point, which is $q_{\Gamma,s}$ under $Q_s$ and $p^c_\Gamma$ under $Q_0$.
The path gain is therefore $\mathbb E_{q_{\Gamma,s}}T_\Gamma-\mathbb E_{p^c_\Gamma}T_\Gamma$, which is the gain $A[q_{\Gamma,s}]$ of \cref{def:gain-cost}.
\end{proof}

\begin{proof}[Proof of \cref{eq:path-endpoint-bridge,eq:endpoint-cost-bound} in \cref{thm:sampler-below}]
\Cref{lem:supp:endpoint} settles the gain.
The two remaining steps are about the cost: the path cost is first reduced to the cost of the end-point law, and this cost is then compared with the frontier.

\paragraph{Step 2: path cost = unusual places + unusual routes.}
The path cost $C_Q(s)=\mathrm{KL}(Q_s\|Q_0)$ compares CFG paths with conditional paths.
As in Step 1, a path is its end point and its bridge, $Y\leftrightarrow(y,\beta)$ with $y=Y_\Gamma$, and both path laws factor:
\[
\begin{aligned}
    dQ_0(Y)&=p^c_\Gamma(y)\,dy\;d\mathsf{Br}_\Gamma(\beta),\\
    dQ_s(Y)&=q_{\Gamma,s}(y)\,dy\;dQ_s(\beta\mid y).
\end{aligned}
\]
The first row is Step 1: the conditional sampler takes the reference routes to every end point.
The second row only names the two factors: $q_{\Gamma,s}$ is the law of the end point of a CFG path, and $Q_s(\cdot\mid y)$ is the law of its bridge given the end point $y$, that is, the routes that CFG takes to $y$.
The ratio of the two rows has one factor for the end point and one for the route:
\[
    \frac{dQ_s}{dQ_0}(Y)
    =\frac{q_{\Gamma,s}(y)}{p^c_\Gamma(y)}\cdot
    \frac{dQ_s(\cdot\mid y)}{d\mathsf{Br}_\Gamma}(\beta).
\]
Both factors exist because $Q_s$ has a density with respect to $Q_0$ (\cref{lem:supp:girsanov}).
The path cost is the mean of the logarithm under $Q_s$, and the logarithm turns the product into a sum:
\begin{align*}
    C_Q(s)&=\mathbb E_{Q_s}\log\frac{dQ_s}{dQ_0}\\
    &=\mathbb E_{y\sim q_{\Gamma,s}}\log\frac{q_{\Gamma,s}(y)}{p^c_\Gamma(y)}\\
    &\quad+\mathbb E_{y\sim q_{\Gamma,s}}\,\mathbb E_{\beta\sim Q_s(\cdot\mid y)}\log\frac{dQ_s(\cdot\mid y)}{d\mathsf{Br}_\Gamma}(\beta)\\
    &=\mathrm{KL}\big(q_{\Gamma,s}\big\|p^c_\Gamma\big)\\
    &\quad+\mathbb E_{y\sim q_{\Gamma,s}}\mathrm{KL}\big(Q_s(\cdot\mid y)\big\|\mathsf{Br}_\Gamma\big).
\end{align*}
This is the chain rule of the KL divergence~\cite[Thm.~2.5.3]{cover2006elements}.
The first term is the cost $C[q_{\Gamma,s}]$ of \cref{def:gain-cost}: the cost of ending in unusual places.
The second term is the cost of getting there by unusual routes, and it is the path slack:
\begin{equation}
    S_s=\mathbb E_{y\sim q_{\Gamma,s}}\mathrm{KL}\big(Q_s(\cdot\mid y)\big\|\mathsf{Br}_\Gamma\big)\ \ge0 .
    \label{eq:supp:slack}
\end{equation}
It compares the routes of CFG to $y$ with the reference routes to $y$, on average over the end points that CFG produces.
With these two names,
\begin{equation}
    \boxed{C_Q(s)=C[q_{\Gamma,s}]+S_s .}
    \label{eq:supp:cost-ineq}
\end{equation}
The frontier is about the first term only, and the slack cannot be seen in the generated sample.
Since $S_s\ge0$, the path cost can only overstate the cost of the samples; this is the data-processing inequality of \cref{sec:supp:info-measures} with its dropped term written out.

\paragraph{Step 3: end-point cost = frontier cost + shortfall.}
Steps 1 and 2 express the gain and the cost of the sampler through its end-point law $q_{\Gamma,s}$.
The frontier is traced by the tilts (\cref{thm:ideal-tilt}).
The tilt at a scale $s'$, its gain and its cost are
\[
\begin{gathered}
    p_{\Gamma,s'}=p^c_\Gamma\,e^{s'T_\Gamma-F(\Gamma,s')},\\
    A(s')=A[p_{\Gamma,s'}],\qquad C(s')=C[p_{\Gamma,s'}],
\end{gathered}
\]
where $F(\Gamma,s')=\log\mathbb E_{p^c_\Gamma}e^{s'T_\Gamma}$ is the log-normalizer.
We compare the sampler with the tilt that has the same gain.

\emph{Same gain.}
The scale $s^\ast$ is defined by $A(s^\ast)=A_Q(s)$: it is the scale at which the tilt reaches the same gain as the sampler at its scale $s$.
By Step 1, $A_Q(s)=A[q_{\Gamma,s}]$ (\cref{eq:supp:gain-equal}).
A gain is a mean of the alignment score minus the constant $\mathbb E_{p^c_\Gamma}T_\Gamma$ (\cref{def:gain-cost}).
Equal gains are therefore equal mean scores:
\begin{equation}
    \mathbb E_{p_{\Gamma,s^\ast}}T_\Gamma=\mathbb E_{q_{\Gamma,s}}T_\Gamma .
    \label{eq:supp:matched-mean}
\end{equation}

\emph{Compare the costs.}
We write $q=q_{\Gamma,s}$ and insert the tilt between $q$ and $p^c_\Gamma$:
\begin{align*}
    C[q]&=\mathbb E_q\log\frac{q}{p^c_\Gamma}\\
    &=\mathbb E_q\log\frac{q}{p_{\Gamma,s^\ast}}
     +\mathbb E_q\log\frac{p_{\Gamma,s^\ast}}{p^c_\Gamma}\\
    &=\mathrm{KL}\big(q\big\|p_{\Gamma,s^\ast}\big)
     +s^\ast\,\mathbb E_qT_\Gamma-F(\Gamma,s^\ast).
\end{align*}
The same line with the tilt itself in place of $q$ has no KL term:
\[
    C(s^\ast)=s^\ast\,\mathbb E_{p_{\Gamma,s^\ast}}T_\Gamma-F(\Gamma,s^\ast).
\]
These are \cref{eq:supp:gibbs-identity,eq:supp:gibbs-equality} at the level $\Gamma$ and the scale $s^\ast$.
By the matched mean \cref{eq:supp:matched-mean}, the two right sides differ only by the KL term:
\begin{equation}
    \boxed{C[q_{\Gamma,s}]=C(s^\ast)+\mathrm{KL}\big(q_{\Gamma,s}\big\|p_{\Gamma,s^\ast}\big).}
    \label{eq:supp:cost-split}
\end{equation}
In words: among all laws with the same mean score, the tilt is the cheapest, and any other law pays in addition its KL divergence from the tilt.

\emph{Two technical remarks.}
The scale $s^\ast$ is unique, because the mean score of the tilt increases with the scale, $\partial_{s'}\mathbb E_{p_{\Gamma,s'}}T_\Gamma=\chi_\Gamma(s')>0$ (\cref{lem:supp:moments,prop:frontier-shape}); the theorem takes its existence as given, with \cref{asm:finite} at $(\Gamma,s^\ast)$.
All terms above are finite: $C[q_{\Gamma,s}]\le C_Q(s)<\infty$ by \cref{eq:supp:cost-ineq}, and $T_\Gamma(Y_\Gamma)=\Lambda$ is integrable under $Q_s$ (\cref{eq:supp:Lambda-under-s}).

\paragraph{Conclusion.}
Inserting \cref{eq:supp:cost-split} into \cref{eq:supp:cost-ineq} gives \cref{eq:path-endpoint-bridge},
\[
    C_Q(s)=C(s^\ast)+\mathrm{KL}\big(q_{\Gamma,s}\big\|p_{\Gamma,s^\ast}\big)+S_s .
\]
The first term is the cost that any law must pay for this gain (\cref{thm:ideal-tilt}).
The second is cost wasted on places: the end points differ from the tilt in ways that buy no gain.
The third is cost wasted on routes.
Both wasted terms are nonnegative, so \cref{eq:supp:cost-split,eq:supp:cost-ineq} give \cref{eq:endpoint-cost-bound}, $C(s^\ast)\le C[q_{\Gamma,s}]=C_Q(s)-S_s$.

The same holds when the frontier is read as a gain at a given cost.
The law $q_{\Gamma,s}$ is feasible in \cref{eq:alignment-budget} at the budget $B=C[q_{\Gamma,s}]$, and a larger budget enlarges the feasible set, so $A^\ast$ is nondecreasing.
With \cref{eq:supp:gain-equal,eq:supp:cost-ineq},
\begin{equation}
    A_Q(s)=A[q_{\Gamma,s}]
    \le A^\ast\big(C[q_{\Gamma,s}]\big)
    \le A^\ast\big(C_Q(s)\big).
    \label{eq:path-frontier-bound}
\end{equation}
The sampler is on or below the frontier, at its end-point cost and hence at its path cost.

This proves \cref{eq:path-endpoint-bridge,eq:endpoint-cost-bound}, with the slack defined by \cref{eq:supp:slack}.
The statement also gives a formula for the slack, \cref{eq:slack-formula}; \cref{sec:supp:slack-formula} derives it, which completes the proof of the statement.
\end{proof}

\par\noindent\runin{The frontier on path space}
The frontier of \cref{def:gain-cost} is also the frontier for laws of trajectories.
No path law beats it: its gain is the gain of its end-point law (Step 1), and its cost is at least the cost of that law (Step 2).
And it is attained, by the path law that ends at the tilt $p_{\Gamma,s}$ and takes the reference routes, which has the gain $A(s)$ and the path cost $C(s)$.
Comparing the path cost of the sampler with the frontier is therefore like for like.

\par\noindent\runin{Path form of the alignment score}
\Cref{eq:supp:Lambda} writes $\Lambda$ through the drifts, and \cref{eq:supp:endpoint} through the end point.
Equating the two gives, almost surely under $Q_0$,
\begin{equation}
    T_\Gamma(Y_\Gamma)
    =\int_0^\Gamma g\cdot dB^{(0)}_\gamma+\frac12\int_0^\Gamma\|g\|^2\,d\gamma .
    \label{eq:supp:T-path}
\end{equation}
This is the pointwise identity \cref{eq:supp:pointwise} for the CFG pair, with the innovation $B^{(0)}$ of the conditional sampler in place of the channel noise.
Under $Q_0$ the innovation is a Brownian motion and the level-$\gamma$ marginal is $p^c_\gamma$, so its mean is $\frac12\int_0^\Gamma\mathbb E_{p^c_\gamma}\|g\|^2\,d\gamma$, the integral of \cref{eq:kl-immse}.

\par\noindent\runin{Numerical check}
Block (b) of \cref{tab:supp:toy-exact} compares the pathwise score of \cref{eq:supp:T-path} with the exact $T_\Gamma$ at the end point of each path.
\Cref{fig:supp:toy-bridge} checks \cref{eq:supp:gain-equal,eq:supp:cost-ineq}: the path gain equals the end-point gain, and the path cost is at least the end-point cost.
\Cref{fig:supp:toy-levels} places the sampler points against the frontier.

\subsubsection{Calculating the Path Slack}
\label{sec:supp:slack-formula}

\Cref{sec:supp:proof-sampler-below} defined the slack as the cost of the routes (\cref{eq:supp:slack}).
Here we calculate it, which gives \cref{eq:slack-formula} and completes the proof of \cref{thm:sampler-below}, and then estimate it with a finite number of bridges.

\par\noindent\runin{Definitions}
Let $g=\hat x_c-\hat x$ and $\kappa=\tfrac12s(1+s)>0$.
Let $Y$ be a trajectory with the end point $y$, and let $\tilde Y_\gamma=\frac\gamma\Gamma\,y+\beta_\gamma$, $\beta\sim\mathsf{Br}_\Gamma$, be a reference route to the same end point.
We write
\begin{equation}
\begin{aligned}
    \|g\|^2_Y&=\int_0^\Gamma\|g(Y_\gamma,\gamma)\|^2\,d\gamma,\\
    m(y)&=\mathbb E_{\beta\sim\mathsf{Br}_\Gamma}e^{-\kappa\|g\|^2_{\tilde Y}} .
\end{aligned}
    \label{eq:supp:W}
\end{equation}
The first is the disagreement accumulated along a trajectory, as in the statement of \cref{thm:sampler-below}.
The second is the mean of the weight $e^{-\kappa\|g\|^2}$ over the reference routes to $y$.
With $y\sim q_{\Gamma,s}$, the two numbers of \cref{eq:slack-formula} are
\[
    I_s=\mathbb E_y\,\mathbb E_{Q_s}\big[\|g\|^2_Y\,\big|\,y\big],
    \qquad
    \tilde I_s=\mathbb E_y\Big[-\frac1\kappa\log m(y)\Big].
\]
Both are a disagreement accumulated on the way to the end points of CFG: $I_s$ is its mean over the routes that CFG takes (the $I_s$ of \cref{prop:sampler}), and $\tilde I_s$ is its exponential mean over the reference routes.
The claim $S_s=\kappa(\tilde I_s-I_s)$ thus says that the slack is $\kappa$ times how much more the two denoisers disagree along the reference routes than along the routes of CFG.

\begin{proof}[Proof of \cref{eq:slack-formula} in \cref{thm:sampler-below}]
\emph{The weight of a CFG path.}
The drifts of $Q_s$ and $Q_0$ differ by $s\,g$, so by \cref{lem:supp:girsanov,eq:supp:innovation}
\begin{equation}
    \log\frac{dQ_s}{dQ_0}
    =s\int_0^\Gamma g\cdot dB^{(0)}_\gamma-\frac{s^2}2\,\|g\|^2_Y .
    \label{eq:supp:Qs-Q0}
\end{equation}
\Cref{eq:supp:Lambda} gives $\int_0^\Gamma g\cdot dB^{(0)}_\gamma=\Lambda-\tfrac12\|g\|^2_Y$, and $\Lambda=T_\Gamma(Y_\Gamma)$ by \cref{eq:supp:endpoint}, so
\begin{equation}
    \log\frac{dQ_s}{dQ_0}
    =s\,T_\Gamma(Y_\Gamma)-\kappa\,\|g\|^2_Y .
    \label{eq:supp:path-weight}
\end{equation}
The first term reads only the end point of the path, and the second reads the whole path.

\emph{CFG routes are reweighted reference routes.}
Step 2 of \cref{sec:supp:proof-sampler-below} wrote the same ratio as $\frac{q_{\Gamma,s}(y)}{p^c_\Gamma(y)}\cdot\frac{dQ_s(\cdot\mid y)}{d\mathsf{Br}_\Gamma}(\beta)$.
Only its second factor depends on the route, and the mean of that factor over $\beta\sim\mathsf{Br}_\Gamma$ is one.
Comparing with \cref{eq:supp:path-weight},
\begin{align}
    \frac{dQ_s(\cdot\mid y)}{d\mathsf{Br}_\Gamma}(\beta)
    &=\frac{e^{-\kappa\|g\|^2_Y}}{m(y)},
    \label{eq:supp:cfg-bridge}\\
    q_{\Gamma,s}(y)
    &=p^c_\Gamma(y)\,e^{sT_\Gamma(y)}\,m(y).
    \label{eq:supp:endpoint-law}
\end{align}
Given where it ends, CFG takes a reference route, reweighted so that routes with a large disagreement are rarer.
The mean weight $m(y)$ normalizes this law of routes, and it is also the factor by which the end-point law differs from the tilt $p_{\Gamma,s}\propto p^c_\Gamma e^{sT_\Gamma}$.

\emph{The slack.}
The logarithm of \cref{eq:supp:cfg-bridge} is $-\kappa\|g\|^2_Y-\log m(y)$.
Its mean over the routes of CFG to $y$ is the KL divergence between the two laws of routes, and the mean of that over $y\sim q_{\Gamma,s}$ is the slack of \cref{eq:supp:slack}:
\begin{align}
    &\mathrm{KL}\big(Q_s(\cdot\mid y)\big\|\mathsf{Br}_\Gamma\big)\notag\\
    &\qquad=-\kappa\,\mathbb E_{Q_s}\big[\|g\|^2_Y\,\big|\,y\big]-\log m(y),
    \label{eq:supp:bridge-kl}\\
    &S_s=-\kappa\,I_s-\mathbb E_y\log m(y)
    =\kappa\,\big(\tilde I_s-I_s\big).
    \label{eq:supp:slack-m}
\end{align}
This is \cref{eq:slack-formula}, and $S_s\ge0$ because it is a KL divergence.
The proof of \cref{thm:sampler-below} is complete.
\end{proof}

\par\noindent\runin{In practice: $n$ bridges}
The formula is exact.
Its estimate is not part of the theorem, and the inequality below comes from a finite number of bridges, not from the sampler.
We replace $m(y)$ by a mean over $n$ reference routes $\tilde Y^{(r)}_\gamma=\frac\gamma\Gamma\,y+\beta_{r,\gamma}$, with independent $\beta_1,\dots,\beta_n\sim\mathsf{Br}_\Gamma$:
\begin{equation}
\begin{aligned}
    \hat m_n(y)&=\frac1n\sum_{r=1}^{n}e^{-\kappa\|g\|^2_{\tilde Y^{(r)}}},\\
    \mathbb E\,\hat m_n(y)&=m(y).
\end{aligned}
    \label{eq:supp:m-hat}
\end{equation}
This estimate is unbiased, but its logarithm is not: the logarithm is concave, so by Jensen's inequality
\begin{equation}
    \mathbb E\log\hat m_n(y)\ \le\ \log\mathbb E\,\hat m_n(y)=\log m(y).
    \label{eq:supp:jensen}
\end{equation}
Inserted into \cref{eq:supp:slack-m}, this makes the estimate of the slack biased upward:
\begin{equation}
    S_s\ \le\ -\kappa\,I_s-\mathbb E\log\Big(\frac1n\sum_{r=1}^{n}e^{-\kappa\|g\|^2_{\tilde Y^{(r)}}}\Big),
    \label{eq:supp:slack-bound}
\end{equation}
with the routes $\tilde Y^{(r)}$ drawn to the end point $Y_\Gamma$ of a CFG trajectory.
We call the right side the slack bound; \cref{alg:sampler} computes the sum $\|g\|^2_{\tilde Y^{(r)}}$ on its level grid and calls it $\tilde W_r$.
Unlike the inequalities of the theorem, which hold exactly, this one holds on average over the bridges, and its gap is due to the finite $n$: $\hat m_n(y)\to m(y)$ as $n$ grows.
An estimated end-point cost, $C_Q(s)$ minus the estimated slack, can therefore fall below the frontier cost $C(s^\ast)$; this signals too few bridges and does not contradict \cref{eq:endpoint-cost-bound}.
With learned denoisers, \cref{eq:supp:path-kl,eq:supp:Lambda} still hold under \cref{asm:girsanov}, but the innovations step of \cref{lem:supp:endpoint} fails, so the measured path curve and the fixed-level curve keep their separate meanings, and the slack bound, which rests on the same step, is then an approximation.

\par\noindent\runin{When the two parts of the gap vanish}
The slack is zero at $s=0$ and at $s=-1$, where $\kappa=0$: these are the conditional and the unconditional sampler, which take the reference routes by \cref{lem:supp:endpoint}.
For other scales $Q_s$ need not be the law of the data path of any clean law, and its trajectories may carry information beyond their end points.
It is such a law for the equal-covariance Gaussian pair of \cref{prop:supp:gaussian}, where $g$ does not depend on $y$ and $\|g\|^2_Y$ is a constant; there the weight in \cref{eq:supp:cfg-bridge} is one, so CFG takes the reference routes, the slack is zero, the end-point law is a tilt, and the sampler is on the frontier.
The shortfall $\mathrm{KL}(q_{\Gamma,s}\|p_{\Gamma,s^\ast})$ is zero only if the end-point law is the tilt at $s^\ast$.

\par\noindent\runin{Numerical check}
\Cref{sec:supp:toy-exact} checks the slack bound \cref{eq:supp:slack-bound} against the exact slack, and \cref{sec:supp:slack-bound} reports it for an image model.

\subsection{A Diversity Index for CFG}
\label{sec:supp:diversity-theory}

This subsection follows \cref{sec:diversity:theory}.
It gives the coverage index and its meaning, then the baseline and the loss due to guidance for the ideal tilt, and then the same two quantities for the CFG sampler.

\subsubsection{The Coverage Index}
\label{sec:supp:coverage-definition}

\replayed{\defCoverage*}

\paragraph{The index through the log-normalizer.}
Let \cref{asm:finite} hold.
The coverage loss is a function of $F$ and its first derivative.
By \cref{eq:cfg-tilt}, the log-ratio of the tilt to the conditional law is $sT_\gamma-F(\gamma,s)$, and the log-ratio of the conditional law to the unconditional law is $T_\gamma$.
The log-ratio of the tilt to the unconditional law is their sum:
\begin{align}
    \log\frac{p_{\gamma,s}}{p_\gamma}
    &=\log\frac{p_{\gamma,s}}{p^c_\gamma}+\log\frac{p^c_\gamma}{p_\gamma}\notag\\
    &=\big[sT_\gamma-F(\gamma,s)\big]+T_\gamma\notag\\
    &=(1+s)\,T_\gamma-F(\gamma,s).
    \label{eq:supp:coverage-log-ratio}
\end{align}
We average under the tilt.
The mean alignment score under the tilt is the first derivative of the log-normalizer, $\mathbb E_{p_{\gamma,s}}T_\gamma=\partial_sF(\gamma,s)$ (\cref{eq:normalizer-derivatives} of \cref{lem:supp:moments}, in \cref{sec:supp:proof-log-normalizer}), so
\begin{align}
    V(s)
    &=\mathbb E_{p_{\gamma,s}}\log\frac{p_{\gamma,s}}{p_\gamma}\notag\\
    &=(1+s)\,\mathbb E_{p_{\gamma,s}}T_\gamma-F(\gamma,s)\notag\\
    &=(1+s)\,\partial_sF(\gamma,s)-F(\gamma,s).
    \label{eq:supp:coverage-from-F}
\end{align}
The coefficient $1+s$ has one unit from conditioning and $s$ units from guidance.

\paragraph{Excess code length.}
The coverage loss is the excess code length for describing guided samples with a code built for the unconditional law.
By \cref{eq:supp:info-notation}, $\mathrm{KL}(q\|p)=\mathbb E_q[-\log p]-\mathbb E_q[-\log q]$ is the expected length under the code for $p$ minus that under the code for $q$~\cite{cover2006elements}.
For densities, the resolution terms of a common fine quantization cancel in the difference.

\paragraph{Subset selection.}
\label{lem:supp:subset}
The index $N=e^{-V}$ is a retained fraction.
Let $S$ be a set with $r=p_\gamma(S)>0$, and suppose that the guided law keeps only the outcomes in $S$, with their relative weights: $q=p_\gamma(\cdot\mid S)=p_\gamma\mathbf 1_S/r$.
On $S$ the density ratio is $q/p_\gamma=1/r$, so
\begin{equation}
    V=\mathbb E_q\log\frac{q}{p_\gamma}=\log\frac1r,
    \qquad N=e^{-V}=r.
    \label{eq:supp:subset}
\end{equation}
In general, $e^{V(s)}=\exp\mathbb E_{p_{\gamma,s}}\log(p_{\gamma,s}/p_\gamma)$ is the geometric mean of the density ratio over guided samples, and $N(s)$ is the fraction whose retention would produce the same concentration.

\paragraph{Narrowing and moving.}
Let $H(s)=-\mathbb E_{p_{\gamma,s}}\log p_{\gamma,s}$ be the entropy of the guided law and $\mathrm{CE}(s)=-\mathbb E_{p_{\gamma,s}}\log p_\gamma$ its cross-entropy to the unconditional law.
Splitting the logarithm of the ratio in \cref{eq:guided-coverage} gives \cref{eq:coverage-entropy}:
\begin{align*}
    V(s)
    &=\mathbb E_{p_{\gamma,s}}\log p_{\gamma,s}
        -\mathbb E_{p_{\gamma,s}}\log p_\gamma\\
    &=-H(s)+\mathrm{CE}(s).
\end{align*}
Subtracting the same identity at $s=0$ splits a change of coverage into two parts:
\begin{equation}
    V(s)-V(0)
        =\underbrace{-[H(s)-H(0)]}_{\text{narrowing}}
        +\underbrace{[\mathrm{CE}(s)-\mathrm{CE}(0)]}_{\text{moving}}.
    \label{eq:coverage-entropy-split}
\end{equation}
Entropy sees only the first part.
A guided law can keep its spread and move its mass to outcomes that the unconditional law finds improbable; it then loses coverage while keeping its entropy.
A law that spreads over such outcomes can even gain entropy and cover less of the unconditional law.
For the Gaussian pair of \cref{prop:supp:gaussian} (\cref{sec:supp:toy}), guidance translates the mean, so $H(s)=H(0)$ while $V(s)-V(0)=\chi s+\chi s^2/2$, which is positive for $s>0$ and $\delta\neq0$.

\par\noindent\runin{Numerical check}
Block (d) of \cref{tab:supp:toy-closed} shows that the entropy change of \cref{eq:coverage-entropy-split} is not the cost, and \cref{fig:supp:toy-law} pictures a tilt that removes modes which are valid outcomes of the condition.

\subsubsection{The Baseline $V(0)$}
\label{sec:supp:coverage-baseline}

\paragraph{Definition.}
The baseline is the coverage loss at $s=0$, where the tilt is the conditional law.
By \cref{eq:guided-coverage,eq:supp:coverage-from-F} and $F(\gamma,0)=0$,
\begin{equation}
    V(0)=\mathrm{KL}(p^c_\gamma\|p_\gamma)=\partial_sF(\gamma,0).
    \label{eq:supp:baseline}
\end{equation}
It is the coverage that the condition itself gives up, before any guidance.
It compares conditions: a broad condition covers much of the unconditional law and has a small $V(0)$, and a specific one has a large $V(0)$.

\paragraph{Derivation of \cref{eq:coverage-immse}.}
Let \cref{asm:exact,asm:finite,asm:regular,asm:zero-snr} hold.
The mismatched I-MMSE identity in its CFG form, \cref{eq:kl-immse} (\cref{sec:supp:info-background}), gives the rate of the baseline in the noise level, $\frac{d}{d\gamma}\mathrm{KL}(p^c_\gamma\|p_\gamma)=\frac12\mathbb E_{p^c_\gamma}\|g\|^2$.
We integrate it from zero SNR to the horizon $\Gamma$; \cref{asm:zero-snr} makes the divergence vanish at zero SNR, and for the KL divergence this limit holds automatically (\cref{eq:supp:zero-snr-kl}):
\begin{align}
    V(0)&=\mathrm{KL}(p^c_\Gamma\|p_\Gamma)\notag\\
    &=\lim_{\gamma\downarrow0}\mathrm{KL}(p^c_\gamma\|p_\gamma)
        +\int_0^\Gamma\frac{d}{d\gamma}
        \mathrm{KL}(p^c_\gamma\|p_\gamma)\,d\gamma\notag\\
    &=0+\frac12\int_0^\Gamma
        \mathbb E_{p^c_\gamma}\|g\|^2\,d\gamma ,
    \label{eq:supp:baseline-immse}
\end{align}
which is \cref{eq:coverage-immse} with $g=\hat x_c-\hat x$.
In the notation of \cref{sec:supp:tradeoff-estimators}, the baseline is $V(0)=\frac12G(\Gamma,0)$.

\paragraph{How it is calculated.}
The baseline is $\frac12G(\Gamma,0)$, the $s=0$ column of the estimate of $G$ in \cref{alg:langevin}.
At $s=0$ the tilt is the conditional law $p^c_\gamma$.
The chains start from data noised to level $\gamma$, so they average $\|g\|^2$ under the conditional observation law, and no guidance scale is involved.
The level averages are integrated over the noise grid, with \cref{eq:supp:prior-extension} for the start at finite noise (\cref{sec:supp:tradeoff-estimators}).
At a finite horizon the baseline is the divergence between the noisy laws, which by data processing is a lower bound on the clean $\mathrm{KL}(p^c\|p)$ (\cref{sec:supp:info-measures,sec:supp:horizon}).

\par\noindent\runin{Numerical check}
\Cref{tab:supp:toy-trained,fig:supp:toy-trained-profile} compare the two sides of \cref{eq:coverage-immse}, with exact and with trained denoisers.

\subsubsection{The Loss Due to Guidance}
\label{sec:supp:coverage-frontier}

\paragraph{Definition.}
The loss due to guidance is the coverage that guidance removes on top of the baseline,
\begin{equation}
    \Delta V(s)=V(s)-V(0).
    \label{eq:supp:delta-V}
\end{equation}
It compares scales for one condition, and $N(s)=N(0)\,e^{-\Delta V(s)}$.

\paragraph{Derivation of \cref{eq:coverage-alignment-cost}.}
Let \cref{asm:finite} hold.
We first take any law $q$ at level $\gamma$ with finite integrals.
The log-ratio of $q$ to the unconditional law is its log-ratio to the conditional law plus $T_\gamma$.
Averaging under $q$, the first term is the cost $C[q]$ of \cref{eq:candidate-alignment-cost}.
The second is $\mathbb E_qT_\gamma=A[q]+\mathbb E_{p^c_\gamma}T_\gamma$, by the definition of the gain, and $\mathbb E_{p^c_\gamma}T_\gamma=\mathrm{KL}(p^c_\gamma\|p_\gamma)=V(0)$.
Hence
\begin{align}
    \mathrm{KL}(q\|p_\gamma)
    &=\mathbb E_q\log\frac{q}{p^c_\gamma}
        +\mathbb E_qT_\gamma\notag\\
    &=C[q]+A[q]+V(0),\notag\\
    \mathrm{KL}(q\|p_\gamma)-V(0)&=A[q]+C[q].
    \label{eq:supp:coverage-chain-rule}
\end{align}
This chain rule does not use the form of $q$.
At $q=p_{\gamma,s}$ it reads $\Delta V(s)=A(s)+C(s)$, which is \cref{eq:coverage-alignment-cost}.

The same result follows through $F$.
We subtract \cref{eq:supp:baseline} from \cref{eq:supp:coverage-from-F}, split $(1+s)\,\partial_sF=\partial_sF+s\,\partial_sF$, and recognize the two brackets as the gain and the cost of \cref{eq:alignment-cost}:
\begin{align}
    \Delta V(s)
    &=(1+s)\,\partial_sF(\gamma,s)-F(\gamma,s)
        -\partial_sF(\gamma,0)\notag\\
    &=\big[\partial_sF(\gamma,s)-\partial_sF(\gamma,0)\big]\notag\\
    &\quad+\big[s\,\partial_sF(\gamma,s)-F(\gamma,s)\big]\notag\\
    &=A(s)+C(s).
    \label{eq:supp:delta-V-from-F}
\end{align}

\paragraph{Monotonicity.}
Differentiating \cref{eq:supp:coverage-from-F} in $s$, with $\chi_\gamma(s)=\partial_s^2F(\gamma,s)$ the variance of $T_\gamma$ under the tilt (\cref{eq:normalizer-derivatives}),
\begin{align}
    \partial_sV(s)
    &=\partial_sF+(1+s)\,\partial_s^2F-\partial_sF\notag\\
    &=(1+s)\,\chi_\gamma(s)\ \ge0
    \label{eq:supp:coverage-monotone}
\end{align}
for every $s>-1$.
Raising the scale of ideal guidance can therefore only lower the coverage index $N(s)$.

\paragraph{How it is calculated.}
$\Delta V$ needs no new estimator.
It is the sum of the gain and the cost of the frontier, which are computed from \cref{eq:alignment-cost} with the estimates of $F$ and $\partial_sF$ of \cref{alg:langevin} (\cref{sec:supp:tradeoff-estimators}).
\Cref{eq:supp:AC-from-G} (\cref{sec:supp:ideal-law}) is the equivalent closed form in terms of $G$.

\par\noindent\runin{Numerical check}
Panel (b) of \cref{fig:supp:toy-closed} places the two sides of \cref{eq:coverage-alignment-cost} on the diagonal, at two horizons and for scales up to $s=3$.

\subsubsection{The Sampler}
\label{sec:supp:coverage-sampler}

We repeat the two items for the CFG sampler.
The sampler defines a law $Q_s$ on trajectories, not a density at one noise level, so its coverage is measured on path space.

\paragraph{Definition.}
The coverage loss of the sampler is its path divergence from the unconditional sampler,
\begin{equation}
\begin{aligned}
    V_Q(s)&=\mathrm{KL}(Q_s\|Q_{-1}),
    \qquad N_Q(s)=e^{-V_Q(s)},\\
    \Delta V_Q(s)&=V_Q(s)-V_Q(0).
\end{aligned}
\label{eq:supp:VQ}
\end{equation}
The baseline $V_Q(0)=\mathrm{KL}(Q_0\|Q_{-1})$ is the divergence of the conditional sampler from the unconditional one.
The index $N_Q$ is the path-space counterpart of $N$ in \cref{eq:guided-coverage}.

\paragraph{Derivation of \cref{eq:sampler-coverage}.}
Let \cref{asm:girsanov} hold.
By \cref{eq:supp:drift-diff} with $s'=-1$, the drifts of $Q_s$ and $Q_{-1}$ differ by $(1+s)\,g$.
The path divergence \cref{eq:supp:path-kl} (\cref{sec:supp:pathspace}, from \cref{lem:supp:girsanov}), $\mathrm{KL}(Q_s\|Q_{s'})=\frac{(s-s')^2}2I_s$, at $s'=-1$ gives
\begin{equation}
    V_Q(s)=\frac{(1+s)^2}2\,I_s,
    \qquad
    V_Q(0)=\frac12\,I_0 .
    \label{eq:supp:VQ-from-I}
\end{equation}
We subtract the baseline, split $(1+s)^2=s^2+(1+2s)$, and recognize the cost and the gain of \cref{eq:sampler-point}:
\begin{align}
    \Delta V_Q(s)
    &=\frac{(1+s)^2}2\,I_s-\frac12\,I_0\notag\\
    &=\frac{s^2}2\,I_s
        +\Big[\frac{1+2s}2\,I_s-\frac12\,I_0\Big]\notag\\
    &=C_Q(s)+A_Q(s),
    \label{eq:supp:delta-VQ}
\end{align}
which is \cref{eq:sampler-coverage}.
This is the chain rule \cref{eq:supp:coverage-chain-rule} on path space, with $\Lambda$ in place of $T_\gamma$ and $Q_0,Q_{-1}$ in place of $p^c_\gamma,p_\gamma$.

\paragraph{How it is calculated.}
Both $V_Q(0)$ and $\Delta V_Q(s)$ are functions of $I_0$ and $I_s$, which are estimated by \cref{eq:supp:Is-hat} from the values of $\|g\|^2$ recorded along the trajectories.
A guided run evaluates both denoisers at every step, so this needs no extra model evaluations.

\paragraph{What $V_Q$ is and is not.}
$V_Q$ is a quantity on trajectories.
The generated image is a function of the trajectory, so by data processing (\cref{sec:supp:info-measures}) the KL divergence between the generated images and the unconditional samples is at most $V_Q(s)$.
$V_Q$ is also not the coverage of the ideal tilt, which the sampler does not reach.
With exact denoisers the two baselines agree at a common horizon: \cref{lem:supp:endpoint} gives $\Lambda=T_\Gamma(Y_\Gamma)$, so
\begin{equation}
    V_Q(0)=\mathbb E_{Q_0}\Lambda
    =\mathbb E_{p^c_\Gamma}T_\Gamma=V(0).
    \label{eq:supp:baselines-agree}
\end{equation}
With learned denoisers the two baseline estimators need not agree, and the paper reports which one each figure uses.

\par\noindent\runin{Numerical check}
\Cref{sec:supp:toy-exact} compares the sampler baseline $V_Q(0)=\tfrac12I_0$ with the closed-form $V(0)$, and the coverage panel of \cref{fig:supp:toy-levels} shows the coverage loss of the sampler next to that of the tilt.

\newpage
\section{Numerical Check on Gaussian Mixtures}
\label{sec:supp:toy}

\begin{figure*}[t]
    \centering
    \includegraphics[width=\textwidth]{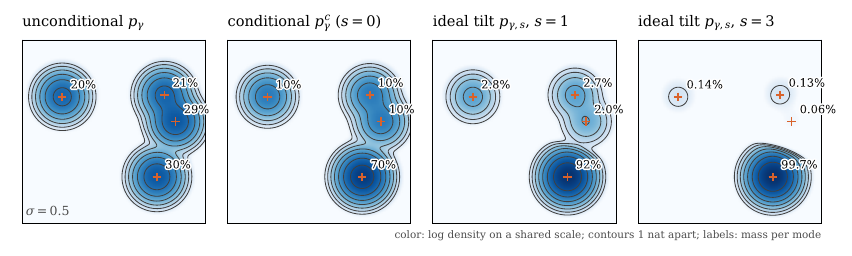}
    \caption{\textbf{The two-dimensional class law of the numerical check} at noise level $\sigma=0.5$.
    Color is log density on one scale shared by the four panels, eight nats below the largest peak, and contours are one nat apart.
    Crosses mark the four component means, and each label is the mass of the law in the region closest to that mean.
    Gaussian mixtures are closed under the Gaussian channel, so the unconditional law, the conditional law and the ideal tilt $p_{\gamma,s}$ are exact at every level.
    Guidance moves the conditional mass onto the dominant component of the class: the three minor components hold $30\%$ of the class at $s=0$, $8\%$ at $s=1$ and $0.3\%$ at $s=3$.}
    \label{fig:supp:toy-law}
\end{figure*}

Gaussian mixtures are closed under the Gaussian channel.
For a mixture pair $(p^c,p)$ the observation densities, the alignment score $T_\gamma$, the denoisers, the tilt and its normalizer are therefore available in closed form at every noise level, and fixed-level averages reduce to quadrature.
This makes mixtures the one setting where every quantity in the paper has a ground truth.
We use it to check the identities at three levels, each one removing an assumption of the previous one.
\begin{enumerate}
    \item \textbf{Closed form.} Every quantity is computed from the densities by quadrature, and the identities are checked as equalities between numbers. No denoiser and none of our estimators is involved.
    \item \textbf{Exact denoisers.} The posterior means are evaluated in closed form, and the paper's estimators (fixed-level Langevin, CFG trajectories, pathwise scores) are run on them. This checks that the estimators recover the closed-form values when the identities hold exactly.
    \item \textbf{Trained denoisers.} Small MLP denoisers are trained on samples of the mixtures and used in place of the exact ones. This checks that the pipeline still returns the closed-form divergences when the denoiser is learned, which is the situation of the image models.
\end{enumerate}
The laws are listed in \cref{tab:supp:toy-laws}, and the two-dimensional class law on which all three levels are compared is pictured in \cref{fig:supp:toy-law}.
All are mixtures; the equal-covariance Gaussian pair is the one-component case, in which the frontier is known exactly (\cref{prop:supp:gaussian}).

\Cref{fig:supp:toy-law} shows the ideal tilt concentrating mass on the component that most strongly supports the condition, while suppressing other components that remain valid conditional outcomes.
Thus, the loss of variety is already present in the optimal alignment--cost law.
This provides a concrete picture of the connection between the paper's two applications: buying more alignment also reduces coverage, as quantified by \cref{eq:coverage-alignment-cost}.

\begin{table*}[t]
    \centering
    \footnotesize
    \setlength{\tabcolsep}{4pt}
    \begin{tabular}{lll}
        \toprule
        Name & Conditional $p^c$ & Unconditional $p$ \\
        \midrule
        Narrow & $\mathcal N(0.6,0.3^2)$ & $\tfrac12\mathcal N(-1,0.5^2)+\tfrac12\mathcal N(1,0.45^2)$ \\
        Mode swap & $0.4\,\mathcal N(-1.5,0.25^2)+0.6\,\mathcal N(1.5,0.25^2)$ & $0.15\,\mathcal N(-1.5,0.35^2)+0.85\,\mathcal N(1.5,0.35^2)$ \\
        Two-mode & $0.4\,\mathcal N(-1,0.35^2)+0.6\,\mathcal N(1.2,0.25^2)$ & $0.5\,\mathcal N(-1.3,0.5^2)+0.5\,\mathcal N(1,0.3^2)$ \\
        Five-mode & weights $q$ on $\mathcal N(m_k,0.12^2)$ & weights $\pi$ on the same components \\
        Shift & $\mathcal N(0.5,0.5^2)$ & $\mathcal N(-0.5,0.5^2)$ \\
        Class, $d$ dims & class $0$ of a 3-class, 4-component law & uniform mixture of the 3 classes \\
        \bottomrule
    \end{tabular}
    \caption{\textbf{Mixture pairs used in the numerical checks.}
    Five-mode: $m_k\in\{-3.6,-1.8,0,1.8,3.6\}$, $\pi=(0.1,0.2,0.3,0.25,0.15)$, and $q$ keeps the relative weights of $\pi$ on a subset of modes.
    Class law: four components of standard deviation $0.3$ in $\mathbb R^d$ at pairwise distance about $5$, with class weights $(0.7,0.1,0.1,0.1)$, $(0.1,0.7,0.1,0.1)$, $(0.1,0.1,0.4,0.4)$; the condition is class $0$.
    The class law is the one on which the denoisers of \cref{sec:supp:toy-trained} are trained.}
    \label{tab:supp:toy-laws}
\end{table*}

\subsection{Closed Form: the Identities as Equalities of Numbers}
\label{sec:supp:toy-closed}

\Cref{tab:supp:toy-closed} collects the closed-form checks.
Each block compares the two sides of one identity of the main text, both computed by quadrature on a fine grid in heat time.

\paragraph{R\'enyi I-MMSE.}
Block (a) checks the heat-flow form \cref{eq:supp:renyi-heat} of \cref{eq:renyi-immse}, which is the differential form of \cref{eq:renyi-integral}, on the narrow pair at $\sigma=0.55$.
The left side is a finite difference of the R\'enyi divergence across noise levels; the right side is the squared score difference, which is $\gamma^2\|g\|^2$, averaged under the tilt.
They agree to seven digits for orders below and above one, including the interpolating order $1+s=0.5$.

\paragraph{The frontier.}
Block (b) checks \cref{eq:alignment-cost,eq:frontier-shape} on the narrow pair at $\sigma=0.45$.
The cost is computed by three routes: directly as $\mathrm{KL}(p_{\gamma,s}\|p^c_\gamma)$ by quadrature, as $s^2\partial_sD_{1+s}$, and as $s\,\partial_sF-F$.
The gain is computed directly as the difference of mean alignment scores and as $\partial_sF(s)-\partial_sF(0)$.
All routes agree to six digits, and the slope $dC/dA$ by finite differences equals the scale $s$ at every point.
The susceptibility $\chi$ and the coverage follow from the same $F$.

\paragraph{A closed-form frontier: the equal-covariance Gaussian pair.}
There is one case in which the frontier of \cref{prop:frontier-shape} has a closed form: a pair of Gaussians with a common covariance, of which the shift pair of \cref{tab:supp:toy-laws} is the one-dimensional instance.
The frontier is then a parabola, which is the small-$s$ law of \cref{cor:supp:small-s}, here exact at every scale.
\begin{sproposition}[An equal-covariance Gaussian pair]
\label{prop:supp:gaussian}
Let $X\mid c\sim\mathcal N(\mu_c,\Sigma)$ and $X\sim\mathcal N(\mu,\Sigma)$, with $\delta=\mu_c-\mu$ and $\Sigma_\gamma=\gamma^2\Sigma+\gamma I$.
Then, for every $\gamma>0$ and every real $s$,
\begin{gather}
    T_\gamma(y)=\gamma\,\delta^\top\Sigma_\gamma^{-1}\Big(y-\gamma\,\frac{\mu_c+\mu}2\Big),\notag\\
    p_{\gamma,s}=\mathcal N\big(\gamma(\mu_c+s\delta),\Sigma_\gamma\big),\notag\\
    \chi_\gamma(s)=\delta^\top(\Sigma+\gamma^{-1}I)^{-1}\delta=:\chi,\notag\\
    F(\gamma,s)=\frac{s(1+s)}2\,\chi,\qquad
    \partial_sF(\gamma,s)=\frac{1+2s}2\,\chi,\notag\\
    A(s)=\chi s,\qquad C(s)=\frac{\chi s^2}2,\notag\\
    V(s)=\frac{(1+s)^2}2\,\chi,\qquad
    \Delta V(s)=\chi s+\frac{\chi s^2}2 .
    \label{eq:supp:gaussian-pair}
\end{gather}
For $\delta\neq0$ the frontier is thus the parabola $C=A^2/(2\chi)$, and $A=\sqrt{2\chi C}$ for $s\ge0$.
\end{sproposition}
\begin{proof}
By the channel \cref{eq:channel}, the observations are $Y_\gamma\mid c\sim\mathcal N(\gamma\mu_c,\Sigma_\gamma)$ and $Y_\gamma\sim\mathcal N(\gamma\mu,\Sigma_\gamma)$.
Write $S=\Sigma_\gamma^{-1}$.
The two log-densities share their normalizing constant and their quadratic term in $y$, so, using $\mu_c^\top S\mu_c-\mu^\top S\mu=\delta^\top S(\mu_c+\mu)$,
\begin{align*}
    T_\gamma(y)
    &=-\tfrac12(y-\gamma\mu_c)^\top S(y-\gamma\mu_c)\\
    &\quad+\tfrac12(y-\gamma\mu)^\top S(y-\gamma\mu)\\
    &=\gamma\,\delta^\top Sy-\tfrac{\gamma^2}2\,\delta^\top S(\mu_c+\mu)\\
    &=\gamma\,\delta^\top S\Big(y-\gamma\,\frac{\mu_c+\mu}2\Big).
\end{align*}
The tilt \cref{eq:cfg-tilt} adds $sT_\gamma$ to the conditional log-density, and $T_\gamma$ is affine in $y$, so
\begin{align*}
    \log p_{\gamma,s}(y)
    &=-\tfrac12(y-\gamma\mu_c)^\top S(y-\gamma\mu_c)\\
    &\quad+s\gamma\,\delta^\top Sy+\mathrm{const}\\
    &=-\tfrac12y^\top Sy+y^\top S\,\gamma(\mu_c+s\delta)+\mathrm{const},
\end{align*}
which is the log-density of $\mathcal N(\gamma(\mu_c+s\delta),\Sigma_\gamma)$: guidance translates the mean and keeps the covariance.
In particular the normalizer is finite for every $s$.
By \cref{eq:normalizer-derivatives}, $\chi_\gamma(s)$ is the variance of $T_\gamma$ under the tilt.
The variance of an affine function with gradient $\gamma S\delta$ under a law with covariance $\Sigma_\gamma$ is
\begin{align*}
    \chi_\gamma(s)
    &=\gamma^2\,\delta^\top S\,\Sigma_\gamma\,S\delta
    =\gamma^2\,\delta^\top\Sigma_\gamma^{-1}\delta\\
    &=\delta^\top(\Sigma+\gamma^{-1}I)^{-1}\delta,
\end{align*}
using $\Sigma_\gamma=\gamma^2(\Sigma+\gamma^{-1}I)$; it does not depend on $s$.

For the log-normalizer, note that under $p^c_\gamma$ the affine score $T_\gamma$ is a Gaussian variable.
Its variance is $\chi$, and its mean is its value at the mean $\gamma\mu_c$ of $p^c_\gamma$,
\[
    \mathbb E_{p^c_\gamma}T_\gamma
    =\gamma\,\delta^\top S\Big(\gamma\mu_c-\gamma\,\frac{\mu_c+\mu}2\Big)
    =\frac{\gamma^2}2\,\delta^\top S\delta
    =\frac\chi2 .
\]
The moment generating function of a Gaussian variable with mean $m$ and variance $v$ is $e^{sm+s^2v/2}$, so
\begin{align*}
    F(\gamma,s)
    &=\log\mathbb E_{p^c_\gamma}e^{sT_\gamma}
    =\frac{s\chi}2+\frac{s^2\chi}2
    =\frac{s(1+s)}2\,\chi,\\
    \partial_sF(\gamma,s)&=\frac{1+2s}2\,\chi .
\end{align*}
Gain and cost follow from \cref{eq:alignment-cost}:
\begin{align*}
    A(s)&=\partial_sF(\gamma,s)-\partial_sF(\gamma,0)\\
        &=\frac{1+2s}2\,\chi-\frac\chi2=\chi s,\\
    C(s)&=s\,\partial_sF(\gamma,s)-F(\gamma,s)\\
        &=\frac{s(1+2s)}2\,\chi-\frac{s(1+s)}2\,\chi
        =\frac{\chi s^2}2 .
\end{align*}
Eliminating $s$ gives $C=A^2/(2\chi)$, hence $A=\sqrt{2\chi C}$ for $s\ge0$.
The coverage loss follows from \cref{eq:supp:coverage-from-F}, with $(1+2s)-s=1+s$:
\begin{align*}
    V(s)&=(1+s)\,\partial_sF(\gamma,s)-F(\gamma,s)\\
        &=\frac{(1+s)(1+2s)}2\,\chi-\frac{s(1+s)}2\,\chi
        =\frac{(1+s)^2}2\,\chi,\\
    \Delta V(s)&=V(s)-V(0)
        =\frac{(1+s)^2-1}2\,\chi
        =\chi s+\frac{\chi s^2}2 ,
\end{align*}
which is $A(s)+C(s)$, as in \cref{eq:coverage-alignment-cost}.
\end{proof}
In this case the CFG sampler is on the frontier.
The posterior mean of the channel \cref{eq:channel} for a Gaussian prior $\mathcal N(m,\Sigma)$ is $(\Sigma^{-1}+\gamma I)^{-1}(\Sigma^{-1}m+y)$, so the CFG direction does not depend on $y$:
\[
    g=(\Sigma^{-1}+\gamma I)^{-1}\Sigma^{-1}\delta=(I+\gamma\Sigma)^{-1}\delta .
\]
Its accumulated square is therefore the same under every law,
\[
    \int_0^\Gamma\|g\|^2\,d\gamma
    =\delta^\top\Big[\int_0^\Gamma(I+\gamma\Sigma)^{-2}\,d\gamma\Big]\delta
    =\delta^\top(\Sigma+\Gamma^{-1}I)^{-1}\delta ,
\]
so $G(\Gamma,s)=I_s=\chi_\Gamma$ for every $s$, and \cref{eq:sampler-point} gives $A_Q(s)=\chi_\Gamma s=A(s)$ and $C_Q(s)=\chi_\Gamma s^2/2=C(s)$.
Moreover, the guided drift $\hat x_c+s\,g$ is the exact posterior mean for the clean law $\mathcal N(\mu_c+s\delta,\Sigma)$, so $Q_s$ is the data path of this law (\cref{lem:supp:endpoint}), and the end-point law of the sampler is the tilt, $q_{\Gamma,s}=\mathcal N(\Gamma(\mu_c+s\delta),\Sigma_\Gamma)=p_{\Gamma,s}$.
As $\gamma\to\infty$, $\chi\to\delta^\top\Sigma^{-1}\delta$, the squared Mahalanobis distance between the means.
Unequal covariances make $\chi$ depend on $s$, and for large $s$ the normalizer can become infinite.

\begin{figure*}[t]
    \centering
    \includegraphics[width=\textwidth]{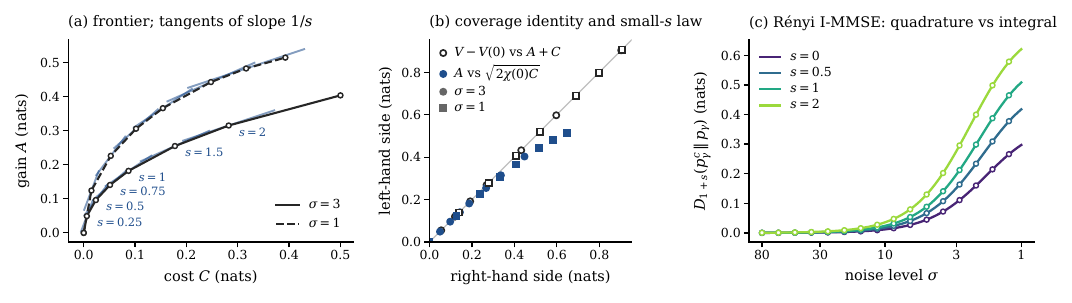}
    \caption{\textbf{Closed-form checks on the class law} (quadrature on the exact densities; no denoiser or estimator).
    (a) The frontier $A(C)$ at two horizons; the light blue line through the point labeled $s$ has slope $1/s$ and is tangent to the frontier there, so the cost is convex in the gain and the guidance scale is its marginal price (\cref{eq:frontier-shape}).
    (b) Left against right side of the coverage identity $V(s)-V(0)=A(s)+C(s)$ (\cref{eq:coverage-alignment-cost}) and of the small-$s$ law $A\approx\sqrt{2\chi(0)C}$ (\cref{sec:supp:frontier-shape}); the diagonal is equality; circles are $\sigma=3$ and squares $\sigma=1$.
    (c) The R\'enyi I-MMSE identity in integral form (\cref{eq:renyi-integral}): $D_{1+s}$ between the noisy laws at each level by quadrature (lines) against the integral of the tilt-averaged squared denoiser difference from the largest noise (circles), for four orders.}
    \label{fig:supp:toy-closed}
\end{figure*}

\paragraph{The same identities on the two-dimensional class law.}
\Cref{fig:supp:toy-closed} repeats the closed-form checks on the class law of \cref{fig:supp:toy-law}, at the horizons $\sigma=3$ and $\sigma=1$ and for scales up to $s=3$, by quadrature on a $400\times400$ grid.
Panel (a) shows the frontier with tangent segments of slope $1/s$ drawn at each scale: the segments are tangent to the curve, which is \cref{eq:frontier-shape} seen directly.
Panel (b) places both sides of the coverage identity \cref{eq:coverage-alignment-cost} on the diagonal; the two sides agree to four digits at every scale and horizon.
The same panel shows the small-$s$ law $A\approx\sqrt{2\chi(0)C}$ of \cref{sec:supp:frontier-shape}: at $s=0.25$, $\sqrt{2\chi(0)C}$ is $0.0495$ against $A=0.0492$ at $\sigma=3$, and the approximation drifts above $A$ as $s$ grows.
Since $A-\sqrt{2\chi_\gamma(0)\,C}=\chi_\gamma'(0)\,s^2/6+O(s^3)$, this is the case $\chi_\gamma'(0)<0$, where $\chi_\gamma'(0)$ is the third cumulant of $T_\gamma$ under $p^c_\gamma$.
Panel (c) checks the R\'enyi I-MMSE identity in its integral form \cref{eq:renyi-integral}: the divergence between the noisy laws at each level, by quadrature, against the integral of the tilt-averaged squared denoiser difference from the largest noise $\sigma_{\max}=80$, for four orders.
At the horizon the two agree to $10^{-4}$ nats at $\sigma=3$ and to $2\times10^{-3}$ at $\sigma=1$, the latter being the trapezoid error of the $64$-level ladder; the differential form \cref{eq:renyi-immse} agrees to six digits at both horizons.

\Cref{fig:supp:toy-closed} connects the shape of the frontier, the loss of coverage, and the denoiser representation of divergence.
The tangent slopes show the increasing marginal cost of alignment; the coverage points follow $V(s)-V(0)=A(s)+C(s)$; and integrated denoiser disagreement agrees with divergence computed directly from the densities.
Together, these checks support the central claim that one information-theoretic construction accounts for both the tradeoff and its diversity consequence.
The small-$s$ approximation is accurate near the origin, as expected for a local description.

\paragraph{Disjoint modes and subset selection.}
Block (c) checks, on the five-mode pair, that for disjoint modes the divergence between two mixtures is the divergence between their weights.
Let $p=\sum_k\pi_k\phi_k$ and $p^c=\sum_kq_k\phi_k$, with the components $\phi_k$ supported on disjoint sets.
On the support of $\phi_k$ the density ratio is $p^c/p=q_k/\pi_k$, so $D_\alpha(p^c\|p)=D_\alpha(q\|\pi)$ at every order $\alpha$.
For uniform weights $\pi_k=1/K$ this gives $Ke^{-D_\alpha(p^c\|p)}=e^{H_\alpha(q)}$, the Hill number of order $\alpha$~\cite{hill1973diversity}, with $H_\alpha(q)=\frac1{1-\alpha}\log\sum_kq_k^\alpha$ the R\'enyi entropy of the weights.
If the condition keeps a subset $S$ of the modes with their relative weights, $q_k=\pi_k/\pi(S)$ on $S$, then $D_\alpha=-\log\pi(S)$ at every order, and the index is the retained probability, $N=e^{-D_\alpha}=\pi(S)$.
The five modes are fifteen standard deviations apart, so they are disjoint to the precision of the table.
In the table, the continuous divergence between the mixtures equals the discrete divergence between their weight vectors at every order, and the order-two Hill number $Ke^{-D_2}$ equals the inverse Simpson index of the weights.
When the condition keeps $K'$ of the five modes with their relative weights, the index returns the retained probability exactly, $N_2=\pi(S)$, which is the subset-selection reading of \cref{eq:guided-coverage} (\cref{lem:supp:subset}).
On the class law, the exact clean divergence is $\mathrm{KL}=0.354$ and $D_{1.5}=0.483$.

\paragraph{Entropy is not coverage.}
Block (d) gives, at $s=1$, the KL cost and the entropy lost by the tilt on five pairs.
The ratio of the two ranges from zero, for the shift, where guidance translates the law without changing its spread, to above four.
This is the point of \cref{eq:coverage-entropy-split}: the entropy change and the cost are different quantities, and no universal factor relates them.

\begin{table*}[t]
    \centering
    \footnotesize
    \setlength{\tabcolsep}{5pt}
    \begin{tabular}{lccc}
        \toprule
        \multicolumn{4}{l}{\emph{(a) R\'enyi I-MMSE, narrow pair, $\sigma=0.55$}} \\
        $s$ & $-\partial_tD_{1+s}$ (finite diff.) & $\tfrac{1+s}2\mathbb E_{p_{\gamma,s}}\|\nabla_zT\|^2$ & rel.\ err. \\
        \midrule
        $-0.5$ & $0.802813$ & $0.802813$ & $7\times10^{-8}$ \\
        $0.5$  & $1.019194$ & $1.019194$ & $1\times10^{-7}$ \\
        $1.0$  & $1.068192$ & $1.068192$ & $1\times10^{-7}$ \\
        $1.5$  & $1.105035$ & $1.105035$ & $9\times10^{-8}$ \\
        \midrule
        \multicolumn{4}{l}{\emph{(b) Frontier, narrow pair, $\sigma=0.45$}} \\
        $s$ & $C$: direct / $s^2\partial_sD_{1+s}$ / $s\partial_sF-F$ & $A$: direct / $\partial_sF(s)-\partial_sF(0)$ & $dC/dA$ \\
        \midrule
        $0.25$ & $0.009544$ / $0.009544$ / $0.009544$ & $0.081459$ / $0.081459$ & $0.2500$ \\
        $0.5$  & $0.030933$ / $0.030933$ / $0.030933$ & $0.139473$ / $0.139473$ & $0.5000$ \\
        $1.0$  & $0.088116$ / $0.088116$ / $0.088116$ & $0.217807$ / $0.217807$ & $1.0000$ \\
        $1.5$  & $0.150945$ / $0.150945$ / $0.150945$ & $0.268719$ / $0.268719$ & $1.5000$ \\
        \midrule
        \multicolumn{4}{l}{\emph{(c) Disjoint modes, five-mode pair, clean laws}} \\
        $\alpha$ & continuous $D_\alpha(p^c\|p)$ & discrete $D_\alpha(q\|\pi)$ & \\
        \midrule
        $1$ & $0.71356$ & $0.71356$ & \\
        $2$ & $0.73397$ & $0.73397$ & $Ke^{-D_2}=2.6316=1/\sum_kq_k^2$ \\
        $3$ & $0.75358$ & $0.75358$ & \\
        $K'$ modes kept & $-\log\pi(S)$ & $D_2$ by quadrature & $N_2=e^{-D_2}$ \\
        \midrule
        $1$ / $2$ / $3$ / $4$ & $2.303$ / $1.204$ / $0.511$ / $0.163$ & $2.303$ / $1.204$ / $0.511$ / $0.163$ & $0.10$ / $0.30$ / $0.60$ / $0.85$ \\
        \midrule
        \multicolumn{4}{l}{\emph{(d) Entropy lost per nat of cost at $s=1$}} \\
        pair & $C(1)$ & $H(p^c_\gamma)-H(p_{\gamma,1})$ & ratio \\
        \midrule
        Shift & $1.1111$ & $0.0000$ & $0.00$ \\
        Nested Gaussians & $0.0540$ & $0.2518$ & $4.66$ \\
        Narrow & $0.0881$ & $0.3132$ & $3.55$ \\
        Mode swap & $0.2244$ & $0.2254$ & $1.00$ \\
        Three-mode collapse & $0.0647$ & $0.2634$ & $4.07$ \\
        \bottomrule
    \end{tabular}
    \caption{\textbf{Closed-form checks.}
    All entries are quadrature on the exact densities; no denoiser or estimator is involved.
    (a) The two sides of the R\'enyi I-MMSE identity.
    (b) Cost and gain by independent routes, and the slope of the frontier.
    (c) The disjoint-mode reduction and the subset-selection value of the index.
    (d) The entropy change is not the KL cost; the two pairs not in \cref{tab:supp:toy-laws} are $\mathcal N(0,0.3^2)$ against $\mathcal N(0,0.8^2)$ and a three-mode collapse with weights $(0.5,0.3,0.2)$ against uniform.}
    \label{tab:supp:toy-closed}
\end{table*}

\subsection{Exact Denoisers: the Estimators Recover the Closed Form}
\label{sec:supp:toy-exact}

The second level runs the paper's estimators with the exact posterior means of the mixtures.
\Cref{tab:supp:toy-exact} reports the checks on the mode-swap pair.

\paragraph{Fixed-level Langevin targets the tilt.}
Block (a) compares the Langevin estimate of $\mathbb E_{p_{\gamma,s}}\|g\|^2$ (\cref{eq:supp:langevin-step}) with quadrature at two levels and two scales.
The chains land within $1$--$3\%$ of the exact tilt average, with the deviation set by the step size, which is the input to the ideal frontier of \cref{sec:tradeoff:theory}.
The alternative of reweighting conditional samples by $e^{sT_\gamma}$ gives the same averages while the effective sample size is healthy, and degrades as $s$ grows (effective sample fractions $0.92$, $0.74$, $0.57$ at $s=0.5$, $1$, $1.5$ at $\sigma=1$), which is why the paper uses Langevin.

\paragraph{The path score is the endpoint score.}
Block (b) checks the endpoint identity of \cref{lem:supp:endpoint} pointwise, in its path form \cref{eq:supp:T-path} (\cref{sec:supp:proof-sampler-below}).
The pathwise score of \cref{eq:supp:T-path} is accumulated along the channel path of each conditional sample and compared with the exact $T_\gamma$ at the path's endpoint.
The two agree with correlation above $0.998$ and a root-mean-square difference of about $0.03$ that shrinks with the step size as $\sqrt{\Delta\gamma}$; their means agree with the exact KL to three digits.
Block (c) checks the same identity through the log-normalizer: averaging $e^{s\Lambda}$ over conditional paths estimates $sD_{1+s}(Q_0\|Q_{-1})$ on path space, and with exact denoisers it reproduces the endpoint value $sD_{1+s}(p^c_\Gamma\|p_\Gamma)$ from quadrature, up to the positive step bias of the exponentiated pathwise score.
The Feynman--Kac route~\cite{skreta2025feynman} reweights guided trajectories instead of conditional ones: it averages $\exp\{\frac{s(1+s)}2\int_0^\Gamma\|g\|^2d\gamma\}$ under $Q_s$, which equals $\mathbb E_{Q_0}e^{s\Lambda}$ by \cref{lem:supp:girsanov}, and it gives the same values.

\paragraph{Sampler quantities by Girsanov.}
On the class law in $d=2$, the sampler baseline $V_Q(0)=\mathrm{KL}(Q_0\|Q_{-1})=\tfrac12I_0$, accumulated along conditional CFG trajectories and extrapolated in the step (\cref{eq:supp:Is-hat}), is $0.345$ against $0.346$ for the data-pool estimate of \cref{eq:coverage-immse} and $0.354$ for the exact clean KL; the remaining difference is the finite-noise start and the level grid.
A separate check with deliberately perturbed, non-gradient denoisers on a two-dimensional pair gives $\mathbb E_{Q_0}\Lambda=0.7665$ against $\tfrac12\int\mathbb E_{Q_0}\|g\|^2d\gamma=0.7646$, confirming that the path identities of \cref{eq:supp:path-kl} do not need exact denoisers.
On the shift pair, where the frontier is exactly parabolic (\cref{prop:supp:gaussian}), the path log-normalizer from simulated trajectories reproduces the analytic $sD_{1+s}$ ($0.9016$ against $0.9000$ at $s=0.5$, $2.4042$ against $2.4000$ at $s=1$) with a Jensen gap below $10^{-14}$.

\begin{figure*}[t]
    \centering
    \includegraphics[width=\textwidth]{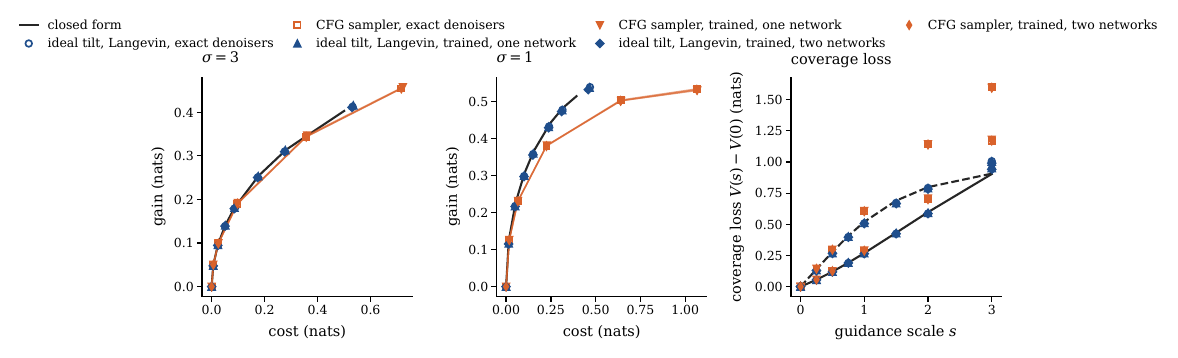}
    \caption{\textbf{The frontier and the coverage loss at the three levels} on the class law.
    Lines: closed form by quadrature.
    Blue markers: the ideal frontier reconstructed from fixed-level Langevin averages of $\|g\|^2$ (\cref{alg:langevin}), with exact denoisers (open) and with trained denoisers (filled).
    Orange markers: the operating points of the stochastic CFG sampler from its own trajectories (\cref{eq:sampler-point}), again with exact and trained denoisers.
    Left and middle: gain against cost at the horizons $\sigma=3$ and $\sigma=1$. Right: the coverage loss $V(s)-V(0)$ against the scale, for the ideal tilt and for the sampler (\cref{eq:sampler-coverage}), at $\sigma=3$ (solid line) and $\sigma=1$ (dashed line).}
    \label{fig:supp:toy-levels}
\end{figure*}

\begin{figure}[t]
    \centering
    \includegraphics[width=\linewidth]{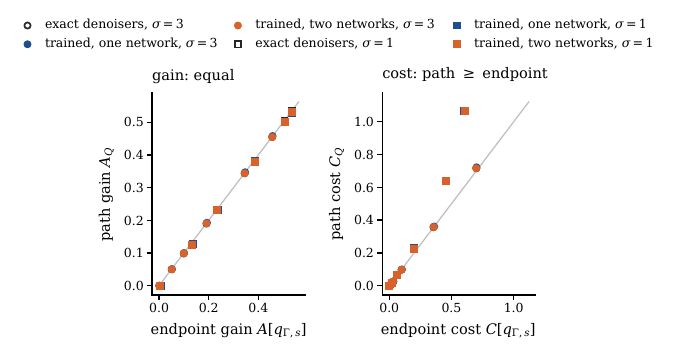}
    \caption{\textbf{Path against endpoint} (\cref{eq:supp:gain-equal,eq:supp:cost-ineq}).
    Each point is one guidance scale at one horizon.
    Left: the path gain $A_Q$ from the trajectories against the endpoint gain $A[q_{\Gamma,s}]$ from the exact alignment score at the sampler's final states; with exact denoisers the two agree.
    Right: the path cost $C_Q$ against the endpoint cost $\mathrm{KL}(q_{\Gamma,s}\|p^c_\Gamma)$, estimated from the final states; the path cost lies on or above the diagonal, and the distance from it is the path slack $S_s$.}
    \label{fig:supp:toy-bridge}
\end{figure}

\paragraph{The frontier, the sampler points and the endpoint bridge.}
\Cref{fig:supp:toy-levels,fig:supp:toy-bridge} run the complete pipeline of \cref{sec:supp:tradeoff-estimators} with the exact denoisers of the class law (open markers), at the horizons $\sigma=3$ and $\sigma=1$.
The ideal frontier is reconstructed from Langevin averages of $\|g\|^2$ on a ladder of $24$ levels between $\sigma_{\max}=80$ and the horizon ($1024$ chains per cell, step $h=0.025\sigma^2$, $600$ burn-in and $1000$ recorded steps), integrated over noise and differentiated across the scale grid by finite differences.
At $\sigma=3$ the reconstructed gain and cost are within $1$--$3\%$ of the closed form at every scale up to $s=2$; at $\sigma=1$ the agreement is the same up to $s=1.5$.
At the last scale, $s=3$, the finite differences on the scale grid dominate: putting the closed-form $F$ through the same differences gives a cost of $0.429$ against the true $0.393$ at $\sigma=1$, and the Langevin value is $0.468$.
The sampler points come from $32\,768$ trajectories of \cref{eq:cfg-sde} on $440$ steps, with a Richardson correction from $220$ steps.
They lie on the closed-form frontier at $\sigma=3$ and below it at $\sigma=1$, where the gap between the sampler and the ideal tilt opens up at larger scales, in agreement with \cref{eq:path-frontier-bound}.
The sampler baseline $V_Q(0)=\tfrac12I_0$ equals the closed-form $V(0)$ to three digits at both horizons ($0.1023$ and $0.2965$ against $0.1023$ and $0.2971$).

\Cref{fig:supp:toy-bridge} checks the endpoint bridge \cref{eq:supp:gain-equal,eq:supp:cost-ineq} directly.
The endpoint gain is the exact alignment score $T_\Gamma$ averaged over the sampler's final states, and it equals the path gain from the trajectories to within $0.001$--$0.006$ nats at every scale and horizon; per trajectory, the pathwise score and the exact endpoint score have correlation above $0.995$ at $\sigma=3$ and above $0.93$ at $\sigma=1$.
The endpoint cost is $\mathrm{KL}(q_{\Gamma,s}\|p^c_\Gamma)$, estimated from the final states by a nearest-neighbor entropy ($k=10$) plus the exact cross term; its bias is visible at $s=0$, where the truth is zero and the estimate is $0.007$ at $\sigma=3$ and $0.002$ at $\sigma=1$.
At $\sigma=3$ the path cost and the endpoint cost coincide within this bias: at such a high-noise horizon the endpoint carries essentially all the information in the trajectory.
At $\sigma=1$ the path cost exceeds the endpoint cost increasingly with the scale, by $0.03$ nats at $s=1$, $0.18$ at $s=2$ and $0.46$ at $s=3$.
This difference is the path slack $S_s$ of \cref{eq:path-endpoint-bridge}: it is positive, and the path cost is a conservative charge for the sampler's departure from the conditional law, as stated in \cref{sec:tradeoff:theory}.

\paragraph{The slack bound.}
A separate check tests the bound \cref{eq:supp:slack-bound} where the slack is known exactly.
It uses one-dimensional mixtures with exact denoisers, a bimodal class law inside a three-part mixture and a two-class pair, at horizons from $\sigma=1$ to $\sigma=0.25$ and scales up to $s=8$.
The exact slack is computed with every law on a grid and no sampling, and the steps of the proof in \cref{sec:supp:proof-sampler-below,sec:supp:slack-formula} hold up to a step-size error that halves when the number of steps doubles.
The bound is then estimated as in \cref{alg:sampler}, from $1000$ sampled end points with up to $1024$ bridges each.
For every number of bridges it is above the exact slack, and it comes close to it as the number grows.
In the hardest case, the first mixture at $\sigma=0.25$ and $s=8$, the exact slack is $2.66$ nats and the bound is $11.57$, $5.45$, $3.54$, $2.92$, $2.74$ and $2.68$ at $n=1$, $4$, $16$, $64$, $256$ and $1024$.
At $\sigma=0.5$ and $s=4$ the exact slack is $0.52$ and the bound is $0.75$ at $n=4$ and $0.58$ at $n=64$.
In these mixtures most of the distance between a sampler point and the frontier is slack: at $\sigma=0.5$ and $s=4$ the distance is $0.57$ nats, of which $0.52$ is slack.
For DiT-XL/2 the bound gives the opposite at moderate scales (\cref{sec:supp:slack-bound}): up to $s=2$, at least about $40\%$ of the distance is a shortfall of the end-point law in every class, so the split of the gap depends on the pair of laws.

In \cref{fig:supp:toy-levels}, the frontier reconstructed from denoiser disagreement follows the closed-form curve, and the trained-head markers closely track the exact-head markers.
At $\sigma=1$, CFG departs visibly from the ideal frontier even with exact denoisers.
This controlled example supports the main paper's distinction between the ideal tilt and the sampler: the separation can arise from the sampling dynamics themselves.
The coverage panel further shows how the same measurements account for the accompanying loss of coverage, using endpoint quantities for the tilt and path quantities for CFG.

\Cref{fig:supp:toy-bridge} shows agreement between path and endpoint alignment gains, while the path cost can exceed the endpoint cost, especially at the lower-noise horizon and stronger guidance.
This checks the two distinct statements of \cref{eq:supp:gain-equal,eq:supp:cost-ineq}: the trajectories recover the endpoint gain for exact denoisers, and charge a conservative cost for the departure of the endpoint law from conditional sampling.
Consequently, the plotted gap to the frontier is the sum of two parts, as \cref{thm:sampler-below} states: the divergence of the endpoint law from the ideal tilt of the same gain, and the path slack, which the right panel shows directly.

\begin{table}[t]
    \centering
    \footnotesize
    \setlength{\tabcolsep}{2.6pt}
    \begin{tabular}{lccc}
        \toprule
        \multicolumn{4}{l}{\emph{(a) Langevin tilt averages $\mathbb E_{p_{\gamma,s}}\|g\|^2$, mode-swap pair}} \\
        $\sigma$, $s$ & Langevin & quadrature & rel.\ err. \\
        \midrule
        $1.00$, $0.5$ & $0.1227$ & $0.1209$ & $+1.4\%$ \\
        $1.00$, $1.0$ & $0.1250$ & $0.1226$ & $+2.0\%$ \\
        $0.41$, $0.5$ & $0.0060$ & $0.0058$ & $+3.4\%$ \\
        $0.41$, $1.0$ & $0.0052$ & $0.0052$ & $+1.0\%$ \\
        \midrule
        \multicolumn{4}{l}{\emph{(b) Pathwise score vs.\ exact $T_\gamma$ at the endpoint, $5\times10^4$ paths}} \\
        $\sigma$ & RMS$(\hat T-T)$ & corr & mean $\hat T$ / mean $T$ \\
        \midrule
        $1.41$ & $0.0255$ & $0.9985$ & $0.0921$ / $0.0916$ \\
        $1.00$ & $0.0313$ & $0.9984$ & $0.1358$ / $0.1353$ \\
        $0.71$ & $0.0352$ & $0.9984$ & $0.1679$ / $0.1673$ \\
        $0.41$ & $0.0368$ & $0.9985$ & $0.1963$ / $0.1956$ \\
        \midrule
        \multicolumn{4}{l}{\emph{(c) $sD_{1+s}$ at $\sigma=0.41$: path routes vs.\ quadrature}} \\
        $s$ & quadrature & conditional paths (ESS) & guided paths (ESS) \\
        \midrule
        $0.5$ & $0.1542$ & $0.1583$ ($0.89$) & $0.1548$ ($0.98$) \\
        $1.0$ & $0.4197$ & $0.4289$ ($0.69$) & $0.4224$ ($0.86$) \\
        $2.0$ & $1.2026$ & $1.2225$ ($0.46$) & $1.2101$ ($0.32$) \\
        $3.0$ & $2.1589$ & $2.1890$ ($0.39$) & $2.1605$ ($0.02$) \\
        \bottomrule
    \end{tabular}
    \caption{\textbf{Estimators with exact denoisers} on the mode-swap pair.
    (a) Unadjusted Langevin at a fixed level ($2000$ chains, $4000$ steps, burn-in $1000$) against quadrature.
    (b) The pathwise score accumulated along channel paths ($\Delta\gamma=0.005$) against the exact log-ratio at the endpoint; the mean of either is the KL.
    (c) The log-normalizer from $\mathrm{logmeanexp}(s\Lambda)$ over conditional paths and from Feynman--Kac weights over guided paths, against $sD_{1+s}(p^c_\Gamma\|p_\Gamma)$ by quadrature; ESS is the effective sample fraction of the weights.}
    \label{tab:supp:toy-exact}
\end{table}

\subsection{Trained Denoisers: the Pipeline on Learned Heads}
\label{sec:supp:toy-trained}
\begin{figure*}[t]
    \centering
    \includegraphics[width=\textwidth]{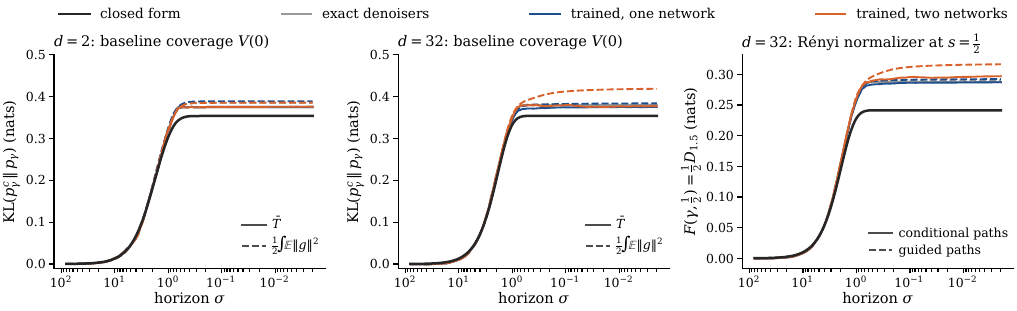}
    \caption{\textbf{Trained denoisers across horizons} on the class law in $d=2$ (left) and $d=32$ (middle, right).
    Left and middle: the baseline $V(0)=\mathrm{KL}(p^c_\gamma\|p_\gamma)$ as a function of the horizon, closed form (black) against the two estimators of \cref{sec:supp:toy-trained}: the mean pathwise score $\bar T$ (solid) and the integrated squared denoiser difference (dashed), run with exact denoisers (gray) and with MLP denoisers trained as one network with label dropout (blue) or as two separate networks (orange).
    Right: the R\'enyi normalizer $F(\gamma,\tfrac12)$ by the conditional-path and guided-path routes.
    The offsets of the exact-denoiser curves from the closed form are the discretization of the $110$-level grid; the one-network heads reproduce them, and the two-network pair drifts at low noise.}
    \label{fig:supp:toy-trained-profile}
\end{figure*}

The third level replaces the exact posterior means by MLP denoisers trained on samples of the class law, in $d=2$ and $d=32$.
Three designs are compared, because the image models differ in this respect.
\emph{Shared}: one network with label dropout provides both heads, as in DiT and Stable Diffusion.
\emph{Separate}: two networks, one conditional and one unconditional, as in EDM2.
\emph{Conservative}: one network that outputs a scalar potential whose gradient is the denoiser, so that $g$ is exactly a gradient.
All use the same backbone (four layers of width $512$, noise range $\sigma\in[0.002,80]$) and training budget ($20\,000$ steps, batch $4096$).
The exact heads are run through the same pipeline as a control, so that discretization error can be separated from head error.

\paragraph{Divergences from learned denoisers.}
\Cref{fig:supp:toy-trained-profile} follows the baseline KL and the R\'enyi normalizer across every horizon from $\sigma=80$ to the clean end, and \cref{tab:supp:toy-trained} reports, at the paper's horizon $\sigma=3$ and at the clean end, the baseline KL by its two estimators, the mean pathwise score $\bar T$ and the integrated squared denoiser difference of \cref{eq:coverage-immse}, and the R\'enyi normalizer $F(\gamma,\tfrac12)=\tfrac12D_{1.5}$ by the conditional-path and guided-path routes, each against the closed-form value.
In the figure, the shared-network estimates remain close to the exact-head estimates across horizons; their common offset from the density-based truth is numerical error already present with exact heads, while the separate-network curves in $d=32$ show an additional low-noise deviation.
With exact heads, both KL estimators sit within one standard error of the truth at $\sigma=3$, and about $0.02$ to $0.03$ nats above it at the clean end, which is the discretization error of the $110$-level grid.
The trained shared and conservative heads reproduce the exact-head numbers to within $0.01$ nats in $d=2$ and $d=32$, at both horizons.
The separate pair is also accurate in $d=2$; in $d=32$ its squared-difference estimator is inflated by $0.04$ nats at the clean end, the first sign of the fragility discussed below.
In all cases the learned denoisers are non-conservative, with a Jacobian asymmetry of $g$ between $0.3$ and $1.1$ at low noise, as large as on the image models, and the identities still hold to within the head error.
What matters is the size of the head error, not whether $g$ is a gradient.

\paragraph{Langevin and sampler checks with learned heads.}
On the $d=2$ law, long Langevin chains driven by the learned heads agree with the stationary law of their own drift, solved on a grid, to within $0.1$--$2.5\%$ at $\sigma=3$ and $\sigma=1$, for every head that is not under-trained; the fixed-level estimator therefore reaches the law it targets also for learned, non-gradient fields.
The sampler baseline $\tfrac12I_0$ along conditional CFG trajectories agrees with the data-pool KL to $0.3$--$1.5\%$ for every head trained for at least $3000$ steps.

\paragraph{What breaks, and how it shows.}
Under-trained heads ($300$ steps) fail visibly: the shared design overestimates the KL by $44\%$ and the separate design by $144\%$, the guided-path normalizer of the separate design explodes, and the mean pathwise score becomes negative and non-monotone in the horizon.
At the full budget the separate design remains the fragile one: across dimensions $d\in\{32,128,256,512,1024\}$, widths and seeds, its squared-difference estimator is inflated by anywhere between $0.04$ and $4$ nats, whereas the shared and conservative designs stay within $0.07$ nats of the exact heads in every configuration tried.
This is the basis for the caution in the main text that learned denoisers give approximations, and for reporting all image-model quantities at a finite horizon.

\begin{table*}[t]
    \centering
    \footnotesize
    \setlength{\tabcolsep}{3.5pt}
    \begin{tabular}{llcccccccccc}
        \toprule
        & & \multicolumn{5}{c}{$\sigma=3$} & \multicolumn{5}{c}{$\sigma=0.002$ (clean end)} \\
        \cmidrule(lr){3-7}\cmidrule(lr){8-12}
        $d$ & heads & $\mathrm{KL}^\ast$ & $\bar T$ & $\tfrac12\!\int\!\mathbb E\|g\|^2$ & $F^\ast$ & $F$: cond.\ / guided & $\mathrm{KL}^\ast$ & $\bar T$ & $\tfrac12\!\int\!\mathbb E\|g\|^2$ & $F^\ast$ & $F$: cond.\ / guided \\
        \midrule
        2  & exact        & $0.104$ & $0.099\pm0.007$ & $0.104$ & $0.076$ & $0.076$ / $0.077$ & $0.354$ & $0.374\pm0.014$ & $0.385$ & $0.241$ & $0.279$ / $0.285$ \\
        2  & shared       & & $0.100\pm0.007$ & $0.106$ & & $0.077$ / $0.078$ & & $0.376\pm0.014$ & $0.389$ & & $0.280$ / $0.286$ \\
        2  & separate     & & $0.099\pm0.007$ & $0.104$ & & $0.076$ / $0.077$ & & $0.376\pm0.014$ & $0.385$ & & $0.279$ / $0.283$ \\
        2  & conservative & & $0.101\pm0.007$ & $0.106$ & & $0.077$ / $0.078$ & & $0.374\pm0.014$ & $0.386$ & & $0.278$ / $0.280$ \\
        \midrule
        32 & exact        & $0.114$ & $0.119\pm0.008$ & $0.114$ & $0.083$ & $0.087$ / $0.084$ & $0.354$ & $0.379\pm0.015$ & $0.381$ & $0.241$ & $0.287$ / $0.290$ \\
        32 & shared       & & $0.117\pm0.008$ & $0.113$ & & $0.086$ / $0.083$ & & $0.375\pm0.015$ & $0.384$ & & $0.287$ / $0.292$ \\
        32 & separate     & & $0.118\pm0.008$ & $0.114$ & & $0.087$ / $0.084$ & & $0.377\pm0.015$ & $0.418$ & & $0.297$ / $0.317$ \\
        32 & conservative & & $0.118\pm0.008$ & $0.113$ & & $0.087$ / $0.083$ & & $0.384\pm0.015$ & $0.385$ & & $0.291$ / $0.293$ \\
        \bottomrule
    \end{tabular}
    \caption{\textbf{Trained denoisers on the class law.}
    Starred columns are the closed-form values of the KL and of $F=F(\gamma,\tfrac12)=\tfrac12D_{1.5}(p^c_\gamma\|p_\gamma)$ at the horizon.
    $\bar T$ is the mean pathwise score over $4096$ noised class samples with its standard error; the next column is the integrated squared denoiser difference of \cref{eq:coverage-immse}; the last two columns estimate $F(\gamma,\tfrac12)$ from conditional paths and from Feynman--Kac weighted guided paths.
    The exact-head rows are the same pipeline with the closed-form denoisers, so their offset from the starred values is discretization ($110$ levels, prior-extension start).
    Trained heads: MLP, four layers of width $512$, $20\,000$ steps.}
    \label{tab:supp:toy-trained}
\end{table*}

\paragraph{The frontier and the sampler with trained denoisers.}
The filled markers in \cref{fig:supp:toy-levels,fig:supp:toy-bridge} repeat the exact-denoiser pipeline of \cref{sec:supp:toy-exact} with the two trained designs in $d=2$, on the same chains, trajectories and noise.
The reconstructed frontier, the sampler points, the coverage loss and the endpoint bridge all land on the exact-denoiser values: the gain and cost from Langevin differ from the exact-head reconstruction by at most $0.005$ nats at $\sigma=3$ and $0.01$ nats at $\sigma=1$, the accumulated disagreement $I_s$ differs by less than $2\%$, and the path gain still equals the endpoint gain to within $0.01$ nats, although nothing in the derivation of that identity holds for a learned field.
In two dimensions at this training budget the head error is simply too small to show; the profile figure in $d=32$ is where the two-network design begins to drift.
The trained-head runs share every setting with the exact-head runs, except that the prior means at the start are the networks' own mean predictions at the largest noise.

\newpage
\section{Experimental Details}
\label{sec:supp:details}

This section gives the settings of all experiments, those of the main text and those of \cref{sec:supp:experiments}.
It is organized by setting, so each setting is stated once and holds for every experiment that uses it.
Gain and cost are in nats and are computed in the latent space of each model.
Throughout, $s=0$ is the conditional baseline, so the usual CFG weight is $1+s$.
The estimators are derived in \cref{sec:supp:tradeoff-estimators}, and the models are listed in \cref{tab:supp:models}.

\subsection{Models and Latent Spaces}
\label{sec:supp:details-models}

\begin{table*}[t]
    \centering
    \footnotesize
    \setlength{\tabcolsep}{5pt}
    \begin{tabular}{lllllll}
        \toprule
        Model & Images & Latent & Unconditional prediction & Steps & $\sigma$ range & Horizon $\sigma$ \\
        \midrule
        DiT-XL/2~\cite{peebles2023scalable} & $256\times256$ & $32{\times}32{\times}4$ & null class & 110 & $157$ to $0.01$ & $2.99$ \\
        EDM2-XS~\cite{karras2024analyzing} & $512\times512$ & $64{\times}64{\times}4$ & separate network & 110 & $80$ to $0.002$ & $3.02$ \\
        SD1.5~\cite{rombach2022high} & $512\times512$ & $64{\times}64{\times}4$ & empty prompt & 106 & $14.6$ to $0.029$ & $2.94$ \\
        SD3.5 Medium~\cite{esser2024scaling} & $512\times512$ & $64{\times}64{\times}16$ & empty prompt & 128 & $117$ to $0.077$ & $2.99$ \\
        FLUX.2 Klein-base 4B and 9B~\cite{bfl2025flux2} & $256\times256$ & $16{\times}16{\times}128$ & empty prompt & 128 & $248$ to $0.103$ & $2.92$ \\
        \bottomrule
    \end{tabular}
    \caption{\textbf{Models.}
    Image size, latent shape, the source of the unconditional prediction, and the noise grid of each model, with the grid level at which gain, cost and baselines are reported (\cref{sec:supp:details-horizon}).
    These settings hold for every experiment that uses the model.
    The only exception is the nested zebra descriptions of \cref{fig:supp-zebra-chain}, where SD3.5 runs at $256\times256$.}
    \label{tab:supp:models}
\end{table*}

DiT-XL/2 and EDM2 are conditioned on ImageNet classes.
SD1.5, SD3.5 Medium and FLUX.2 are conditioned on text prompts.
We use the undistilled Klein-base versions of FLUX.2, with 4B and 9B parameters.
EDM2 appears only in the classification and baseline experiments of \cref{sec:supp:tradeoff-classification,sec:supp:diversity-models}.
These experiments also test a larger variant, which pairs the S conditional network with the XS unconditional one.

SD3.5 and FLUX.2 are flow models that predict a velocity $v_\theta(z_t,t)$.
We convert it to a clean-image estimate by $\hat x=z_t-t\,v_\theta(z_t,t)$, at the noise level $\sigma=t/(1-t)$.
All other steps are then the same as for the diffusion models.

\subsection{Conditions and Image Pools}
\label{sec:supp:details-conditions}

\paragraph{ImageNet classes.}
The CFG sweep of DiT-XL/2 covers ten classes: zebra, volcano, coral reef, goldfish, balloon, golden retriever, Labrador retriever, tabby cat, Egyptian cat, and hook.
The classification experiment (\cref{sec:supp:tradeoff-classification}) and \cref{fig:supp-baseline-models}(a) use the same ten.
The ideal frontier (\cref{fig:supp-ac-dit}) is also estimated for ten classes.
Seven are shared with the sweep, and hourglass, clog and football helmet replace goldfish and the two retrievers.
The condition comparison of \cref{fig:supp-conditions} adds 20 classes drawn at random: jay, American chameleon, garden spider, whippet, toy poodle, dung beetle, clog, coffee mug, Dutch oven, football helmet, hourglass, limousine, loudspeaker, maypole, planetarium, soccer ball, submarine, window screen, cheeseburger, and potpie.

\Cref{fig:supp-baseline-models}(b) and the ImageNet diversity study (\cref{sec:supp:diversity-imagenet}) share one set of 100 classes.
It holds the ten sweep classes, 40 classes chosen to span narrow and broad classes, and 50 drawn at random from the rest.
The part of the study that follows the guidance scale uses 30 classes, the ten of the sweep and the 20 random ones above, which are not among the 100.

\paragraph{Text prompts.}
\Cref{tab:supp:prompts} lists the prompts of the tradeoff experiments and of \cref{fig:diversity-flux}.
The prompts of \cref{fig:supp-diversity-one-scale,fig:supp-diversity-sets-sd35,fig:supp-diversity-sets-flux} are listed in \cref{tab:supp:diversity-prompt-sets}.

\begin{table*}[t]
    \centering
    \footnotesize
    \setlength{\tabcolsep}{5pt}
    \begin{tabular}{@{}p{.25\textwidth} p{.72\textwidth}@{}}
        \toprule
        Experiment & Prompts \\
        \midrule
        SD1.5 tradeoff (\cref{fig:supp-ac-sd15,fig:supp-conditions}) &
        ``a photo of \dots'': a dog, a zebra, a yellow school bus, a pizza, a dog in space, a pizza with whipped cream, a cat, a bus, a horse, a train \\
        \addlinespace
        FLUX.2 Klein tradeoff (\cref{fig:supp-conditions,fig:supp-flux-apg-images,fig:supp-flux-slack}) &
        4B: the first five SD1.5 prompts, and ``a photo of a corgi dog in space riding a bicycle'' and ``a photo of a zebra in a snowy forest'' for the ideal frontier. 9B: ``a photo of a zebra'' \\
        \addlinespace
        SD3.5 tradeoff (\cref{fig:supp-conditions}) &
        ``a photo of \dots'': a raccoon wearing a crown on a throne made of pizza, an octopus playing the piano in an underwater concert hall, a black cat astronaut floating inside a space station, a black cat napping on a windowsill, a fox sitting in the snow \\
        \addlinespace
        Zebra chain (\cref{fig:supp-zebra-chain}) &
        ``a photo of a zebra in a snowy forest''. Each of the next five prompts appends one clause to the one before it: ``, wearing pilot goggles'', ``, pulling a wooden sled'', ``, with a flamingo sitting on the sled'', ``, the flamingo holding a lantern'', ``, both made of gold'' \\
        \addlinespace
        FLUX.2 diversity (\cref{fig:diversity-flux}) &
        ``A breathtaking view of a distant house in beautiful scenery'', ``A photo of a dog playing in a park'', ``An oil painting of an old sailor'', ``A photo of a yorkshire terrier playing in a park'', ``a photo of a yorkshire terrier dog in space riding a bicycle'', ``a photo of a corgi dog in space riding a bicycle'' \\
        \bottomrule
    \end{tabular}
    \caption{\textbf{Prompts of the text-conditional experiments.}
    All models of the zebra chain (SD1.5, SD3.5 and FLUX.2 Klein 4B) use the same six prompts.}
    \label{tab:supp:prompts}
\end{table*}

\paragraph{Prompt ladders.}
The ladders of \cref{fig:diversity-prompts,fig:supp-diversity-prompts-vendi} are run on SD1.5.
Every ladder starts at ``a photo'', and every other prompt begins ``a photo of''.
We list the rest of each prompt, and ``+'' marks words appended to the prompt before it.
The three main ladders are:
an animal, a dog, a golden retriever, a golden retriever puppy on grass;
food, a pizza, a pepperoni pizza on a wooden table;
a vehicle, a bus, a red double-decker bus in London.
The ten ladders suggested by an LLM are:
furniture, a couch, a leather couch in a living room;
electronics, a laptop, + on a desk, + next to a cup of coffee;
a kitchen appliance, a refrigerator, a stainless steel refrigerator in a kitchen;
sports, a skateboarder, + doing a trick, + on a ramp;
sports, a tennis player, + on a clay court;
a person, a woman, + holding an umbrella, + in the rain;
an animal, a cat, a black cat, + lying on a bed;
food, a cake, a birthday cake, + with candles;
a vehicle, an airplane, a passenger jet, + on a runway;
a vehicle, a train, a steam train, + with smoke.
Prompts shared by several ladders are run once, which gives 43 prompts.

\paragraph{Real-image pools.}
The baseline comparison across models (\cref{sec:supp:diversity-models}) and the ImageNet diversity study use 256 ImageNet training images per class, and both models encode the same image files.
The classification experiment scores 30 validation images per class, and its deployed classifier uses 32 per class.
The SD1.5 check of the deployed classifier uses 32 COCO~\cite{lin2014microsoft} images for each of four captions (giraffe, pizza, dog, and bus).
COCO images that match the prompt also start the Langevin chains of SD1.5 and FLUX.2 where such images exist (\cref{sec:supp:details-langevin}).
For the prompt ladders, each prompt has a pool of 64 or 128 matched COCO images (\cref{sec:supp:details-pools}).

\subsection{Noise Grids and Horizon}
\label{sec:supp:details-horizon}

\paragraph{Noise grids.}
Each model has one noise grid, and all experiments of that model use it.
We place 256 levels with equal ratios of the SNR over the $\sigma$ range of the model and keep every second level.
For DiT-XL/2 and SD1.5, each level is moved to the nearest trained timestep and repeated timesteps are merged, which leaves slightly fewer steps.
For EDM2 we place 221 levels in place of 256, which gives 110 steps.
\Cref{tab:supp:models} gives the number of steps and the $\sigma$ range of each grid.

\paragraph{Reporting horizon.}
The paper reports gain, cost and baselines at a horizon of $\sigma=3$ (\cref{sec:supp:horizon}).
In practice this is the grid level nearest $\sigma=3$, and \cref{tab:supp:models} gives its exact value for each model.
There are two exceptions.
The ImageNet diversity study of \cref{sec:supp:diversity-imagenet} reads all quantities at $\sigma\approx2$, and the classification experiment of \cref{sec:supp:tradeoff-classification} reports the accuracy at every grid level.
The horizon applies only to the reported quantities.
Every displayed image is decoded from a trajectory that runs to the last level of the grid, while runs that are used only for gain and cost may stop at the horizon.

\paragraph{Other horizons.}
For \cref{fig:supp:horizon-sweep}, the analysis of the ten DiT-XL/2 classes is repeated with the integrals stopped at $\sigma=10$, $6$, $4.5$, $3.75$, $3.3$ and $3$.
The frontier at a horizon uses the Langevin levels of \cref{sec:supp:details-langevin} at and above it, and the sampler is read along its trajectory at the grid level nearest to the horizon.
Below $\sigma=3$ we add Langevin chains at $\sigma=2.6$, $2.3$ and $2$ for four classes (zebra, coral reef, hook and football helmet), with the same settings, at the guidance scales $0$, $0.25$, $0.5$, $1$, $2$, $4$ and $6$, and for zebra also at $1.5$ and $3$.
The error bars are one standard error, from a bootstrap over the Langevin chains and the sampler trajectories.
The example of \cref{fig:supp:toy-horizon} is a two-dimensional Gaussian mixture, computed in closed form by quadrature.
Its three classes are centered on a circle of radius $10$, and each class is a ring of radius $0.15$ of six components of standard deviation $0.03$.
The unconditional law gives equal weight to the classes and to the components; the conditional law gives the classes the weights $(0.7,0.15,0.15)$ and, inside every class, the components the weights $(0.4,0.25,0.15,0.1,0.05,0.05)$.

\paragraph{Finite-noise start.}
The integrals start at zero SNR, while the grids start at a finite noise level.
The sampler, frontier and diversity estimates therefore add the constant prior extension of \cref{eq:supp:prior-extension}, computed from the mean conditional and unconditional predictions at the largest noise level.
The path estimates on real images in the classification and baseline experiments (\cref{sec:supp:details-pools}) do not add it and start from zero at the largest noise level.

\subsection{CFG Trajectory Runs}
\label{sec:supp:details-paths}

All sampler points in the paper and in this supplement are measured on trajectories of the stochastic sampler of \cref{eq:cfg-sde}, simulated with \cref{alg:sampler}.
\Cref{tab:supp:cfg-runs} lists the number of trajectories and the guidance scales of each set of runs.
The other guidance rules change only the drift of this sampler (\cref{sec:supp:details-guidance}).

\begin{table}[t]
    \centering
    \footnotesize
    \setlength{\tabcolsep}{3pt}
    \begin{tabular}{@{}llrl@{}}
        \toprule
        Runs & Conditions & Traj. & Scales $s$ \\
        \midrule
        DiT-XL/2, sweep & 10 classes & 256 & 37 values in $[0,20]$ \\
        DiT-XL/2, random & 20 classes & 256 & 20 values in $[0,8]$ \\
        SD1.5 & 10 prompts & 128 & 30 values in $[0,20]$ \\
        FLUX.2 4B & 6 prompts & 64 & $0,0.25,0.5,1,2,\ldots,20$ \\
        FLUX.2 9B & zebra & 64 & $0,1,2,4,8,16$ \\
        SD3.5 & 5 prompts & 64 & $0,1.5,3,5.25,8,12$ \\
        Zebra descriptions & 6 per model & 128/64 & $0,2,4,8,16$ \\
        \Cref{fig:diversity-flux} & 6 prompts & 128/64 & see text \\
        Diversity prompt sets & \cref{tab:supp:diversity-prompt-sets} & 64 & $0,1.5,3,5.25,8,12$ \\
        \bottomrule
    \end{tabular}
    \caption{\textbf{CFG trajectory runs.}
    Trajectories per condition and scale, and the guidance scales.
    The 37 DiT-XL/2 scales are $0$, $0.05$, $0.1$, $0.15$, $0.2$, $0.25$, $0.375$, $0.5$, $0.75$, $1$, $1.25$, $1.5$, then steps of $0.5$ up to $8$ and integers up to $20$.
    The 20-scale grid is the same up to $4$, followed by $5$, $6$ and $8$.
    The 30 SD1.5 scales are $0$, $0.25$, $0.5$, $1$, then steps of $0.5$ up to $8$ and integers up to $20$.
    The nested zebra descriptions (\cref{fig:supp-zebra-chain}) use 128 trajectories for SD1.5 and 64 for SD3.5 and FLUX.2 4B.
    \Cref{fig:diversity-flux} uses FLUX.2 4B, and the diversity prompt sets are run with FLUX.2 4B and SD3.5.
    The conditions are listed in \cref{sec:supp:details-conditions}.}
    \label{tab:supp:cfg-runs}
\end{table}

\paragraph{Shared noise.}
Within each condition, the initial noise and the Brownian increments are shared across scales and across guidance methods.
For the nested descriptions of \cref{fig:supp-zebra-chain}, they are also shared across the descriptions within each model.

\paragraph{Trajectories and scales.}
The six FLUX.2 4B prompts of \cref{tab:supp:cfg-runs} are the five prompts of \cref{fig:supp-conditions} and the corgi prompt of \cref{fig:supp-flux-slack}.
In \cref{fig:diversity-flux}, the house, dog in a park and sailor prompts use 128 trajectories per scale, and the other three prompts use 64.
Its image rows are at $s\in\{0,1.5,5.25,12\}$ ($w=1,2.5,6.25,13$).
Its curves are drawn at $s\in\{0,0.5,1,1.5,2,3,4,5,5.25\}$ for the house prompt and at $s\in\{0,1.5,3,5.25\}$ for the other five.

\paragraph{Estimates.}
Gain and cost are computed from the left-point sums in SNR of \cref{eq:supp:Is-hat}, averaged over trajectories.
The errors of the alignment gain use paired differences against the $s=0$ run with the same noise.
For \cref{fig:diversity-flux} and the diversity prompt sets, we report $A_Q+C_Q$ relative to the $s=0$ run of the same prompt.
The reporting horizon and the finite-noise start are in \cref{sec:supp:details-horizon}.

\paragraph{Slack bound.}
The slack bound of \cref{fig:supp-ac-dit}(a), of the bars in \cref{fig:tradeoff-imagenet}(a) and of \cref{fig:supp-flux-slack} uses the optional loop of \cref{alg:sampler}.
The bridges start from the end points of the sampler runs at the horizon, and each bridge sum uses the same grid as the trajectory sum.
For DiT-XL/2, we run $n=64$ bridges on each of $64$ end points at $s\in\{1,2,4,8\}$, for the ten classes of \cref{fig:supp-ac-dit} (the hourglass also at $s=0.5$).
Zebra is the exception, with $n=4$ bridges on each of its $256$ end points at $s\in\{0.25,0.5,1,1.5,2\}$ and $n=64$ bridges on $64$ end points at $s\in\{3,4,6,8,12\}$.
For the FLUX.2 4B corgi prompt, we run $n=64$ bridges on each of $8$ end points at $s\in\{0.5,1,1.5,2,3,4,5,6,7,8\}$, and for the zebra prompt at $s\in\{1,2,4,6,8\}$.
There the $n$ bridges of an end point are run together as a particle system.
At each noise level every bridge is weighted by $e^{-\kappa\|g\|^2\Delta_k}$ for that level, with the step $\Delta_k$ of \cref{eq:supp:Is-hat}, and the bridges are then resampled in proportion to these weights.
The estimate of $m(y)$ is the product over the levels of the mean weight of the $n$ bridges, which is the normalizing-constant estimate of a particle filter.
Like the plain mean $\hat m_n(y)$ of \cref{eq:supp:m-hat}, it is unbiased for $m(y)$ for every $n$, so \cref{eq:supp:jensen} applies to it and the slack bound holds on average.
The resampling only lowers its variance, which makes the bound tighter at the same $n$.
Inside the bound, $C_Q$ and $I_s$ are taken on the end points that carry the bridges, and the standard errors are over end points.

\paragraph{Outside distance.}
For \cref{fig:supp-cost-kid}, DiT-XL/2 uses 100 ImageNet classes at $s\in\{0,0.5,1,2,4\}$ with 256 trajectories.
They are the ten sweep classes, the 20 random classes of \cref{sec:supp:details-conditions} and 70 more drawn at random.
The unguided reference of each class is a second $s=0$ run with independent noise.
SD1.5 uses 30 prompts ``a photo of a $\langle$category$\rangle$'', for 30 COCO categories, at $s\in\{0,0.25,0.5,0.75,1,1.5,2,3,4\}$ with 256 trajectories, and the guided samples of 128 trajectories are compared with the unguided samples of the other 128.
The KID~\cite{bińkowski2018demystifying} is computed on the class-token features of DINOv2-base~\cite{oquab2024dinov} for the decoded samples.
We use its unbiased estimate with the usual cubic kernel $(x^\top y/d+1)^3$, where $d$ is the feature dimension, without rescaling.

\subsection{Other Guidance Rules}
\label{sec:supp:details-guidance}

APG~\cite{sadat2025eliminating}, CFG++~\cite{chung2025cfgpp} and Annealing Guidance~\cite{yehezkel2025navigating} are compared with CFG in \cref{fig:supp-ac-methods,fig:supp-ac-dit,fig:supp-ac-dit-images,fig:supp-flux-apg-images} and in \cref{fig:tradeoff-imagenet}(d).
We do not run these methods in their original samplers.
Each rule only changes the drift of our stochastic sampler (\cref{alg:sampler}), on the grid, the trajectories and the noise of the CFG runs (\cref{sec:supp:details-paths}).
We write the drift as $\hat x_c+u$, where CFG is $u=s\,g$.
Along each trajectory we sum $\|u\|^2/2$ for the cost and $\|g\|^2/2+g^\top u$ for the gain, with the weights of \cref{eq:supp:Is-hat}, and subtract the gain sum of the conditional run ($u=0$).
For $u=s\,g$ these sums are the CFG estimates.

\paragraph{APG.}
APG changes the direction of the guidance: $u=s\,(g_\perp+\eta\,g_\parallel)$, where $g_\parallel$ is the projection of $g$ onto $\hat x_c$ and $g_\perp=g-g_\parallel$.
We use this projection with $\eta=0.25$, the setting with the largest gain over CFG in a pilot on the zebra class.
In the same pilot, the norm threshold of APG moved the points along the same curve, and its momentum, which acts per step, lowered the gain on our finer grid; we therefore leave both out.
DiT-XL/2 uses the integer scales $s=0,\ldots,20$.
FLUX.2 4B uses $s=1,\ldots,20$ on the dog, zebra, school bus and pizza prompts, and FLUX.2 9B is run on the zebra prompt.

\paragraph{CFG++.}
In its original DDIM sampler, CFG++ is CFG with a scale that changes along the trajectory: one step from $\sigma$ to $\sigma_{\mathrm{next}}$ with parameter $\lambda$ equals a CFG step with the scale $1+s_{\mathrm{eff}}=\lambda/(1-\rho)$, $\rho=\sigma_{\mathrm{next}}/\sigma$.
We run this scale schedule in our sampler on DiT-XL/2, with $\rho$ from the 50-step DDIM grid of the original method.
The values are $\lambda=0.05,0.1,\ldots,1$ and $\lambda\in\{1.2,1.4,1.6\}$.

\paragraph{Annealing Guidance.}
Annealing Guidance replaces $\lambda$ in the CFG++ schedule by the output of a small learned scheduler.
The released scheduler is trained for SDXL, and none exists for DiT-XL/2, the model on which we compare all rules.
We therefore train one with the released code, keeping its network, loss and training settings and replacing the prompt by the class label.
It is trained on ImageNet training latents from 100 classes, with the diffusion model frozen.
We use its output as it is, for $\lambda\in\{0.05,0.4,0.7,0.8,1\}$.

\subsection{Ideal-Frontier Runs}
\label{sec:supp:details-langevin}

The ideal frontiers come from Langevin chains at fixed noise levels, run separately from the sampler trajectories (\cref{alg:langevin}, with the theory in \cref{sec:supp:tradeoff-estimators}).
\Cref{tab:supp:langevin} lists the runs.

\paragraph{Chains.}
The step size at noise level $\sigma$ is $h=0.025\,\sigma^2$ in the coordinate of \cref{eq:supp:langevin-step}.
We set the chain length with a rough relaxation time of $(0.7+\sigma^2)/h$ steps.
This is the relaxation time of a Gaussian with variance $0.7+\sigma^2$, where $0.7$ stands for the variance of a clean latent coordinate.
A chain runs for 30 relaxation times for DiT-XL/2 (about 1,200 to 1,300 steps) and for 20 for SD1.5 and FLUX.2 (about 800 to 860 steps).
The first third of each chain is discarded.

The chains start from real images of the condition, noised to the level.
For DiT-XL/2 these are ImageNet training images of the class, and for the text models they are COCO images that match the prompt (\cref{sec:supp:details-conditions}).
Two SD1.5 prompts have no matching images, the dog in space and the pizza with whipped cream.
Their chains start from end points of the conditional sampler ($s=0$).
The corgi prompt of FLUX.2 starts from COCO dog images, and its zebra prompt from COCO zebra images.
Within a condition, the same starting images and the same noise are used at every scale.

\begin{table}[t]
    \centering
    \footnotesize
    \setlength{\tabcolsep}{3pt}
    \newcommand*{\rr}{\raggedright\hspace{0pt}}%
    \begin{tabular}{@{}p{1.7cm}rp{2.45cm}p{2.65cm}@{}}
        \toprule
        \rr Model and conditions & Chains & Noise levels $\sigma$ & Scales $s$ \\
        \midrule
        \rr DiT-XL/2, zebra & $64$ & $40$, $20$, $10$, $6$, $4.5$, $3.75$, $3.3$, $3$ & $0$, $0.1$, $0.25$, $0.5$, $0.75$, $1$, $1.5$, $2$, $3$, $4$, $6$, $8$ \\
        \addlinespace
        \rr DiT-XL/2, nine classes & $64$ & as above & $0$, $0.25$, $0.5$, $1$, $2$, $4$, $6$, \ldots, $20$ \\
        \addlinespace
        \rr SD1.5, ten prompts & $32$ & $14.6$, $9$, $6$, $4.5$, $3.75$, $3.3$, $3$ & $0$, $1$, \ldots, $20$ \\
        \addlinespace
        \rr FLUX.2 4B, corgi prompt & $32$ & $248$, $60$, $15$, $9$, $6$, $4.5$, $3.75$, $3.3$, $3$ & $0$, $1$, \ldots, $20$ \\
        \addlinespace
        \rr FLUX.2 4B, zebra prompt & $32$ & as above & $0$, $1$, $2$, $4$, $6$, $8$, $10$ \\
        \bottomrule
    \end{tabular}
    \caption{\textbf{Ideal-frontier runs.} Chains per noise level and scale, noise levels, and scales. The classes and prompts are listed in \cref{sec:supp:details-conditions}.}
    \label{tab:supp:langevin}
\end{table}

\paragraph{Reconstruction.}
The chains give the mean squared disagreement at each noise level and scale.
We interpolate it between the levels, linearly in log--log coordinates against the SNR, and integrate by the trapezoid rule to get $G$, adding the prior extension of \cref{sec:supp:details-horizon}.
The slopes of $F=s(1+s)G/2$ between neighboring scales are then fitted by isotonic regression, weighted by their inverse variances, anchored at $F'(0)=G(0)/2$, and $F$ is the integral of the fitted slopes.
The frontier is $A=F'-F'(0)$ and $C=sF'-F$, up to the last midpoint of the scale grid, so the largest scale gives no frontier point.
Errors come from a bootstrap that resamples the chains, the same chains at every level and scale, and repeats the whole reconstruction.
The fit imposes convexity (\cref{sec:supp:tradeoff-estimators}), so the shape of the frontier is not a test of convexity.
The proposition is checked without this constraint on the mixtures of \cref{sec:supp:toy-closed}, where the slope $dC/dA$ by finite differences equals the scale at every point.
On DiT-XL/2, the slopes before the fit are nondecreasing at every scale for six of the ten classes; in the other four they decrease at $8$ of the $118$ pairs of neighboring slopes in total, each time by at most $1.1$ bootstrap standard errors.

\subsection{Estimates on Real Images}
\label{sec:supp:details-pools}

Some experiments estimate the score $T_\gamma$ or the baseline $V(0)$ on a fixed pool of images and not on guided trajectories.
The pool holds real images or conditional samples of the model (\cref{sec:supp:details-conditions}).
We use two estimates, both on the noise grid of the model (\cref{sec:supp:details-horizon}).

\paragraph{Path estimate.}
Each image is noised along one path on the grid.
The pointwise identity \cref{eq:supp:pointwise} is accumulated along this path, which gives a score for each image at every level.
In the classification experiment (\cref{sec:supp:tradeoff-classification}), each validation image is scored under all ten classes on one shared path.
The predicted class of an image at a noise level is the class with the largest score there.
For the baseline across models (\cref{sec:supp:diversity-models}), $V(0)$ of a class is the mean score at the horizon over the training images of the class, and the error bars are standard errors over images.

\paragraph{Pool estimate.}
The pool is noised to each grid level, and the two predictions are evaluated there.
Following \cref{eq:coverage-immse}, $V(0)$ is half the integral in SNR of the mean squared disagreement over the pool, computed by the trapezoid rule up to the horizon, plus the prior term of \cref{eq:supp:prior-extension}.
The prompt ladders (\cref{fig:diversity-prompts,fig:supp-diversity-prompts-vendi}) use this estimate with independent noise at each level.
The pool of a prompt holds 128 conditional samples drawn with the $s=0$ sampler of \cref{sec:supp:details-paths}.
The dashed curve and the hollow points of \cref{fig:diversity-prompts} use a pool of real COCO images instead, 64 per prompt on the three main ladders and 128 on the ten ladders, or all matching images when there are fewer.
A broad prompt is matched by its COCO supercategory or category and a more specific prompt by words in the captions, with the rules of \cref{sec:supp:diversity-coco}.
The ImageNet diversity study (\cref{sec:supp:details-vendi}) uses the same estimate, with each image noised along one path.

\paragraph{Linear probe and deployed classifier.}
The two reference classifiers of \cref{fig:supp-classifier} follow previous work and do not use the estimates above.
The linear probe is a logistic regression on the clean latents of 256 training images per class.
The deployed classifier computes the likelihood of a clean image under the conditional probability-flow ODE of each class.
We solve the ODE with Heun's method at 40 and 80 steps, combine the two solutions by Richardson extrapolation, and estimate the divergence with Hutchinson probes.
The larger EDM2 conditional network and the SD1.5 check on COCO images use the same solver.

\subsection{Diversity Measures and Statistics}
\label{sec:supp:details-vendi}

\paragraph{Vendi score.}
We compute the Vendi score~\cite{friedman2023vendi} with a cosine kernel on the class-token features of DINOv2-base~\cite{oquab2024dinov}.
Each score uses 256 decoded images in the ImageNet study and 128 in the prompt ladders, where the images are the samples on which $V(0)$ is estimated (\cref{sec:supp:details-pools}).

\paragraph{ImageNet study.}
The study of \cref{sec:supp:diversity-imagenet} uses DiT-XL/2 and the 100 classes of \cref{sec:supp:details-conditions}.
Each class has two pools, 256 real training images and 256 conditional samples drawn with the $s=0$ sampler.
The baseline $V(0)$ is the pool estimate of \cref{sec:supp:details-pools} on the real images, and it is compared with the Vendi score of both pools.
All coverage terms are read at $\sigma\approx2$ (\cref{sec:supp:details-horizon,sec:supp:horizon}).
The sampler rows of \cref{tab:diversity-checks} use 256 trajectories per class at $s\in\{0,0.5,1\}$, with noise shared across scales, and $A_Q+C_Q$ is estimated as in \cref{sec:supp:details-paths}.
The diversity lost at scale $s$ is the change in log Vendi from the $s=0$ samples to the samples at $s$.
Rank agreement is the Spearman correlation across classes, with $95\%$ percentile bootstrap intervals from 2000 resamples of the classes.
The intervals of the baseline across models (\cref{sec:supp:diversity-models}) are computed in the same way.

\paragraph{Ideal-tilt samples.}
The ideal-tilt rows use Langevin chains (\cref{alg:langevin}) at $\sigma\approx2$ and $s\in\{0.5,1\}$, with step $h=0.05\sigma^2$.
Each class has 32 chains, started from training images of the class noised to this level.
After 100 burn-in steps we record 768 steps and keep every 96th state, which gives 256 states per class.
Each state is denoised in one step with the guided prediction $\hat x_c+sg$ and decoded.
The $s=0$ reference is the 256 real images of the class, noised to the same level and passed through the same step.
The predictor $A+C$ comes from the fixed-level estimate of \cref{sec:supp:details-langevin}, with 32 chains per class at $\sigma\in\{20,10,6,4,3,2\}$.
With only two positive scales, $F$ is the cubic through $F(0)=0$ and the measured $F(0.5)$ and $F(1)$, with its slope at zero set to the baseline.

\paragraph{Along the guidance scale.}
The bottom rows of \cref{tab:diversity-checks} and \cref{fig:diversity-imagenet}(c) follow the 30 classes of \cref{sec:supp:details-conditions}, the ten of the sweep and the 20 random ones.
Each class and scale has 256 trajectories, run as in \cref{sec:supp:details-paths}.
The sweep classes use the 16 positive scales of their grid up to $s=4$, and the 20 random classes use $s\in\{0.25,0.5,0.75,1,1.5,2,3,4\}$, which gives 320 (class, scale) cells with $s>0$.
$A_Q+C_Q$ is read at $\sigma\approx2$, and Vendi is computed on the 256 decoded samples of each cell.
The per-class row is the Spearman correlation between $-(A_Q+C_Q)$ and Vendi over the scales of one class, and the table reports its median over the classes.

\paragraph{Vendi gauge.}
\Cref{eq:vendi-gauge} has one floor $\mathrm{Vendi}_\infty$ per class and one rate $\nu$ shared by all classes, and $\mathrm{Vendi}(0)$ is the measured value.
Both parameters are fitted by least squares in log Vendi over the cells of the 30 classes.
The floor of a class is kept between one and the smallest Vendi measured for that class.
For the prediction of a class, $\nu$ is fitted on the other 29 classes, and the floor is then fitted on the class itself.
The last row of \cref{tab:diversity-checks} compares these predictions with the measured Vendi over the 320 cells.

\subsection{Displayed Images}
\label{sec:supp:details-images}

The DiT-XL/2 images of \cref{fig:tradeoff-imagenet}(e) and \cref{fig:supp-budget-images} are selected by cost budget.
For each budget in $\{0,7,25,60,110,175,250\}$ nats, we interpolate the scale that reaches it on the measured cost curve of the class and show the nearest measured scale.
Every image in a row is the first trajectory of the class, so the row shares its noise.
The FLUX.2 rows of \cref{fig:tradeoff-imagenet}(e) show the first five trajectories of each model at $s=1$.

In \cref{fig:supp-ac-dit-images,fig:supp-flux-apg-images}, we chose the noise realization of each class or prompt by hand, and all the methods share it.
The red frame in \cref{fig:supp-ac-dit-images} marks the method with the highest measured gain $A$ in its column.
In \cref{fig:supp-flux-apg-images}, APG uses $s\in\{2,4,\dots,14\}$ and each CFG partner is the integer scale nearest the CFG scale of equal cost.
The pictured pairs are therefore close in cost but not equal, and the curves give the comparison at matched cost.

In \cref{fig:diversity-flux,fig:supp-diversity-one-scale,fig:supp-diversity-sets-sd35,fig:supp-diversity-sets-flux}, the images are the decoded end points of the trajectories on which the index is estimated.
We drew six trajectories at random once and show the same six for every prompt and model, so no image is chosen by hand.

\subsection{Numerical Accuracy and Limitations}
\label{sec:supp:details-accuracy}

All the estimates carry finite-step, mixing and learned-score errors.
We checked the step size on DiT-XL/2 by rerunning zebra and hook trajectories with twice as many steps.
At the horizon, $I_s$ changes by up to about $8\%$ for zebra and by less than $5\%$ for hook.
We checked the mixing of the Langevin chains by comparing the two halves of the recorded window.
For zebra, the two halves differ by more than two standard errors in 13 of the 96 cells, and by at most $3.3$.
The ideal curves are therefore numerical estimates computed with learned predictions, while the identities of the paper are exact under their stated assumptions.

\subsection{Compute}
\label{sec:supp:details-compute}

\Cref{tab:supp:compute} gives the cost of the core computations for one condition, on the grids of \cref{sec:supp:details-paths,sec:supp:details-langevin}.
A network evaluation is one conditional or unconditional pass on one sample, so each step of a guided sampler or of a Langevin chain costs two.
GPU-hours are rounded estimates for one NVIDIA A6000.
The diversity index of \cref{sec:diversity} is read off the same trajectories as the sampler curve and needs no extra runs, and all the analyses that follow the runs take minutes on a CPU.

The ideal frontier dominates the cost.
It takes about $20$ GPU-hours per DiT-XL/2 class (about $200$ for the ten classes of \cref{fig:supp-ac-dit}) and about $200$ for the FLUX.2 corgi prompt (about $60$ for its zebra prompt on a coarser grid of scales), against $1$--$7$ GPU-hours for a full sampler curve (CFG or APG) and $2$--$8$ for a slack bound.
The sampler curve and the slack bound therefore scale to many conditions, while the frontier was computed only for the conditions shown in the figures.

\begin{table}[t]
    \centering
    \footnotesize
    \setlength{\tabcolsep}{3pt}
    \begin{tabular}{@{}llrr@{}}
        \toprule
        Model & Computation & Net.\ evals & GPU-h \\
        \midrule
        DiT-XL/2 & ideal frontier & $1.8\times10^7$ & $20$ \\
                 & CFG sampler curve & $5\times10^5$ & $1$--$2$ \\
                 & APG sampler curve & $1.2\times10^6$ & $1.5$ \\
                 & slack bound ($4$ scales) & $2\times10^6$ & $2.5$ \\
        \midrule
        SD1.5 & ideal frontier & $8\times10^6$ & $30$ \\
              & CFG sampler curve & $8\times10^5$ & $4$ \\
        \midrule
        FLUX.2 4B & ideal frontier & $10^7$ & $200$ \\
                  & CFG sampler curve & $4\times10^5$ & $7$ \\
                  & APG sampler curve & $3\times10^5$ & $6$ \\
                  & slack bound ($6$ scales) & $8\times10^5$ & $8$ \\
        \bottomrule
    \end{tabular}
    \caption{\textbf{Approximate compute per condition} (one class or prompt), for the runs of \cref{sec:supp:details-paths,sec:supp:details-langevin}. Ideal frontier: all noise levels and scales, with chains of about $1250$ steps (DiT-XL/2, the $14$ scales of the nine classes) or $830$ steps (SD1.5 and FLUX.2, $21$ scales). CFG sampler curve: for DiT-XL/2, the $20$-scale grid of the random classes with trajectories stopped at the horizon ($52$ steps), where the $37$-scale sweep run to the last level takes about four times as many evaluations; $30$ scales and $106$ steps for SD1.5; $23$ scales and $128$ steps for FLUX.2. APG sampler curve: $21$ scales and $110$ steps for DiT-XL/2, $20$ scales and $128$ steps for FLUX.2. Slack bound: $64$ bridges on each of $64$ (DiT-XL/2) or $8$ (FLUX.2) end points per scale. GPU-hours on one A6000.}
    \label{tab:supp:compute}
\end{table}

\end{document}